\documentclass[conference]{IEEEtran}
\AtBeginDocument{%
  \providecommand\BibTeX{{%
    \normalfont B\kern-0.5em{\scshape i\kern-0.25em b}\kern-0.8em\TeX}}}

\usepackage{cite}
\usepackage{algorithmic}
\usepackage{graphicx}
\usepackage{textcomp}
\usepackage{xcolor}
\usepackage{multirow}
\usepackage{subcaption}

\usepackage{siunitx}
\usepackage{booktabs}
\usepackage{chemformula}
\usepackage{collcell}
\usepackage{supertabular}

\newcommand{\fan}[1]{{\color{cyan}(Fan: #1)}}

\newcommand{\cz}[1]{{\color{red} #1}}
\newcommand{\m}[1]{{\color{black}#1}}

\usepackage{amsfonts}

\usepackage{soul}
\usepackage[normalem]{ulem}
\renewcommand\sout{\bgroup\markoverwith
{\textcolor{brown}{\rule[.5ex]{2pt}{1pt}}}\ULon}

\usepackage{bbm}
\usepackage{booktabs}
\usepackage{makecell}

\usepackage{graphicx}
\usepackage{xcolor,colortbl}

\colorlet{tabgray}{gray!30}

\usepackage[ruled,vlined,linesnumbered]{algorithm2e}
\SetKwComment{Comment}{$\triangleright$\ }{}

\usepackage[shortlabels]{enumitem}

\usepackage{hyperref}
\usepackage[nameinlink]{cleveref}
\crefname{ineq}{inequality}{inequalities}
\creflabelformat{ineq}{#2{\upshape(#1)}#3} 

\usepackage{longtable}
\usepackage{supertabular}

\usepackage{amsthm}

\usepackage[super]{nth}

\theoremstyle{definition}
\newtheorem{definition}{Definition}
\theoremstyle{plain}
\newtheorem{theorem}{Theorem}
\theoremstyle{plain}
\newtheorem{lemma}{Lemma}
\theoremstyle{plain}
\newtheorem{proposition}{Proposition}

\makeatletter
\def\old@comma{,}
\catcode`\,=13
\def,{%
  \ifmmode%
    \old@comma\discretionary{}{}{}%
  \else%
    \old@comma%
  \fi%
}
\makeatother

\newcommand{\eg}[0]{\textit{e.g.}}
\newcommand{\etal}[0]{\textit{et al.}\xspace}
\newcommand{\ie}[0]{\textit{i.e.}}

\newcommand{\eps}[0]{\varepsilon}

\newcommand{\rzero}[0]{\rightarrow0}

\newcommand{\linkteller}[0]{\textsc{LinkTeller}\xspace}
\newcommand{\edgerand}[0]{\textsc{EdgeRand}\xspace}
\newcommand{\lapgraph}[0]{\textsc{LapGraph}\xspace}

\newcommand{\edgedp}[0]{edge differential privacy\xspace}

\newcommand{\eedgedp}[0]{$\eps$-edge differential privacy\xspace}
\newcommand{\eedgedpshort}[0]{$\eps$-edge DP\xspace}
\newcommand{\Eedgedp}[0]{$\eps$-edge differentially private\xspace}

\newcommand{\norm}[1]{\lVert{#1}\rVert}
\newcommand{\dotp}[2]{\langle {#1},{#2} \rangle}
\newcommand{\bb}[1]{\mathbb{#1}}

\renewcommand{\cal}[1]{\mathcal{#1}}
\renewcommand{\bf}[1]{\mathbf{#1}}

\usepackage{float}
\usepackage{diagbox}

\newcommand{\PreserveBackslash}[1]{\let\temp=\\#1\let\\=\temp}
\newcolumntype{C}[1]{>{\PreserveBackslash\centering}p{#1}}
\newcolumntype{R}[1]{>{\PreserveBackslash\raggedleft}p{#1}}
\newcolumntype{L}[1]{>{\PreserveBackslash\raggedright}p{#1}}

\newcommand\given[1][]{\:#1\vert\:}

\makeatletter
\newcommand\footnoteref[1]{\protected@xdef\@thefnmark{\ref{#1}}\@footnotemark}
\makeatother

\everypar{\looseness=-1}
\begin{document}

\title{\linkteller: Recovering Private Edges from Graph Neural Networks
via Influence Analysis}

\author{
Fan Wu$^{1}$
\qquad
Yunhui Long$^{1}$
\qquad
Ce Zhang$^{2}$
\qquad
Bo Li$^{1}$
\\
$^{1}${\small University of Illinois at Urbana-Champaign}
\qquad
$^{2}${\small ETH Zürich}
\\
{\tt\footnotesize \{fanw6,ylong4,lbo\}@illinois.edu}
\qquad
{\tt\footnotesize ce.zhang@inf.ethz.ch}
}

\maketitle

\begin{abstract}

Graph structured data have enabled several successful applications such as recommendation systems and traffic prediction, given the rich node features and edges information.
However, these high-dimensional features and high-order adjacency information are usually heterogeneous and held by different data holders in practice.
Given such vertical data partition (\eg, one data holder will only own either the node features or edge information), different data holders have to develop efficient joint training protocols rather than directly transferring data to each other due to privacy concerns. 
In this paper, we focus on the \textit{edge privacy}, and consider a  training scenario where the data holder Bob with node features will first send  training node features to  Alice who owns the adjacency information. Alice will then train a graph neural network (GNN) with the joint information and provide an inference API to Bob. During inference time, Bob is able to provide test node features and query the API to obtain the predictions for test nodes.
Under this setting, we first propose a privacy attack \linkteller via influence analysis to infer the private edge information held by Alice via designing adversarial queries for Bob. We then empirically show that \linkteller is able to recover a significant
amount of private edges in different settings, \m{both including inductive (8 datasets) and transductive (3 datasets), under different graph densities}, significantly outperforming  existing baselines. 
To further evaluate the privacy leakage for edges, we adapt an existing algorithm for differentially private graph convolutional network (DP GCN) training as well as propose a new DP GCN mechanism \lapgraph based on Laplacian mechanism to evaluate \linkteller. 
\m{We show that these DP GCN mechanisms are not always resilient against \linkteller empirically under mild privacy guarantees ($\eps >5$)}.
Our studies will shed light on future research  towards designing more resilient privacy-preserving GCN models; 
in the meantime,
provide an in-depth understanding about the tradeoff between GCN model utility and robustness against potential privacy attacks.

\end{abstract}


\section{Introduction}

Graph neural networks (GNNs) have been widely applied to different domains owing to their ability of modeling the high-dimensional feature and high-order adjacency information on both homogeneous and heterogeneous graph structured data~\cite{long2015fully,wei2019mmgcn,zhou2020privacy}.
The high-quality graph structured data have enabled a range of successful applications, including traffic prediction~\cite{zhao2020traffic}, recommendation systems~\cite{ying2018recommendation}, and abnormal access detection~\cite{jiang2019abnormal}.
As these applications are becoming more and more prevalent, privacy concerns in these applications are non-negligible given the sensitive information in the graph data. Thus, undesirable outcomes may arise due to lack of understanding of the models and application scenarios.

\begin{figure}[t!]
\centering
\includegraphics[width=1.0\columnwidth]{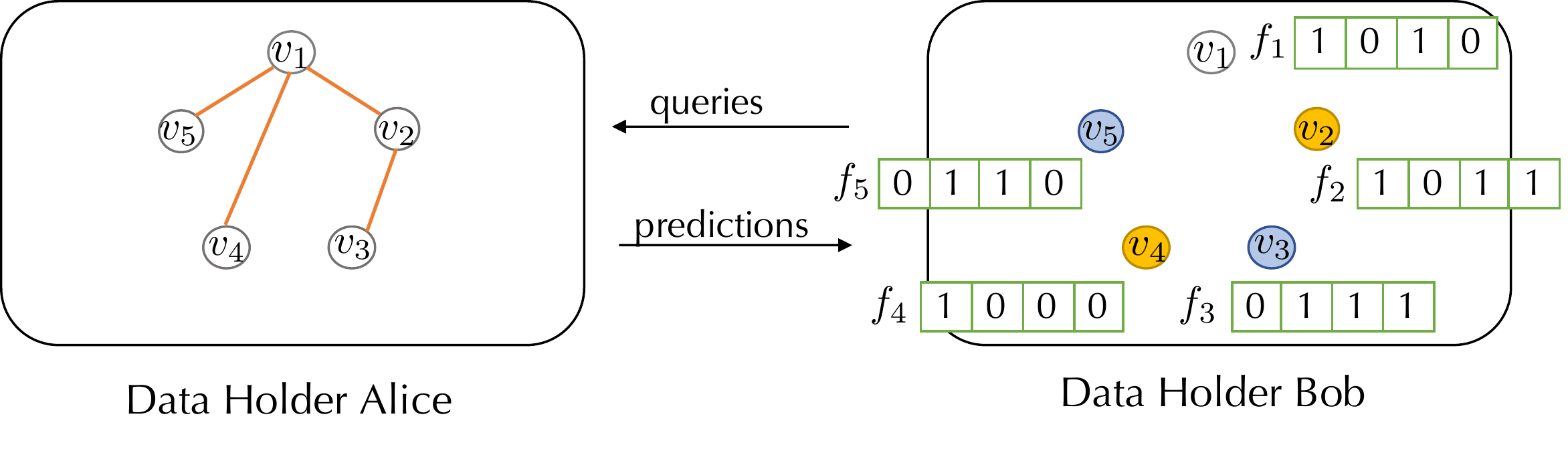}
\vspace{-1.5em}
\caption{\small Vertically partitioned graph  data for different data holders.}
\label{fig:pip}
\end{figure}

In this paper, we aim at
understanding the \textit{edge privacy} in applications
of GNN models. We 
focus on one specific 
 scenario as 
\textit{training/serving GNN models 
in a vertically partitioned graph 
setting}. As illustrated in~\Cref{fig:pip}, 
in this setting, node features
and adjacency information (edges) are isolated or hosted by different data holders.
Our interest in 
this setting is inspired 
not only by recent academic research
(\eg, Zhou \etal~\cite{zhou2020privacy}
proposed a
privacy-preserving GNN (PPGNN) mechanism via homomorphic encryption
under this setting)
but also a real-world 
industrial example we see and
the potential privacy risks it incurs. In such an example, an international 
Internet company hopes to 
train a single GNN model jointly 
using data collected by two of its subdivisions (namely, data holder Alice and Bob). Because these subdivisions focus on different products, their data are heterogeneous in nature. Specifically, in this example,  Alice collects user interaction (social network) data (\ie, adjacency information), and 
 Bob collects user behavior data (\ie, node features).
Noticing the potential benefit of 
integrating user interaction
data into its predictive model,
 Bob hopes to enrich the 
model using data collected by
 Alice. Although they belong to 
the same company, directly copying 
the user interaction data from Alice to Bob is not allowed due to privacy concerns. 
Thus, Bob will first send training data containing node features and labels to Alice, and Alice will train a GNN model jointly with her edge information. 
Then Alice will release an inference API to Bob. During inference, Bob would send a new set of test nodes with their features to query the API and obtain corresponding predictions.
\m{Different users can query the API to enjoy the service from Alice.}
\m{For instance, in practice there are  several ML/AI service platforms that provide similar interactions---taking the training data from users to train an ML model and providing inference APIs for users to make queries with their test data---such as Vertex AI~\cite{VertexAI} from Google Cloud, ParlAI platform~\cite{ParlAI} from Facebook, and InfoSphere Virtual Data Pipeline~\cite{InfoSphe} from IBM.}


During this type of interaction, \textit{the fundamental question is to understand the risks of edge privacy 
(mainly for data holder Alice) for training and releasing a GNN inference API on graph data, as well as possible ways to 
amortise such risks.} 

\vspace{0.3em}
\noindent
\textbf{\underline{{Challenges and Problem Formulation.}}}\quad
The main motivation and challenge of the problem attributes to
the heterogeneity of data---one data holder owns the features of users (\ie, node features), while the other holds the ``connections'' or ``interactions'' among users (\ie, adjacency information) as shown in~\Cref{fig:pip}. 
Inspired by this real-world example, we abstract it into 
the following technical 
problem. 
Let there be $n$ users and 
$A \in \{0,1\}^{n\times n}$
be the adjacency information. Data holder Alice has full access
to adjacency information $A$ while it is kept secret
from the data holder Bob. Bob interacts 
with Alice during both \textit{training} and \textit{inference} stages. 
\begin{enumerate}[leftmargin=*]
\item 
Training: During 
training, \m{Bob (or some other users)
collects (training) node features and labels 
for a subset of users}, forming
a feature matrix $X$ with 
label vector $y$, and sends them to Alice. 
Alice then trains a GNN model using all the node features from Bob and her collected adjacency information $A$, and releases an
inference API to Bob.
\item Inference: During
inference, Bob
collects features for another (test)
subset of users $X'$, and sends them to Alice via the inference
API, who will run inference 
using the trained GNN model, and return corresponding
predictions.
\end{enumerate}
Given this interaction model,
we aim to ask: Whether 
the inference API will leak private
information of the adjacency information to a potentially malicious user Bob indirectly? How can we better protect the adjacency information
from privacy leakage while preserving high model utility?

Apart from this specific case, there have been similar concerns from different real-world cases. For instance, the advertisement department of Facebook would usually hold certain public features of individuals (\ie, node features), and needs to query the predictions from another department that holds the social network connection information which is private. Thus, how to protect the edge privacy in this setting is critical.
However, directly conducting such privacy attacks is challenging. For instance, given a large graph, naively comparing the similarities between nodes to infer their connections is clearly not enough. On the other hand, it is known that the trained GNN is based on the node influence propagation~\cite{klicpera2018predict}: If two nodes are connected, there is a high chance that changing the features of one would affect the prediction of the other. 
Thus, we hope to address the research question: \textit{Is it possible to design an effective edge re-identification attack against GNNs based on the node influence information?}

Different from existing work~\cite{he2020stealing} which collects node pairs with and without connections to train a model to infer the existence of an edge, in this paper, we aim to analyze and leverage the \textit{node influence} to predict potential edge connections.
In particular, we first propose an attack strategy \linkteller under such a vertically data partitioned setting based on the node influence analysis, and explore how much the private adjacency information could be revealed from Alice via answering queries from Bob. 
Then we will evaluate the proposed \linkteller attack against both an existing and a proposed differentially private graph convolutional network (DP GCN)  mechanisms to analyze whether the  \linkteller could further attack the privacy preserving GCN models. In addition, we explore what is the \textit{safe} privacy budget to choose in order to protect the trained GCN models from being attacked by privacy attacks such as \linkteller on different datasets via extensive empirical evaluation.


\vspace{0.3em}
\noindent
\textbf{\underline{Technical Contributions.}}\quad 
In this paper, we focus on 
understanding the \textit{edge privacy} risk 
and the strength of the privacy protection mechanisms (\eg, DP) for vertically partitioned graph learning.
Specifically, we make following contributions.
\begin{enumerate}[leftmargin=*]
\item We propose the \textit{first} query based edge re-identification attack \linkteller against GNN models
by considering the influence propagation in GNN training. We 
show that
it is possible to re-identify private edges  effectively in a vertically partitioned graph learning setting.
\item \looseness=-1 We explore and evaluate the proposed \linkteller attack against different DP GCN mechanisms as countermeasures.
Since there is no DP GCN mechanism proposed so far, we evaluate  \linkteller against a standard DP strategy \edgerand on graph, and a proposed DP GCN approach \lapgraph. 
\item
\looseness=-1
We provide formal privacy analysis for the two DP GCN approaches and an upper bound for general edge re-identification attack success rate on DP GCN mechanisms.
\item
\m{We design extensive experiments on eight datasets under the inductive setting and three datasets under the transductive setting to show that the proposed \linkteller is able to achieve high attack precision and recall, and significantly outperforms the random attack and two state of the art methods.
We show that both DP GCN approaches 
are not always resilient against \linkteller empirically under mild privacy guarantees.}
\item We systematically depict the empirical tradeoff space between (1) 
\textit{model utility}---the quality of
the trained GCN model, and (2) 
\textit{privacy vulnerability}---the risk of a GCN model being successfully attacked.  We 
carefully analyze different regimes
under which a data holder
might want to take different 
actions \m{via evaluating a range of privacy budgets, and we also analyze such tradeoff by selecting a privacy budget via a validation dataset}. 
\end{enumerate}



\section{Preliminaries}

\vspace{-1mm}
\subsection{Graph Neural Networks}
\vspace{-1mm}

Graph Neural Networks (GNNs)~\cite{wu2020comprehensive} are commonly used in semi-supervised node classification tasks on graphs. Given a graph $G=(V,E)$ with $V$ denoting the nodes ($n=|V|$) and $E$ the edges, the adjacency matrix $A\subseteq \{0,1\}^{n\times n}$ is a sparse matrix, where $A_{ij}=1$ denotes the existence of an edge from node $i$ to node $j$. 
Since Graph Convolutional Network (GCN)~\cite{Kipf2016gcn} is one most representative class of GNN, we next introduce GCN, which is a stack of multiple graph convolutional layers as defined below:
\begin{align}
\label{eq:gcn-layer}
    H^{l+1}=\sigma(\widehat A { H}^{l} W^l),
\end{align} where $\widehat A$ is the normalized adjacency matrix derived using a certain normalization technique and $\sigma$ is the activation function.
For the $l$-th graph convolutional layer, we denote the input node embeddings by ${H}^l$, the output by ${H}^{l+1}$, and the learnable weight by $W^l$. 
Each graph convolutional layer constructs the embeddings for each node by aggregating the embeddings of the node's neighbors from the previous layer. Specifically, ${H}^0$ is the node feature matrix $X$.

GNNs were first proposed for transductive training where training and testing occur on the same graph. 
Recently, inductive learning has been widely studied and applied~\cite{hamilton17graphsage,rong2020dropedge,zeng20graphsaint,xu2018representation}, which is a setting where the trained GNNs are tested on unseen nodes/graphs. 
There are two main application scenarios for the inductive setting: 1) training on an evolving graph (\eg, social networks, citation networks) for future use when more nodes arise in the graph; 2) training on one graph belonging to a group of similarly structured graphs, and transfer the model to other graphs from the similar distribution.
We consider both 
the inductive and transductive settings in this paper.

\vspace{-1mm}
\subsection{Differential Privacy}
\vspace{-1mm}
Differential privacy~\cite{dwork14dp} is a privacy notion that ensures an algorithm only outputs general information about its training data without revealing the information of individual records.
\begin{definition}[Differential Privacy]
    A randomized algorithm $\mathcal{M}$ with domain $\mathbb N^{|\cal X|}$ is $(\eps,0)$-differentially private if for all $\cal S\subseteq $ Range($\cal M$) and for all $x,y\in \bb N^{|\cal X|}$ such that $\lVert x-y\lVert_1\leq 1$:
    \begin{align*}
        \operatorname{Pr}[\mathcal{M}(x) \in\cal  S] \leq \exp (\varepsilon) \operatorname{Pr}[\mathcal{M}(y) \in\cal  S]
    \end{align*}
\end{definition}
There are two extensions of differential privacy to preserve private information in graph data. Edge differential privacy~\cite{karwa2014private} protects the edge information, while node differential privacy~\cite{kasiviswanathan2013analyzing} protects the existence of nodes. 
A recent work has proposed an algorithm to generate synthetic graphs under edge local differential privacy~\cite{qin2017generating}, which provides privacy protection when the graph data is distributed among different users. 
We consider a practical privacy model in the data partitioning scenario where one data holder only owns either the edge or node information, and we aim to protect the edge information from being leaked during the training and inference processes.



\section{\linkteller: Link Re-identification Attack}
In this section, we focus on understanding the risk of edge
privacy leakage caused by 
exposing a GNN trained on private graph data via an inference API.
We first describe the \textbf{interaction model} between data holders, and then 
the \linkteller
algorithm that 
probes the \textit{inference
values} between pairs of nodes and uses these
values as our confidence 
on whether edges 
exist between pairs of nodes.
As we will see, this attack allows us to 
recover a significant number of edges 
from the private graph structured data.

\vspace{-1mm}
\subsection{Interaction Model between Data Holders}
\vspace{-1mm}
We consider an ML application based on graph structured data, where
different data holders have access to different information of the graphs (\eg, nodes or edges). 
More specifically, the graph edge information is not available to everyone, since the edge connections or interactions between the node entities usually contain sensitive information, which can be exploited for malicious purposes.
Thus, we first make the following abstraction of the data holder interaction.

\begin{figure}
    \centering
    \includegraphics[width=0.9\columnwidth]{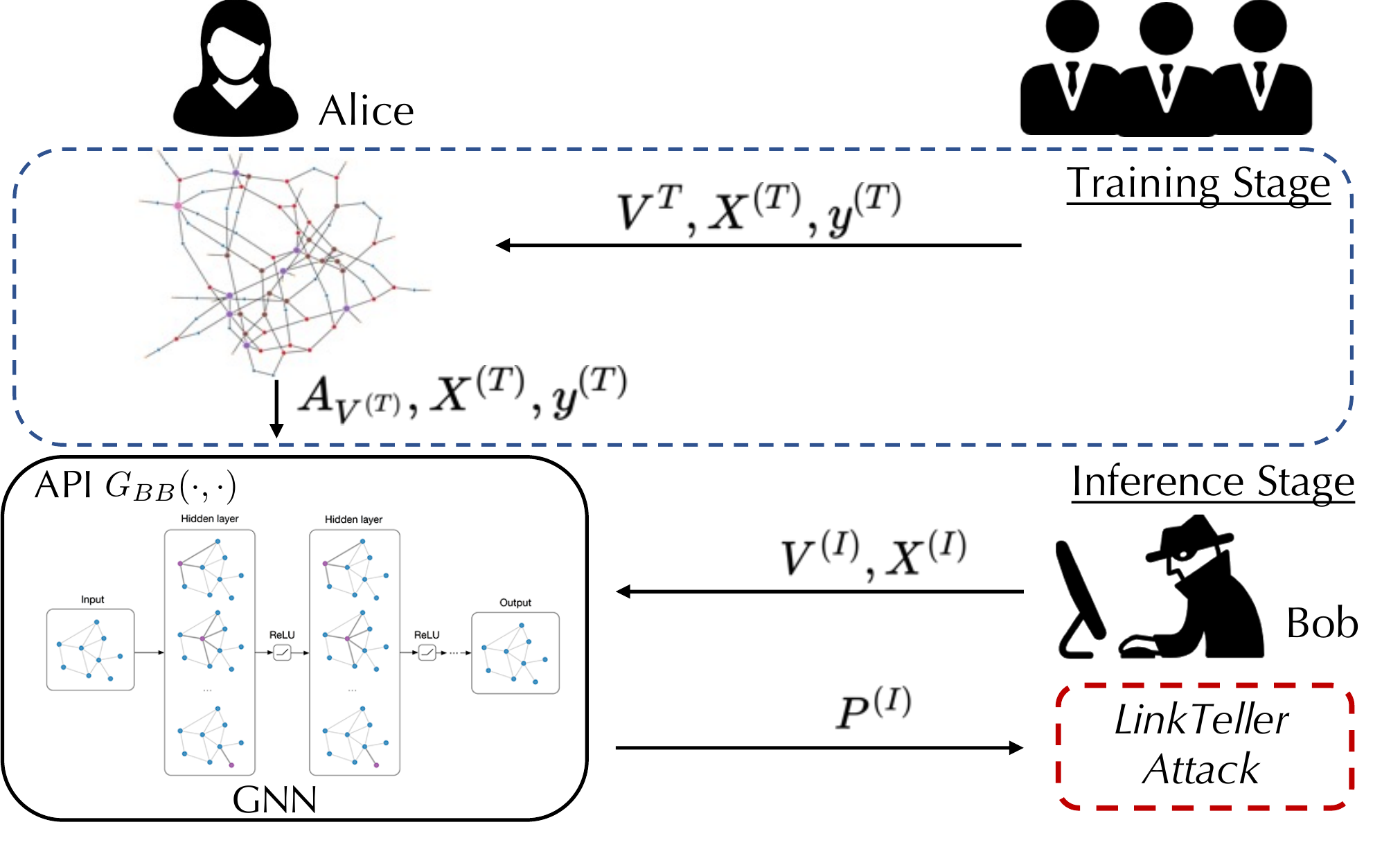}
    \caption{\small \textbf{The interaction model.} \m{In the training stage, first, some users send Alice the node set $V^{(T)}$, the associated features $X^{(T)}$ and labels $y^{(T)}$. Next, Alice trains a GNN model with corresponding adjacency matrix $A_{V^{(T)}}$, $X^{(T)}$, and $y^{(T)}$. In the inference stage, an adversarial user Bob queries Alice with a test node set $V^{(I)}$ and associated features $X^{(I)}$. Alice outputs the prediction matrix $P^{(I)}$.
    }}
    \label{fig:int-model}
\end{figure}

As shown in~\Cref{fig:int-model},
the data holder Alice holds private edge information of a graph, while \m{other  users hold the node information, and due to privacy concerns, the sensitive edge information from Alice cannot be directly shared.}
During the \textit{training stage}, in order to jointly train a GNN model on the graph data, \m{some users will first send the node features of the training graph $X^{(T)}$ together with their labels $y^{(T)}$ for a set of nodes $V^{(T)}$ to Alice; and Alice will train a GNN model together with her edge connection information for future inference purpose by releasing an inference API to external users.
}
During the \textit{inference stage}, \m{a potential adversarial user Bob will collect the node features of the inference graph $X^{(I)}$ (\eg, patients in the next month) and obtain their predicted labels from Alice via the inference API. }
Alice will then send the prediction matrix $P^{(I)}$ formed by the prediction vector for each node to Bob.
\m{Without loss of generality, in the following we will use ``Bob" to denote both the general users during training and inference time and the adversary during inference, although they are usually independent users in practice. }


Next, we will define such training interaction formally. In particular, we will use the lower case letter to denote a vector and the upper case letter to denote a matrix.
We denote the set of nodes of the training graph held by Bob as $V^{(T)}=\{v^{(T)}_1,v^{(T)}_2,\ldots, v^{(T)}_n\} \subseteq V$. 
In the training stage, data holder Bob will first send the corresponding node features 
${ X}^{(T)}$
and labels 
${ y}^{(T)}$ 
to Alice.
Here each label $y^{(T)}_i$ takes the value from a set of $c$ classes.
Then the data holder Alice who holds the node connection information will generate the adjacency matrix $A_{V^{(T)}}\subseteq \{0,1\}^{n\times n}$, where ${A_{V^{(T)}}}_{ij} = 1$ if and only if there is one edge between node $v^{(T)}_i$ and $v^{(T)}_j$. 
This way, Alice can leverage the node features and labels from Bob together with her adjacency information to train a GNN model and provide the model as a \textbf{blackbox} API, $G_{BB}(\cdot, \cdot)$, for Bob.
The learned model parameters are denoted as $\{{W}^i\}$, where
$W^i\in \bb R^{d_i\times d_{i+1}}$ 
represents the weight of the $i$-th layer.

During the inference stage, the data holder Bob who owns another set of nodes from the inference graph will query the inference API for node prediction. In particular, given a set of inference nodes $V^{(I)} \subseteq V$, Bob will send the associated node features $X^{(I)}$ to the trained GNN API $G_{BB}(\cdot, \cdot)$.
Then together with the private adjacency matrix of inference graph $A_{V^{(I)}}$, the API from Alice will make inference on the nodes, and following the standard commercial ML inference services
such as Clarifai~\cite{clarifai} and Google Vision API~\cite{VisionAI43}, Alice will send the logits information back to Bob as below.
\begin{align*}
    {G_{BB}}&(V^{(I)}, {X}^{(I)})=\mbox{GNN}(A_{V^{(I)}},{X}^{(I)}, \{W^i\}).
\end{align*}
For ease of reference, we denote the output \textit{prediction matrix} of $G_{BB}(V^{(I)}, {X}^{(I)})$ as $P^{(I)}$, which is of shape $|V^{(I)}|\times c$.
Each row of the prediction matrix corresponds to one node in $V^{(I)}$, and each column corresponds to the confidence for one class.
Alice will then send $P^{(I)}$ back to Bob. 

We discuss more 
on the properties of $V^{(T)}$, $V^{(I)}$, and $V$. 
It is worth noting that $V$ may not necessarily be a fixed set. New nodes and edges may arise with time elapsing, though Alice always has an up-to-date view of the graph structure. In this case, $V^{(T)}$ can be nodes in the stale graph, and $V^{(I)}$ can be the newly arisen nodes. 
There is also no restriction that all nodes in $V$ should form a connected component. Rather, $V$ can contain nodes in a group of graphs, as long as the grouping makes logical and practical sense. Under this setting, $V^{(T)}$ and $V^{(I)}$ can be the nodes of different graphs in the group.

\vspace{-1mm}
\subsection{Overview of the Attack}
\label{sec:attack}
\vspace{-1mm}

We will first introduce the capability/knowledge of the attacker, and then provide overview of the attack method.
During the attack, the attacker has access to a set of node features and their labels which are required during training. During inference, Bob is able to query the trained API for multiple times with the subset of nodes that are of interest.
That is to say, the attacker's \textbf{capability} includes the query access to a blackbox GNN model and the obtained prediction probability for a set of nodes during inference. 
Note that the attacker has no information about the API model except 
that it is a GNN model with unknown architecture and parameters.
Unlike He \etal~\cite{he2020stealing} which assumes the knowledge of partial graphs or a shadow dataset, here, we have no such additional assumptions.

The \textbf{overview} of the proposed link re-identification attack \linkteller is as follows. The attacker plays the role of Bob in the interaction model (\Cref{fig:int-model}). 
The \textit{goal} of the attacker is to recover the connections among the  inference node entities. 
Concretely, during inference, attacker Bob will query the GNN API with a set of inference nodes. With the returned prediction probability vectors, Bob will infer the connections between certain pairs of nodes.
The attack succeeds if the attacker can correctly determine whether two given nodes are connected by a link or not. 
We use the standard \textbf{metrics} \textit{precision}, indicating what fraction of pairs inferred to be connected are indeed connected in the graph; and \textit{recall}, indicating what fraction of the connected pairs are revealed,
to measure the attack success rate.
We also evaluate the \textit{AUC scores}.

\paragraph*{Intuitions}
Our attack is inspired by the intuition behind the training of 
a GNN model---during training, 
if two nodes $u$
and $v$ are connected by an edge, GNN would ``propagate''
the information of $u$ to
$v$. As a result, 
if there is an edge from $u$
to $v$, we would expect that
changing the feature vector of 
$u$ would impact the prediction of $v$.
Thus, if we can
compute the \textit{influence} of one node on the other, 
we could use it to guess whether
there is an edge between the two nodes:
If the \textit{influence value} is ``large'',  we would be more confident on the existence of an edge; if the \textit{influence value} is ``small'', we would be more confident that the nodes are not directly connected. 
Below, we describe 
a concrete algorithm 
to approximate such an influence 
value by probing the trained GNN inference API.

\vspace{-1mm}
\subsection{\linkteller: Edge Influence Based Attack}
\label{sec:oracles}
\vspace{-1mm}

\linkteller attack proceeds in
two phases. 
First, given a collection of 
nodes $V^{(I)}$ at the inference time
and the inference API, $G_{BB}(\cdot,\cdot)$,
the attacker tries to 
calcluate the  
\textit{influence value}
between each pair of nodes in $V^{(I)}$.
Second, \linkteller then sorts 
all pairs of nodes by their 
influence value, and predict 
the first 
$m=\hat k\cdot \frac{n(n-1)}{2}$ node 
pairs with highest influence 
values as ``\texttt{\small with edge}''
and all other pairs as ``\texttt{\small without edge}''. Here $\hat{k}$ is a 
hyperparameter specified by the attacker,
which indicates his prior ``belief'' of 
the graph density. \m{We call $\hat{k}$
the \textit{density belief}, which is a key hyper-parameter (details in \Cref{alg:attack}). In our experiments,
we observe that 
the attack performance will decrease slightly given
the discrepancy between estimated $\hat{k}$ and ground truth $k$.
Nevertheless, we show that 
\linkteller remains more 
effective than the state-of-the-art attacks even with  inaccurate estimation for $\hat{k}$.}
 
\m{In practice, it is also possible for attackers to further estimate $\hat k$ or the influence value \textit{threshold} for edge re-identification with additional knowledge.
For instance, 
if the attacker has partial graph information, she can either estimate $\hat k$, or directly
calculate the influence values for known connected/unconnected pairs and estimate the threshold for distinguishing them. 
More concrete descriptions of such actionable strategies are deferred to~\Cref{sec:append-belief}, which we hope can inspire more effective attacks as interesting future work.
}

\paragraph*{Measuring Influence Values via the Inference API}
Here we describe the calculation of
the influence value between a node pair.
Recall that in the interaction model, for any inquiry involving a set of nodes and their features, the attacker Bob is given a prediction vector for each node.
With the hope of taking advantage of the prediction vectors to obtain the influence value, we look into the structure of graph convolutional layers and analyze the influence of an edge on its incident nodes.

We characterize the influence of one node $v$ to the other node $u$ 
by measuring the change in the prediction for node $u$ when the features of the node $v$ get reweighted.
Formally, let $V^{(I)}$ be the set of nodes involved in the inference stage and 
${X}=[{x}_1^\top,\ldots, {x}_v^\top,\ldots]^\top$
be the corresponding feature matrix. By upweighting the features of the node $v$ by a small value $\Delta$, the attacker generates a new feature matrix $
{X}'=[{x}_1^\top,\ldots, (1+\Delta){x}_v^\top,\ldots]^\top
$. The difference between the two predictions $P^{(I)}$ and $P'^{(I)}$ with respect to $\Delta$ denotes the influence of reweighting $v$ on the prediction of all other nodes. 
We define the influence matrix of $v$ on other nodes as $I_v=\lim_{\Delta\rzero}\left(P'^{(I)}-P^{(I)}\right)/\Delta$ with size $|V^{(I)}| \times c$
Its $u$-th row ${i_{v}}_u \in I_v$ represents the prediction influence vector of $v$ on $u$ for each class dimension. 
Finally, we compute the $\ell_2$ norm of the corresponding influence vector as the influence value of $v$ on $u$ as $\norm{{i_{v}}_u}$.
Since computing the influence matrix $I_v$ 
yields the influence of one node $v$ on \textit{all} other nodes, to compute the influence value between  $n^2$ pairs of nodes ($n$ is the number of nodes of interest), we only need to compute $n$ influence matrices, each for one node in the interested node set.
This requires $2n$ forward passes of the trained network in all, which does not constitute a significant overhead during inference time. We report the running time of \linkteller in~\Cref{sec:append-runtime}.

Next, we will theoretically show that the influence value $\norm{{i_{v}}_u}$ comes with a nice property for GCN models, that is, \textit{two nodes that are at least $k+1$ hops away have no influence on each other in a $k$-layer graph convolutional network}.
We start from the simple case of a 1-layer GCN in~\Cref{thm:1-layer-gcn}.

\vspace{-1mm}
\begin{proposition}[Influence value of a 1-layer GCN]
\label{thm:1-layer-gcn}
For a 1-layer trained GCN model with parameters $W$, when its input adjacency matrix is $A$ and feature matrix is $X$, when there is no edge between node $u$ and node $v$, the influence value $\norm{{i_{v}}_u}=0$.
\end{proposition}

\vspace{-1mm}
We omit the proof in~\Cref{sec:append-proof-thm-1-layer-gcn}
and next present a natural extension of the above conclusion for a $k$-layer GCN. 
\vspace{-1mm}
\begin{theorem}[Influence value for a $k$-layer GCN]
\label{thm:k-layer-gcn}
For a $k$-layer trained GCN model,
when node $u$ and node $v$ are at least $k+1$ hops away,
the influence value $\norm{{i_{v}}_u}=0$.
\end{theorem}
\vspace{-1mm}
The complete proof is provided in~\Cref{sec:append-proof-thm-k-layer-gcn}.
With the guarantee provided by~\Cref{thm:1-layer-gcn}, we can identify the connected pairs against a 1-layer GCN with high confidence; the criterion is that the pair is connected if and only if the influence value is non-zero.
For GCNs with more layers, though this criterion does not directly apply, 
\Cref{thm:k-layer-gcn} can help to rule out nodes that are $k+1$ hops apart, thus eliminating a significant number of negative examples (\ie, unconnected pairs). 
Moreover, the node pairs that are directly connected would have higher influence values,
observed by studies on local neighborhood properties~\cite{xu2018representation}.
Although there is no strict guarantee that the influence values of the connected pairs are the largest since the values also depend on the features and the learned weights, in practice, the learned weights will generally display a preference for connected pairs for better label propagation, and thus the corresponding influence values of connected pairs are larger.

\begin{small}
\begin{algorithm}[t!]
\algsetup{linenosize=\tiny}
\footnotesize
\DontPrintSemicolon
\KwIn{A set of nodes of interest $V^{(C)} \subseteq V^{(I)}$; the associated node features $X$;
the inference API $G_{BB}(\cdot,\cdot)$; density belief $\hat k$, reweighting scale $\Delta$}
\KwOut{a $0/1$ value for each pair of nodes, indicating the absence/presence of edge}
\SetKwFunction{calc}{InfluenceValue}
\SetKwFunction{infmat}{InfluenceMatrix}
\SetKwProg{Fn}{Function}{:}{}
  \Fn{\infmat{$V^{(I)},X,G_{BB}(\cdot,\cdot),v$}}{
    $P=G_{BB}(V^{(I)},X)$\;
    $X'=[{x}_1^\top,\ldots, (1+\Delta){x}_v^\top,\ldots, ]^\top$\;
    $P'=G_{BB}(V^{(I)},X')$\; 
    $I=\frac{1}{\Delta}(P'-P)$\;
    \KwRet $I$
  }
  \;
\For {each node $v\in V^{(C)}$} {
    $I \leftarrow$ \infmat($V^{(I)},X,G_{BB}(\cdot,\cdot),v$)\;
    \For {each node $u\in V^{(C)}$} {
        $i_{uv}\leftarrow \norm{I[u,:]}$\Comment*[r]{\scriptsize The norm of the $u$-th row of $I$}
    }
}
Sort all $i_{uv}$ in a descending order\;
$n\leftarrow$ $|V|$\;
$m\leftarrow$ $\hat k\cdot \frac{n(n-1)}{2}$\;
Assign $1$ to the first $m$ pairs, and $0$ to the remaining\;
 \caption{\small Link Re-identification Attack ({\linkteller})}\label{alg:attack}
\end{algorithm}

\end{small}

\section{Countermeasures of \linkteller: Differentially Private GCN}
In this section, we aim to evaluate to what extent the proposed \linkteller in~\Cref{sec:attack} reveals the private connection information effectively through a trained GCN, as well as the sufficient conditions of the attack, via considering different countermeasure approaches.
The most direct countermeasure or defense against such an attack would be a differentially private GCN model.
However, so far there is no existing work directly training differentially private GCN models to our best knowledge.
As a result, we first revisit the \textbf{general framework} and principles of developing a differentially private (DP) GCN against such edge re-identification attacks (\Cref{sec:defense-overview}).
We then formally define the DP GCN, followed by two proposed \textbf{practical algorithms} to train a DP GCN. We also discuss the upper bound of the precision of general edge re-identification attacks on DP GCNs. Importantly, we point out that the theoretical guarantee of differential privacy is insufficient in preserving both privacy and utility for a GCN. It is equally important to empirically choose an appropriate privacy budget to strike a better privacy-utility balance. 

\vspace{-1mm}
\subsection{Overview of DP GCN Framework}
\label{sec:defense-overview}
\vspace{-1mm}

In the following sections, we review the definition of edge differential privacy for graph algorithms~\cite{karwa2014private} and present two practical DP GCN training algorithms via graph structure input perturbation. Input perturbation for GCN is a non-trivial problem since naively adding noise to the graph structure would destroy the sparsity of the adjacency matrix. The loss of sparsity greatly  limits a GCN's performance and increases its memory and computation cost. 

To preserve the sparsity of the adjacency matrix, we discuss two approaches for GCN input perturbation: \edgerand and \lapgraph. \edgerand adapts the idea in Mülle \etal~\cite{muile2015pig} and randomly flips each entry in the adjacency matrix according to a Bernoulli random variable. \lapgraph improves upon \edgerand by pre-calculating the original graph density using a small privacy budget and using that density to clip the perturbed adjacency matrix. Compared to \edgerand, \lapgraph preserves the sparsity of the adjacency matrix under a small privacy budget. 

\paragraph*{Differentially Private GCN}
In edge differential privacy~\cite{karwa2014private}, two undirected graphs are said to be neighbors if one graph can be obtained by the other by adding/removing one edge. \Cref{def:nbr} defines the neighboring relation using the adjacency matrix representation. 


\begin{definition}[Neighboring relation]
    \label{def:nbr}
    Let $\cal A$ be the set of adjacency matrices of undirected graphs. Any pair of two \textit{symmetric} matrices $A,A'\in \cal A$ are said to be neighbors when the graph represented by $A'$ could be obtained by adding/removing one edge from  graph  $A$, denoted as $A\sim A'$. Further, we denote the differing edge as $e=A\oplus A'$.
\end{definition}


\begin{definition}[\eedgedp]
    \label{def:edgedp}
    A mechanism $\mathcal{M}$ is \Eedgedp~if for all valid matrix $A\in\cal A$, and $A'\sim A$, and any subset of outputs $\cal S\subseteq Range(\cal M)$, the following holds:
    \begin{align}
        \label[ineq]{eq:dp}
        \Pr[\mathcal{M}(A)\in \cal S]\leq \exp(\eps)\cdot \Pr[\mathcal{M}(A')\in \cal S].
    \end{align}
    The probability $\Pr$ is taken over the randomness of $\mathcal{M}$.
\end{definition}
\Cref{def:edgedp} formally presents the definition of \eedgedp (\eedgedpshort). It guarantees that the outputs of a mechanism $\mathcal{M}$ should be indistinguishable on any pair of neighboring input graphs differing in one edge. 


Next, we apply \eedgedpshort to a GCN. To protect against link re-identification attacks, we need to guarantee that a GCN's inference results do not reveal the edge information of its input graph. Specifically, in the inductive training, the edge information of both the training graph and the inference graph should be protected. In addition, the privacy protection should hold even if the attacker submits an infinite number of inference queries. 

Based on the above criteria, we first perturb the graphs before training a GCN. Since the training and the inference steps are both post-processing on the perturbed DP graphs, the edge information is protected for both the training and inference graphs. Moreover, submitting more queries would not reveal sensitive edge information.

\m{We present the detailed algorithm of \textit{perturbation}, \textit{training}, and \textit{inference} in a differentially private GCN framework in~\Cref{alg:dpgcn} in~\Cref{sec:append-algo-dp-gcn}.}
First, the adjacency matrix $A$ is perturbed to meet the DP guarantee. Second, a DP GCN model is trained on a subset of training nodes in the perturbed graph. Finally, during inference, the DP GCN model is used to predict the labels for a subset of inference nodes in the perturbed graph. 
Essentially, \eedgedpshort is achieved in the step of adjacency matrix perturbation (line 1-2), and the guarantee is provided by the privacy guarantee of the perturbation mechanism. Since the same perturbed graph is used for both the training and inference steps, making multiple inferences would not consume additional privacy budget.

The following theorem provides differential privacy guarantee for the training procedure in~\Cref{alg:dpgcn} for GCN models. 

\begin{theorem}[\Eedgedp GCN]
\label{thm:DP GCN}
The DP GCN model trained by~\Cref{alg:dpgcn} is \Eedgedp if the perturbation mechanism $M_\varepsilon$ is \Eedgedp.
\end{theorem}

We omit all proofs for the DP guarantees in~\Cref{sec-append-dp-proof}. 
Next, we show that the \texttt{\small Inference} step in \Cref{alg:dpgcn} guarantees \eedgedpshort for edges in both the training and testing graph ($A_{V^{(T)}}$ and $A_{V^{(I)}}$). To prove this privacy guarantee, we first introduce the parallel composition property of \eedgedpshort.

\begin{lemma}[Parallel composition of \eedgedpshort]
\label{lemma:parallel_composition}
If the perturbation mechanism $M_{\varepsilon}$ is \Eedgedp and $A_1, A_2, \dots, A_m$ are adjacency matrices with non-overlapping edges, the combination of $M_{\varepsilon}(A_1), M_{\varepsilon}(A_2), \dots, M_{\varepsilon}(A_m)$ is also \Eedgedp.
\end{lemma}

The following theorem guarantees differential privacy for any inference using the DP GCN model. 

\begin{theorem}[\Eedgedp GCN inference]
\label{thm:dp-infer}
The \texttt{\small Inference} step in~\Cref{alg:dpgcn} is \Eedgedp~for any $V^{(I)} \subseteq V$.
\end{theorem}

The above analysis of the general DP GCN framework provides  privacy guarantees for GCN models trained following the principles in Algorithm~\ref{alg:dpgcn}. Next, we will introduce two such concrete training mechanisms.




\vspace{-1mm}
\subsection{Practical DP GCN}
\vspace{-1mm}

In \Cref{alg:dpgcn}, the perturbation step $M_\varepsilon$ takes the adjacency matrix of the input graph and adds noise to the adjacency matrix to guarantee \eedgedpshort. In this section, we present two practical DP mechanisms for this process. 

The intuition behind perturbing the adjacency matrix is to add enough noise in the adjacency matrix to guarantee the indistinguishability between any pair of neighboring adjacency matrices $A$ and $A'$---the ratio of the probability of getting the same perturbed matrix from $A$ and $A'$ should be bounded by a small constant $\varepsilon$. The smaller $\varepsilon$ is, the stronger the protection is. 

In addition to the privacy requirements, the perturbed adjacency matrix $A'$ also needs to satisfy the following two requirements in order to be used as a training/inference graph for DP GCN. First, for large graphs, $A'$ needs to preserve a reasonable level of sparsity to avoid huge memory consumption when training a GCN model. Second, each row in the perturbed adjacency matrix $A'$ should represent the same node as its corresponding row in the original adjacency matrix $A$. This requirement ensures that the node features and labels can be associated with the right graph structure information in the perturbed adjacency matrix during training and inference. 

However, the second requirement is often not satisfied by prior work on DP synthetic graph generation~\cite{sala2011sharing,wang2013preserving,xiao2014differentially,brunet2016novel,qin2017generating}.
This line of work aims at generating graphs that share similar statistics with the original graphs.
Though the desired statistics of the graphs are preserved, the nodes in the generated graph and the original graph are intrinsically unrelated. Therefore, the new DP graph structure cannot be connected with the node features and labels to train a DP GCN model. More discussions on prior works are provided in the related work section.




To satisfy the privacy and utility requirements for DP GCN, we introduce two perturbation methods that directly add noise to the adjacency matrix. 

\subsubsection{Edge Randomization (\edgerand)}
We set out with a discrete perturbation method proposed in Mülle \etal~\cite{muile2015pig}. This algorithm was originally proposed as a pre-processing step for DP node clustering. Since the algorithm naturally preserves the sparse structure of the adjacency matrix, we adopt it as the input perturbation algorithm for DP GCN and name it \edgerand. 
\m{We present the algorithm for \edgerand in~\Cref{alg:pig} in~\Cref{sec:append-algo-dp-mech}.}
We first randomly choose the cells to perturb and then randomly choose the target value from $\{0,1\}$ for each cell to be perturbed. 

In \edgerand, the level of the sparsity of the perturbed adjacency matrix is purely determined by the sampling parameter $s$, which can be conveniently controlled to adapt to the given privacy budget $\eps$. The relationship between $s$ and $\eps$ is characterized in~\Cref{thm:dppig}.

\begin{theorem}
    \label{thm:dppig}
    \edgerand guarantees \eedgedpshort for $\eps\geq\ln \left(\frac{2}{s}-1\right)$, $s\in (0,1]$.
\end{theorem}

\edgerand guarantees differential privacy for the perturbed adjacency matrix. However, the privacy protection comes at the cost of changing the density of the perturbed graph. 
Let the density of the input graph to \edgerand be $k$, the expectation of the density of the output graph is $k'=(1-s)k+s/2$.
Take $\eps=1$ as an example, in this case, $s$ shall be at least $0.5379$ according to~\Cref{thm:dppig}, and $k'$ is therefore larger than $1/4$.
As such, when $\eps$ is small, the perturbed graph generated by \edgerand could have a much higher density compared to the original one. This would increase the memory consumption for training DP GCNs on large graphs and may cause memory errors when the perturbed adjacency matrix becomes too dense to fit into the memory.

\subsubsection{Laplace Mechanism for Graphs (\lapgraph)}
\edgerand is not applicable to large graphs under small privacy budgets due to the huge memory consumption caused by a dense adjacency matrix. Therefore, we propose \lapgraph to address this problem. 

The classical idea of adding Laplace noise to the private value is also applicable to our scenario.
The difference is that, in traditional scenarios, Laplace noise is applied to entities such as a database entry, while in our case, the private entity is the adjacency matrix. Therefore, additional care shall be taken to tailor the Laplace mechanism to the graph scenario.

By the definition of Laplace mechanism~\cite{dwork14dp}, adding a certain amount of noise to each cell in the adjacency matrix will lead to any two neighboring adjacency matrices being indistinguishable. 
However, directly applying this mechanism will add a huge amount of continuous noise to each cell of the adjacency matrix, which inevitably undermines the sparse property of the matrix. The loss of sparse property introduces two problems: First, it drastically increases the computation and memory cost of training a GCN. Second, adding the continuous noise in the adjacency matrix is equivalent to adding new weighted edges between almost \emph{every} pair of nodes in the graph, which greatly impairs the utility of the adjacency matrix and, consequently, the GCN trained on it.

To retain the sparsity, after adding noise, we only keep the largest $T$ cells as existing edges in the perturbed graph. To preserve the original graph structure, we set $T$ to be the approximation of the number of edges in the original graph using a small portion of the differential privacy budget. 
We name the perturbation method \lapgraph and \m{present the details in~\Cref{alg:laplace} in~\Cref{sec:append-algo-dp-mech}. }
The privacy guarantee for this method is given in~\Cref{thm:dplap}.

Compared with \edgerand, \lapgraph has the advantage of better preserving the density of the original graph, especially for large graphs and small $\varepsilon$. Since the number of edges in a large graph is often orders of magnitude higher than the sensitivity of adding/removing a single edge, it is possible to estimate $T$ even under a very limited privacy budget. Thus, the density of the perturbed graph is much closer to the original one than \edgerand. This improvement makes it possible to train DP GCN on large graphs under small privacy budgets without causing memory errors. 

\begin{theorem}
\label{thm:dplap}
    \lapgraph guarantees \eedgedpshort.
\end{theorem}

Due to the lack of DP GCN approaches, here we focus on the existing technique \edgerand and the proposed \lapgraph to provide DP guarantees for GCN as countermeasures to further evaluate the proposed attack \linkteller. We have provided the formal analysis for the privacy guarantees for \edgerand and \lapgraph above, and next, we will discuss a general upper bound of edge privacy on DP GCN models.

\subsection{Discussion: Upper Bound of Edge Re-Identification Attack Performance on DP GCN}
\label{sec:dp-bound}
\vspace{-1mm}

As implied by \eedgedpshort in~\Cref{def:edgedp}, it is generally difficult to tell, among the two neighboring adjacency matrices $A$ and $A'$, which one leads to the observed prediction. The direct consequence of the indistinguishability is that the existence of the differing edge $e=A\oplus A'$ cannot be inferred. In this section, we aim to analyze the upper bound of edge re-identification attacks against  DP GCN.

Same as the attack model introduced in \Cref{sec:attack}, we assume the attacker has access to a set of node features and their labels without any knowledge about the GCN structure and parameters.

To start with, we formalize the link re-identification attack proposed in \Cref{sec:attack} as the following game between the graph owner Alice and the attacker Bob:
\begin{enumerate}[leftmargin=*]
    \item Let $V$ be a set of nodes and $\mathcal{A}_V$ be the set of all possible adjacency matrices for graphs with nodes $V$. First, Alice selects an adjacency matrix $A \in \mathcal{A}_V$ uniformly at random and uses it to generate a graph.
    \item Bob selects a set of training nodes $V^{(T)} \subseteq V$. He sends $V^{(T)}$ with the features and labels of $V^{(T)}$ to Alice. 
    \item Alice then trains an \Eedgedp GCN model and exposes the inference API $G_{BB}$ to Bob.
    \item Bob selects a set of inference nodes $V^{(I)} \subseteq V$ and nodes of interests $V^{(C)} \subseteq V^{(I)}$. Let $k^{(C)}$ denote the graph density over $V^{(C)}$. For each pair of nodes $<u,v>\in V^{(C)}\times V^{(C)}$, Bob launches a link re-identification attack $\mathcal{R}_{G_{BB}}(u, v)$ to infer whether an edge exists between nodes $u$ and
    $v$, and $\mathcal{R}_{G_{BB}}(u, v) \in \left\{0,1\right\}$.
\end{enumerate}

To obtain an upper bound for the above attack, we assume the attacker  knows the inference node density $k^{(C)}$. 
Formally, we bound the expected precision of the link re-identification attack $\mathcal{R}$ by the following theorem.

\begin{theorem}
\label{thm:dp-bound}
The precision of $\mathcal{R}_{G_{BB}}$ over nodes of interests $V^{(C)}$ with density $k^{(C)}$ is upper-bounded by:
\begin{small}
\begin{equation*}
    \underset{<u,v>\in V^{(C)}\times V^{(C)}}{\Pr}\left[A_{uv} = 1 \mid \mathcal{R}_{G_{BB}}(u, v) = 1 \right] \leq \exp(\eps) \cdot k^{(C)}, 
\end{equation*}
\end{small}
where the probability is calculated over the randomness in the graph selection, the noise introduced by the DP GCN training, and the selection of node pair $<u, v>$. 
\end{theorem}

\textbf{Proof Sketch.} \quad Based on \Cref{def:edgedp} and Bayes' theorem, the ratio between the posterior probability $\Pr[A_{uv}=1 \mid G_{BB} \in S]$ and the prior belief on $\Pr[A_{uv}=1]$ is bounded by $\exp(\varepsilon)$. Since the precision of a random guess based on the prior probability (\ie, the graph density) is at most $k^{(C)}$, the upper bound for the precision of a link re-identification attack on an \Eedgedp GCN is $\exp(\varepsilon)\cdot k^{(C)}$. The complete proof is provided in~\Cref{sec:append-proof-thm-dp-bound}.

Although~\Cref{thm:dp-bound} provides a theoretical upper bound for the precision of an edge re-identification attack, it may not be sufficiently tight to provide the best privacy-utility trade-off. For example, given a graph with $1\%$ density, the attack precision is bounded below $2\%$ (\ie, no more than two times higher than random guessing using the prior probability) if and only if $\eps \leq \ln{2}$. However, in practice, the same empirical protection might be achieved by a model with weaker privacy protection (\ie, higher privacy budget) and therefore better utility. Thus, in~\Cref{sec:eval-dp-attack}, we empirically evaluate the privacy-utility trade-off of DP GCN across multiple datasets. 

\m{In addition to DP GCN approaches, it may also
be possible to leverage some heuristics to detect such attacks. For instance, one may distinguish the abnormal behavior of querying the same set of inference nodes $V^{(I)}$ multiple times (with the node features of one node slightly altered in each query). 
The defender could also optimize a query limit $Q$ which decreases the attack performance while maintaining reasonable benign query accuracy, although there is no guarantee for such detection. More discussions on the detection strategies are deferred to~\Cref{sec:append-stealthy}, and in this paper, we will focus on the DP GCN mechanisms with privacy guarantees.
}

\section{Evaluation of \linkteller}
\label{sec:eval-attack}
We evaluate the effectiveness of the \linkteller attack on multiple graph datasets under various scenarios compared with three baselines. 
In particular, we investigate how different factors such as node degree affect the attack performance.

\vspace{-1mm}
\subsection{Datasets}
\label{sec:datasets}
\vspace{-1mm}

\m{We evaluate \linkteller on eight datasets in the \textit{inductive} setting and {three} datasets in the \textit{transductive} setting (\Cref{sec:append-datasets}) and provide a brief description of the data below.}
\m{Under the inductive setting, 
the first dataset is the \textit{twitch dataset}~\cite{rozemberczki2019multiscale} which is composed of 6 graphs as disjoint sets of nodes.} Each of the graphs represents a set of people in one country; the nodes within a graph represent users in one country, and the links represent mutual friendships between users. 
The dimension of the features is the same across different graphs and each dimension has the same semantic meaning. Some sampled features include games they like, location, and streaming habits. The task is a binary classification task which classifies whether a streamer uses explicit language.
This dataset is proposed for transfer learning, \ie, applying the model learned on one graph to make inferences on the other graphs corresponding to different countries. 
In our evaluation, we train the GNN model on the graph twitch-ES, and transfer it to other five countries (RU, DE, FR, ENGB, PTBR).
\textit{PPI}~\cite{hamilton17graphsage} and \textit{Flickr}~\cite{zeng20graphsaint} are another two standard datasets used in graph inductive learning setting. PPI is a dataset for multi-label classification task, which aims to categorize the function of proteins across various biological protein-protein interaction graphs.
Flickr is an evolving graph for the classification task, which contains descriptions and common properties of images as node features.
For both PPI and Flickr, we use the standard splits for training and testing following the previous works.
\m{Under the transductive setting, we adopt three standard datasets (Cora, Citeseer, and Pubmed). More details of the data can be found in~\Cref{sec:append-datasets}.}

\vspace{-1mm}
\subsection{Models}
\label{sec:models}
\vspace{-1mm}


We mainly experiment with GCN models. 
The configurations/hyperparameters 
include the normalization techniques applied to the adjacency matrix, the number of hidden layers, the number of input units, hidden units, and output units, as well as the dropout rate. For each combination of hyperparameters, we train the network to minimize the cross-entropy loss for the intended tasks.
We performed grid search to get the best set of hyperparameters on the validation set. The search space for the hyperparameters and the formulae for different normalization techniques are provided in~\Cref{sec:append-eval-detail}. 
To measure the 
performance of a GCN model, we follow previous work and use F1 score for their corresponding binary classification tasks. We leave the description of the best hyper-parameters we achieve in~\Cref{sec:append-best-para}.
In addition to the 2-layer GCNs evaluated in the main paper, in~\Cref{sec:append-3-layer}, we also experimented with the 3-layer GCNs and include a discussion about GCNs of 1 layer and more than 3 layers. 
We conclude that \linkteller is a successful attack against most practical GCN models.
In addition, we evaluate \linkteller on Graph Attention Networks (GATs). The details are in~\Cref{sec:attack-gat}.

\vspace{-1mm}
\subsection{Setup of the Evaluation}
\vspace{-1mm}
In this section, we first describe the metrics we use to evaluate the attack effectiveness of \linkteller.
We then present the baseline attack methods.

\subsubsection{Evaluation Metrics of the attack}
\label{sec:eval-metric}
We use the standard metrics: \textbf{precision} (the fraction of existing edges among the pairs recognized as true by Bob) and \textbf{recall} (the fraction of edges discovered by Bob over all existing edges among the subset of nodes). 
We also compute the \textbf{F1 score} (the harmonic mean of precision and recall).
The reason we adopt the metric is that our problem here (distinguishing connected pairs from unconnected ones) is an imbalanced binary classification problem where the minority (the connected pair) is at the core of concern. See~\Cref{sec:append-eval-metrics} for more details.
Additionally, for fair comparison with baselines, we follow the evaluation in He \etal~\cite{he2020stealing} and compute the \textbf{AUC score}.

\subsubsection{Baseline Attacks}
\label{sec:baseline-attack}

We compare \linkteller with two baselines: random attack and LSA2 attacks
in He \etal~\cite{he2020stealing}.

For the random attack, 
we follow the standard notion 
and construct a 
random classifier 
as a Bernoulli random variable with parameter $p$ which predicts true if and only if the random variable takes the value $1$~\cite{flach2015precision}.
Given a set of instances where $a$ of them are true and $b$ are false, the precision of this classifier is $a/(a+b)$ and the recall is $p$. 
In our case, $a$ is the number of connected pairs of nodes, while $a+b$ is the number of all pairs.
Therefore, \textit{precision} is exactly the \textit{density} $k$ of the subset, which we formally define as
$
    k = 2 m/(n(n-1))
$, where $n=|\m{V^{(C)}}|$ is the size of the set of interest and $m$ is the number of connections among the set \m{$V^{(C)}$}.
The recall of 
such a random classifier 
will be the 
density belief $\hat{k}$.

We also compare \linkteller with the state of the art LSA2 attacks~\cite{he2020stealing}.
In the paper, the authors discussed several types of background knowledge including node attributes, partial graph, and a shadow dataset for attackers.
Among the combinations, their \textit{Attack-2} is closest to our scenario where the attacker has only access to the target graph's node features.
We follow their best practices, computing the \textit{correlation distance} between 1) \textit{posteriors} given by the target model and 2) \textit{node attributes}, 
referred to as LSA2-post and LSA2-attr attacks.

\vspace{-1mm}
\subsection{Evaluation Protocol}
\label{sec:attack-setup}
\vspace{-1mm}



Think about the paparazzi who are fanatical about exploiting the connections among celebrities, or 
the indiscriminate criminals that are maliciously targeted at the mass mediocre majority, their targets are substantially different.
Consequently,
the subsets they gather for attack have diverse node degree distributions.
Catering to the need of evaluating our attack against \textit{nodes of different degree distributions}, we design the scenario as follows.
We consider three types of subsets that are of potential interest to the attacker: nodes of low degree, unconstrained degree, and high degree.
\m{For each type, we randomly sample a fixed number $\m{n^{(C)}}$ of nodes to form a subset \m{$V^{(C)}$} for evaluation. When sampling nodes of low (or high) degree, we place a threshold value $d_{\rm low}$ (or $d_{\rm high}$) and sample from nodes whose degrees are no larger than $d_{\rm low}$ (or no smaller than $d_{\rm high}$). The value $d_{\rm low}$ and $d_{\rm high}$ are chosen empirically based on the graph. 
When sampling nodes of unconstrained degree, we sample nodes from the entire test set uniformly at random.}

More specifically, for all datasets, we choose $\m{n^{(C)}}=|\m{V^{(C)}}|=500$.
For twitch datasets, 
to form the unconstrained subset, we sample from each entire testing graph.
For the low degree subset and high degree subset, the threshold $d_{\rm low}$ and $d_{\rm high}$ are set to $5$ and $10$, respectively. 
We set the $d_{\rm low}$ value to $10$ for twitch-PTBR, since the graph is much denser with abundant connections among a small number of nodes.
For PPI and Flickr graphs, the subsets for testing are sampled from the testing graphs/nodes that are not involved in training. 
We set $d_{\rm low}$ as $15$ and $d_{\rm high}$ as $30$ for these two large graphs.

We also evaluate different \textit{density belief} $\hat k\in\{k/4,k/2,k,2k,4k\}$, where $k$ is the true density. In the experiments, we round the density $k$ to the closest value in its most significant bit (\eg, 5.61e-5 rounded to 6e-5).
As we will see, the effectiveness of \linkteller does not heavily depend on the exact knowledge of the density $k$.

\vspace{-1mm}
\subsection{Evaluation for \linkteller}
\label{sec:attack-eval}
\vspace{-1mm}

{
\setlength{\tabcolsep}{4pt} 

\begin{table}[]
    \centering
    \renewrobustcmd{\bfseries}{\fontseries{b}\selectfont}
    \sisetup{detect-weight,mode=text,group-minimum-digits = 4}

    \caption{\small \textbf{Attack Performance (Precision and Recall)} of \linkteller on different datasets, compared with two baseline methods LSA2-\{post, attr\}~\cite{he2020stealing}. Each table corresponds to a dataset. We sample nodes of low, unconstrained, and high degrees as our targets. Groups of rows represent different \textit{density belief} $\hat k$ of the attacker.
    }
    \label{tab:att-np}

\begin{subtable}[]{\columnwidth}
    \centering
    \vspace{0.5em}
    \resizebox{\columnwidth}{!}{%
        \begin{tabular}{cc
  S[table-format=3.1]
  @{\tiny${}\pm{}$}
  >{\tiny}S[table-format=2.1]<{\endcollectcell}
  S[table-format=3.1]
  @{\tiny${}\pm{}$}
  >{\tiny}S[table-format=2.1]<{\endcollectcell}
  S[table-format=3.1]
  @{\tiny${}\pm{}$}
  >{\tiny}S[table-format=2.1]<{\endcollectcell}
  S[table-format=3.1]
  @{\tiny${}\pm{}$}
  >{\tiny}S[table-format=2.1]<{\endcollectcell}
  S[table-format=3.1]
  @{\tiny${}\pm{}$}
  >{\tiny}S[table-format=2.1]<{\endcollectcell}
  S[table-format=3.1]
  @{\tiny${}\pm{}$}
  >{\tiny}S[table-format=2.1]<{\endcollectcell}
  S[table-format=3.1]
  @{\tiny${}\pm{}$}
  >{\tiny}S[table-format=2.1]<{\endcollectcell}
}
    \toprule
    \multicolumn{2}{c}{\textbf{twitch-RU}} &  \multicolumn{4}{c}{\small low}  & \multicolumn{4}{c}{\small unconstrained} & \multicolumn{4}{c}{\small high}\\
    \cmidrule(lr){3-6}\cmidrule(lr){7-10}\cmidrule(lr){11-14}
    \makecell{\small $\hat k$} & Method   & \multicolumn{2}{c}{\small precision}  & \multicolumn{2}{c}{\small recall} & \multicolumn{2}{c}{\small precision} & \multicolumn{2}{c}{\small recall} & \multicolumn{2}{c}{\small precision} & \multicolumn{2}{c}{\small recall} \\\midrule
\rowcolor{tabgray}
\cellcolor{white}{\multirow{3}{*}{$k/4$}}
            & Ours  & \bfseries 100.0 & 0.0  & \bfseries 33.0 & 2.8  & \bfseries 95.1 & 1.1  & \bfseries 26.0 & 1.0  & \bfseries 98.9 & 0.2  & \bfseries 18.1 & 1.3 \\
            & LSA2-post  & 0.0 & 0.0  & 0.0 & 0.0  & 0.0 & 0.0  & 0.0 & 0.0  & 0.2 & 0.3  & 0.0 & 0.1 \\
            & LSA2-attr  & 0.0 & 0.0  & 0.0 & 0.0  & 1.3 & 0.5  & 0.4 & 0.1  & 2.5 & 1.3  & 0.4 & 0.2 \\[0.05em]\midrule[0.05em]
\rowcolor{tabgray}
\cellcolor{white}{\multirow{3}{*}{$k/2$}}
            & Ours  & \bfseries 100.0 & 0.0  & \bfseries 61.3 & 5.1  & \bfseries 87.9 & 0.4  & \bfseries 48.1 & 2.3  & \bfseries 97.1 & 0.3  & \bfseries 35.6 & 2.6 \\
            & LSA2-post  & 0.0 & 0.0  & 0.0 & 0.0  & 0.0 & 0.0  & 0.0 & 0.0  & 0.1 & 0.2  & 0.0 & 0.1 \\
            & LSA2-attr  & 0.0 & 0.0  & 0.0 & 0.0  & 1.7 & 0.2  & 0.9 & 0.1  & 2.5 & 0.5  & 0.9 & 0.1 \\[0.05em]\midrule[0.05em]
\rowcolor{tabgray}
\cellcolor{white}{\multirow{3}{*}{$k$}}
            & Ours  & \bfseries 78.7 & 1.9  & \bfseries 92.6 & 5.5  & \bfseries 71.8 & 2.2  & \bfseries 78.5 & 2.4  & \bfseries 89.7 & 1.7  & \bfseries 65.7 & 3.9 \\
            & LSA2-post  & 0.0 & 0.0  & 0.0 & 0.0  & 0.0 & 0.0  & 0.0 & 0.0  & 0.1 & 0.1  & 0.0 & 0.1 \\
            & LSA2-attr  & 0.0 & 0.0  & 0.0 & 0.0  & 1.1 & 0.1  & 1.2 & 0.1  & 2.2 & 0.6  & 1.6 & 0.3 \\[0.05em]\midrule[0.05em]
\rowcolor{tabgray}
\cellcolor{white}{\multirow{3}{*}{$2k$}}
            & Ours  & \bfseries 42.7 & 3.4  & \bfseries 100.0 & 0.0  & \bfseries 43.5 & 1.9  & \bfseries 95.0 & 0.5  & \bfseries 62.9 & 4.2  & \bfseries 91.8 & 1.3 \\
            & LSA2-post  & 0.7 & 0.9  & 1.8 & 2.5  & 0.0 & 0.0  & 0.0 & 0.0  & 0.0 & 0.0  & 0.0 & 0.1 \\
            & LSA2-attr  & 1.3 & 0.9  & 3.2 & 2.3  & 0.8 & 0.1  & 1.8 & 0.3  & 2.0 & 0.3  & 2.8 & 0.3 \\[0.05em]\midrule[0.05em]
\rowcolor{tabgray}
\cellcolor{white}{\multirow{3}{*}{$4k$}}
            & Ours  & \bfseries 21.3 & 1.7  & \bfseries 100.0 & 0.0  & \bfseries 22.5 & 1.1  & \bfseries 98.1 & 0.6  & \bfseries 33.6 & 2.5  & \bfseries 98.0 & 0.4 \\
            & LSA2-post  & 0.3 & 0.5  & 1.8 & 2.5  & 0.0 & 0.0  & 0.1 & 0.1  & 0.0 & 0.0  & 0.0 & 0.1 \\
            & LSA2-attr  & 0.7 & 0.5  & 3.2 & 2.3  & 0.7 & 0.1  & 3.0 & 0.5  & 1.6 & 0.3  & 4.6 & 0.5 \\\bottomrule
\end{tabular}%
}
\end{subtable}

\begin{subtable}[]{\columnwidth}
    \centering
    \vspace{0.5em}
    \resizebox{\columnwidth}{!}{%
        \begin{tabular}{cc
  S[table-format=3.1]
  @{\tiny${}\pm{}$}
  >{\tiny}S[table-format=2.1]<{\endcollectcell}
  S[table-format=3.1]
  @{\tiny${}\pm{}$}
  >{\tiny}S[table-format=2.1]<{\endcollectcell}
  S[table-format=3.1]
  @{\tiny${}\pm{}$}
  >{\tiny}S[table-format=2.1]<{\endcollectcell}
  S[table-format=3.1]
  @{\tiny${}\pm{}$}
  >{\tiny}S[table-format=2.1]<{\endcollectcell}
  S[table-format=3.1]
  @{\tiny${}\pm{}$}
  >{\tiny}S[table-format=2.1]<{\endcollectcell}
  S[table-format=3.1]
  @{\tiny${}\pm{}$}
  >{\tiny}S[table-format=2.1]<{\endcollectcell}
  S[table-format=3.1]
  @{\tiny${}\pm{}$}
  >{\tiny}S[table-format=2.1]<{\endcollectcell}
}
    \toprule
    \multicolumn{2}{c}{\textbf{twitch-FR}} &  \multicolumn{4}{c}{\small low}  & \multicolumn{4}{c}{\small unconstrained} & \multicolumn{4}{c}{\small high}\\
    \cmidrule(lr){3-6}\cmidrule(lr){7-10}\cmidrule(lr){11-14}
    \makecell{\small $\hat k$} & Method   & \multicolumn{2}{c}{\small precision}  & \multicolumn{2}{c}{\small recall} & \multicolumn{2}{c}{\small precision} & \multicolumn{2}{c}{\small recall} & \multicolumn{2}{c}{\small precision} & \multicolumn{2}{c}{\small recall} \\\midrule
\rowcolor{tabgray}
\cellcolor{white}{\multirow{3}{*}{$k/4$}}
            & Ours  & \bfseries 100.0 & 0.0  & \bfseries 28.3 & 2.4  & \bfseries 97.2 & 0.9  & \bfseries 22.7 & 0.6  & \bfseries 99.4 & 0.5  & \bfseries 24.1 & 2.1 \\
            & LSA2-post  & 0.0 & 0.0  & 0.0 & 0.0  & 0.2 & 0.2  & 0.0 & 0.1  & 0.5 & 0.2  & 0.1 & 0.0 \\
            & LSA2-attr  & 0.0 & 0.0  & 0.0 & 0.0  & 0.6 & 0.4  & 0.1 & 0.1  & 1.9 & 0.7  & 0.5 & 0.1 \\[0.05em]\midrule[0.05em]
\rowcolor{tabgray}
\cellcolor{white}{\multirow{3}{*}{$k/2$}}
            & Ours  & \bfseries 100.0 & 0.0  & \bfseries 50.0 & 0.0  & \bfseries 95.0 & 1.0  & \bfseries 44.3 & 1.3  & \bfseries 98.3 & 1.0  & \bfseries 47.7 & 4.5 \\
            & LSA2-post  & 0.0 & 0.0  & 0.0 & 0.0  & 0.1 & 0.1  & 0.0 & 0.1  & 0.3 & 0.1  & 0.1 & 0.0 \\
            & LSA2-attr  & 0.0 & 0.0  & 0.0 & 0.0  & 0.6 & 0.2  & 0.3 & 0.1  & 1.4 & 0.2  & 0.7 & 0.0 \\[0.05em]\midrule[0.05em]
\rowcolor{tabgray}
\cellcolor{white}{\multirow{3}{*}{$k$}}
            & Ours  & \bfseries 92.5 & 5.4  & \bfseries 92.5 & 5.4  & \bfseries 84.1 & 3.7  & \bfseries 78.2 & 1.9  & \bfseries 83.2 & 1.4  & \bfseries 80.6 & 6.7 \\
            & LSA2-post  & 0.0 & 0.0  & 0.0 & 0.0  & 0.0 & 0.1  & 0.0 & 0.1  & 0.1 & 0.0  & 0.1 & 0.0 \\
            & LSA2-attr  & 0.0 & 0.0  & 0.0 & 0.0  & 0.8 & 0.2  & 0.7 & 0.2  & 1.4 & 0.2  & 1.3 & 0.1 \\[0.05em]\midrule[0.05em]
\rowcolor{tabgray}
\cellcolor{white}{\multirow{3}{*}{$2k$}}
            & Ours  & \bfseries 51.1 & 1.6  & \bfseries 100.0 & 0.0  & \bfseries 51.3 & 2.1  & \bfseries 95.3 & 1.3  & \bfseries 49.1 & 2.7  & \bfseries 94.8 & 3.2 \\
            & LSA2-post  & 0.0 & 0.0  & 0.0 & 0.0  & 0.0 & 0.0  & 0.0 & 0.1  & 0.1 & 0.0  & 0.1 & 0.0 \\
            & LSA2-attr  & 1.7 & 2.4  & 3.3 & 4.7  & 0.8 & 0.1  & 1.4 & 0.2  & 1.6 & 0.2  & 3.1 & 0.3 \\[0.05em]\midrule[0.05em]
\rowcolor{tabgray}
\cellcolor{white}{\multirow{3}{*}{$4k$}}
            & Ours  & \bfseries 25.6 & 0.8  & \bfseries 100.0 & 0.0  & \bfseries 26.5 & 1.1  & \bfseries 98.3 & 1.0  & \bfseries 25.4 & 1.8  & \bfseries 97.6 & 1.6 \\
            & LSA2-post  & 0.0 & 0.0  & 0.0 & 0.0  & 0.0 & 0.0  & 0.0 & 0.1  & 0.0 & 0.0  & 0.1 & 0.0 \\
            & LSA2-attr  & 0.8 & 1.2  & 3.3 & 4.7  & 1.0 & 0.2  & 3.7 & 0.8  & 1.5 & 0.1  & 5.7 & 0.1 \\\bottomrule
\end{tabular}%
}
\end{subtable}

\begin{subtable}[]{\columnwidth}
    \centering
    \vspace{0.5em}
    \resizebox{\columnwidth}{!}{%
        \begin{tabular}{cc
  S[table-format=3.1]
  @{\tiny${}\pm{}$}
  >{\tiny}S[table-format=2.1]<{\endcollectcell}
  S[table-format=3.1]
  @{\tiny${}\pm{}$}
  >{\tiny}S[table-format=2.1]<{\endcollectcell}
  S[table-format=3.1]
  @{\tiny${}\pm{}$}
  >{\tiny}S[table-format=2.1]<{\endcollectcell}
  S[table-format=3.1]
  @{\tiny${}\pm{}$}
  >{\tiny}S[table-format=2.1]<{\endcollectcell}
  S[table-format=3.1]
  @{\tiny${}\pm{}$}
  >{\tiny}S[table-format=2.1]<{\endcollectcell}
  S[table-format=3.1]
  @{\tiny${}\pm{}$}
  >{\tiny}S[table-format=2.1]<{\endcollectcell}
  S[table-format=3.1]
  @{\tiny${}\pm{}$}
  >{\tiny}S[table-format=2.1]<{\endcollectcell}
}
    \toprule
    \multicolumn{2}{c}{\textbf{PPI}} &  \multicolumn{4}{c}{\small low}  & \multicolumn{4}{c}{\small unconstrained} & \multicolumn{4}{c}{\small high}\\
    \cmidrule(lr){3-6}\cmidrule(lr){7-10}\cmidrule(lr){11-14}
    \makecell{\small $\hat k$} & Method   & \multicolumn{2}{c}{\small precision}  & \multicolumn{2}{c}{\small recall} & \multicolumn{2}{c}{\small precision} & \multicolumn{2}{c}{\small recall} & \multicolumn{2}{c}{\small precision} & \multicolumn{2}{c}{\small recall} \\\midrule
\rowcolor{tabgray}
\cellcolor{white}{\multirow{3}{*}{$k/4$}}
            & Ours  & \bfseries 100.0 & 0.0  & \bfseries 26.1 & 2.2  & \bfseries 99.5 & 0.7  & \bfseries 25.9 & 2.7  & \bfseries 99.7 & 0.3  & \bfseries 21.6 & 0.8 \\
            & LSA2-post  & 0.0 & 0.0  & 0.0 & 0.0  & 0.0 & 0.0  & 0.0 & 0.0  & 1.4 & 0.6  & 0.3 & 0.1 \\
            & LSA2-attr  & 0.0 & 0.0  & 0.0 & 0.0  & 0.0 & 0.0  & 0.0 & 0.0  & 1.3 & 0.7  & 0.3 & 0.1 \\[0.05em]\midrule[0.05em]
\rowcolor{tabgray}
\cellcolor{white}{\multirow{3}{*}{$k/2$}}
            & Ours  & \bfseries 100.0 & 0.0  & \bfseries 47.6 & 4.7  & \bfseries 99.5 & 0.8  & \bfseries 51.5 & 5.4  & \bfseries 99.7 & 0.2  & \bfseries 43.3 & 1.6 \\
            & LSA2-post  & 0.0 & 0.0  & 0.0 & 0.0  & 0.3 & 0.4  & 0.1 & 0.2  & 1.6 & 0.5  & 0.7 & 0.2 \\
            & LSA2-attr  & 0.0 & 0.0  & 0.0 & 0.0  & 0.0 & 0.0  & 0.0 & 0.0  & 0.6 & 0.3  & 0.3 & 0.1 \\[0.05em]\midrule[0.05em]
\rowcolor{tabgray}
\cellcolor{white}{\multirow{3}{*}{$k$}}
            & Ours  & \bfseries 98.7 & 1.9  & \bfseries 89.2 & 7.9  & \bfseries 89.5 & 6.5  & \bfseries 91.9 & 3.7  & \bfseries 98.0 & 0.3  & \bfseries 85.1 & 3.2 \\
            & LSA2-post  & 0.0 & 0.0  & 0.0 & 0.0  & 0.3 & 0.2  & 0.3 & 0.2  & 2.1 & 0.6  & 1.8 & 0.6 \\
            & LSA2-attr  & 0.0 & 0.0  & 0.0 & 0.0  & 0.1 & 0.2  & 0.1 & 0.2  & 0.3 & 0.2  & 0.3 & 0.1 \\[0.05em]\midrule[0.05em]
\rowcolor{tabgray}
\cellcolor{white}{\multirow{3}{*}{$2k$}}
            & Ours  & \bfseries 56.7 & 6.6  & \bfseries 100.0 & 0.0  & \bfseries 49.0 & 5.4  & \bfseries 100.0 & 0.0  & \bfseries 57.7 & 2.3  & \bfseries 100.0 & 0.0 \\
            & LSA2-post  & 0.0 & 0.0  & 0.0 & 0.0  & 0.3 & 0.2  & 0.5 & 0.4  & 2.1 & 0.1  & 3.6 & 0.3 \\
            & LSA2-attr  & 0.0 & 0.0  & 0.0 & 0.0  & 0.1 & 0.1  & 0.1 & 0.2  & 0.2 & 0.1  & 0.3 & 0.1 \\[0.05em]\midrule[0.05em]
\rowcolor{tabgray}
\cellcolor{white}{\multirow{3}{*}{$4k$}}
            & Ours  & \bfseries 28.3 & 3.3  & \bfseries 100.0 & 0.0  & \bfseries 24.5 & 2.7  & \bfseries 100.0 & 0.0  & \bfseries 28.8 & 1.2  & \bfseries 100.0 & 0.0 \\
            & LSA2-post  & 0.0 & 0.0  & 0.0 & 0.0  & 0.3 & 0.1  & 1.3 & 0.3  & 2.0 & 0.0  & 7.0 & 0.1 \\
            & LSA2-attr  & 0.3 & 0.5  & 1.1 & 1.6  & 0.0 & 0.0  & 0.1 & 0.2  & 0.1 & 0.0  & 0.3 & 0.1 \\\bottomrule
\end{tabular}%
}
\end{subtable}

\begin{subtable}[]{\columnwidth}
    \centering
    \vspace{0.5em}
    \resizebox{\columnwidth}{!}{%
        \begin{tabular}{cc
  S[table-format=3.1]
  @{\tiny${}\pm{}$}
  >{\tiny}S[table-format=2.1]<{\endcollectcell}
  S[table-format=3.1]
  @{\tiny${}\pm{}$}
  >{\tiny}S[table-format=2.1]<{\endcollectcell}
  S[table-format=3.1]
  @{\tiny${}\pm{}$}
  >{\tiny}S[table-format=2.1]<{\endcollectcell}
  S[table-format=3.1]
  @{\tiny${}\pm{}$}
  >{\tiny}S[table-format=2.1]<{\endcollectcell}
  S[table-format=3.1]
  @{\tiny${}\pm{}$}
  >{\tiny}S[table-format=2.1]<{\endcollectcell}
  S[table-format=3.1]
  @{\tiny${}\pm{}$}
  >{\tiny}S[table-format=2.1]<{\endcollectcell}
  S[table-format=3.1]
  @{\tiny${}\pm{}$}
  >{\tiny}S[table-format=2.1]<{\endcollectcell}
}
    \toprule
    \multicolumn{2}{c}{\textbf{Flickr}} &  \multicolumn{4}{c}{\small low}  & \multicolumn{4}{c}{\small unconstrained} & \multicolumn{4}{c}{\small high}\\
    \cmidrule(lr){3-6}\cmidrule(lr){7-10}\cmidrule(lr){11-14}
    \makecell{\small $\hat k$} & Method   & \multicolumn{2}{c}{\small precision}  & \multicolumn{2}{c}{\small recall} & \multicolumn{2}{c}{\small precision} & \multicolumn{2}{c}{\small recall} & \multicolumn{2}{c}{\small precision} & \multicolumn{2}{c}{\small recall} \\\midrule
\rowcolor{tabgray}
\cellcolor{white}{\multirow{3}{*}{$k/4$}}
            & Ours  & \bfseries 83.3 & 23.6  & \bfseries 26.1 & 5.5  & \bfseries 63.9 & 30.7  & \bfseries 18.4 & 9.0  & \bfseries 14.9 & 3.8  & \bfseries 3.8 & 1.3 \\
            & LSA2-post  & 0.0 & 0.0  & 0.0 & 0.0  & 0.0 & 0.0  & 0.0 & 0.0  & 1.4 & 2.0  & 0.4 & 0.6 \\
            & LSA2-attr  & 0.0 & 0.0  & 0.0 & 0.0  & 0.0 & 0.0  & 0.0 & 0.0  & 0.7 & 1.0  & 0.2 & 0.3 \\[0.05em]\midrule[0.05em]
\rowcolor{tabgray}
\cellcolor{white}{\multirow{3}{*}{$k/2$}}
            & Ours  & \bfseries 63.9 & 10.4  & \bfseries 38.3 & 10.3  & \bfseries 60.0 & 22.5  & \bfseries 29.7 & 11.7  & \bfseries 19.6 & 2.8  & \bfseries 9.9 & 1.9 \\
            & LSA2-post  & 0.0 & 0.0  & 0.0 & 0.0  & 0.0 & 0.0  & 0.0 & 0.0  & 1.8 & 1.1  & 0.9 & 0.6 \\
            & LSA2-attr  & 0.0 & 0.0  & 0.0 & 0.0  & 0.0 & 0.0  & 0.0 & 0.0  & 0.4 & 0.5  & 0.2 & 0.3 \\[0.05em]\midrule[0.05em]
\rowcolor{tabgray}
\cellcolor{white}{\multirow{3}{*}{$k$}}
            & Ours  & \bfseries 51.0 & 7.0  & \bfseries 53.3 & 4.7  & \bfseries 33.8 & 13.3  & \bfseries 32.1 & 13.3  & \bfseries 18.2 & 4.5  & \bfseries 18.5 & 6.1 \\
            & LSA2-post  & 0.0 & 0.0  & 0.0 & 0.0  & 0.0 & 0.0  & 0.0 & 0.0  & 2.3 & 0.7  & 2.3 & 0.9 \\
            & LSA2-attr  & 0.0 & 0.0  & 0.0 & 0.0  & 0.0 & 0.0  & 0.0 & 0.0  & 0.3 & 0.4  & 0.3 & 0.4 \\[0.05em]\midrule[0.05em]
\rowcolor{tabgray}
\cellcolor{white}{\multirow{3}{*}{$2k$}}
            & Ours  & \bfseries 34.5 & 6.7  & \bfseries 71.1 & 15.0  & \bfseries 27.3 & 8.4  & \bfseries 50.3 & 16.8  & \bfseries 13.3 & 1.7  & \bfseries 26.8 & 5.6 \\
            & LSA2-post  & 0.0 & 0.0  & 0.0 & 0.0  & 0.0 & 0.0  & 0.0 & 0.0  & 1.6 & 0.6  & 3.2 & 1.3 \\
            & LSA2-attr  & 0.0 & 0.0  & 0.0 & 0.0  & 0.0 & 0.0  & 0.0 & 0.0  & 0.4 & 0.4  & 0.8 & 0.9 \\[0.05em]\midrule[0.05em]
\rowcolor{tabgray}
\cellcolor{white}{\multirow{3}{*}{$4k$}}
            & Ours  & \bfseries 21.7 & 2.4  & \bfseries 86.1 & 10.4  & \bfseries 19.8 & 3.0  & \bfseries 71.9 & 10.6  & \bfseries 9.2 & 0.8  & \bfseries 37.3 & 7.1 \\
            & LSA2-post  & 0.0 & 0.0  & 0.0 & 0.0  & 0.0 & 0.0  & 0.0 & 0.0  & 1.3 & 0.3  & 5.3 & 1.2 \\
            & LSA2-attr  & 0.0 & 0.0  & 0.0 & 0.0  & 0.0 & 0.0  & 0.0 & 0.0  & 0.3 & 0.2  & 1.3 & 0.9 \\\bottomrule
\end{tabular}%
}
\end{subtable}

\end{table}
}
{
\setlength{\tabcolsep}{4pt} 

\begin{table}[]
    \centering
    \renewrobustcmd{\bfseries}{\fontseries{b}\selectfont}
    \sisetup{detect-weight,mode=text,group-minimum-digits = 4}    
    \caption{\small \textbf{AUC} of \linkteller comparing with two baselines LSA2-\{post, attr\}. Each column corresponds to one dataset. Groups of rows represent sampled nodes of different degrees.
    }
    \label{tab:att-np-auc}
\begin{subtable}[]{\columnwidth}
    \centering
    \resizebox{\columnwidth}{!}{%
        \begin{tabular}{cc
  S[table-format=1.2]
  @{\tiny${}\pm{}$}
  >{\tiny}S[table-format=1.2]<{\endcollectcell}
  S[table-format=1.2]
  @{\tiny${}\pm{}$}
  >{\tiny}S[table-format=1.2]<{\endcollectcell}
  S[table-format=1.2]
  @{\tiny${}\pm{}$}
  >{\tiny}S[table-format=1.2]<{\endcollectcell}
  S[table-format=1.2]
  @{\tiny${}\pm{}$}
  >{\tiny}S[table-format=1.2]<{\endcollectcell}
  S[table-format=1.2]
  @{\tiny${}\pm{}$}
  >{\tiny}S[table-format=1.2]<{\endcollectcell}
  S[table-format=1.2]
  @{\tiny${}\pm{}$}
  >{\tiny}S[table-format=1.2]<{\endcollectcell}
  S[table-format=1.2]
  @{\tiny${}\pm{}$}
  >{\tiny}S[table-format=1.2]<{\endcollectcell}
}
    \toprule
    \makecell{\small Degree} & {\small Method}   & \multicolumn{14}{c}{\small Dataset}\\
    \cmidrule(lr){3-16}
    & & \multicolumn{2}{c}{\small RU}  & \multicolumn{2}{c}{\small DE} & \multicolumn{2}{c}{\small FR} & \multicolumn{2}{c}{\small ENGB} & \multicolumn{2}{c}{\small PTBR} & \multicolumn{2}{c}{\small PPI} & \multicolumn{2}{c}{\small Flickr} \\\midrule
\rowcolor{tabgray}
\cellcolor{white}{\multirow{3}{*}{\small low}}
            & Ours  & \bfseries 1.00 & 0.00  & \bfseries 1.00 & 0.00  & \bfseries 1.00 & 0.00  & \bfseries 1.00 & 0.00  & \bfseries 1.00 & 0.00  & \bfseries 1.00 & 0.00  & \bfseries 1.00 & 0.00 \\
            & LSA2-post  & 0.58 & 0.04  & 0.58 & 0.09  & 0.67 & 0.06  & 0.56 & 0.02  & 0.59 & 0.01  & 0.70 & 0.05  & 0.65 & 0.09 \\
            & LSA2-attr  & 0.72 & 0.03  & 0.77 & 0.08  & 0.82 & 0.02  & 0.62 & 0.05  & 0.74 & 0.00  & 0.48 & 0.08  & 0.62 & 0.14 \\[0.05em]\midrule[0.05em]
\rowcolor{tabgray}
\cellcolor{white}{\multirow{3}{*}{\small\makecell{ uncon-\\strained}}}
            & Ours  & \bfseries 1.00 & 0.00  & \bfseries 1.00 & 0.00  & \bfseries 0.99 & 0.00  & \bfseries 1.00 & 0.00  & \bfseries 1.00 & 0.00  & \bfseries 1.00 & 0.00  & \bfseries 1.00 & 0.00 \\
            & LSA2-post  & 0.51 & 0.00  & 0.52 & 0.03  & 0.51 & 0.01  & 0.54 & 0.01  & 0.51 & 0.01  & 0.64 & 0.00  & 0.70 & 0.08 \\
            & LSA2-attr  & 0.53 & 0.03  & 0.51 & 0.02  & 0.53 & 0.01  & 0.61 & 0.02  & 0.49 & 0.01  & 0.48 & 0.02  & 0.49 & 0.04 \\[0.05em]\midrule[0.05em]
\rowcolor{tabgray}
\cellcolor{white}{\multirow{3}{*}{\small high}}
            & Ours  & \bfseries 1.00 & 0.00  & \bfseries 1.00 & 0.00  & \bfseries 0.99 & 0.01  & \bfseries 0.99 & 0.00  & \bfseries 0.99 & 0.00  & \bfseries 1.00 & 0.00  & \bfseries 0.97 & 0.00 \\
            & LSA2-post  & 0.52 & 0.01  & 0.51 & 0.01  & 0.52 & 0.01  & 0.54 & 0.01  & 0.51 & 0.00  & 0.57 & 0.00  & 0.69 & 0.01 \\
            & LSA2-attr  & 0.46 & 0.01  & 0.50 & 0.02  & 0.50 & 0.01  & 0.55 & 0.01  & 0.46 & 0.01  & 0.48 & 0.01  & 0.51 & 0.01 \\\bottomrule
\end{tabular}%
}
\end{subtable}
\end{table}
}


We first evaluate 
the precision, recall, and AUC of \linkteller on eight datasets in the inductive setting, 
under 3 sampling strategies (low, unconstrained, and high degree), using 5 density beliefs ($k/4,k/2,k,2k,4k$), compared with different baselines.
For each scenario, the reported results are averaged over 3 runs using different random seeds for node sampling.

We report the precision, recall, and AUC results on some datasets in~\Cref{tab:att-np} and~\Cref{tab:att-np-auc} and the remaining datasets in~\Cref{sec:append-eval-results}
due to the space limit.
We leave the results of the weak random attack baseline in~\Cref{sec:append-rand-baseline}.
As a brief summary,
\linkteller significantly outperforms the random attack baseline. 
We mainly focus on the comparison with LSA2 attacks~\cite{he2020stealing}.
We show that \linkteller significantly outperforms these two baselines.
In~\Cref{tab:att-np}, LSA2-\{post, attr\} fail to attack in most of the scenarios, while \linkteller attains fairly high precision and recall.
\m{The AUC scores in~\Cref{tab:att-np-auc} also demonstrate the advantage of \linkteller.}
\m{Since the baselines LSA2-\{post, attr\} are  only performed under transductive setting in He \etal~\cite{he2020stealing}, 
to demonstrate the generality of \linkteller, we also compare with them  following the same evaluation protocol as in He \etal~\cite{he2020stealing}  on three datasets in the transductive setting. 
The results are reported
in~\Cref{sec:append-trans}.}
\m{We can see that the inductive setting is indeed more challenging: the baselines always fail to attack in the inductive setting while \linkteller is effective; the baselines are able to re-identify some private edges in the transductive setting, while \linkteller is consistently more effective.}

\m{Intuitively, the high attack effectiveness of \linkteller compared to baselines is because that LSA2-\{post, attr\} only leverage node-level information (posteriors or node attributes) to perform the edge re-identification attack.} 
\m{Although these node-level features can be correlated with the graph structure in some graphs, this correlation is not guaranteed, especially in the inductive setting. 
In comparison, \linkteller leverages the graph-structure information inferred from the inter-node influence in a GCN model according to \Cref{thm:k-layer-gcn}.}
\m{We defer more detailed comparison and analysis in~\Cref{sec:append-baseline-cmp}.}


\m{In addition, 
it is clear that given an accurate estimation of the density ($\hat k=k$), \linkteller achieves very high precision and recall across different node degree distributions and datasets. 
It is interesting to see that even when the density estimation is inaccurate (\eg, $\hat k\in\{k/4,k/2,2k,4k\}$), the attack is still effective.
Concretely, when the belief is smaller ($\hat k=k/2$), 
the precision values increase in all cases; when the belief is larger ($\hat k=2k$), almost all recall values are above 90\% except for Flickr.
Even under extremely inaccurate estimations such as $\hat k=k/4$ (or $\hat k=4k$), the precision values (or the recall values) are mostly higher than $95\%$.
This observation demonstrates the generality of \linkteller.}
We notice that \linkteller's performance on Flickr is
slightly poorer than other datasets. 
This may be because that 
the trained GCN on Flickr does not achieve good performance given its highly sparse structure. This implies that the parameters of the trained network on Flickr may not capture the graph structure very well,
and thus negatively influencing the attack performance.


\vspace{-1mm}
\subsection{Beyond GCNs: \linkteller on GATs}
\label{sec:attack-gat}
\vspace{-1mm}
In this section, we aim to study the effectiveness of \linkteller on other GNNs.
Since the rule of information propagation holds almost ubiquitously in GNNs, we hypothesize that our influence analysis based \linkteller can also successfully attack other types of GNNs.
We directly apply~\Cref{alg:attack} on another classical model---Graph Attention Networks (GATs)~\cite{velivckovic2017graph}\m{, aiming to investigate the \textit{transferability} of our influence analysis based attack from GCNs.}

We evaluate the attack on the two large datasets PPI and Flickr introduced in~\Cref{tab:dataset}. For both datasets, we train a 3-layer GAT. We leave details of the architecture and hyperparameters to~\Cref{sec:append-best-para-gat} and report the result in~\Cref{tab:att-gat}.
\m{Although \linkteller still significantly outperforms the baselines, it is less effective than that on GCNs. This is mainly due to the different structures of GCNs and GATs, which leads to different inﬂuence calculations for the two models (one related to the graph convolution and the other related to the attention mechanism).
We provide more discussion on conveniently adapting \linkteller to other architectures in~\Cref{sec:append-limit}.
}

{
\setlength{\tabcolsep}{4pt} 

\begin{table}[]
    \centering
    \renewrobustcmd{\bfseries}{\fontseries{b}\selectfont}
    \sisetup{detect-weight,mode=text,group-minimum-digits = 4}  
    \vspace{-4mm}
    \caption{\small \textbf{Attack Performance (Precision and Recall)}
    of \linkteller on GAT.
    }
    \label{tab:att-gat}
\begin{subtable}[]{\columnwidth}
    \centering
    \vspace{0.5em}
    \resizebox{\columnwidth}{!}{%
        \begin{tabular}{cc
  S[table-format=3.1]
  @{\tiny${}\pm{}$}
  >{\tiny}S[table-format=2.1]<{\endcollectcell}
  S[table-format=3.1]
  @{\tiny${}\pm{}$}
  >{\tiny}S[table-format=2.1]<{\endcollectcell}
  S[table-format=3.1]
  @{\tiny${}\pm{}$}
  >{\tiny}S[table-format=2.1]<{\endcollectcell}
  S[table-format=3.1]
  @{\tiny${}\pm{}$}
  >{\tiny}S[table-format=2.1]<{\endcollectcell}
  S[table-format=3.1]
  @{\tiny${}\pm{}$}
  >{\tiny}S[table-format=2.1]<{\endcollectcell}
  S[table-format=3.1]
  @{\tiny${}\pm{}$}
  >{\tiny}S[table-format=2.1]<{\endcollectcell}
  S[table-format=3.1]
  @{\tiny${}\pm{}$}
  >{\tiny}S[table-format=2.1]<{\endcollectcell}
}
    \toprule
    \multicolumn{2}{c}{\textbf{GAT, PPI}} &  \multicolumn{4}{c}{\small low}  & \multicolumn{4}{c}{\small unconstrained} & \multicolumn{4}{c}{\small high}\\
    \cmidrule(lr){3-6}\cmidrule(lr){7-10}\cmidrule(lr){11-14}
    \makecell{\small $\hat k$} & Method   & \multicolumn{2}{c}{\small precision}  & \multicolumn{2}{c}{\small recall} & \multicolumn{2}{c}{\small precision} & \multicolumn{2}{c}{\small recall} & \multicolumn{2}{c}{\small precision} & \multicolumn{2}{c}{\small recall} \\\midrule
\rowcolor{tabgray}
\cellcolor{white}{\multirow{3}{*}{$k/4$}}
            & Ours  & \bfseries 8.3 & 11.8  & \bfseries 2.1 & 2.9  & \bfseries 21.2 & 9.7  & \bfseries 5.8 & 3.3  & \bfseries 36.0 & 5.6  & \bfseries 7.8 & 0.9 \\
            & LSA2-post  & 0.0 & 0.0  & 0.0 & 0.0  & 0.0 & 0.0  & 0.0 & 0.0  & 3.4 & 0.5  & 0.7 & 0.1 \\
            & LSA2-attr  & 0.0 & 0.0  & 0.0 & 0.0  & 0.0 & 0.0  & 0.0 & 0.0  & 1.3 & 0.7  & 0.3 & 0.1 \\[0.05em]\midrule[0.05em]
\rowcolor{tabgray}
\cellcolor{white}{\multirow{3}{*}{$k/2$}}
            & Ours  & \bfseries 14.3 & 11.7  & \bfseries 5.3 & 5.3  & \bfseries 19.5 & 9.3  & \bfseries 9.9 & 4.4  & \bfseries 26.6 & 1.4  & \bfseries 11.5 & 0.1 \\
            & LSA2-post  & 0.0 & 0.0  & 0.0 & 0.0  & 0.3 & 0.4  & 0.1 & 0.2  & 3.8 & 0.8  & 1.7 & 0.4 \\
            & LSA2-attr  & 0.0 & 0.0  & 0.0 & 0.0  & 0.0 & 0.0  & 0.0 & 0.0  & 0.6 & 0.3  & 0.3 & 0.1 \\[0.05em]\midrule[0.05em]
\rowcolor{tabgray}
\cellcolor{white}{\multirow{3}{*}{$k$}}
            & Ours  & \bfseries 20.5 & 3.6  & \bfseries 12.5 & 4.5  & \bfseries 12.7 & 6.4  & \bfseries 12.8 & 5.8  & \bfseries 18.5 & 2.1  & \bfseries 16.0 & 1.7 \\
            & LSA2-post  & 0.0 & 0.0  & 0.0 & 0.0  & 0.4 & 0.3  & 0.4 & 0.4  & 3.3 & 0.8  & 2.8 & 0.6 \\
            & LSA2-attr  & 0.0 & 0.0  & 0.0 & 0.0  & 0.1 & 0.2  & 0.1 & 0.2  & 0.3 & 0.2  & 0.3 & 0.1 \\[0.05em]\midrule[0.05em]
\rowcolor{tabgray}
\cellcolor{white}{\multirow{3}{*}{$2k$}}
            & Ours  & \bfseries 10.7 & 1.9  & \bfseries 12.5 & 4.5  & \bfseries 7.5 & 3.8  & \bfseries 15.1 & 6.5  & \bfseries 12.5 & 1.0  & \bfseries 21.7 & 1.4 \\
            & LSA2-post  & 0.0 & 0.0  & 0.0 & 0.0  & 0.5 & 0.2  & 1.1 & 0.6  & 3.0 & 0.4  & 5.3 & 0.5 \\
            & LSA2-attr  & 0.0 & 0.0  & 0.0 & 0.0  & 0.1 & 0.1  & 0.1 & 0.2  & 0.2 & 0.1  & 0.3 & 0.1 \\[0.05em]\midrule[0.05em]
\rowcolor{tabgray}
\cellcolor{white}{\multirow{3}{*}{$4k$}}
            & Ours  & \bfseries 5.3 & 0.9  & \bfseries 12.5 & 4.5  & \bfseries 5.4 & 1.5  & \bfseries 21.7 & 3.8  & \bfseries 7.9 & 0.7  & \bfseries 27.5 & 1.3 \\
            & LSA2-post  & 0.7 & 0.9  & 2.1 & 2.9  & 0.9 & 0.1  & 3.6 & 0.9  & 2.7 & 0.2  & 9.2 & 0.5 \\
            & LSA2-attr  & 0.3 & 0.5  & 1.1 & 1.6  & 0.0 & 0.0  & 0.1 & 0.2  & 0.1 & 0.0  & 0.3 & 0.1 \\\bottomrule
\end{tabular}%
}
\end{subtable}

\begin{subtable}[]{\columnwidth}
    \centering
    \vspace{0.5em}
    \resizebox{\columnwidth}{!}{%
        \begin{tabular}{cc
  S[table-format=3.1]
  @{\tiny${}\pm{}$}
  >{\tiny}S[table-format=2.1]<{\endcollectcell}
  S[table-format=3.1]
  @{\tiny${}\pm{}$}
  >{\tiny}S[table-format=2.1]<{\endcollectcell}
  S[table-format=3.1]
  @{\tiny${}\pm{}$}
  >{\tiny}S[table-format=2.1]<{\endcollectcell}
  S[table-format=3.1]
  @{\tiny${}\pm{}$}
  >{\tiny}S[table-format=2.1]<{\endcollectcell}
  S[table-format=3.1]
  @{\tiny${}\pm{}$}
  >{\tiny}S[table-format=2.1]<{\endcollectcell}
  S[table-format=3.1]
  @{\tiny${}\pm{}$}
  >{\tiny}S[table-format=2.1]<{\endcollectcell}
  S[table-format=3.1]
  @{\tiny${}\pm{}$}
  >{\tiny}S[table-format=2.1]<{\endcollectcell}
}
    \toprule
    \multicolumn{2}{c}{\textbf{GAT, Flickr}} &  \multicolumn{4}{c}{\small low}  & \multicolumn{4}{c}{\small unconstrained} & \multicolumn{4}{c}{\small high}\\
    \cmidrule(lr){3-6}\cmidrule(lr){7-10}\cmidrule(lr){11-14}
    \makecell{\small $\hat k$} & Method   & \multicolumn{2}{c}{\small precision}  & \multicolumn{2}{c}{\small recall} & \multicolumn{2}{c}{\small precision} & \multicolumn{2}{c}{\small recall} & \multicolumn{2}{c}{\small precision} & \multicolumn{2}{c}{\small recall} \\\midrule
\rowcolor{tabgray}
\cellcolor{white}{\multirow{3}{*}{$k/4$}}
            & Ours  & \bfseries 33.3 & 47.1  & \bfseries 8.3 & 11.8  & \bfseries 8.3 & 11.8  & \bfseries 2.4 & 3.4  & \bfseries 14.5 & 3.2  & \bfseries 3.6 & 0.3 \\
            & LSA2-post  & 0.0 & 0.0  & 0.0 & 0.0  & 0.0 & 0.0  & 0.0 & 0.0  & 0.4 & 0.5  & 0.1 & 0.1 \\
            & LSA2-attr  & 0.0 & 0.0  & 0.0 & 0.0  & 0.0 & 0.0  & 0.0 & 0.0  & 0.7 & 1.0  & 0.2 & 0.3 \\[0.05em]\midrule[0.05em]
\rowcolor{tabgray}
\cellcolor{white}{\multirow{3}{*}{$k/2$}}
            & Ours  & \bfseries 16.7 & 23.6  & \bfseries 8.3 & 11.8  & \bfseries 4.8 & 6.7  & \bfseries 2.4 & 3.4  & \bfseries 7.3 & 1.7  & \bfseries 3.6 & 0.3 \\
            & LSA2-post  & 0.0 & 0.0  & 0.0 & 0.0  & 0.0 & 0.0  & 0.0 & 0.0  & 0.4 & 0.3  & 0.2 & 0.1 \\
            & LSA2-attr  & 0.0 & 0.0  & 0.0 & 0.0  & 0.0 & 0.0  & 0.0 & 0.0  & 0.4 & 0.5  & 0.2 & 0.3 \\[0.05em]\midrule[0.05em]
\rowcolor{tabgray}
\cellcolor{white}{\multirow{3}{*}{$k$}}
            & Ours  & \bfseries 8.3 & 11.8  & \bfseries 8.3 & 11.8  & \bfseries 5.9 & 4.3  & \bfseries 5.7 & 4.2  & \bfseries 4.2 & 1.0  & \bfseries 4.2 & 0.7 \\
            & LSA2-post  & 0.0 & 0.0  & 0.0 & 0.0  & 0.0 & 0.0  & 0.0 & 0.0  & 0.2 & 0.2  & 0.2 & 0.1 \\
            & LSA2-attr  & 0.0 & 0.0  & 0.0 & 0.0  & 0.0 & 0.0  & 0.0 & 0.0  & 0.3 & 0.4  & 0.3 & 0.4 \\[0.05em]\midrule[0.05em]
\rowcolor{tabgray}
\cellcolor{white}{\multirow{3}{*}{$2k$}}
            & Ours  & \bfseries 4.2 & 5.9  & \bfseries 8.3 & 11.8  & \bfseries 3.0 & 2.2  & \bfseries 5.7 & 4.2  & \bfseries 2.6 & 0.6  & \bfseries 5.1 & 0.9 \\
            & LSA2-post  & 0.0 & 0.0  & 0.0 & 0.0  & 0.0 & 0.0  & 0.0 & 0.0  & 0.4 & 0.2  & 0.7 & 0.4 \\
            & LSA2-attr  & 0.0 & 0.0  & 0.0 & 0.0  & 0.0 & 0.0  & 0.0 & 0.0  & 0.4 & 0.4  & 0.8 & 0.9 \\[0.05em]\midrule[0.05em]
\rowcolor{tabgray}
\cellcolor{white}{\multirow{3}{*}{$4k$}}
            & Ours  & \bfseries 2.2 & 3.1  & \bfseries 8.3 & 11.8  & \bfseries 1.5 & 1.1  & \bfseries 5.7 & 4.2  & \bfseries 1.3 & 0.3  & \bfseries 5.1 & 0.9 \\
            & LSA2-post  & 0.0 & 0.0  & 0.0 & 0.0  & 0.0 & 0.0  & 0.0 & 0.0  & 0.3 & 0.0  & 1.2 & 0.2 \\
            & LSA2-attr  & 0.0 & 0.0  & 0.0 & 0.0  & 0.0 & 0.0  & 0.0 & 0.0  & 0.3 & 0.2  & 1.3 & 0.9 \\\bottomrule
\end{tabular}%
}
\end{subtable}

\end{table}
}

\section{Evaluation of Differentially Private GCN}
\label{sec:dp-gcn}
In this section, we aim to understand the capability of \linkteller attack
by experimenting with potential ways to defend against it.
In particular, we examine whether it is possible to weaken the effectiveness of \linkteller by ensuring the \eedgedpshort guarantee of the GCN model.
We further investigate the utility of the DP GCN models.
In the end, we demonstrate the tradeoff between privacy and utility, which may be of interest to practitioners who wish to use DP GCNs to defend against \linkteller.


In particular,  we aim to evaluate the attack effectiveness of \linkteller and the model utility on four types of models: DP GCN models derived using DP mechanisms \edgerand and \lapgraph, vanilla GCN models which have no privacy guarantee, as well as multi-layer perceptron (MLP) models
with only node features.
We note that MLP can be viewed as ``perfectly" private since the edge information is not involved.

\subsection{Datasets and Models}
\vspace{-1mm}

We use the datasets described in~\Cref{sec:datasets}.
The DP GCN models are derived using DP mechanisms \edgerand and \lapgraph under various privacy guarantees. 
For each privacy budget $\eps$, we execute the procedure outlined in~\Cref{alg:dpgcn}, first getting a perturbed copy of the adjacency matrix, then using the perturbed training graph to train a GCN. We follow the criteria in~\Cref{sec:models} 
for parameter searching and model training, 
and
leave more descriptions to~\Cref{sec:append-search-space}. 

We provide an evaluation of the model utility,
aiming to characterize the tradeoff between model utility and the success rate of \linkteller.
In evaluating the utility of the DP GCNs, we compare with two baseline models: 1) vanilla GCNs which are expected to have higher utility, though vulnerable to \linkteller as previously shown; 
2) MLPs trained only on node features which may achieve lower classification utility but provide perfect protection of the edge information due to the non-involvement of edge information in the model.

\renewcommand{\thesubfigure}{\alph{subfigure}}
\newcommand{\mycaption}[1]
{\refstepcounter{subfigure}\textbf{(\thesubfigure) }{\ignorespaces #1}}

\begin{figure*}

\newlength{\utilheight}
\settoheight{\utilheight}{\includegraphics[width=.138\linewidth]{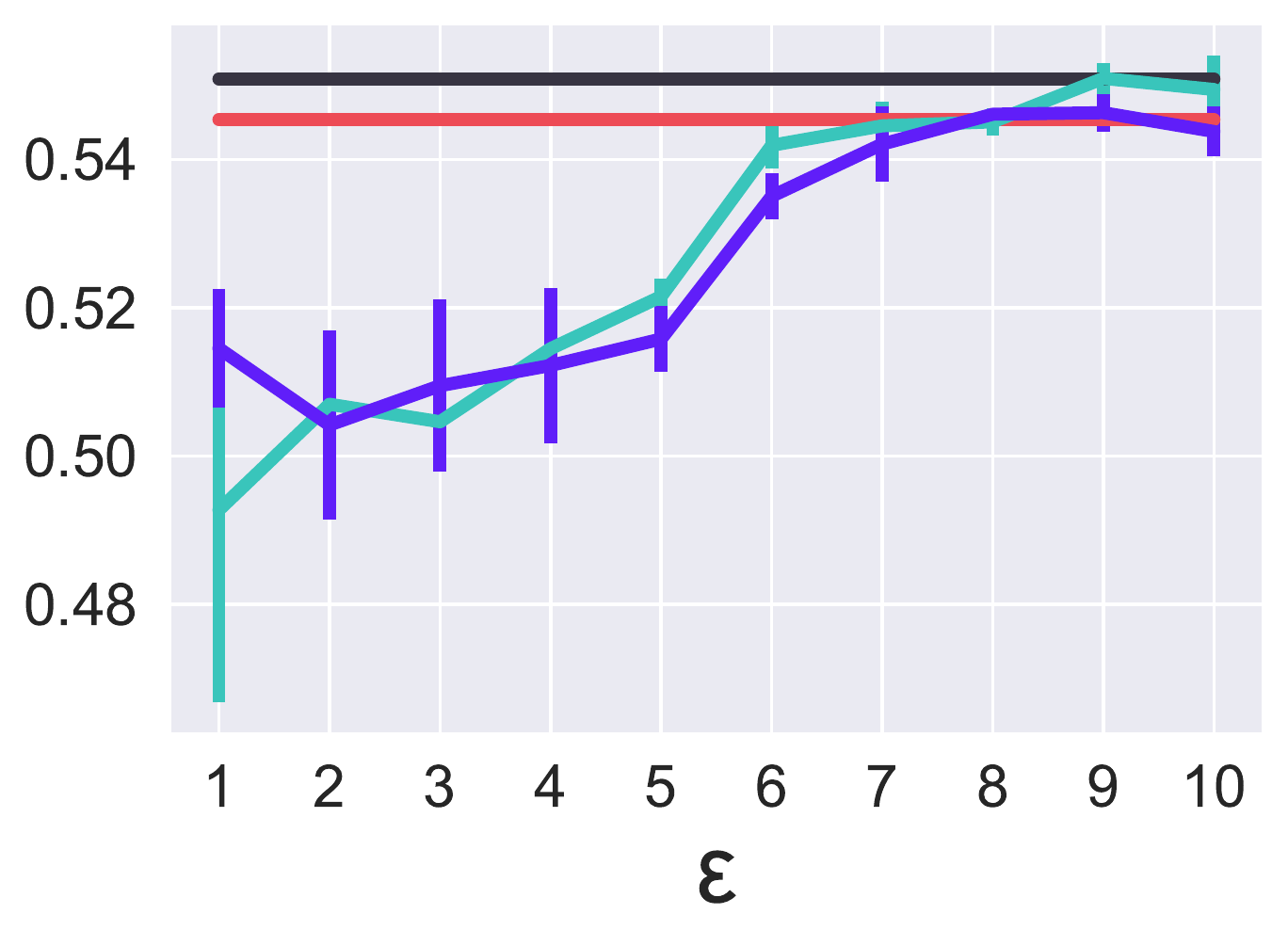}}%

\newlength{\tempdima}
\settoheight{\tempdima}{\includegraphics[width=.138\linewidth]{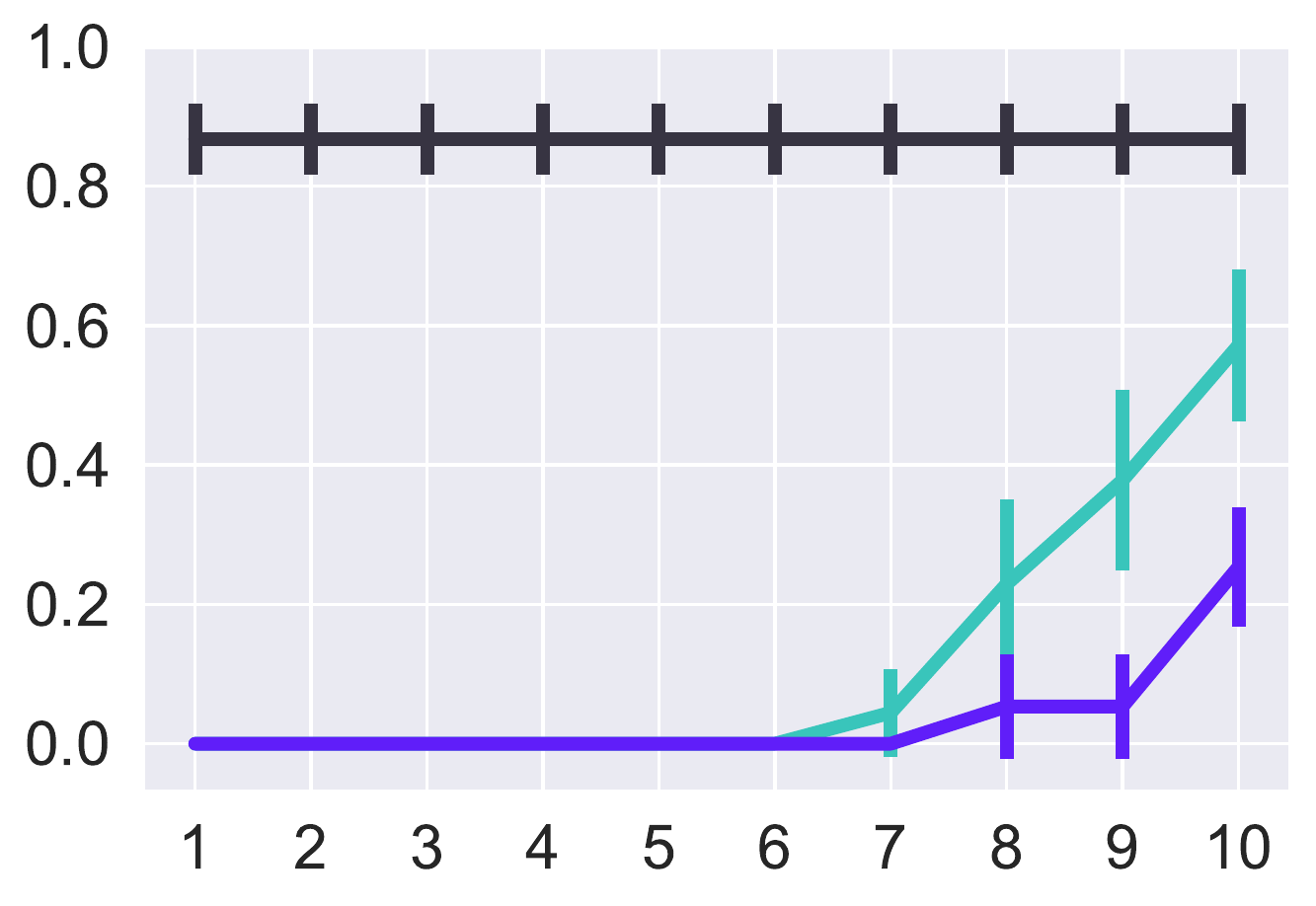}}%

\newlength{\attackheighta}
\settoheight{\attackheighta}{\includegraphics[width=.138\linewidth]{figures/twitch-DE_low-degree.pdf}}%

\newlength{\attackheightb}
\settoheight{\attackheightb}{\includegraphics[width=.138\linewidth]{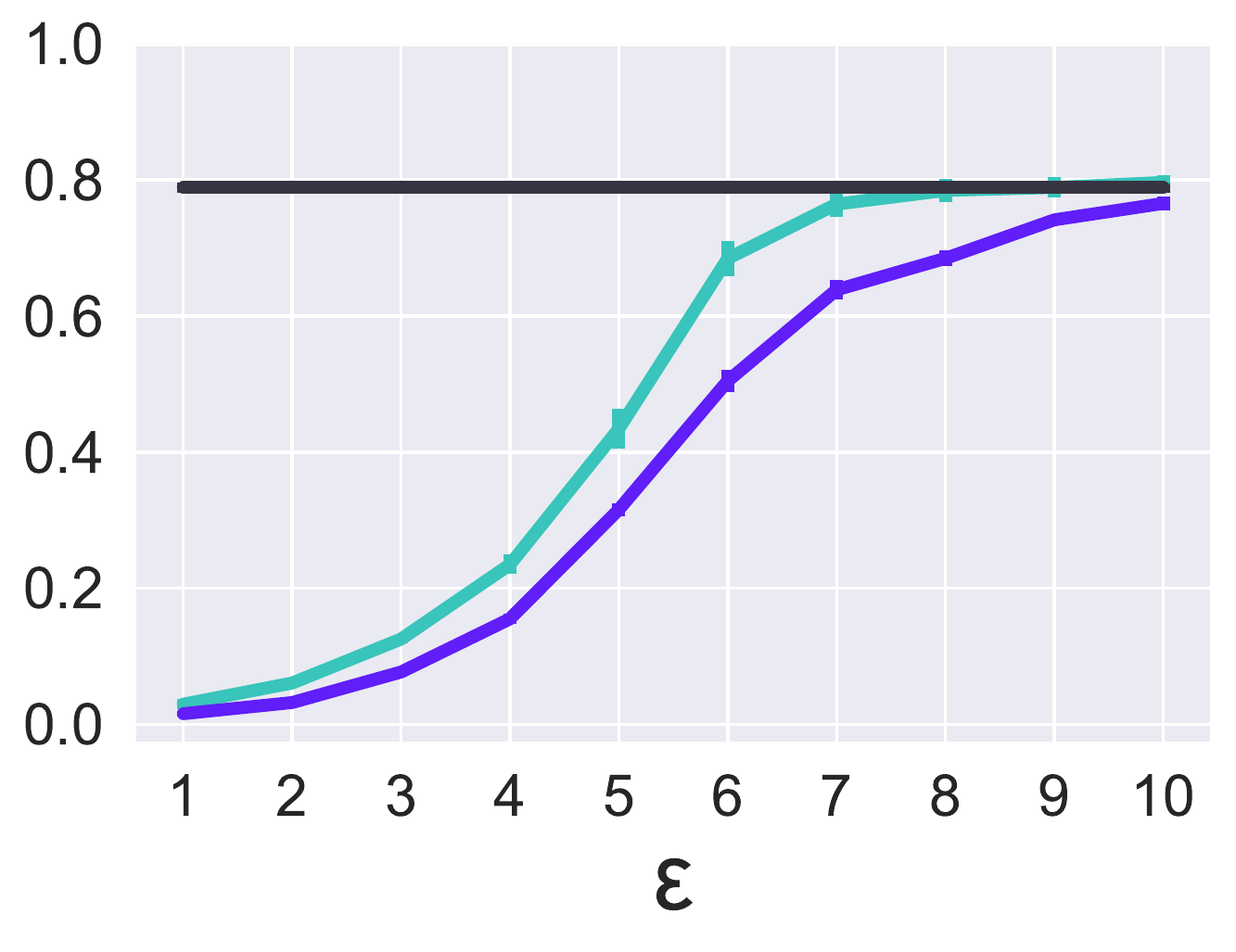}}%

\newlength{\legendheight}
\setlength{\legendheight}{0.3\attackheighta}%

\newcommand{\rowname}[1]
{\rotatebox{90}{\makebox[\tempdima][c]{\footnotesize #1}}}

\centering

{
\renewcommand{\tabcolsep}{10pt}

\begin{subtable}[]{\linewidth}
\begin{tabular}{l}
\includegraphics[height=\legendheight]{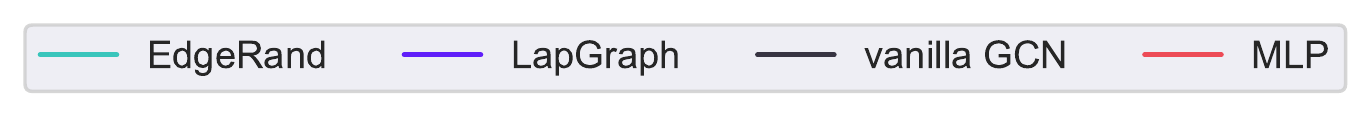}
\end{tabular}
\end{subtable}

\begin{subtable}[]{\linewidth}
\centering
    \begin{tabular}{@{}p{5mm}@{}c@{}c@{}c@{}c@{}c@{}c@{}c@{}}
        & \makecell{\footnotesize{twitch-RU}}
        & \makecell{\footnotesize{twitch-DE}}
        & \makecell{\footnotesize{twitch-FR}}
        & \makecell{\footnotesize{twitch-ENGB}}
        & \makecell{\footnotesize{twitch-PTBR}}
        & \makecell{\footnotesize{PPI}}
        & \makecell{\footnotesize{Flickr}}
        \vspace{-1.7pt}\\
        \rowname{\quad\footnotesize{}}&
        \includegraphics[height=\utilheight]{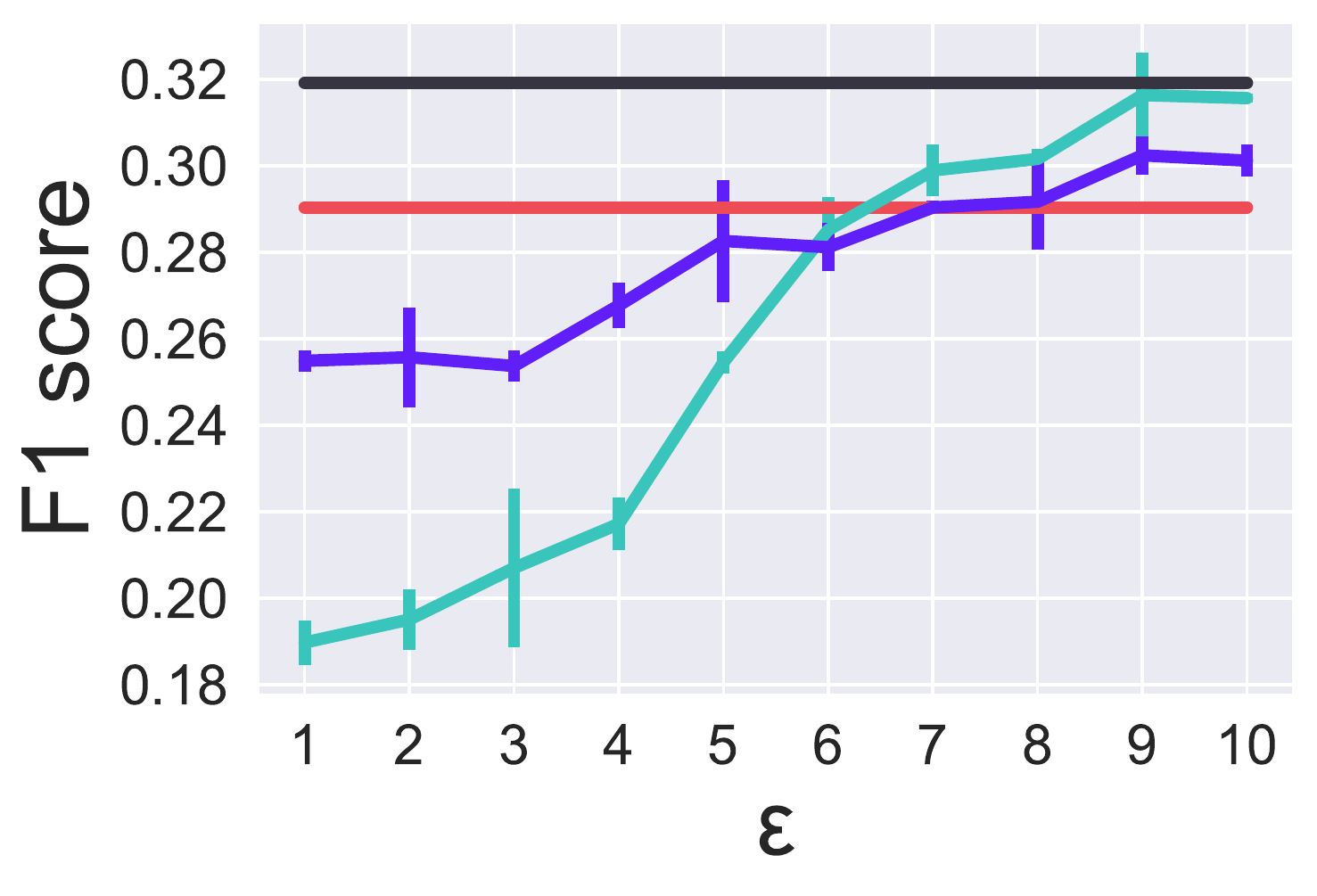}&
        \includegraphics[height=\utilheight]{figures/twitch-DE.pdf}&
        \includegraphics[height=\utilheight]{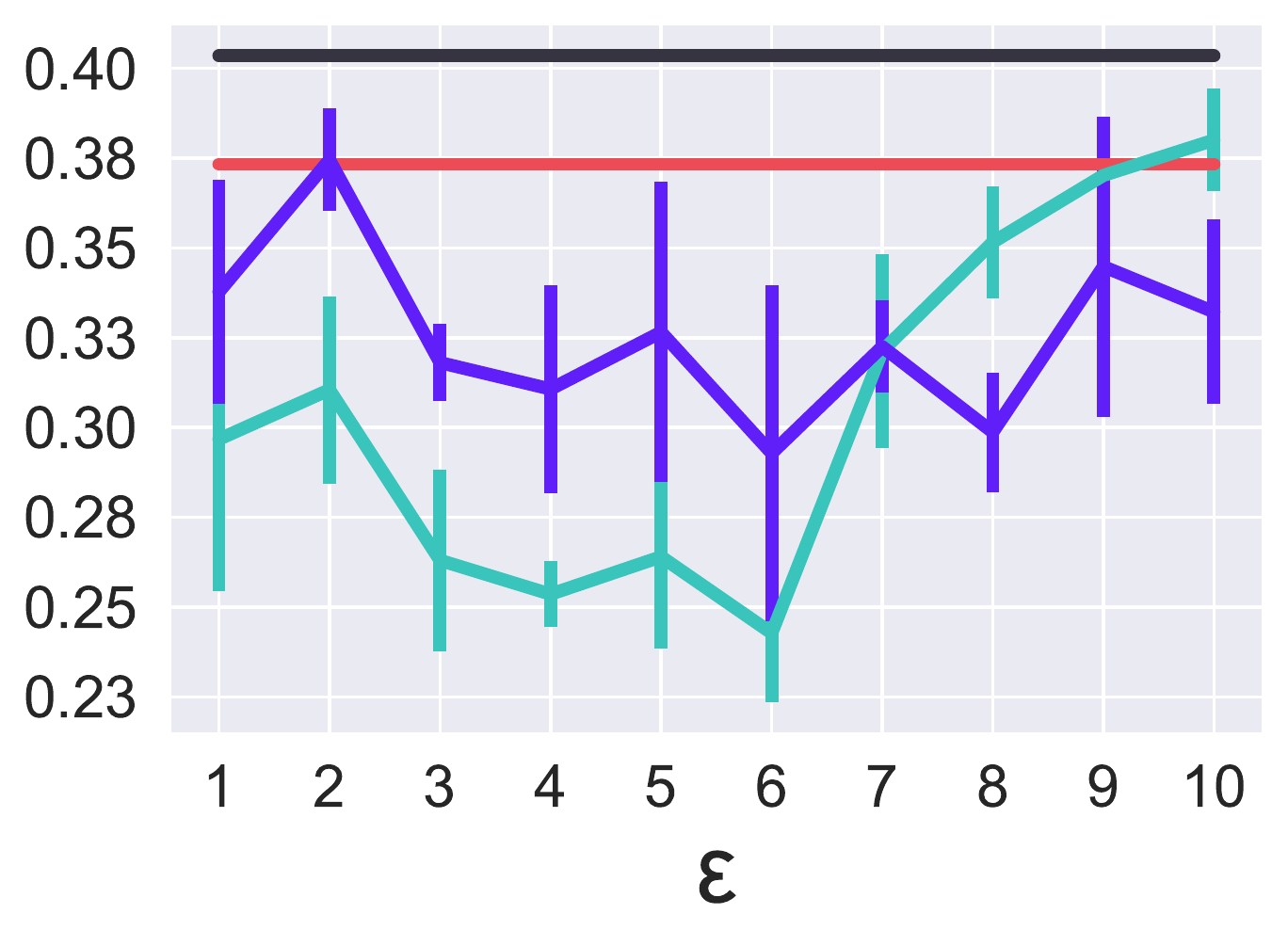}&
        \includegraphics[height=\utilheight]{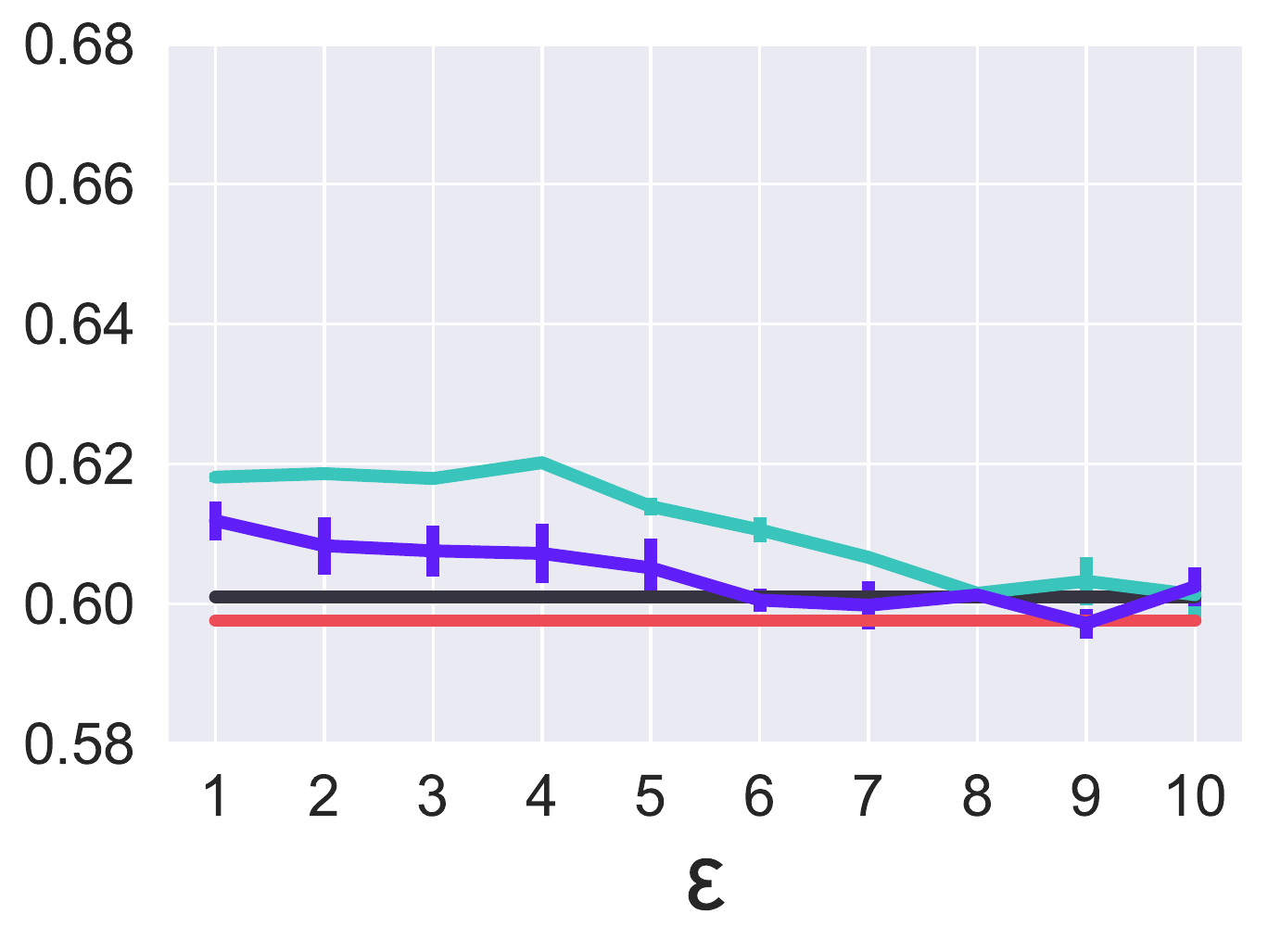}&
        \includegraphics[height=\utilheight]{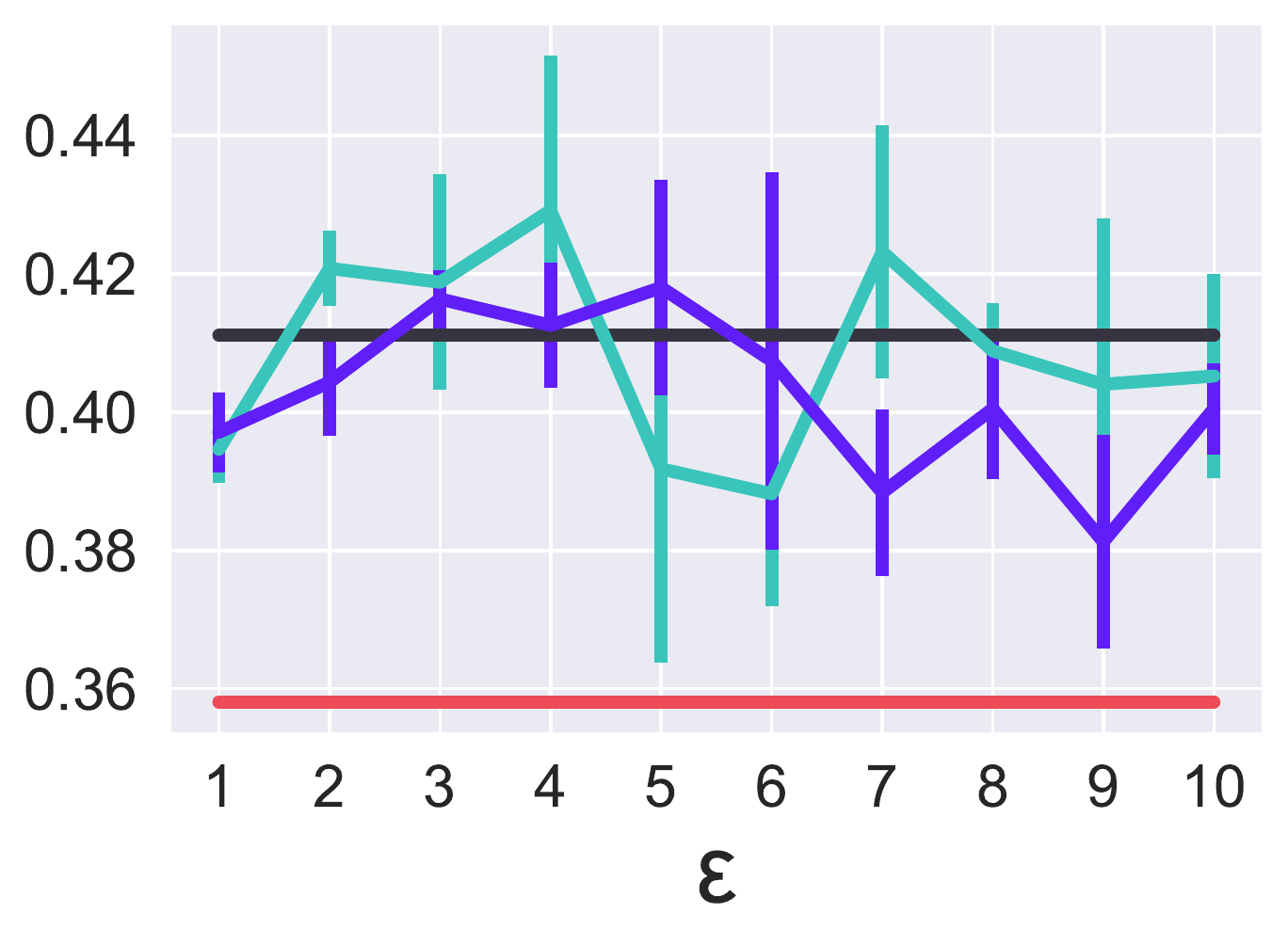}&
        \includegraphics[height=\utilheight]{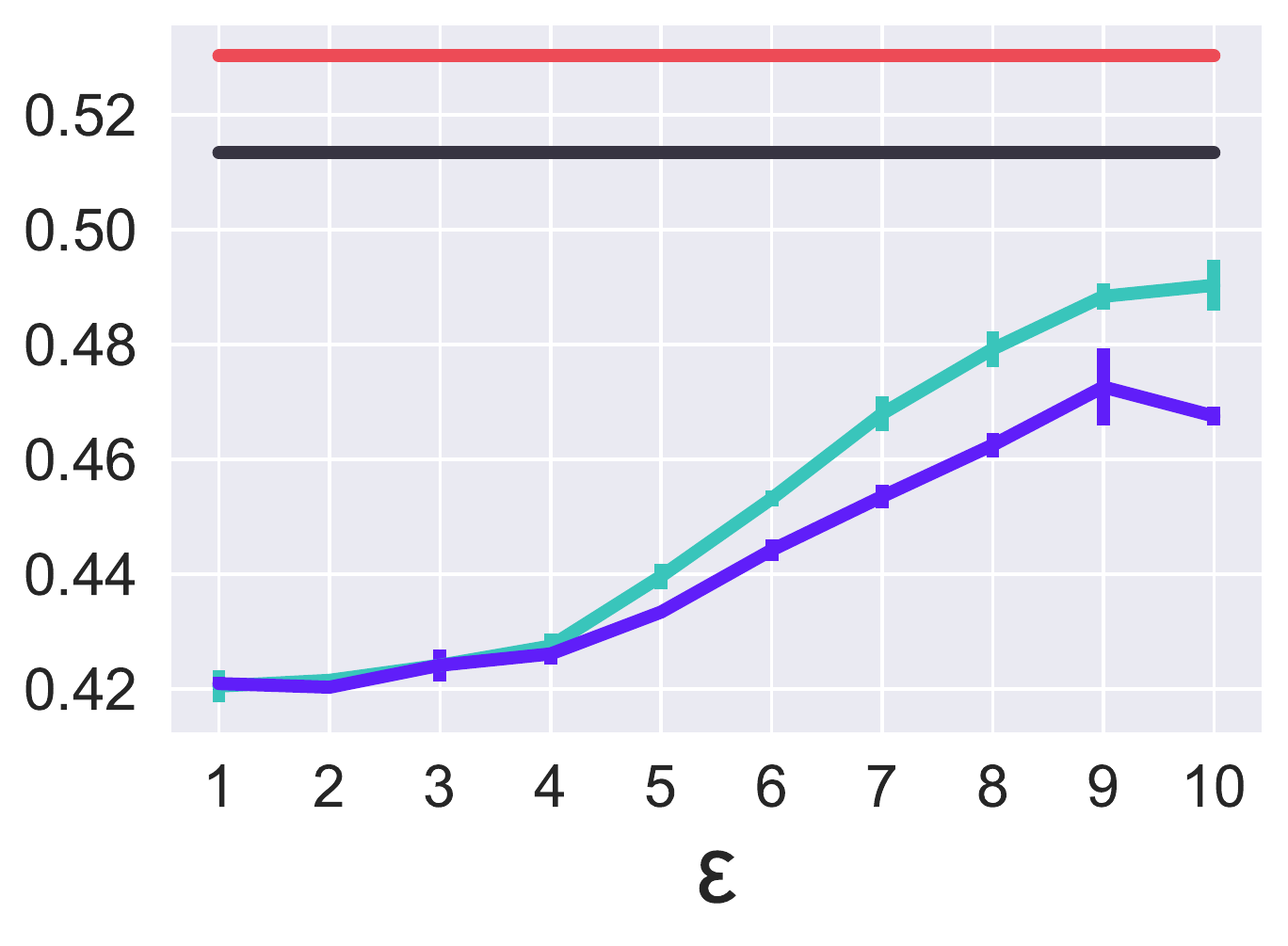}&
        \includegraphics[height=\utilheight]{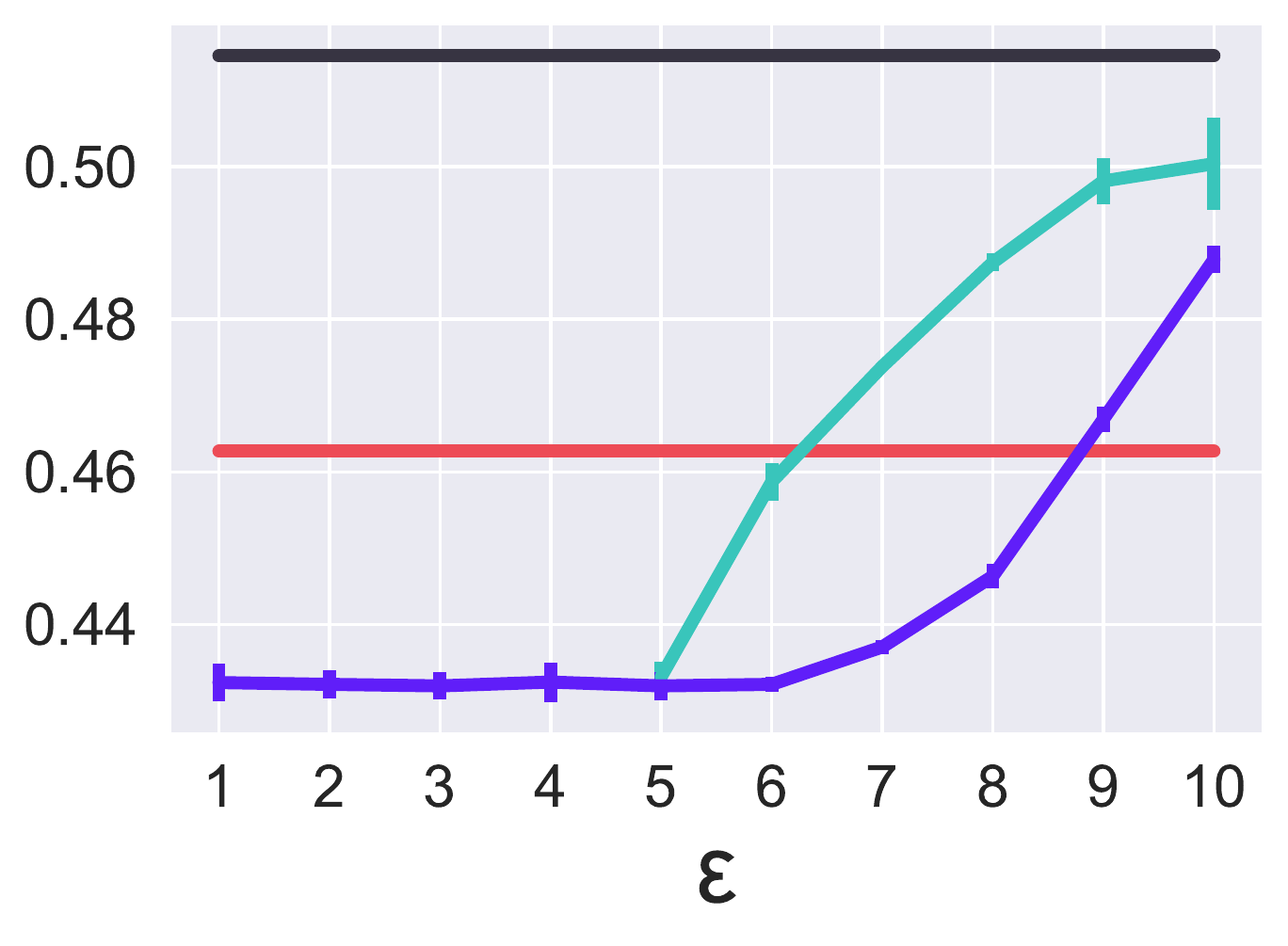}\\[-1.2ex]
    \end{tabular}
\caption{Model utility}\label{tab:utils}
\end{subtable}

\begin{subtable}[]{\linewidth}
\begin{tabular}{l}
\includegraphics[height=\legendheight]{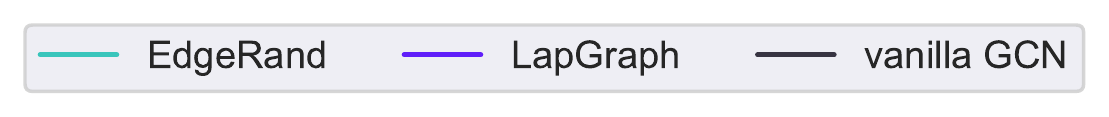}
\end{tabular}
\end{subtable}

\begin{subtable}[]{\linewidth}
\centering
\begin{tabular}{@{}p{5mm}@{}c@{}c@{}c@{}c@{}c@{}c@{}c@{}}
        & \makecell{\footnotesize{twitch-RU}}
        & \makecell{\footnotesize{twitch-DE}}
        & \makecell{\footnotesize{twitch-FR}}
        & \makecell{\footnotesize{twitch-ENGB}}
        & \makecell{\footnotesize{twitch-PTBR}}
        & \makecell{\footnotesize{PPI}}
        & \makecell{\footnotesize{Flickr}}
        \vspace{-1.7pt}\\
\rowname{\makecell{low}}&
\includegraphics[height=\attackheighta]{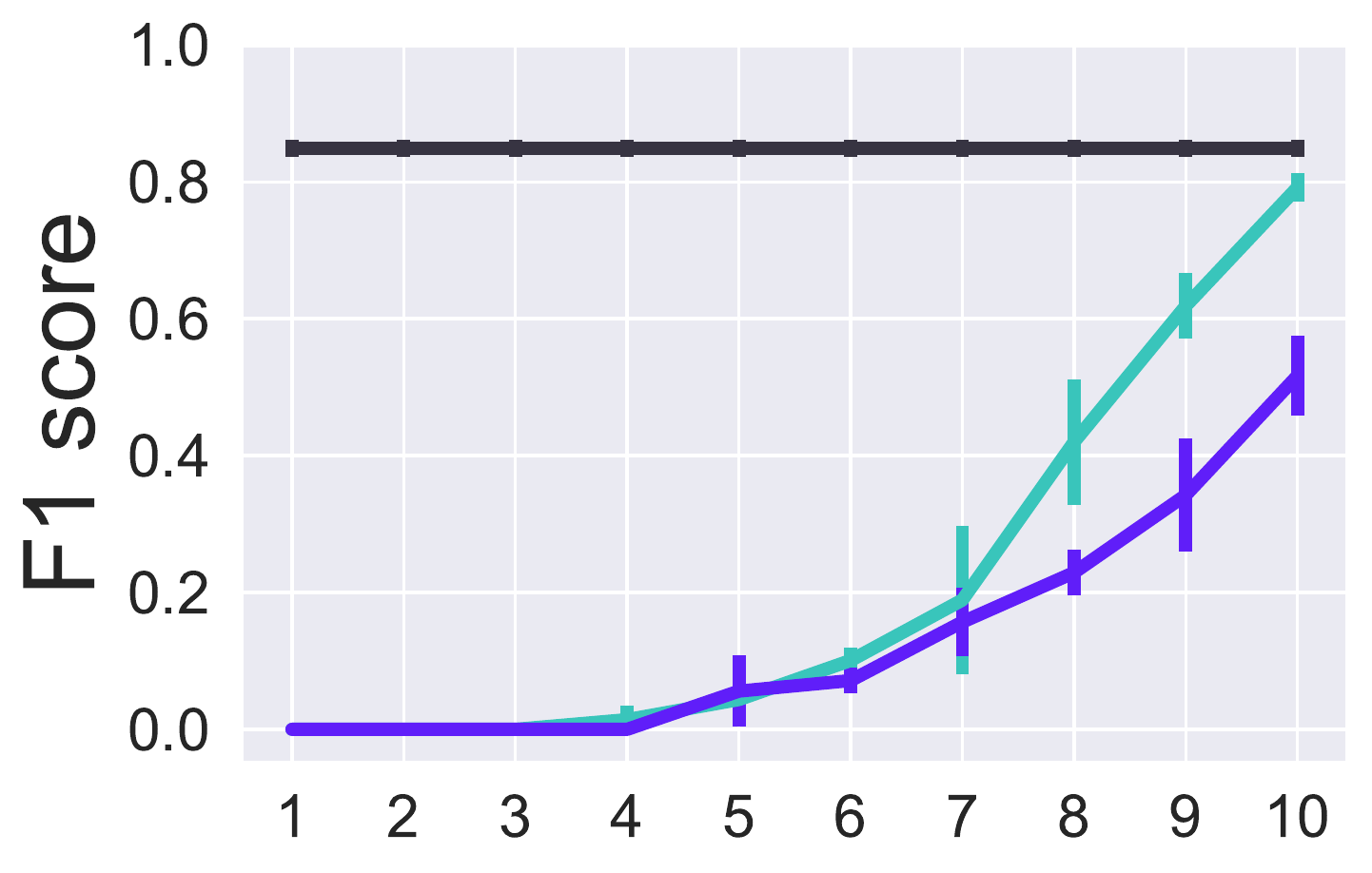}&
\includegraphics[height=\attackheighta]{figures/twitch-DE_low-degree.pdf}&
\includegraphics[height=\attackheighta]{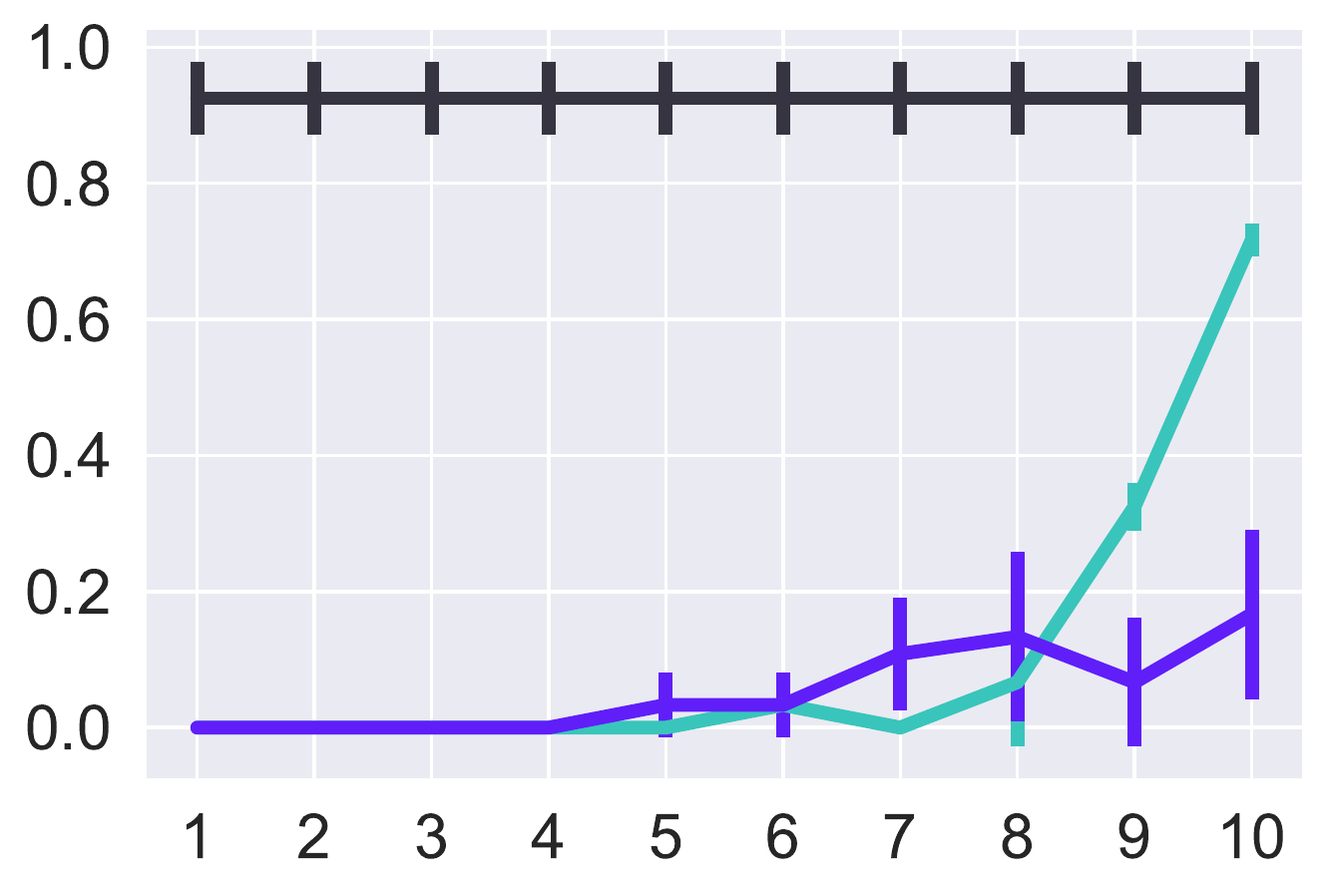}&
\includegraphics[height=\attackheighta]{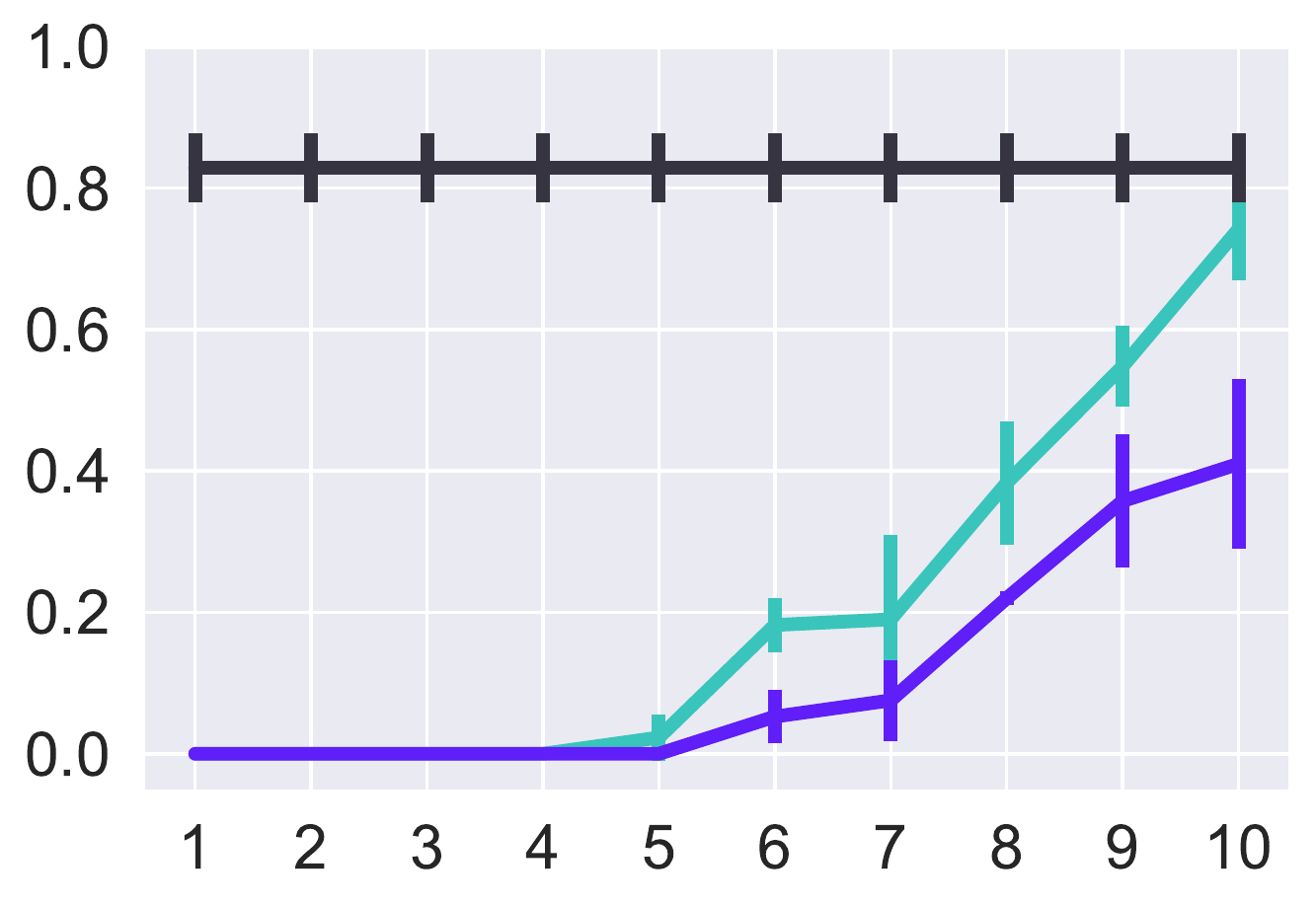}&
\includegraphics[height=\attackheighta]{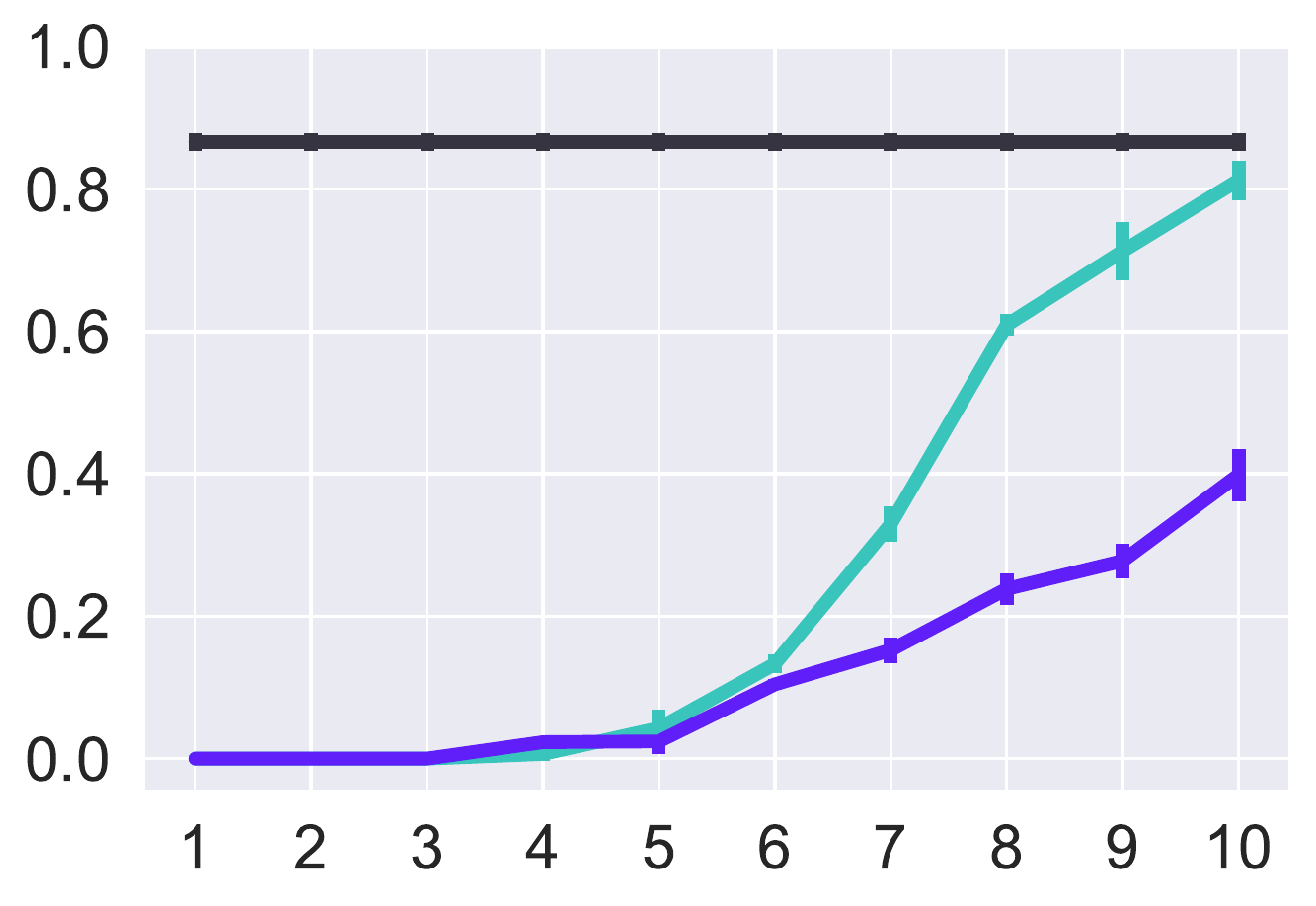}&
\includegraphics[height=\attackheighta]{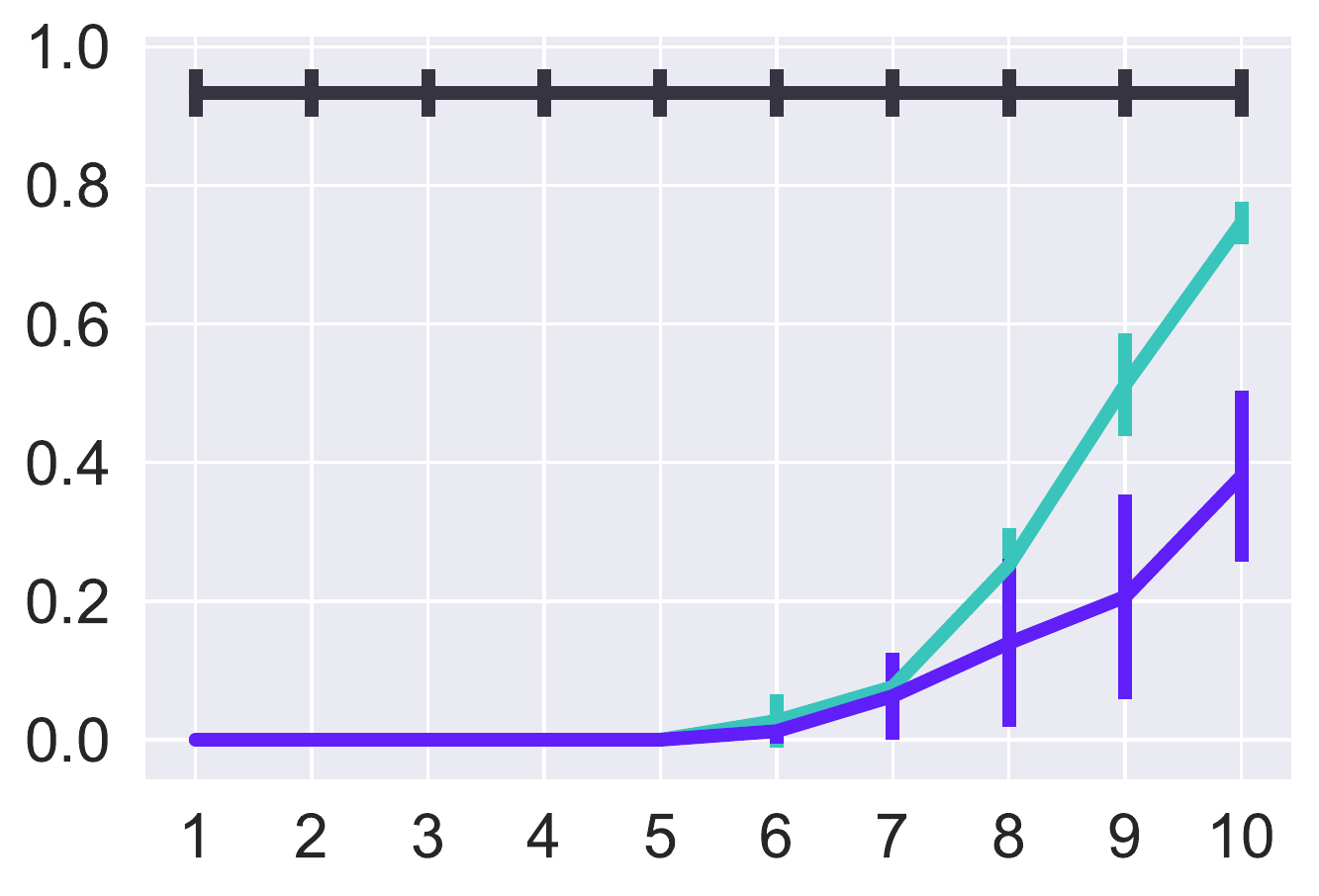}&
\includegraphics[height=\attackheighta]{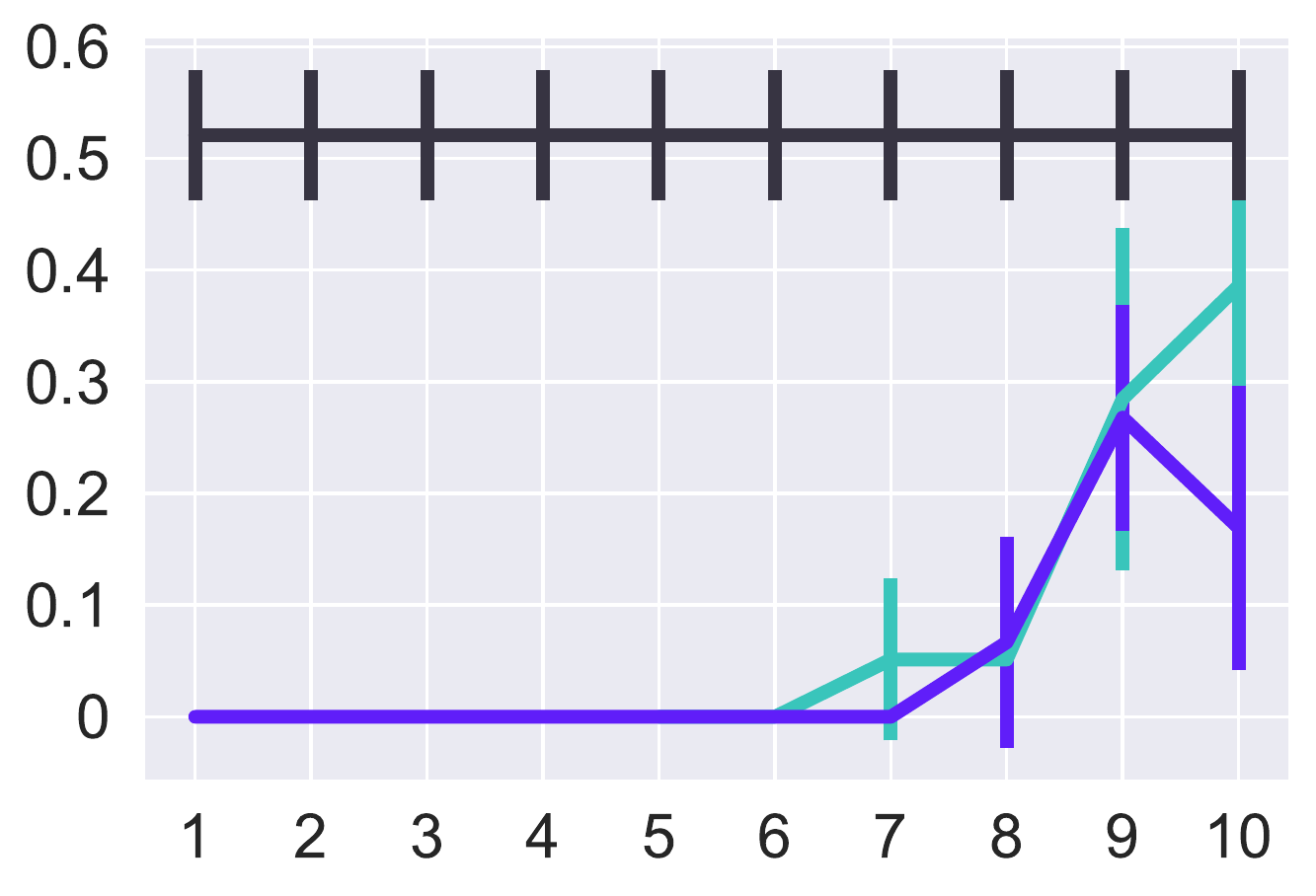}\\[-1.2ex]
\rowname{\makecell{unconstrained}}&
\includegraphics[height=\attackheighta]{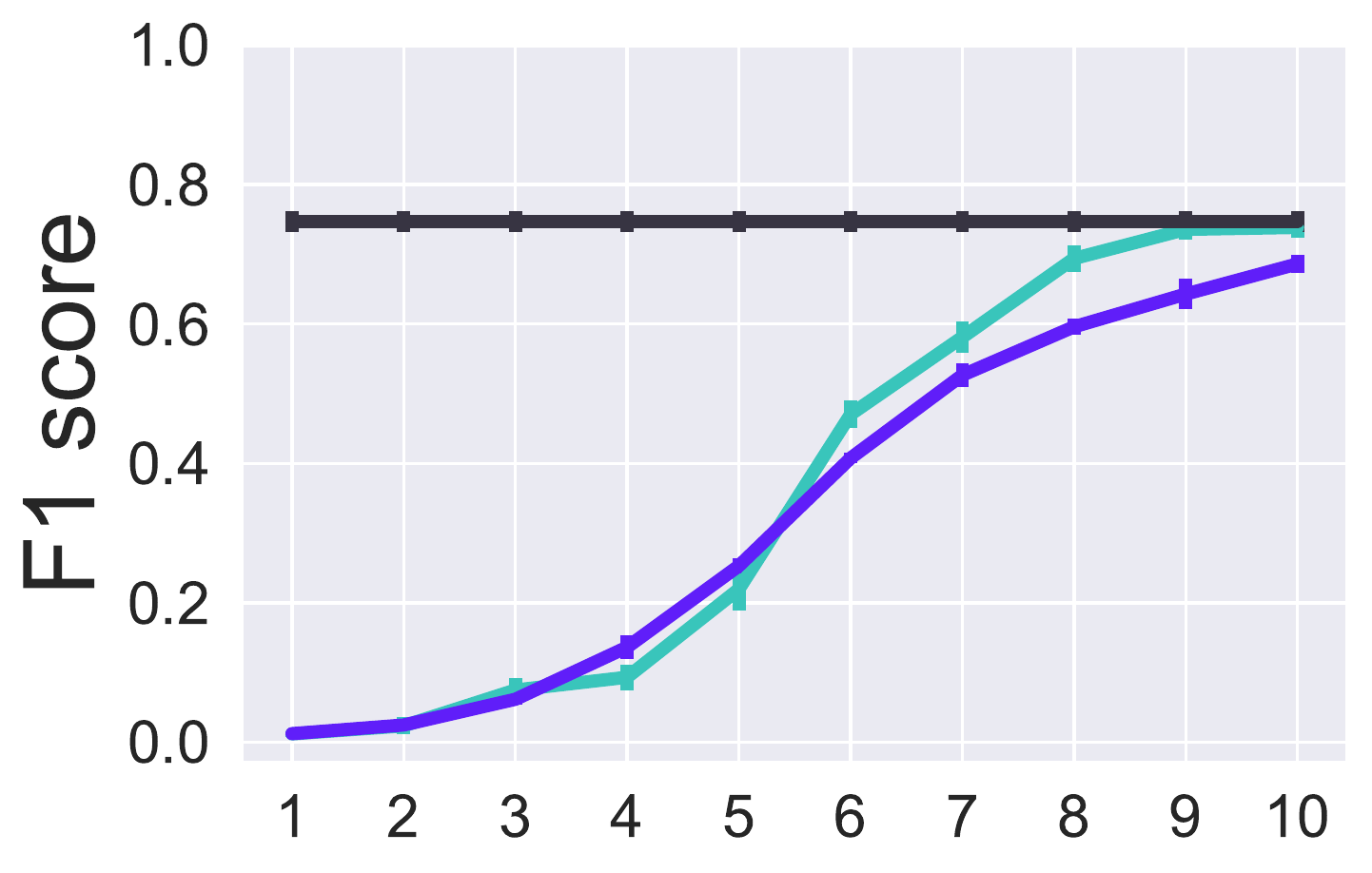}&
\includegraphics[height=\attackheighta]{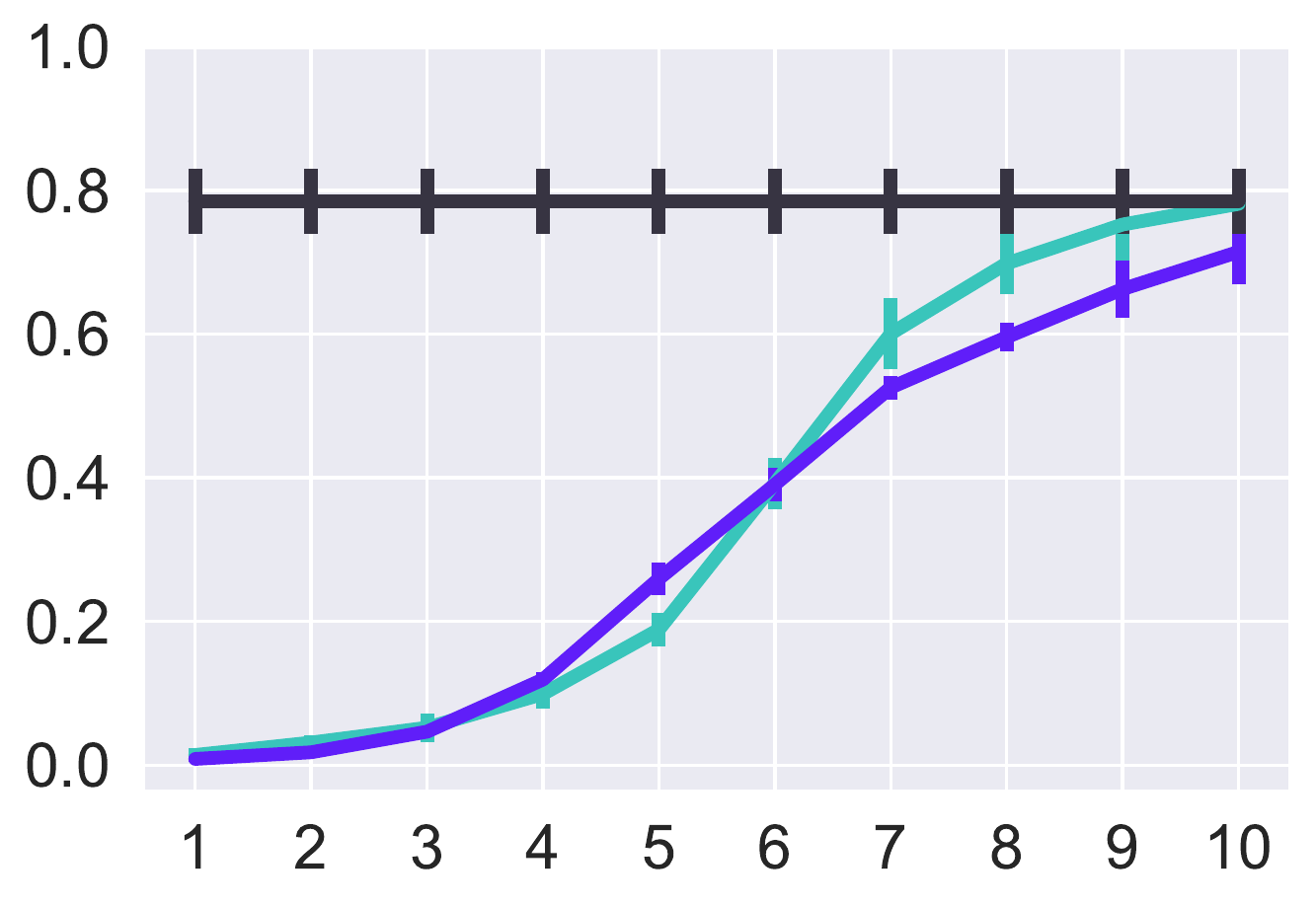}&
\includegraphics[height=\attackheighta]{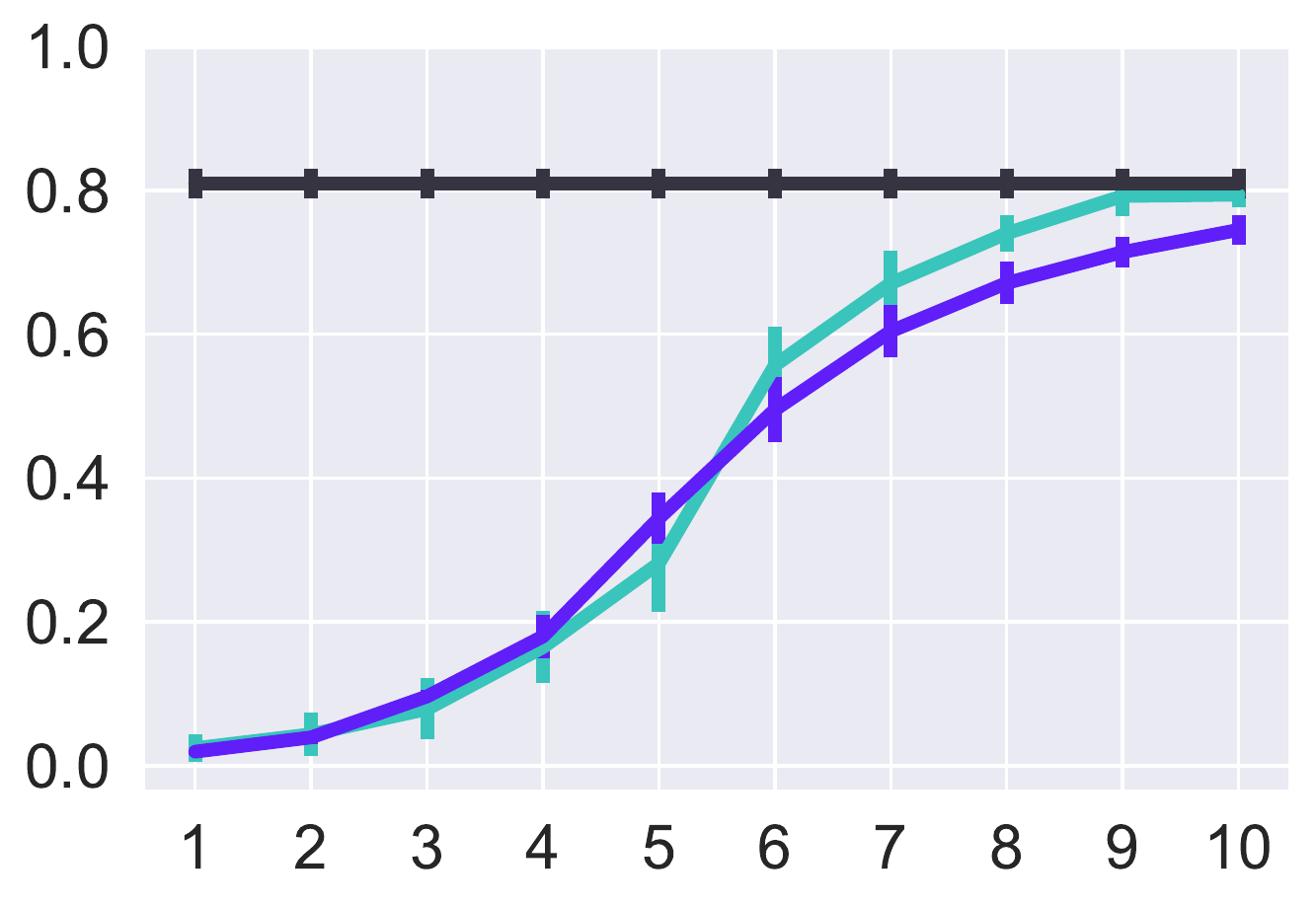}&
\includegraphics[height=\attackheighta]{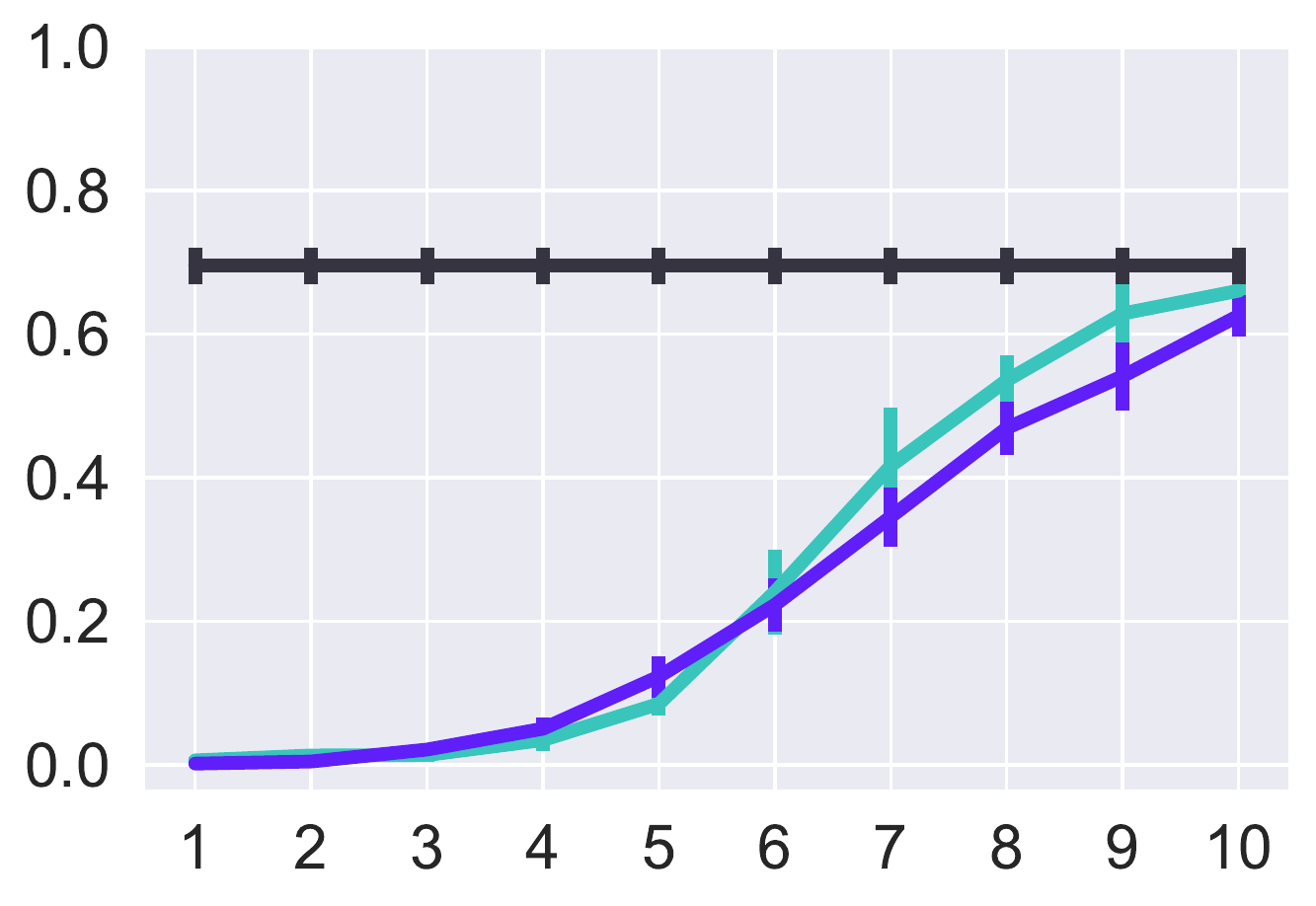}&
\includegraphics[height=\attackheighta]{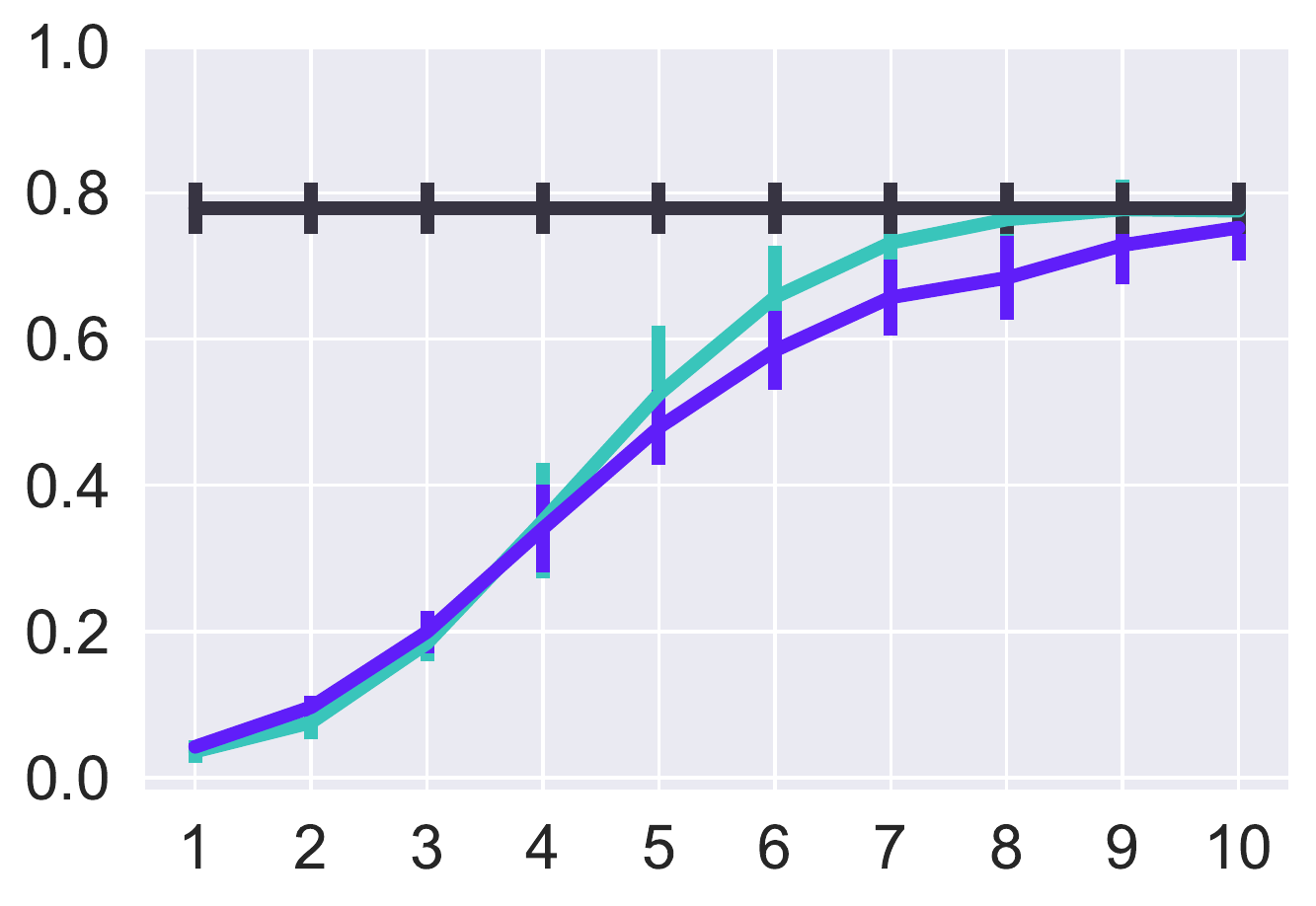}&
\includegraphics[height=\attackheighta]{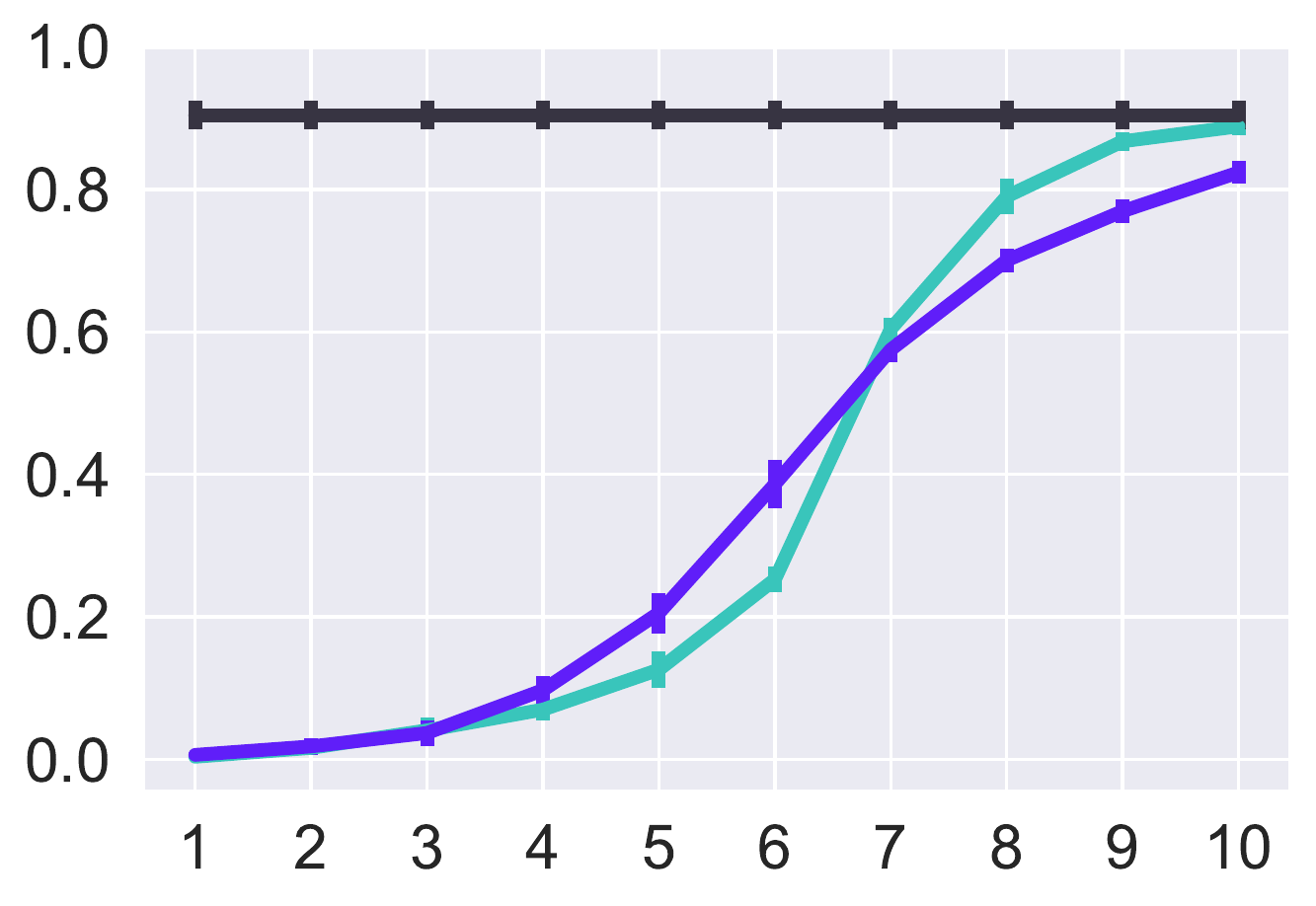}&
\includegraphics[height=\attackheighta]{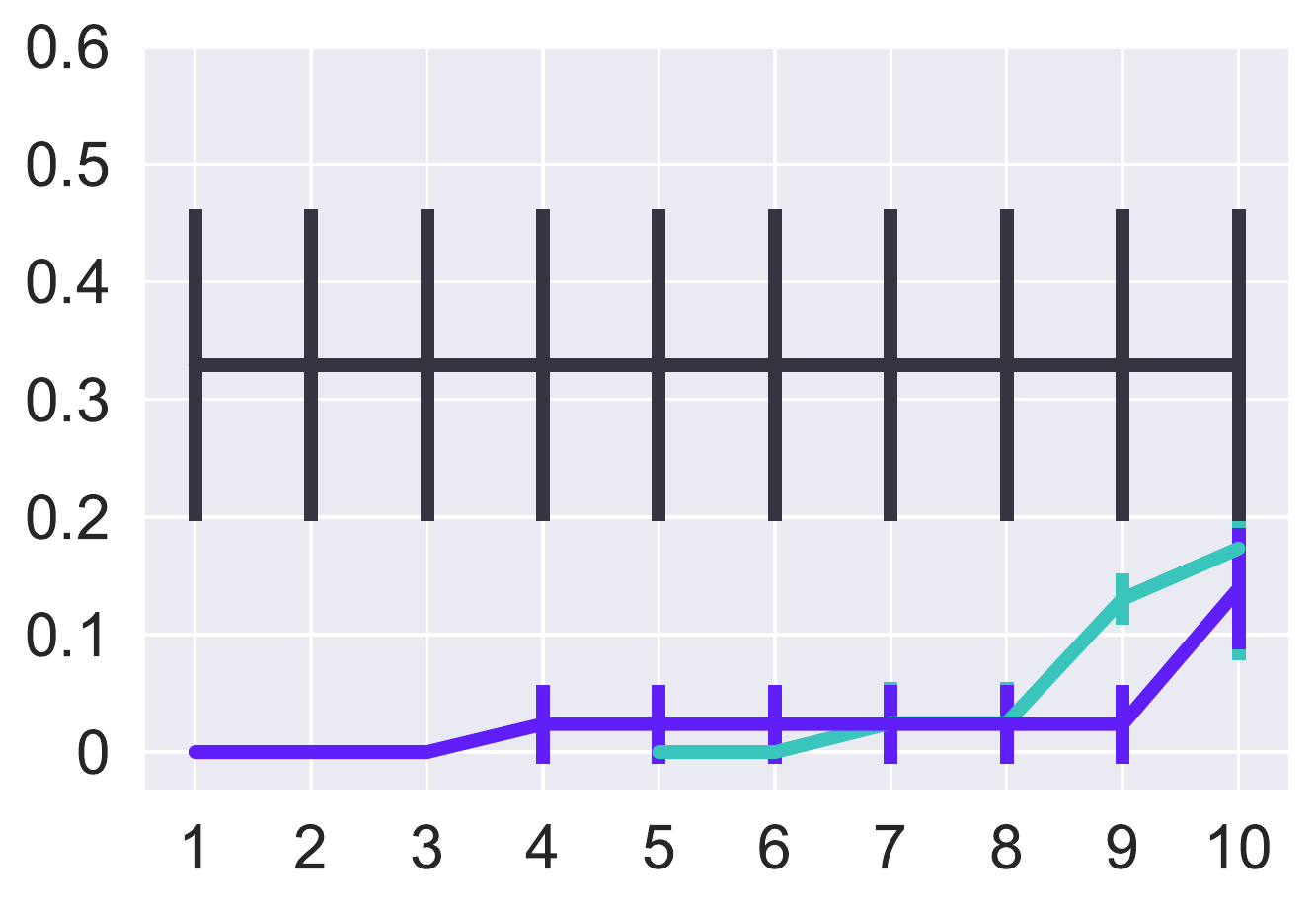}\\[-1.2ex]
\rowname{\quad\makecell{high}}&
\includegraphics[height=\attackheightb]{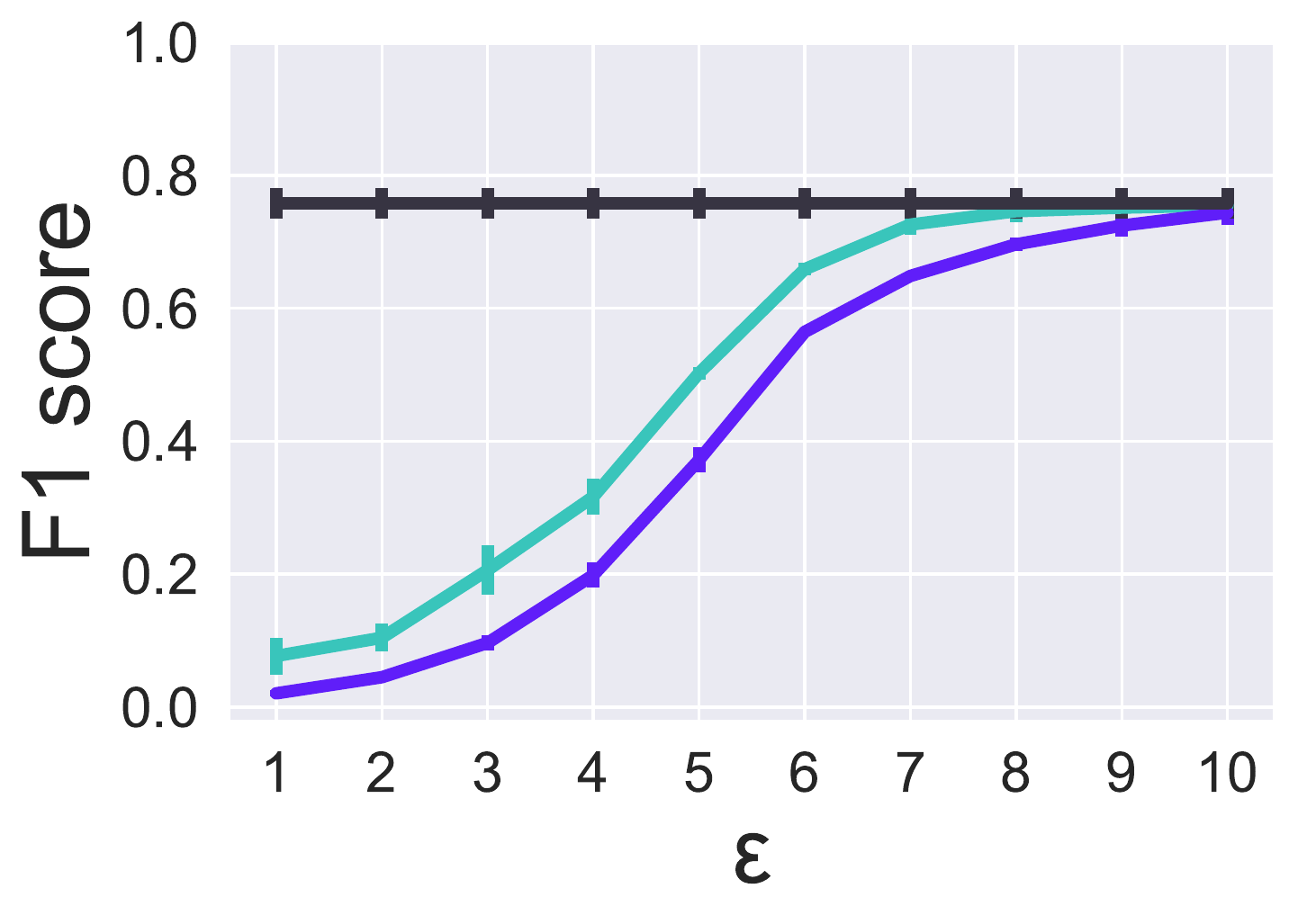}&
\includegraphics[height=\attackheightb]{figures/twitch-DE_high-degree.pdf}&
\includegraphics[height=\attackheightb]{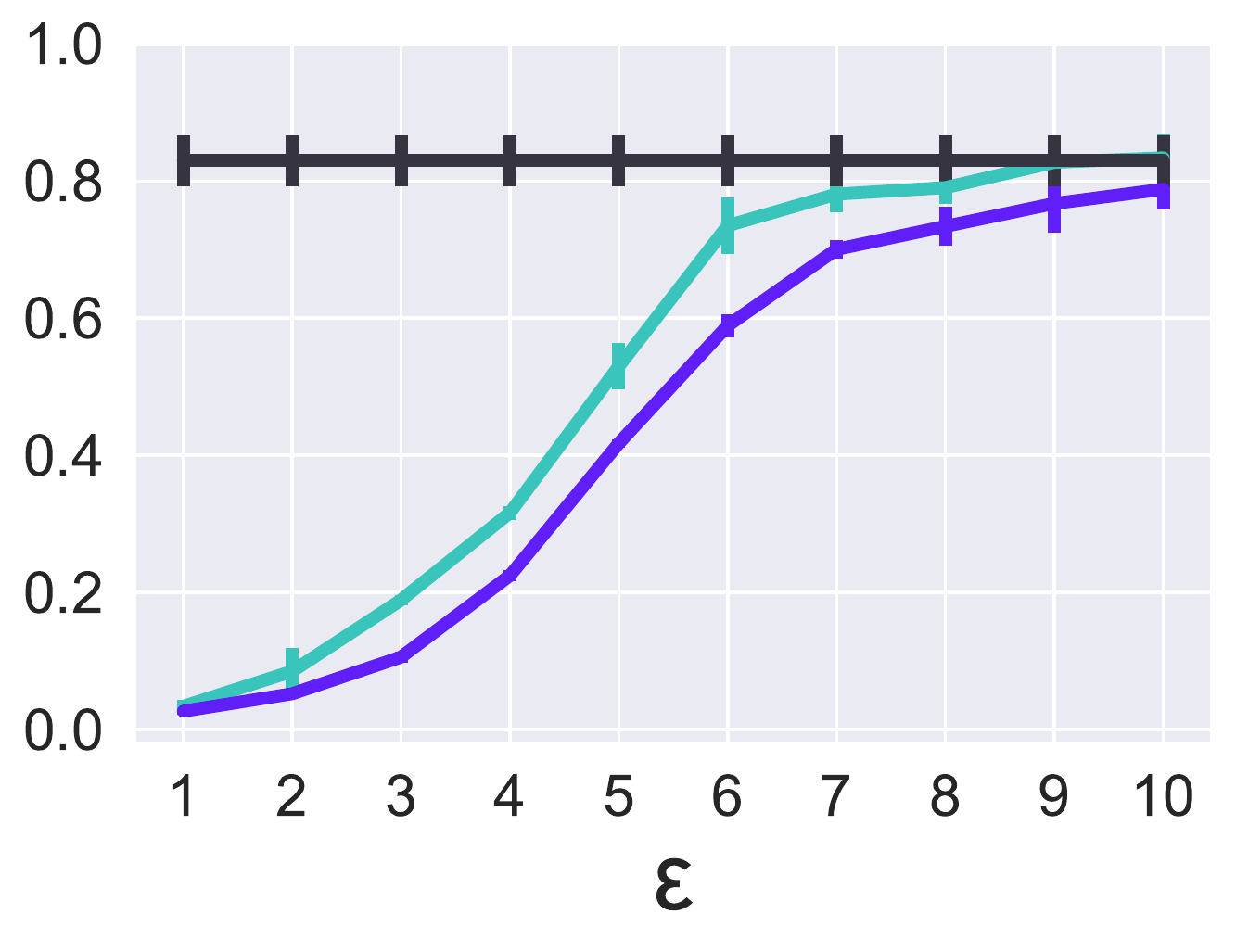}&
\includegraphics[height=\attackheightb]{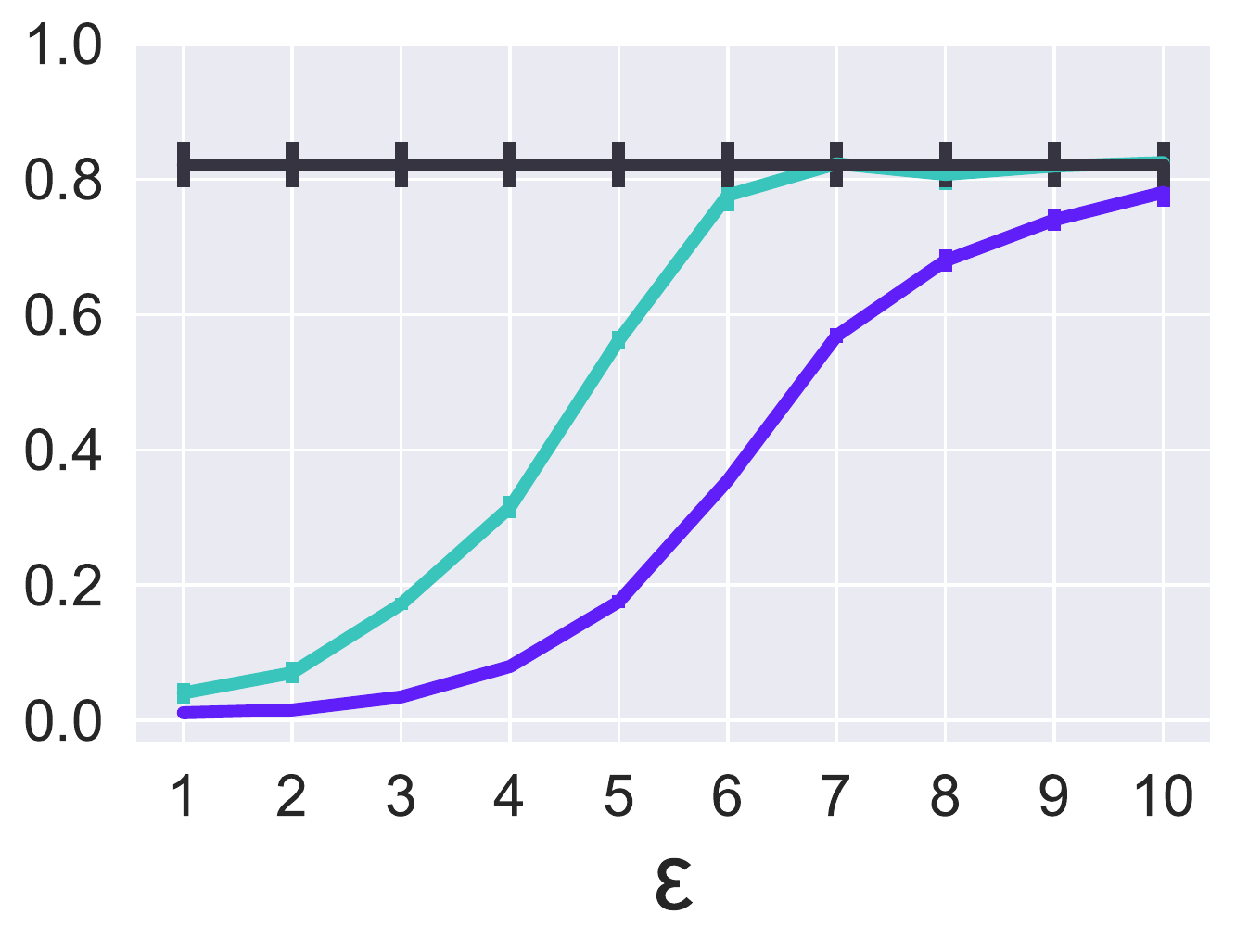}&
\includegraphics[height=\attackheightb]{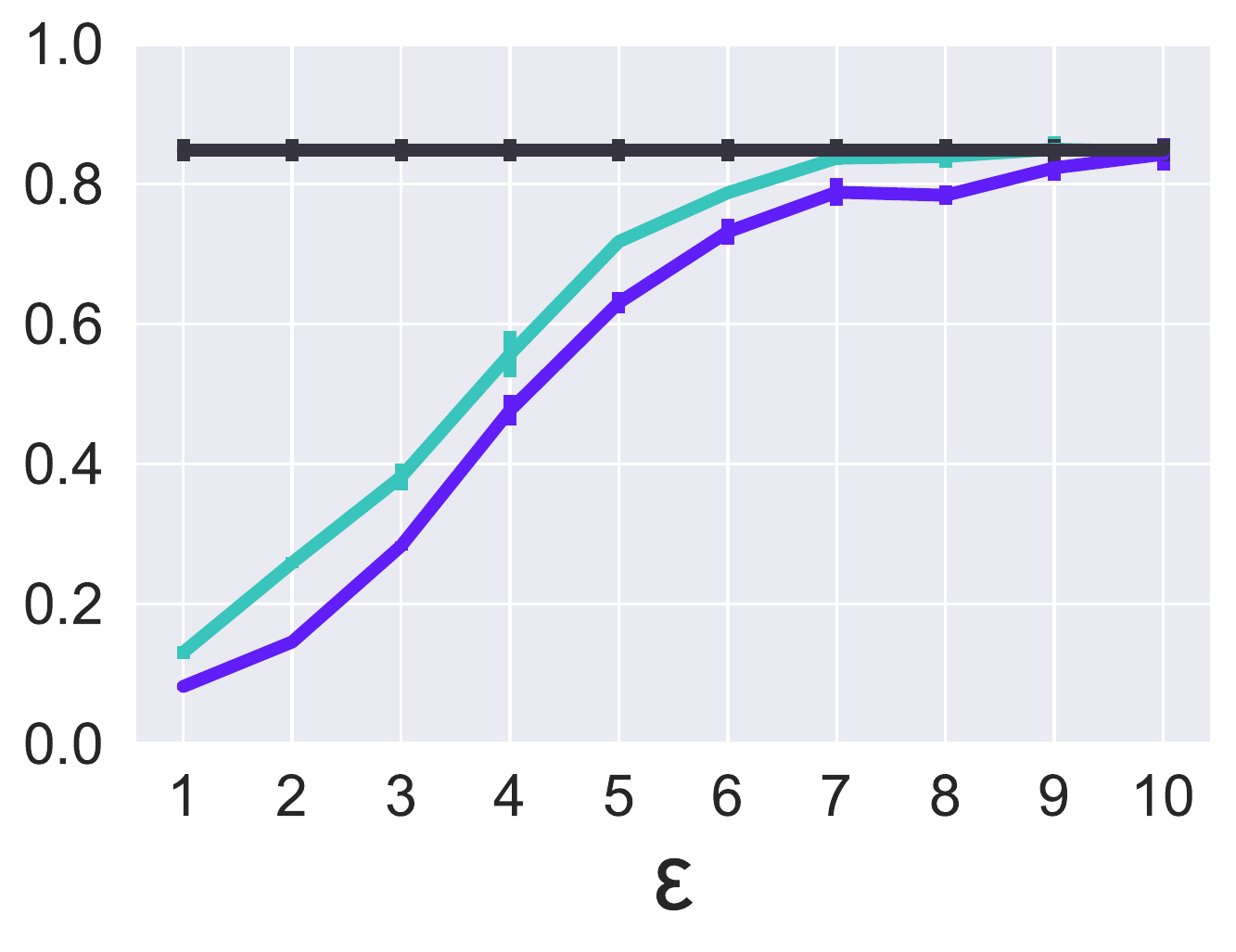}&
\includegraphics[height=\attackheightb]{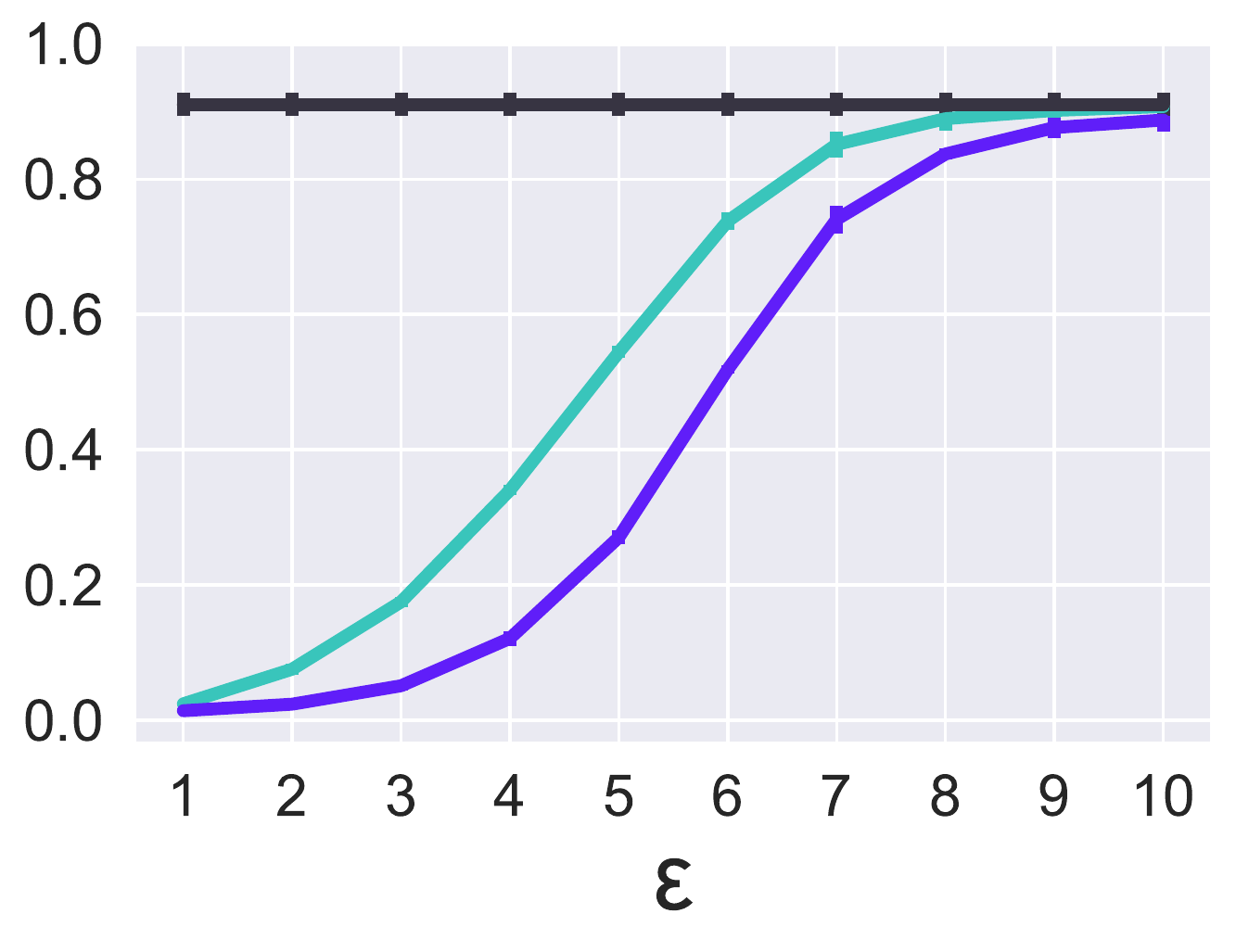}&
\includegraphics[height=\attackheightb]{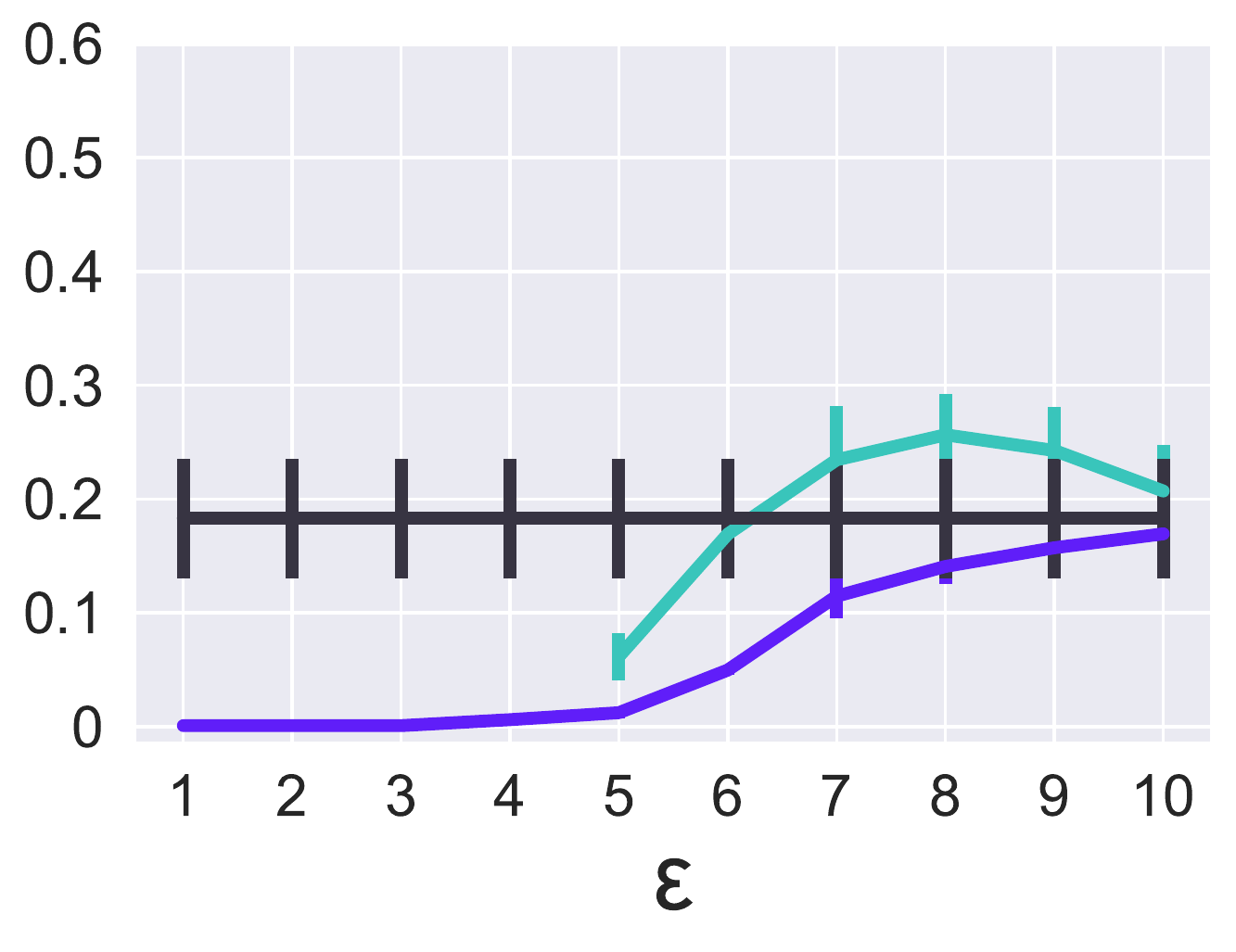}\\[-1.2ex]
\end{tabular}
\caption{Attack effectiveness on different models and node degree distributions (low, unconstrained, and high)}\label{tab:attack-success}
\end{subtable}

}

\vspace{-1mm}
\caption{\small (a) Model utility and (b) attack effectiveness on different models ($\hat k=k$). 
Each column corresponds to a dataset.
We consider four types of models: \edgerand{}, \lapgraph{}, vanilla GCN, and MLP, with the first two satisfying DP guarantees. In each figure, the vertical bar represents the standard deviation.}%
\label{fig:dp-attack}
\end{figure*}

\vspace{-1mm}
{
\renewcommand{\thesubfigure}{\alph{subfigure}}

\begin{figure*}


\newlength{\tempdimab}
\settoheight{\tempdimab}{\includegraphics[width=.138\linewidth]{figures/twitch-DE_low-degree.pdf}}%

\newlength{\attackheightc}
\settoheight{\attackheightc}{\includegraphics[width=.138\linewidth]{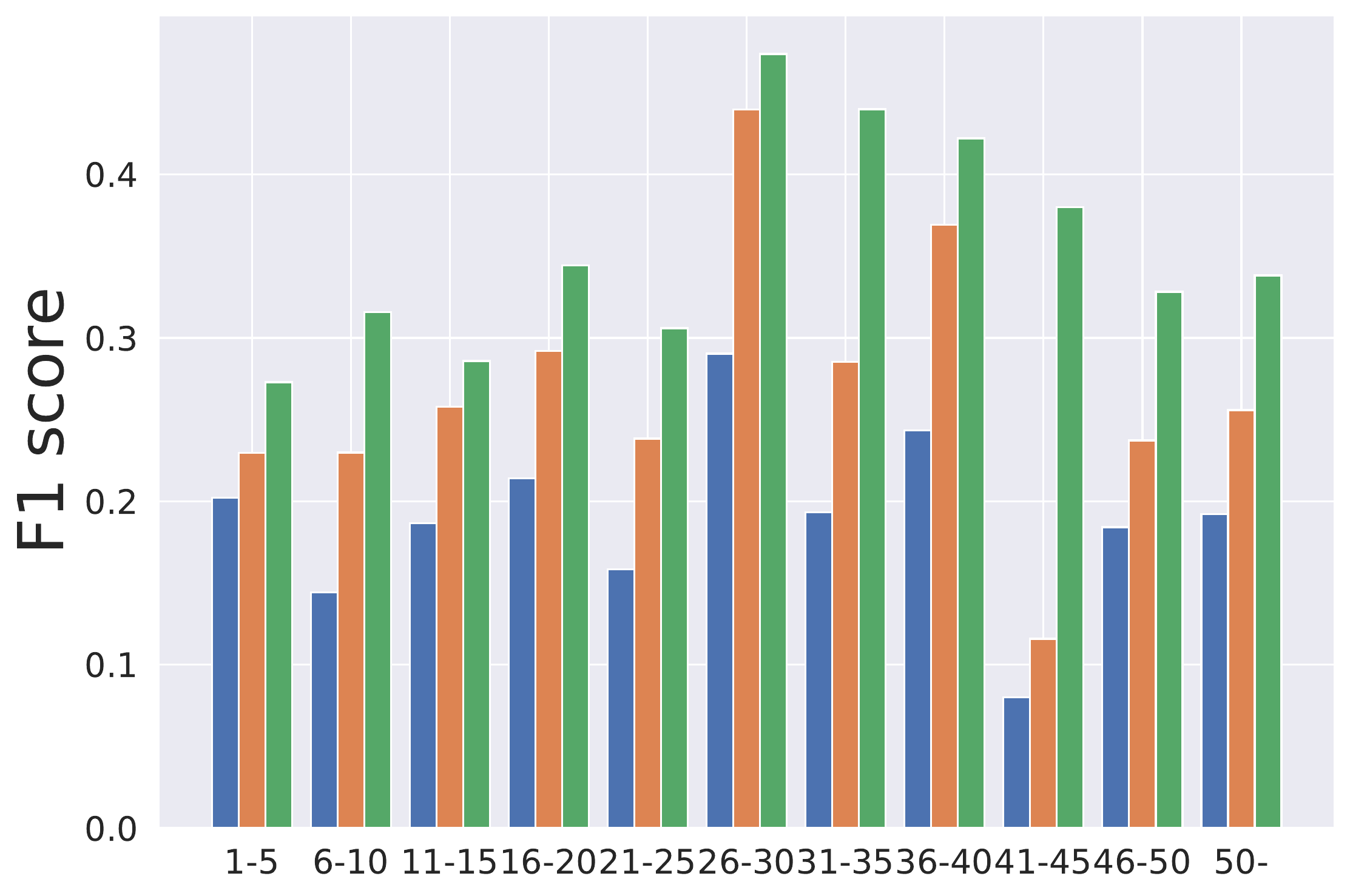}}%


\newlength{\legendheightc}
\setlength{\legendheightc}{0.3\attackheightc}%

\newcommand{\rowname}[1]
{\rotatebox{90}{\makebox[\tempdimab][c]{\footnotesize #1}}}

\centering

{
\renewcommand{\tabcolsep}{10pt}

\begin{subtable}[]{\linewidth}
\begin{tabular}{l}
\includegraphics[height=\legendheightc]{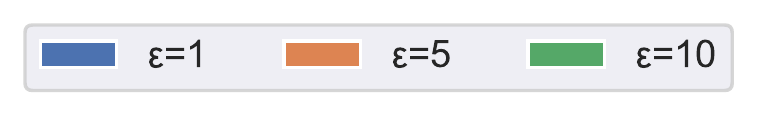}
\end{tabular}
\end{subtable}

\begin{subtable}[]{\linewidth}
\centering
\begin{tabular}{@{}p{5mm}@{}c@{}c@{}c@{}c@{}c@{}c@{}c@{}}
        & \makecell{\footnotesize{twitch-RU}}
        & \makecell{\footnotesize{twitch-DE}}
        & \makecell{\footnotesize{twitch-FR}}
        & \makecell{\footnotesize{twitch-ENGB}}
        & \makecell{\footnotesize{twitch-PTBR}}
        & \makecell{\footnotesize{PPI}}
        & \makecell{\footnotesize{Flickr}}
        \vspace{-1.7pt}\\
\rowname{\makecell{\edgerand{}}}&
\includegraphics[height=\attackheightc]{figures/disc_twitch-RU.pdf}&
\includegraphics[height=\attackheightc]{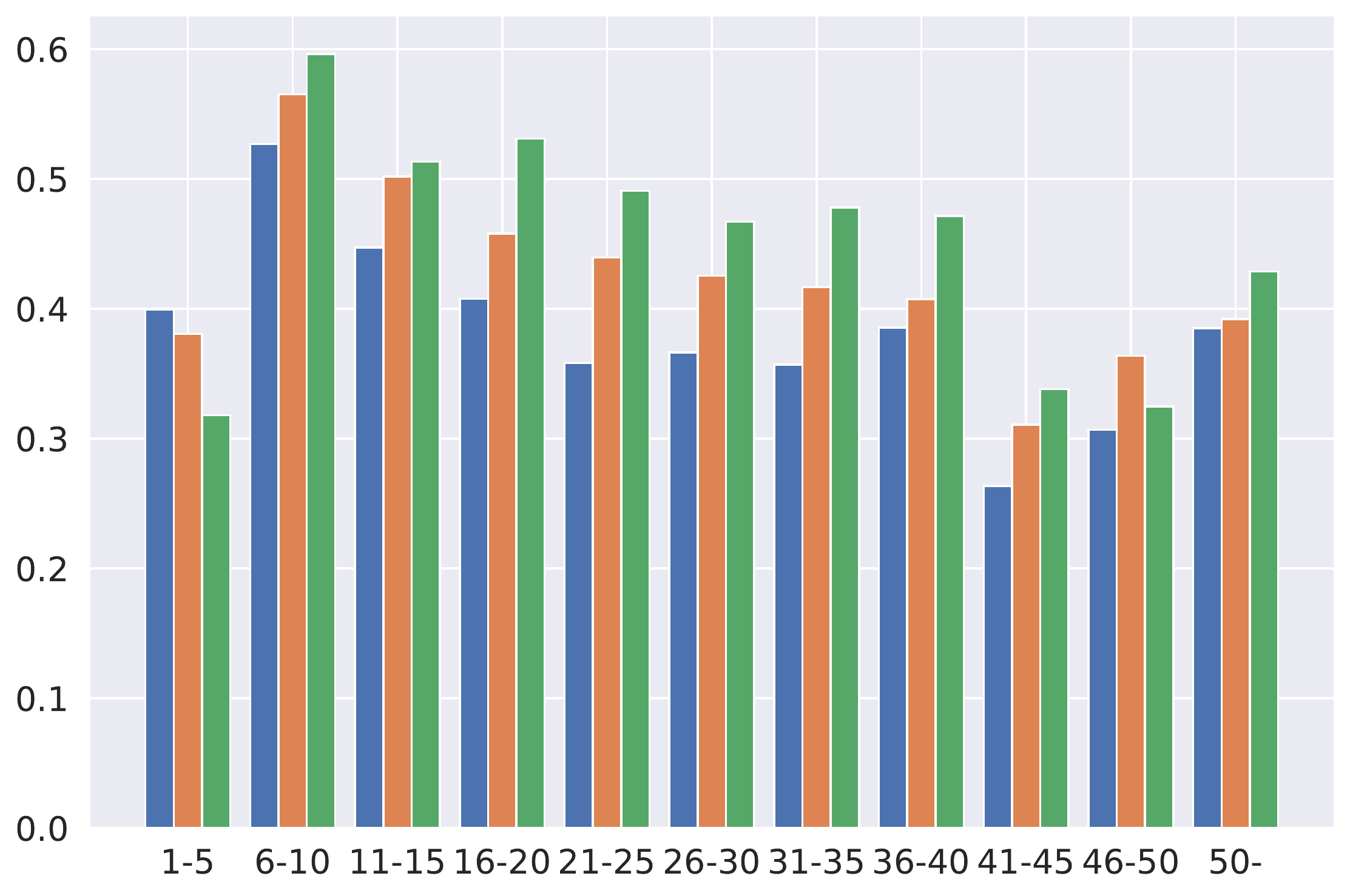}&
\includegraphics[height=\attackheightc]{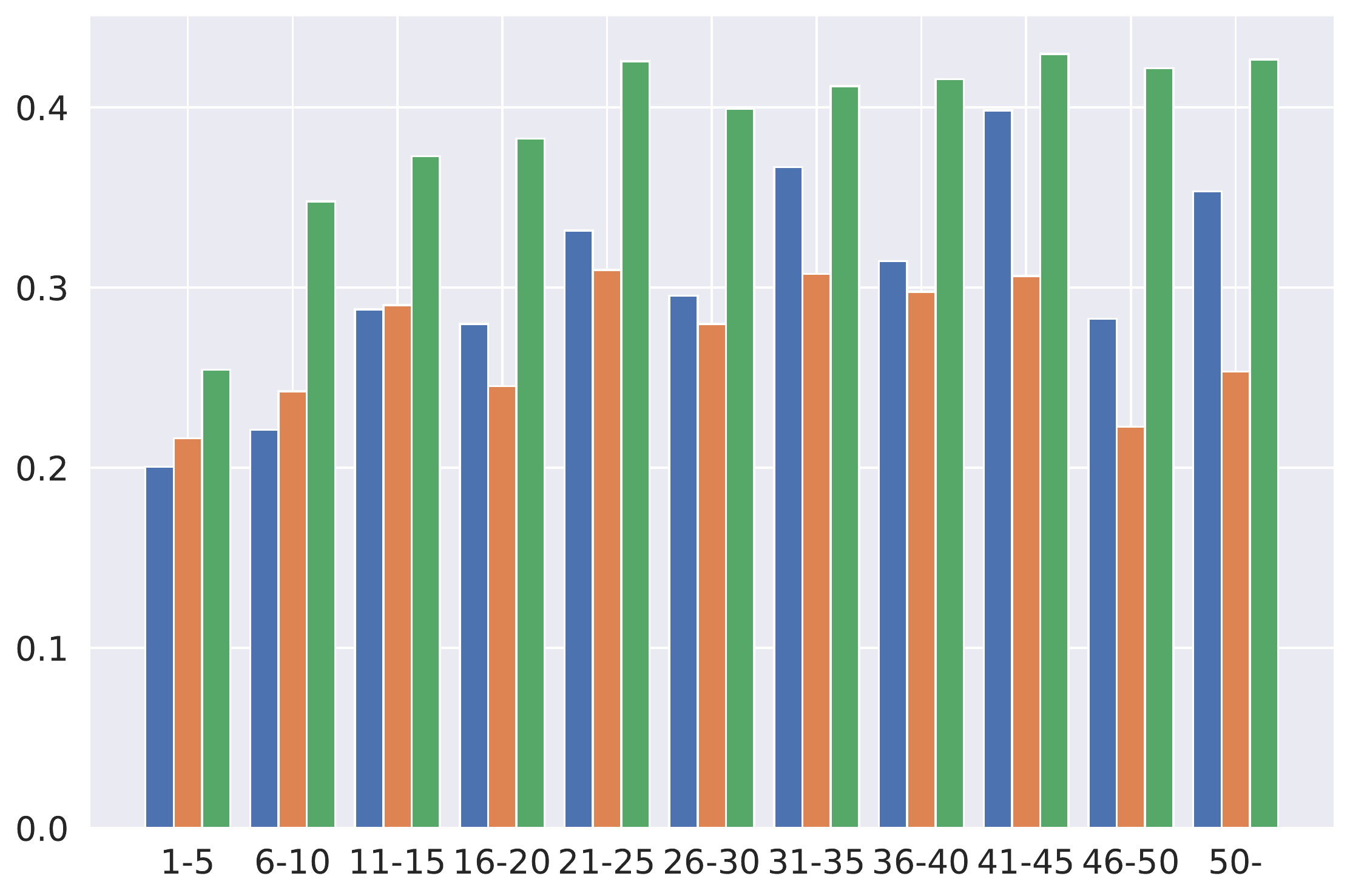}&
\includegraphics[height=\attackheightc]{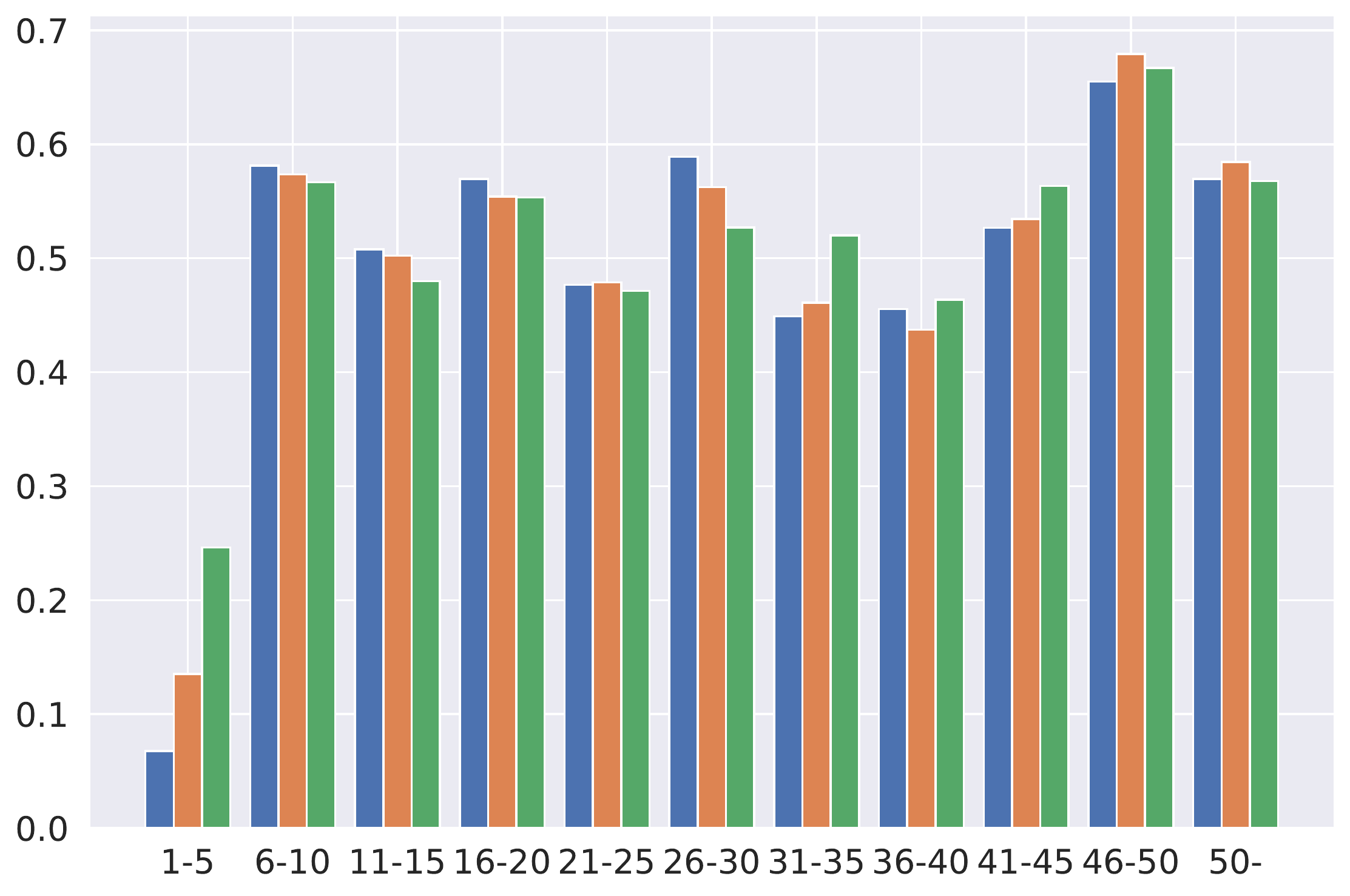}&
\includegraphics[height=\attackheightc]{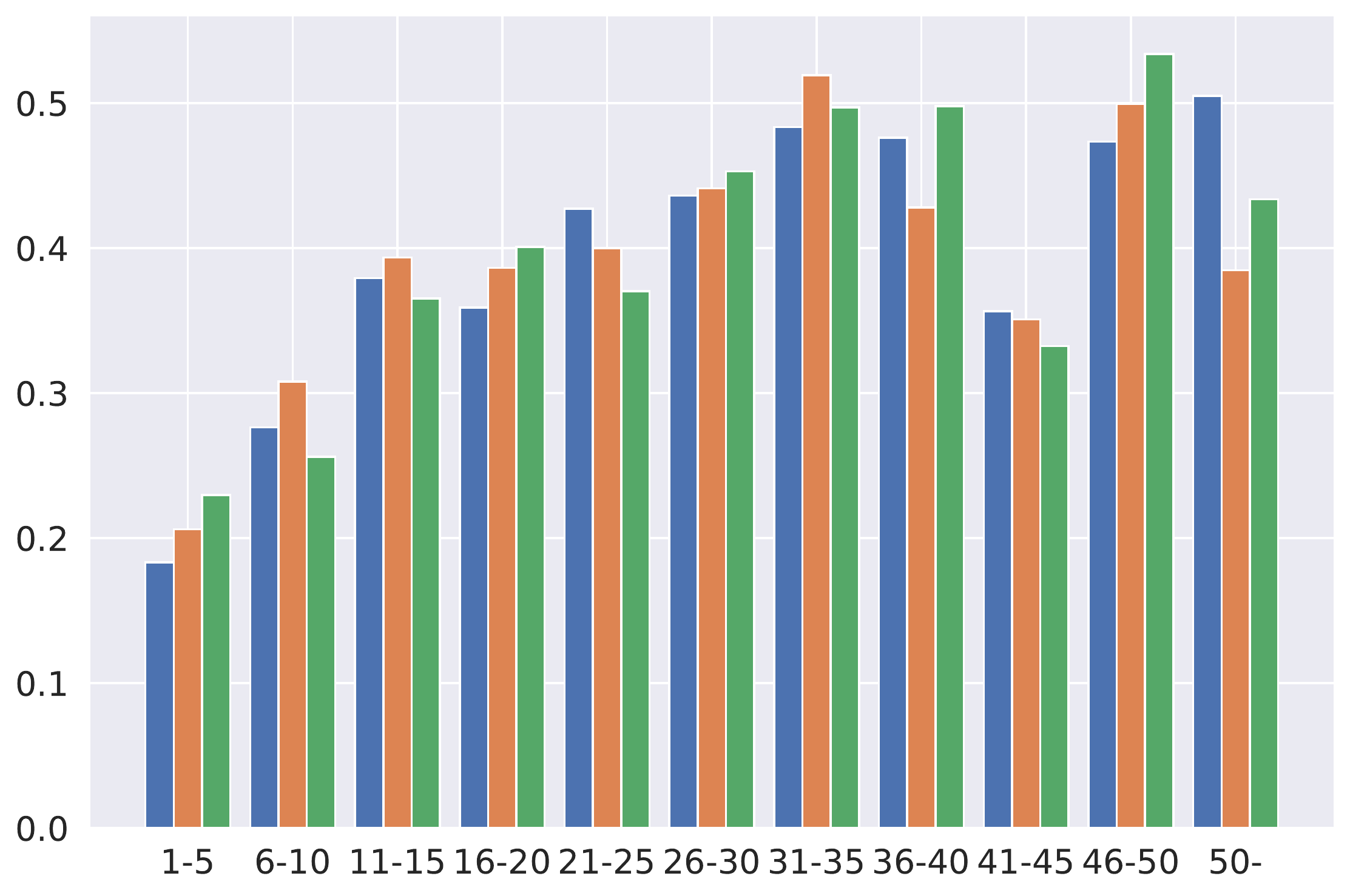}&
\includegraphics[height=\attackheightc]{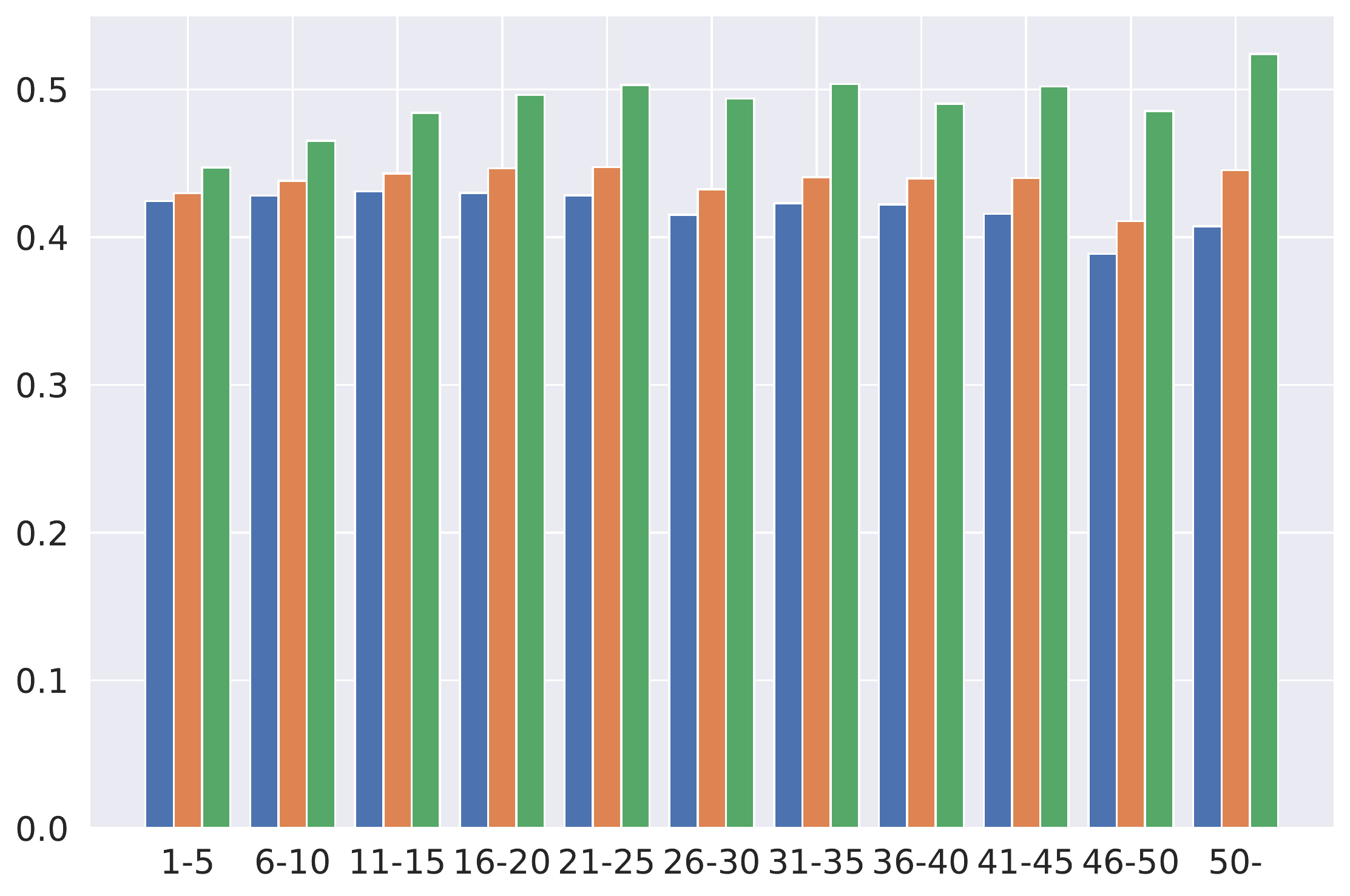}&
\includegraphics[height=\attackheightc]{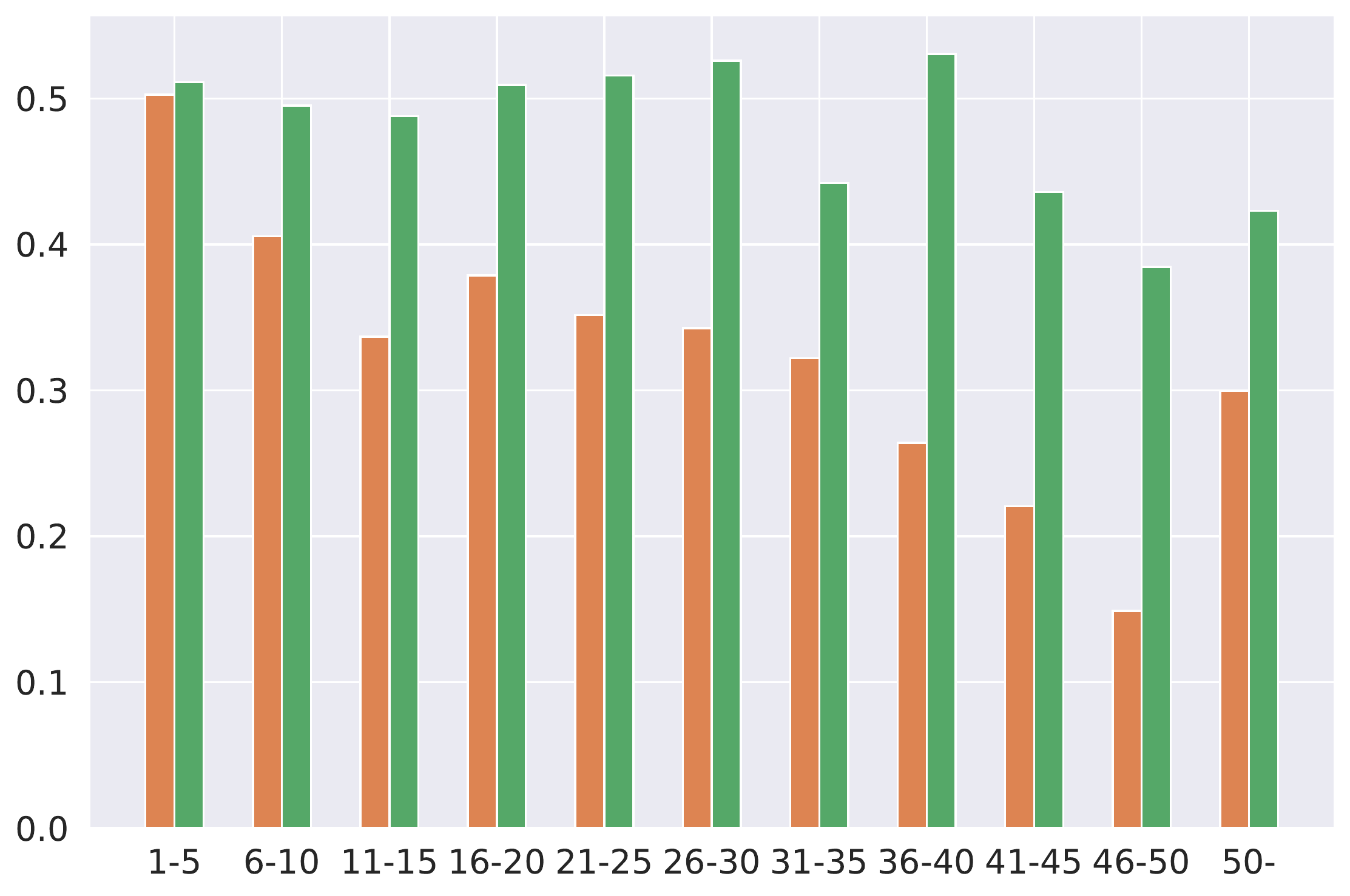}\\[-1.2ex]
\rowname{\makecell{\lapgraph{}}}&
\includegraphics[height=\attackheightc]{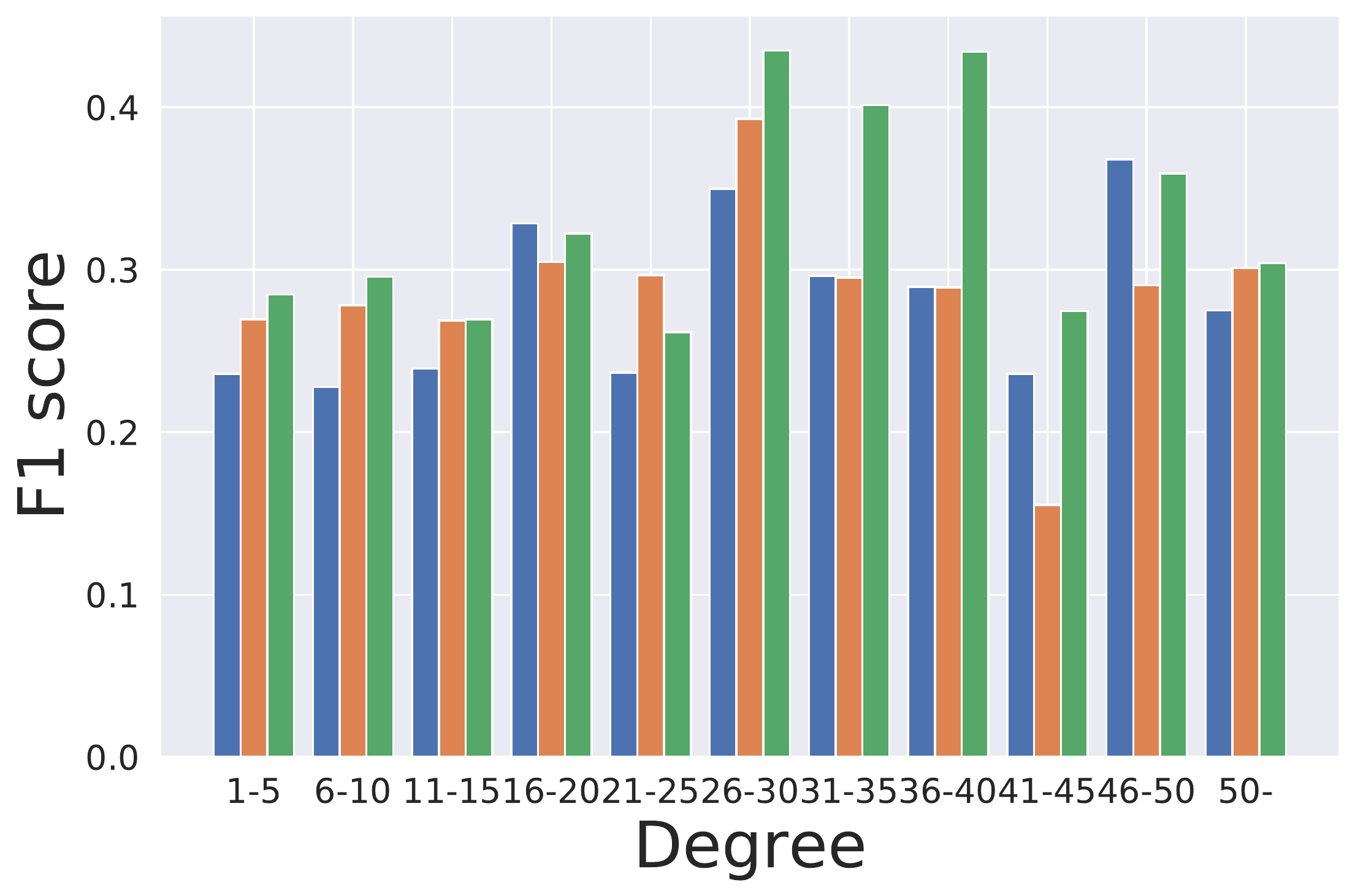}&
\includegraphics[height=\attackheightc]{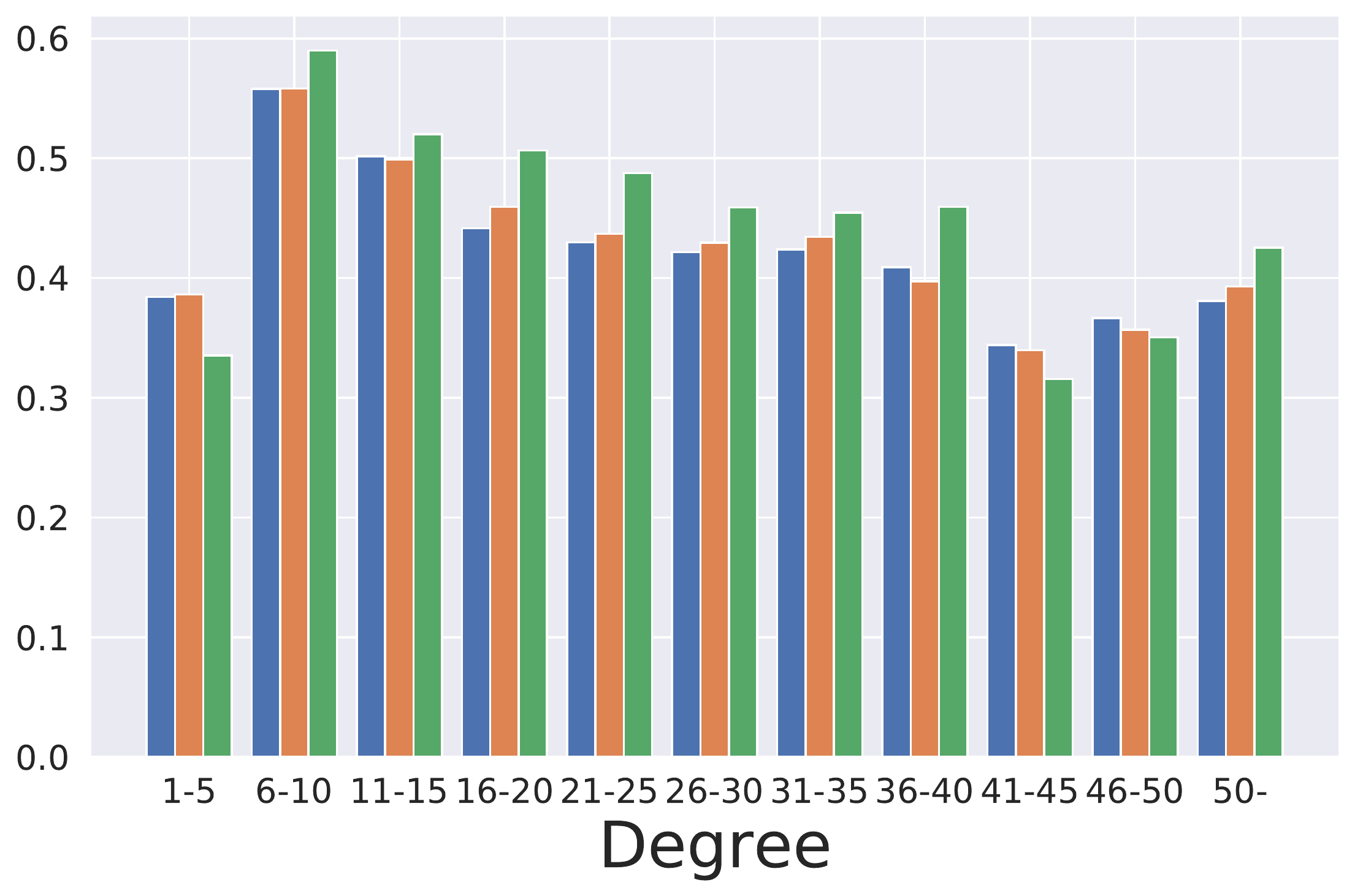}&
\includegraphics[height=\attackheightc]{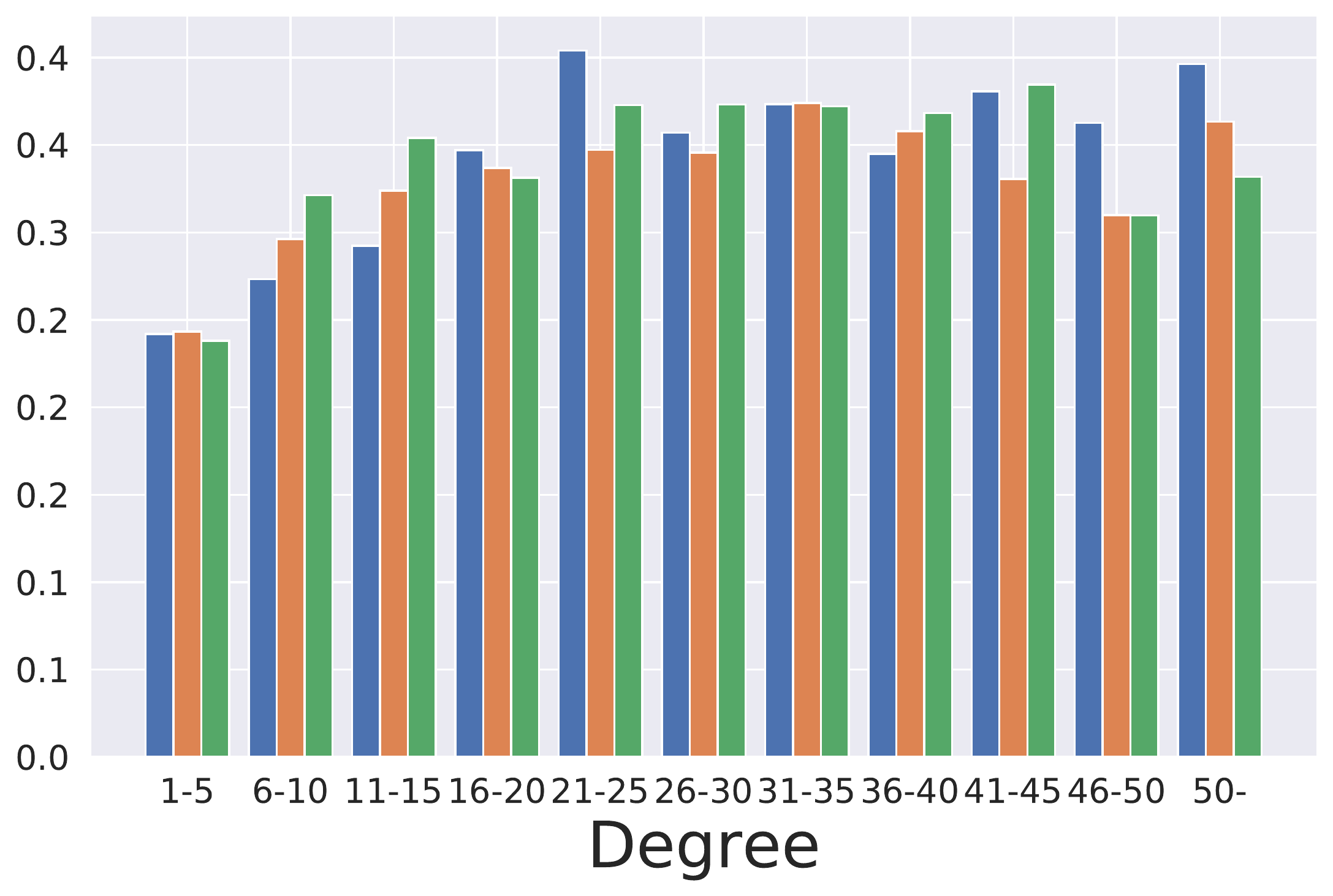}&
\includegraphics[height=\attackheightc]{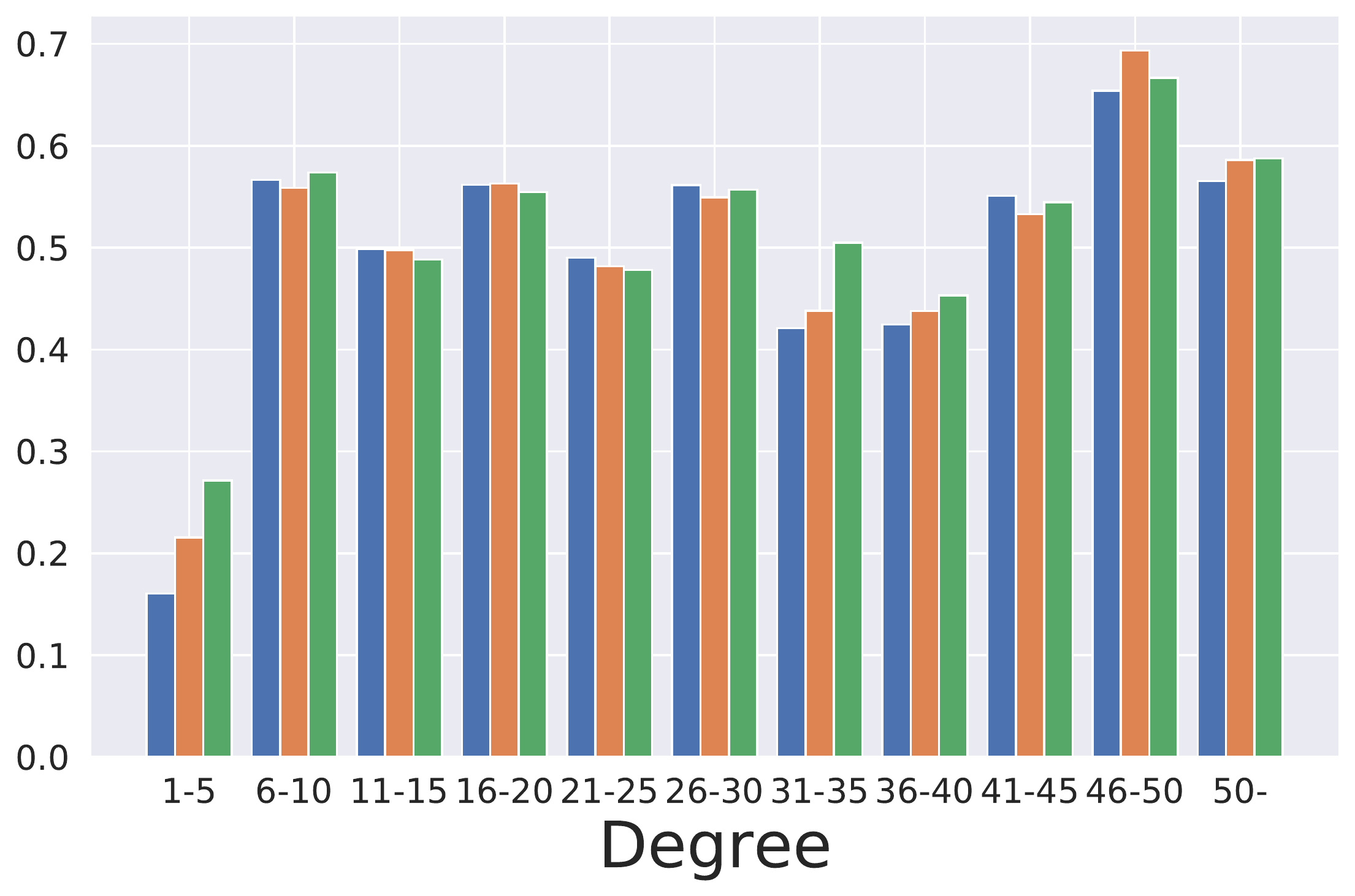}&
\includegraphics[height=\attackheightc]{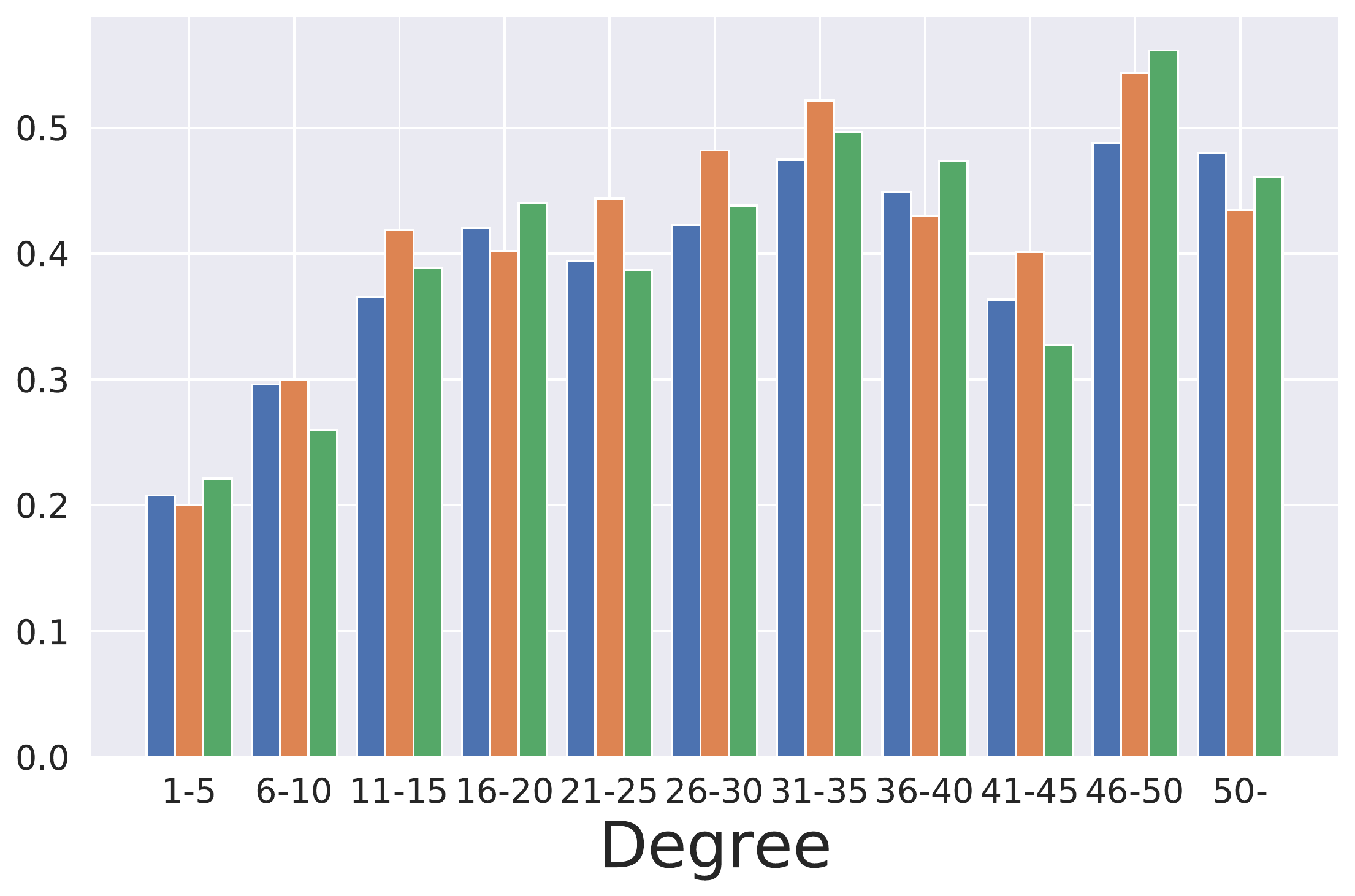}&
\includegraphics[height=\attackheightc]{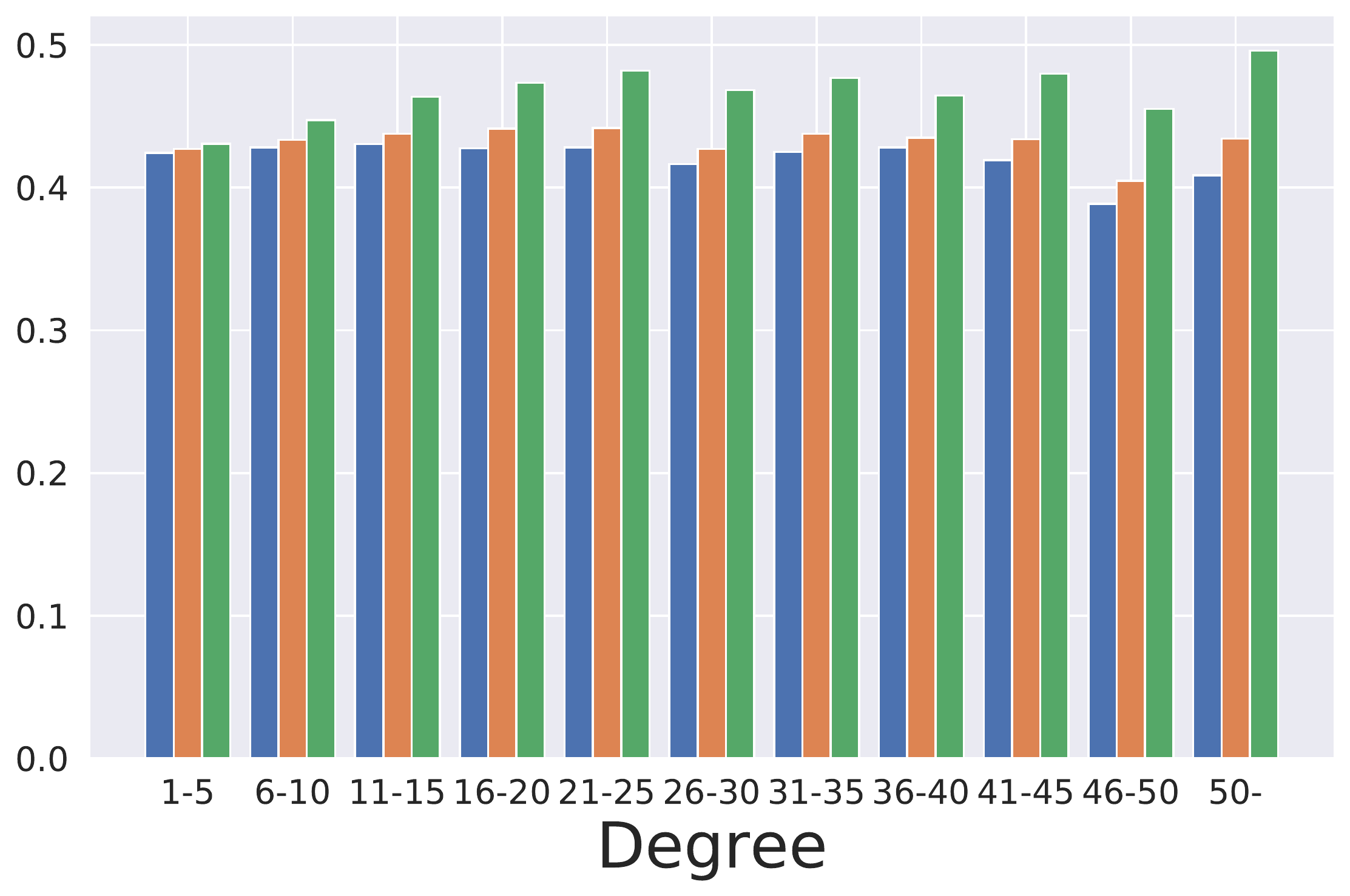}&
\includegraphics[height=\attackheightc]{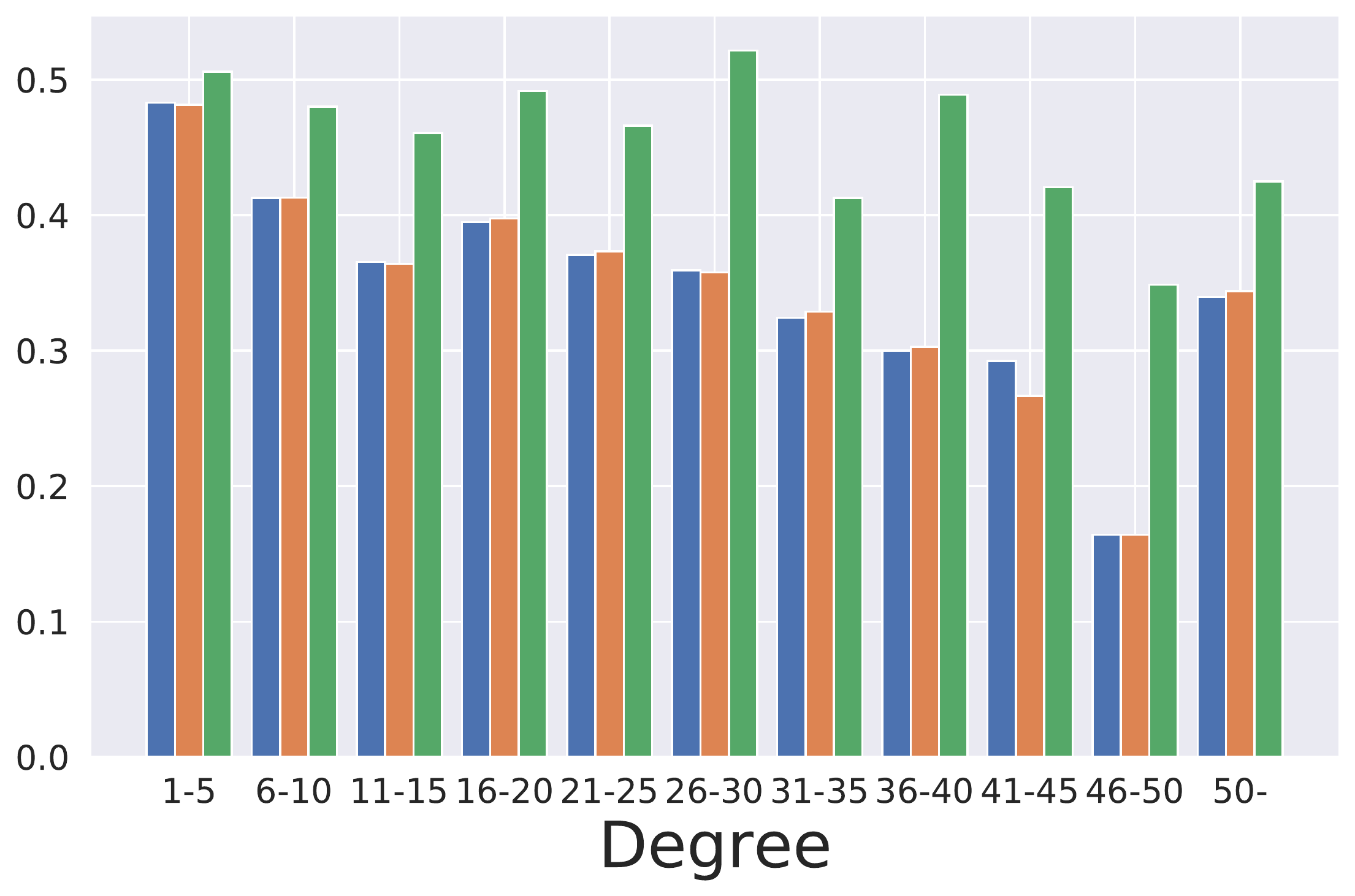}\\[-1.2ex]
\end{tabular}
\end{subtable}

}

\vspace{-1mm}
\caption{\small F1 score of nodes with different degrees under privacy budget $\eps\in\{1,5,10\}$ for two DP mechanisms (\edgerand{} and \lapgraph{}). Each bar group represents a degree range, \eg, 1-5, 6-10, 50-. Each bar within a group corresponds to one privacy budget.}%
\label{fig:dp-degree}
\end{figure*}
}

\subsection{DP GCN against \linkteller}
\label{sec:eval-dp-attack}
\vspace{-1mm}

We validate the effectiveness of \linkteller under various levels of privacy guarantees.
We first provide the experimental setup, followed by concrete results including comparisons of the attack effectiveness of \linkteller on different models.

\subsubsection{Experimental Setup}

We inspect the effectiveness of \linkteller on DP GCN using the same setup as~\Cref{sec:attack-setup}.
Similar to~\Cref{sec:attack-eval}, we use \textit{precision} and \textit{recall} to evaluate the attack.
For each dataset, we consider all combinations of $2$ DP mechanisms (\edgerand and \lapgraph), $10$ privacy budgets ($1.0,2.0,\ldots,10.0$), $3$ sampling strategies (low, unconstrained, and high degree), and $5$ density beliefs ($k/4, k/2, k,2k,4k$). 
The reported result of each scenario is averaged over 3 runs with different random seeds for sampling.


\subsubsection{Evaluation Results}

We leave the full evaluation results for all scenarios 
in~\Cref{sec:append-eval-results} and focus on density belief $\hat k=k$  here. In~\Cref{fig:dp-attack}(b), we plot the \textit{F1 score} of the attack w.r.t. 
DP budget $\eps$.
%
We see that the effectiveness of \linkteller decreases as a result of applying DP. Particularly, the F1 score becomes 
almost 0 when the privacy budget $\eps$ becomes smaller. 
When $\eps$ is large, however, the protection offered by DP is limited. In these cases, \linkteller is able to achieve a success rate close to that of attacking the non-private baseline. 

We also note that the node degree distribution has an impact on the performance of \linkteller.
The trend is clear that as the node degree increases, the attack success rate increases substantially. 
Together with our previous observation in the non-private scenario that the attack success rate does not differ much for varying node degrees, we conclude that DP can offer better protection to low degree nodes than high degree nodes. 
We offer a simple and intuitive explanation as follows.
By the design of \edgerand and \lapgraph,
the perturbations in all cells of the matrix are independent. As a result, nodes of low degree (those incident to fewer edges) are more susceptible to the influence, and therefore better protected by DP.

\subsection{Model Utility Given DP Protection}
\label{sec:eval-dp-util}
\vspace{-1mm}

We next present the evaluation of the model utility, not only to complement the evaluation of the DP GCNs, but also to provide insights about the tradeoff between model utility and robustness against the \linkteller attack. 

\subsubsection{Experimental Setup}

We evaluate the influence of applying DP (\edgerand and \lapgraph) on the utility of the GCN models by comparing the results with two baseline models (a non-private vanilla GCN model and a ``perfectly'' private MLP baseline).
We adopt the same metric to evaluate the utility of all four models:  
F1 score of the rare class for the twitch datasets and micro-averaged F1 score for PPI and Flickr datasets. The rationale is put in~\Cref{sec:append-eval-metrics}.


\subsubsection{Evaluation Results}

The figures for the model utility are presented in~\Cref{fig:dp-attack}(a). We plot the change of the utility with the increase of privacy budget of two DP mechanisms \edgerand{} and \lapgraph{}, as well as the utility of two baseline models independent of the privacy budget.
For each privacy budget $\eps$, the reported results are averaged over 10 runs for different random seeds.
We examine~\Cref{fig:dp-attack}(a) to see how the model utility of DP methods compares with the baselines. 
We first compare the performance of two baseline models: GCN (the black horizontal line) and MLP (the red line).
We note that 
GCN is almost always better than MLP except on the PPI dataset. 
The observation is well suited to our intuition that the knowledge of the graph structure can benefit learning.
For most datasets, the model utility increases with the growth of the privacy budget, since the privacy protection to the graph structure becomes weaker. 
However, \textbf{twitch-ENGB} is an exception that achieves slightly higher utility when more privacy noise is added.
This may be due to the following reason.
For twitch datasets, the model is trained on the graph of twitch-ES and   tested on graphs of other five countries.
When the training and testing graph distribution distance is too large, there is no guarantee that the performance on the training graph can be transferred to the testing graph.
This is the case for twitch-ENGB, which is an extremely sparse graph (see~\Cref{tab:dataset}(a)) compared with the training graph. Thus, with slightly more random noise added, its generalization may be improved.
\m{In addition to the evaluation over a range of $\eps$, we also evaluate such tradeoff of model utility and privacy resiliency by selecting appropriate $\eps$ based on a validation dataset. The detailed setups and results are omitted to~\Cref{sec:append-choice}.}



\vspace{-1mm}
\subsection{Tradeoff between Model Utility and Privacy}
\label{sec:tradeoff}
\vspace{-1.5mm}

We analyze the tradeoff between model utility and the attack performance:
models with high utility tend to be more vulnerable to \linkteller. 
The sweet spot differs across datasets and scenarios. 
Summarizing the observations, we derive a series of conclusions on \textit{how to protect privacy given the gap of the model utility between the vanilla GCN and the MLP model}.
\textbf{First}, if the utility of the vanilla GCN is much higher than the MLP model, then
there is space for performance degradation caused by ensuring privacy. 
We do observe a few cases
where the utility of the DP model is above the MLP baseline, and the attack success rate at that point is relatively low, especially under the low degree case, \eg, the DP model on twitch-RU when $\eps=7$.
In such cases, carefully choosing an $\eps$ will give the practitioner fairly good \textit{utility} and a certain level of \textit{privacy} guarantee simultaneously.
\textbf{Second}, when the performance of the vanilla GCN only exceeds MLP by a small margin, 
almost all DP models that can effectively defend against the attack suffer tremendous utility loss.
We point out that most scenarios fall under this category, where either privacy or utility will be sacrificed. This further substantiates the power of \linkteller.
\textbf{Third}, when the graph structure hurts learning (\eg, PPI), we may avoid using the graph structure in training by using MLP. There might exist other graph neural networks that can achieve better performance on datasets like PPI, and applying 
\linkteller to these models are exciting future work.

\textbf{Utility and privacy of low-degree nodes.}\quad
As noted in~\Cref{sec:eval-dp-attack}, DP GCN offers better protection to nodes of low degree.
A natural question is then: \textit{does better protection imply a degradation of utility of these nodes?}
To answer this question, we separate the nodes into bins by degree (\eg, 1-5, 6-10, $\ldots$, 46-50, 50-), and investigate the F1 score of nodes in each individual bin.
The results on all datasets, two DP mechanisms, with three privacy budgets are presented in~\Cref{fig:dp-degree}.
We can see that the utility for low-degree nodes does not drop faster than high-degree nodes when the privacy budget decreases,
which indicates that DP GCN does not sacrifice the utility of low-degree nodes particularly.

\textbf{Discussion: \edgerand{} or \lapgraph{}.}\quad
\label{sec:discussion}
We further compare the results of the two mechanisms regarding model utility and attack success rate.
When $\eps$ is small, the utility of \edgerand and \lapgraph do not differ much (especially on PPI). 
When $\eps$ is large, \edgerand{} generally has better model utility, while \lapgraph is more robust to \linkteller.
The results for \edgerand are incomplete for the large scale dataset Flickr under tight privacy budgets ($\eps\in\{1,2,3,4\}$) using \edgerand.
Under these cases, the graphs become much denser after perturbation of large magnitudes, and we experience an OOM error using an 11 GB GPU.
In comparison, \lapgraph does not suffer such an issue.




\section{Related Work}
\subsubsection{Privacy Attack on Graphs}
This topic was widely studied~\cite{ZheGet07preserve,hay2007anonymizing,zhang09anonymity,runden12perturb} before graph neural networks came into play.
There are mainly three types of privacy attacks on graphs: identity disclosure, attribute disclosure, and link re-identification~\cite{ZheGet07preserve}, corresponding to different components (nodes, node attributes, and edges) of a graph.
In this paper, we focus on \textit{edge privacy}. Previous endeavors have illustrated 
the feasibility of the link re-identification attack,
whilst 
relying on strong prior knowledge and 
information that arguably might not
always hold or accessible. 
For example, 
when prior knowledge about
the graph is available---\eg, 
nodes with similar attributes or 
predictions are likely connected---He \etal~\cite{he2020stealing} claim that an attacker could infer links 
in the training graph
by applying methods such as 
clustering to predict connections for nodes within the same cluster.
Duddu \textit{et al.}~\cite{duddu2020quantifying}
show that with access to the node embeddings
trained to preserve the graph structure,
one can recover edges by analyzing predictions based on the embeddings.
Apart from the privacy attacks, there exist other adversarial attacks on GNNs, \eg, against node embeddings~\cite{bojchevski2019adversarial} and 
graph- and node-level classifiers~\cite{dai2018adversarial}.
Despite the promising attacks illustrated by these
early endeavors, there is a clear need to 
weaken the assumptions for more reliable, practical settings.
In this paper, we thus answer:  to what extent can we recover 
private edges of a graph by  probing 
a trained blackbox GNN model
\textit{without} strong
prior knowledge? Could we leverage the property of influence propagation among nodes in GNNs to design an effective attack? 

\subsubsection{Differential Privacy for Graphs}

Differential privacy~\cite{dwork14dp} is a notion of privacy that entails that the outputs of the model on neighboring inputs are close. This privacy requirement ends up obscuring the influence of any individual training instance on the model output.
There are a series of works that examine the theoretical guarantee or the practical performance of models under differential privacy guarantees~\cite{li2013membership,Yeom2018PrivacyRI,jayaraman2019evaluating,erlingsson2019that,jayaraman2020revisiting,humphries2020differentially}. 
Depending on the properties of the datasets, \eg, the distinction between the distribution of members and non-members and the underlying correlations within the datasets, there is no trivial answer to this problem.

The extension of differential privacy to the graph setting was first conducted in Hay \etal~\cite{hay2009accurate}.
Since then, there has been extensive research on computing graph statistics such as degree distribution~\cite{hay2009accurate}, cut queries~\cite{blocki2012johnson}, and sub-graph counting queries~\cite{blocki2013differentially} under edge or node differential privacy.
These statistics are useful for graph analysis but insufficient for training a GCN model. Thus, in this paper, to evaluate the strength of the \linkteller attack, we adapt one existing algorithm \edgerand and propose a Laplacian mechanism \lapgraph for training DP GCN models as evaluation baselines. 

\section{Conclusions}

We propose
the first edge re-identification attack \linkteller via influence analysis against GNNs. 
We also evaluate \linkteller against differentially private GNNs trained using an existing  and a proposed DP mechanisms \edgerand and \lapgraph to understand the capability of the attack. 
Extensive experiments on real-world datasets \m{(8 for  inductive  and 3 for  transductive setting)} demonstrate the effectiveness of \linkteller in revealing private edge information, even when there are certain privacy guarantees provided by a DP mechanism.

We believe this work will inspire a range of  future research opportunities 
and lay down a foundation 
for future explorations by providing a clear data isolation problem setup, analysis of edge privacy, together with extensive empirical observations and conclusions.

\noindent\textbf{Acknowledgement.}
This work is partially supported by the
NSF grant No.1910100, NSF CNS 20-46726 CAR, NSF TRASE (ECCS-2020289), and Amazon Research Award.

\newpage

\bibliographystyle{IEEEtran}
\bibliography{references}


\appendix
\vspace{-1mm}
\subsection{Proofs for Influence Analysis in \linkteller}
\vspace{-1mm}
\subsubsection{Proof of \texorpdfstring{\Cref{thm:1-layer-gcn}}{Proposition 1}}
\label{sec:append-proof-thm-1-layer-gcn}
\begin{proof}

Consider the $1$-layer GCN, where
\begin{small}
\begin{align}
    \label{eq:gcn-0}
    \mbox{GCN}(A,X,W)=AXW.
\end{align}
\end{small}
For the simplicity of the proof, we ignore the normalization applied to the adjacency matrix $A$, since it only changes the scale of numbers in the matrix.
We next calculate the influence value of a pair of nodes against this $1$-layer GCN
according to the function \texttt{InfluenceValue} described in~\Cref{alg:attack}.

Following the notation in~\Cref{alg:attack}, let the set of inference nodes be $V^{(I)}$, the node feature matrix associated with the inference node set be $X$. We further denote the adjacency matrix induced on this node set as $A$.
Taking a pair of nodes $u,v$ from the set of nodes of interest $V^{(C)}$, we calculate the influence value of this pair of nodes following the steps in \texttt{InfluenceMatrix}:
\begin{small}
\begin{align*}
    P&=G_{BB}(V^{I},X)={\rm GCN}(A,X,W)=AXW,\\
    X^{\prime}&=\left[x_{1}^\top, \ldots,(1+\Delta) x_{v}^\top, \ldots,\right]^\top,\\
    P'&=G_{BB}(V^{I},X')={\rm GCN}(A,X',W)=AX'W,\\
    P'-P&=A(X'-X)W=AX_{\Delta}W,
\end{align*}
\end{small}
where $X_{\Delta}$ is a matrix of the same size as $X$ that only contains value in each $v$-th row. Specifically, the $v$-th row vector of $X_{\Delta}$ is equal to $\Delta x_v$, where $\Delta$ is the reweighting coefficient.

We next compute the $u$-th row of the influence matrix $I=AX_{\Delta}W$. We start from computing $X_{\Delta}W$
\begin{small}
\begin{align*}
    X_{\Delta}W 
    = \left[\bf 0,\ldots,\Delta x_v^\top,\ldots,\bf 0 \right]^\top W
    = \Delta \left[\bf 0,\ldots,x_v^\top,\ldots,\bf 0 \right]^\top W
\end{align*}
\end{small}
Therefore, for $I$ defined as $I=(P'-P)/\Delta=AX_\Delta W$, its $u$-th row is 
$A_{uv}x_vW$.
When there is no edge between $u$ and $v$, $A_{uv}=0$, and therefore the row vector is an all zero vector. The influence value, which is the $\ell_2$ norm of the row vector, is therefore $0$.
\end{proof}

\subsubsection{Proof of \texorpdfstring{\Cref{thm:k-layer-gcn}}{Theorem 1}}
\label{sec:append-proof-thm-k-layer-gcn}

We first present a lemma.
\begin{lemma}
\label{lem:iter}
    Let $A\in \{0,1\}^{n\times n}$ be the adjacency matrix of the graph, $H_i,H_i'\in \bb R^{n\times d_i}$ be two hidden feature matrices of size
    that differ in $t_i$ rows $\{r_1,\ldots, r_{t_i}\}$ corresponding to the nodes $\{v_{r_1},\ldots, v_{r_{t_i}}\}$, and $W^i\in \bb R^{d_i\times d_{i+1}}$ be the weight matrix of the $i$-th graph convolutional layer,
    then 
    \begin{enumerate}[(1)]
        \item $AH_iW^i$ and $AH_i'W^i$ differ in 
    at most $t_{i+1}$ rows corresponding to the node set $\bigcup_{l=1}^{t_i}\cal N(v_{r_l})$, where $\cal N(u)$ denotes the neighbor set of node $u$.
        \item Further, let $H_{i+1}=\sigma(AH_iW^i)$ and $H_{i+1}'=\sigma(AH_i'W^i)$,
    then 
    $H_{i+1}$ and $H_{i+1}'$ also differ in at most $t_{i+1}$ rows corresponding to the node set $\bigcup_{l=1}^{t_i}\cal N(v_{r_l})$.
    \end{enumerate}
\end{lemma}
We next use~\Cref{lem:iter} to help with the proof of~\Cref{thm:k-layer-gcn}.
\vspace{-\topsep}
\begin{proof}
Consider a $k$-layer GCN which is a stack of $k$ graph convolutional layers defined as below
\begin{small}
\begin{align}
    \label{eq:gcn-1}
    \mbox{GCN}(A,X, \{W^i\})=A\cdots \sigma(A\sigma(AX{W}^1){W}^2)\cdots {W}^k.
\end{align}
\end{small}
Its \textit{input} feature matrices are $H_0=X$ and $H_0'=X'$ that differ in one row; let it correspond to node $u$.
Its \textit{output} feature matrices are $AH_{k-1}W^{k-1}$ and $AH_{k-1}'W^{k-1}$.

Since $H_0$ and $H_0'$ differ only in node $u$, according to~\Cref{lem:iter}-(2), $H_1$ and $H_1'$ differ in the rows corresponding to $\cal N(u)$ (which contains nodes that are $1$ hop away from $u$);
$H_2$ and $H_2'$ differ in the rows corresponding to $\bigcup_{v\in\cal N(u)}\cal N(v)$ (which contains nodes that are at most $2$ hops away from $u$).
Iteratively applying~\Cref{lem:iter}-(2), we see that $H_{k-1}$ and $H_{k-1}'$ differ in rows corresponding to nodes that are at most $k-1$ hops away from $u$.
We finally apply ~\Cref{lem:iter}-(1) and obtain the conclusion that 
$AH_{k-1}W^{k-1}$ and $AH_{k-1}'W^{k-1}$ differ in rows corresponding to nodes that are at most $k$ hops away from $u$.
Thus, the influence matrix 
$$P_\Delta =AH_{k-1}'W^{k-1} - AH_{k-1}W^{k-1}$$
has at most $t_k$ non-zero rows corresponding to nodes that are at most $k$ hops away from $u$.
It thus follows that when $u$ and $v$ are at least $k+1$ hops away, the $v$-th row of the influence matrix of node $u$ is an all-zero row.
Since the influence value of $u$ on $v$ is the norm of the $v$-th row, we thus can conclude that the \textit{influence value} $\norm{i_{v_u}}=0$.
\end{proof}
Finally, we complete the proof for~\Cref{lem:iter}.
\vspace{-\topsep}
\begin{proof}
    First of all, for $H_i$ and $H_i'$ that differs in $t_i$ rows, it is obvious that $H_iW^i$ and $H_i'W^i$ differ in the same $t_i$ rows. For simplicity, we denote $H_iW^i$ as $F_i$ and $H_i'W^i$ as $F_i'$. 
    We next present the condition for $AF_i$ and $AF_i'$ to differ. Consider the element in $j$-th row and $k$-th column of $AF_i$ and $AF_i'$, which are 
    $
    \sum_{p=1}^{n} A_{jp} (F_i)_{pk}
    $
    and
    $
    \sum_{p=1}^{n} A_{jp} (F_i)'_{pk}
    $, respectively. 
    We first note that when $A_{jp}=0$, the difference of $(F_i)_{pk}$ and $(F_i)_{pk}'$ does not matter.
    Next, for all $p$ such that $A_{jp}=1$, only when $(F_i)_{pk}\neq (F_i)_{pk}'$ will the difference of the product contribute to the difference of the sum.
    The two points jointly imply that, 
    if $j\notin \bigcup_{l=1}^{t_i}\cal N(v_{r_l})$, then $(F_i)_{pk}=(F_i)_{pk}'$ for all $k$.
    Thus, $AF_i$ and $AF_i'$ differ in at most $t_{i+1}$ rows, corresponding to the node set $\bigcup_{l=1}^{t_i}\cal N(v_{r_l})$. Hence we establish the first claim.
    
    For the second claim, we notice that the activation layer such as ReLU used in the standard GCN~\cite{Kipf2016gcn} is point-wise operation. Thus $\sigma(AF_i)$ and $\sigma(AF_i')$ cannot differ in more rows, meaning $H_{i+1}$ and $H_{i+1}'$ will also differ at most in the rows corresponding to $\bigcup_{l=1}^{t_i}\cal N(v_{r_l})$.
\end{proof}

\vspace{-1mm}
\subsection{Proofs for Privacy Guarantees of the DP mechanisms}
\label{sec-append-dp-proof}
\vspace{-1mm}
\subsubsection{Proof of \texorpdfstring{\Cref{thm:DP GCN}}{Theorem 2}}
\vspace{-0.2em}
\begin{proof}
Since $\widetilde A_{V^{(T)}}$ is perturbed to meet \eedgedp and other inputs (\eg, node features) in~\Cref{alg:dpgcn} are independent of the graph structure, the DP GCN model is \Eedgedp~due to the post-processing property of differential privacy.
\end{proof}
\vspace{-0.2em}

\subsubsection{Proof of \texorpdfstring{\Cref{lemma:parallel_composition}}{Lemma 1}}
\vspace{-0.2em}
\begin{proof}
\Cref{lemma:parallel_composition} extends the parallel composition property of differential privacy~\cite{mcsherry2009privacy} to \edgedp. The parallel composition property states that if $M_1, M_2, \dots, M_m$ are algorithms that access disjoint datasets $D_1, D_2, \dots, D_m$ such that each $M_i$ satisfies $\varepsilon_i$-differential privacy, then the combination of their outputs satisfies $\varepsilon$-differential privacy with $\varepsilon = \max( \varepsilon_1, \varepsilon_2, \dots, \varepsilon_m)$. 
Since the adjacency matrices $A_1, A_2, \dots, A_m$ have non-overlapping edges, they could be viewed as disjoint datasets of edges. Thus, the combination of $M_{\varepsilon}(A_1), M_{\varepsilon}(A_2), \dots, M_{\varepsilon}(A_m)$ is \Eedgedp.
\end{proof}
\vspace{-0.2em}
\subsubsection{Proof of \texorpdfstring{\Cref{thm:dp-infer}}{Theorem 3}}

\begin{proof}
\vspace{-0.2em}
Under the transductive setting, since the same perturbed matrix $\widetilde A_{V^(T)}$ is used in both training and inference, the inference step does not leak extra graph structure information other than that in the DP GCN model. Therefore, the inference step is \Eedgedp~due to the post-processing property of differential privacy. Under the inductive setting, the perturbation method $M_\varepsilon$ is applied to both $A_{V^{(T)}}$ and $A_{V^{(I)}}$. Since $A_{V^{(T)}}$ and $A_{V^{(I)}}$ contain non-overlapping sets of edges, \Cref{lemma:parallel_composition} guarantees \eedgedp of the \texttt{Inference} step.
\end{proof}
\vspace{-0.2em}

\subsubsection{Proof of \texorpdfstring{\Cref{thm:dppig}}{Theorem 4}}
\vspace{-0.2em}
\begin{proof}
    By definition, \edgerand is \Eedgedp iff. for all symmetric matrices $A'\sim A\in\cal A$ and all subsets $S\subseteq\cal A$,~\Cref{eq:dp} holds. 

    We first note that $\cal M$ operates on each cell $A_{ij}$ independently.
    Therefore, the probability of perturbing the matrix $A$ to get a certain output is the product of the probability of perturbing each cell in $A$ to match the corresponding cell in the desired output.
    Let $e=(u,v)$ be the only differing edge between $A$ and $A'$. For all $(i,j)\neq(u,v)$, $A_{ij}=A'_{ij}$, so the probability of perturbing $A_{ij}$ and $A'_{ij}$ into the same value is the same. 
    For $(u,v)$, however, getting the same outcome means that one of $A_{u,v}$ and $A'_{uv}$ is changed through perturbation while the other remains unchanged.
    The probability of the state change, according to~\Cref{alg:pig}, is $s/2$. Putting the statements together, we derive
    \begin{small}
    \begin{align*}
        \frac{\Pr[\cal M(A)\in S]}{\Pr[\cal M(A')\in S]}
        =\prod_{i,j}\frac{\Pr[\cal M(A_{ij})\in S_{ij}]}{\Pr[\cal M(A'_{ij})\in S_{ij}]}
        &=\frac{\Pr[\cal M(A_{uv})\in S_{uv}]}{\Pr[\cal M(A'_{uv})\in S_{uv}]}\\
        &\leq \frac{1-s/2}{s/2}\leq \exp(\eps)
    \end{align*} when $\eps\geq \ln \left(\frac{2}{s}-1\right)$, $s\in(0,1]$.
    \end{small}
\end{proof}
\vspace{-0.2em}

\subsubsection{Proof of \texorpdfstring{\Cref{thm:dplap}}{Theorem 5}}
\vspace{-0.2em}
\begin{proof}
In~\Cref{alg:laplace}, line 6 is $\eps_1$-edge differentially private and line 7 is $\eps_2$-edge differentially private due to the differential privacy guarantee of the Laplace mechanism. Therefore, from the composition theorem and the post-processing property of differential privacy, we know that \lapgraph guarantees $\eps$-edge differential privacy. 
\end{proof}
\vspace{-0.2em}
\vspace{-1mm}
\subsection{Proof of \texorpdfstring{\Cref{thm:dp-bound}}{Theorem 6}}
\label{sec:append-proof-thm-dp-bound}
\vspace{-1mm}
The work on membership privacy~\cite{li2013membership} shows that differential privacy is a type of positive membership privacy, which prevents the attacker from significantly improving its confidence on a membership inference attack. Similarly, edge differential privacy bounds an attacker's precision in an edge re-identification attack. In this section, we present a formal proof for the upper bound defined in~\Cref{thm:dp-bound}. 


\begin{proof}
With a slight abuse of notations, we use $e_1$ to represent $A_{uv} = 1$, $e_0$ to represent $A_{uv} = 0$, and $\mathcal{R}_{G}$ to represent the attack $\mathcal{R}_{G_{BB}}(u, v)$ where $G_{BB}$ is a black-box GCN.
Based on Bayes' theorem, we have
\begin{small}
\begin{equation}
\begin{aligned}
\label{eq:precision}
    &\Pr\left[e_1 \given \mathcal{R}_{G} = 1 \right] \\
    = & \frac{\Pr\left[\mathcal{R}_{G} = 1 \given e_1 \right] \Pr\left[e_1\right]}
    {\Pr\left[\mathcal{R}_{G} = 1 \given e_1 \right]\Pr\left[e_1\right] + 
    \Pr\left[\mathcal{R}_{G} = 1 \given e_0 \right]\Pr\left[e_0\right]} \\
    = & \frac{\Pr\left[\mathcal{R}_{G} = 1 \given e_1 \right] \Pr\left[e_1\right]}
    {\Pr\left[\mathcal{R}_{G} = 1 \given e_1 \right]\Pr\left[e_1\right] + 
    \Pr\left[\mathcal{R}_{G} = 1 \given e_0 \right]\left(1-\Pr\left[e_1\right]\right)}.
\end{aligned}
\end{equation}
\end{small}
Without loss of generality, let $\mathcal{G}$ represent the set of all GCN models. Since the attacker tries to re-identify the edges through querying the black-box model $G$, we could rewrite $\Pr\left[\mathcal{R}_{G} = 1 \given e_1 \right]$ as follows:
\begin{small}
\begin{equation*}
    \Pr\left[\mathcal{R}_{G} = 1 \given e_1 \right] = \sum_{G_i \in \mathcal{G}} \Pr\left[ \mathcal{R}_{G} = 1 \given G=G_i \right] \Pr\left[G = G_i \given e_1 \right].
\end{equation*}
\end{small}
Similarly,
\begin{small}
\begin{equation*}
    \Pr\left[\mathcal{R}_{G} = 1 \given e_0 \right] = \sum_{G_i \in \mathcal{G}} \Pr\left[ \mathcal{R}_{G} = 1 \given G=G_i \right] \Pr\left[G = G_i \given e_0 \right].
\end{equation*}
\end{small}
Therefore, to calculate the upper bound for Eq.~\ref{eq:precision}, it is sufficient to upper bound the following ratio for any $G_i \in \mathcal{G}$: 

\begin{small}
\begin{equation}
\label{eq:ratio}
    \frac{\Pr\left[G=G_i \given e_1 \right] \Pr\left[e_1\right]}
    {\Pr\left[G=G_i  \given e_1 \right]\Pr\left[e_1\right] + 
    \Pr\left[G=G_i  \given e_0 \right](1-\Pr\left[e_1\right])}.
\end{equation}
\end{small}

Suppose $\mathcal{A}$ is the set of all possible adjacency matrices. Let $A^{(1)}, A^{(0)} \in \mathcal{A}$ be a pair of neighboring adjacency matrices differing by edge $(u,v)$, and $A^{(1)}_{uv}=1$, $A^{(0)}_{uv}=0$. Based on the definition of differential privacy (\Cref{def:edgedp}), for any $G_i \in \mathcal{G}$ and $A^{(1)}, A^{(0)} \in \mathcal{A}$, we have 
\begin{small}
\begin{equation*}
    \Pr[G=G_i \given A=A^{(1)}] \leq \exp(\varepsilon) \Pr[G=G_i \given A=A^{(0)}].
\end{equation*}
\end{small}
Therefore, for any $G_i \in \mathcal{G}$, 
\begin{small}
\begin{equation*}
    \Pr[G=G_i \given e_1] \leq \exp(\varepsilon) \Pr[G=G_i \given e_0].
\end{equation*}
\end{small}
Since $\exp(\varepsilon)>1$ for any positive privacy budget $\varepsilon$, we also have
\begin{small}
\begin{equation*}
    \Pr[G=G_i \given e_1] \leq \exp(\varepsilon) \Pr[G=G_i \given e_1].
\end{equation*}
\end{small}
Therefore, 
\begin{small}
\begin{equation*}
\label{eq:sum}
\begin{aligned}
    &\Pr[G=G_i \given e_1] \\
    \leq &\exp(\varepsilon) \cdot \min (\Pr[G=G_i \given e_0], \Pr[G=G_i \given e_1]) \\
    \leq &\exp(\varepsilon) \left(\Pr\left[G=G_i  \given e_1 \right]\Pr\left[e_1\right] + 
    \Pr\left[G=G_i  \given e_0 \right](1-\Pr\left[e_1\right])\right) 
\end{aligned}
\end{equation*}
\end{small}
The second inequality holds because $\leq 0 \Pr[e] \leq 1$. Therefore, we could compute the upper bound for the ratio in (\ref{eq:ratio}):
\begin{small}
\begin{equation*}
\begin{aligned}
    \frac{\Pr\left[G=G_i \given e_1 \right] \Pr\left[e_1\right]}
    {\Pr\left[G=G_i  \given e_1 \right]\Pr\left[e_1\right] + 
    \Pr\left[G=G_i  \given e_0 \right](1-\Pr\left[e_1\right])}\\
    \leq \exp(\varepsilon)\cdot \Pr\left[e_1\right]
\end{aligned}
\end{equation*}
\end{small}
Because the graph density over $V^{(C)}$ is $k^{(C)}$, by the definition of graph density, we have $\Pr\left[e_1\right] = k^{(C)}$. Therefore,
\begin{small}
\begin{equation*}
    \Pr\left[e_1 \given \mathcal{R}_{G} = 1 \right] \leq \exp(\varepsilon)\cdot k^{(C)}
\end{equation*}
\end{small}
\end{proof}

\m{

\begin{small}

\begin{algorithm}[t!]
\algsetup{linenosize=\tiny}
\footnotesize
\DontPrintSemicolon
  \KwIn{\small perturbation method $M\in\{$\edgerand,\lapgraph$\}$, privacy parameter $\varepsilon$; 
  node set $V^{(T)}$, $V^{(I)}$, adjacency matrix $A_{V^{(T)}}$, $A_{V^{(I)}}$, feature matrix $X^{(T)}$, $X^{(I)}$, and labels $y^{(T)}$. The subscript $^{(T)}$ stands for training and $^{(I)}$ is for inference.}
  \SetKwFunction{perturb}{Perturbation}
  \SetKwFunction{training}{Training}
  \SetKwFunction{inference}{Inference}
  \SetKwProg{Pn}{Procedure}{:}{}
  \SetKwProg{Fn}{Function}{:}{}
  \Pn{\perturb{$A_{V^{(T)}}$, $M$, $\varepsilon$}}{
        $\widetilde A_{V^{(T)}}\leftarrow M_\eps(A_{V^{(T)}})$\;
  }
  \Pn{\training{$\widetilde A$, $V^{(T)}$, $X^{(T)}$, $y^{(T)}$}}{
        GCN $\leftarrow$ a trained model using $\widetilde A_{V^{(T)}},X^{(T)}, y^{(T)}$\;
  }
  \;
  \eIf{$V^{(T)} = V^{(I)}$}{
    $\widetilde A_{V^{(I)}} \leftarrow \widetilde A_{V^{(T)}}$ \;}{
     $\widetilde A_{V^{(I)}}\leftarrow M_\eps(A_{V^{(I)}})$  \;
   }
  \Fn{\inference{$\widetilde A, V^{(I)},X^{(I)}$}}{
        \KwRet $\mbox{GCN}(\widetilde A_{V^{(I)}},X^{(I)}, \{W^i\})$
  }
  \caption{Training and inference of DP GCN}\label{alg:dpgcn}
\end{algorithm}

\end{small}

\begin{small}
\begin{algorithm}[t!]
\algsetup{linenosize=\tiny}
\footnotesize
\DontPrintSemicolon
\KwIn{a symmetric matrix $A$, privacy parameter $s$, randomization generator}
\KwOut{the perturbed outcome $\widetilde A$}
Reset $\widetilde A$ to an all-zero matrix\;
\For{$1\leq i< j\leq n$} {
    $x\leftarrow \text{a sample drawn from } \mbox{Bern}(1-s)$\;
    \eIf{$x=1$} {
        $\widetilde A_{ij}$ and $\widetilde A_{ji}$ are set to $A_{ij}$\Comment*[r]{\scriptsize Preservation}
    } {
        $y\leftarrow\text{a sample drawn from } \mbox{Bern}(1/2)$\;
        $\widetilde A_{ij}$ and $\widetilde A_{ji}$ are set to $y$\Comment*[r]{\scriptsize Randomization}
    }
}
 \caption{Edge Randomization (\edgerand)}\label{alg:pig}
\end{algorithm}

\end{small}
\begin{small}
\begin{algorithm}[t!]
\algsetup{linenosize=\tiny}
\footnotesize
\DontPrintSemicolon
\KwIn{a symmetric matrix $A$, privacy parameter $\eps$, randomization generator}
\KwOut{the perturbed outcome $\widetilde A$}
  $\eps_1\leftarrow 0.01\eps$\Comment*[r]{\scriptsize Distribute privacy budget}
 $\eps_2\leftarrow \eps-\eps_1$\;
 $T\leftarrow$ number of edges in $A$\;
 $T\leftarrow T+\mbox{Lap}(1/\eps_1)$\Comment*[r]{\scriptsize Get a private count}
  $A\leftarrow$ the upper triangular part of $A$\;
 \For{$1\leq i<j\leq n$} {
    $ A_{ij}\leftarrow A_{ij}+\mbox{Lap}(1/\eps_2)$\Comment*[r]{\scriptsize Laplace mechanism}
}
 \Comment*[r]{\scriptsize Postprocess: Keep only the largest $T$ cells}
$S\leftarrow$ the indice set for the largest $T$ cells in $A$\;
Reset $\widetilde A$ to an all-zero matrix\;
\For{$(i,j)\in S$}{
    $\widetilde A_{ij}$ and $\widetilde A_{ji}$ are set to $1$\;
}
 \caption{Laplace Mechanism for Graphs (\lapgraph)}\label{alg:laplace}
\end{algorithm}
\end{small}
\vspace{-1mm}
\subsection{Detailed Algorithms for DP GCN}
\label{sec:append-detail-algo}
\vspace{-1mm}
\subsubsection{Algorithm for the Training and Inference of DP GCN}
\label{sec:append-algo-dp-gcn}

\Cref{alg:dpgcn} presents the perturbation, training, and inference steps in a differentially private GCN framework.
$V^{(T)}$ and $A_{V^{(T)}}$ represent the set of training nodes and the adjacency matrix of the training graph; $V^{(I)}$ and $A_{V^{(I)}}$ represent the set of testing nodes and the adjacency matrix of the testing graph. The DP guarantee holds for both transductive training (\ie, $V^{(T)} = V^{(I)}$) and inductive training (\ie, $V^{(T)} \neq V^{(I)}$). 

\subsubsection{Algorithm for the DP Mechanisms}
\label{sec:append-algo-dp-mech}

\Cref{alg:pig} presents the algorithm for \edgerand. 
We first randomly choose the cells to perturb and then randomly choose the target value from $\{0,1\}$ for each cell to be perturbed. 

\Cref{alg:laplace} presents the algorithm for \lapgraph. 
A small portion of the privacy budget $\varepsilon_1$ is used to compute the number of edges in the graph using the Laplacian mechanism, and the remaining privacy budget $\varepsilon_2 = \varepsilon - \varepsilon_1$ is used to apply Laplacian mechanism on the entire adjacency matrix. 
To preserve the degree of the original graph, the top-elements in the perturbed adjacency matrix are set to 1 and the remaining elements are set to 0. 
\vspace{-1mm}
\subsection{Additional Discussions on the \linkteller Attack}
\label{sec:append-discuss}
\vspace{-1mm}
\subsubsection{Stealthiness and Alternative Detection Strategies}
\label{sec:append-stealthy}

Our \linkteller attack queries the same set of inference nodes $V^{(I)}$ for $2n$ times where $n=\left|V^{(C)}\right|$, with the node features of one node slightly altered in each query.
This abnormal behavior can easily distinguish \linkteller from a benign user and therefore allows the detection of the attack.

In particular, we describe details of potential detection strategies as follows. 
First, a defender can use validation data to evaluate both the attack and benign query performance in terms of the attack F1 score and query node classification accuracy under different query limits. 
Then the defender could optimize a query limit $Q$ which decreases the attack performance while maintaining reasonable benign query accuracy. Such a query limit would depend on the properties of different datasets and how safety-critical the application is. Note that in general limiting the number of queries will not affect the performance for a single user, while it would hurt if several users aim to query about the same set of nodes, thus the query limit could be made for each node. In practice, the defender can directly flag the users who try to exceed the query limit $Q$ for a limited set of nodes as suspicious for further inspection.

\subsubsection{Estimation of the Density Belief $\hat k$}
\label{sec:append-belief}

In this part, we describe a few actionable strategies for the attacker given limited knowledge of the density $k$ and/or strategies to improve the accuracy of the density belief. For example, the attacker could use some similar publicly available graphs (\eg, a similar social network) or partial graphs to estimate $k$. Specifically:
    \begin{enumerate}[(a)]
        \item The attacker could estimate $k$ based on partial graph information. With the prior knowledge of some connected/unconnected pairs, the attacker can calculate the inﬂuence values for each known pair. Then, she can estimate a threshold for distinguishing the connected pairs from the unconnected ones with high conﬁdence, and thus obtain the estimated density belief $\hat k$.
        \item The attacker could estimate $k$ based on the relationship of one or a few particular nodes. The attacker can start from an intentionally low $\hat k$ and increase it until an edge is inferred for the relationship, or until the known existing edges are inferred. The attacker then stops at this specific $\hat k$ and takes it as the density belief.
        \item The attacker could estimate $k$ by running a link prediction algorithm. When a partial graph is available, the attacker can run a link prediction algorithm, \eg, training a link prediction model, to predict all edges in the graph. Based on the predictions, the attacker will then obtain a rough estimate of the density belief $\hat k$ for use in \linkteller.
    \end{enumerate}

\subsubsection{Variations of Our Attack under Different Settings}
\label{sec:append-capable}

We discuss the variations of our attack under different settings, more specifically, different attacker's capabilities or different assumptions on the interaction model.
We present three specific settings below.

\ul{When the attacker has additional knowledge of some edges}: The attacker’s prior knowledge on the existence of some edges can be leveraged to improve the density belief $\hat k$ in our \linkteller. More concretely, based on the knowledge of some edges, the attacker can calculate their inﬂuence values. Then, she can estimate a threshold for distinguishing the edges from the unconnected pairs with high conﬁdence, and then obtain a refined estimation of the density belief $\hat k$. The attack effectiveness will subsequently be improved.

\ul{When the attacker has only partial control over a subset of node features}: In this setting, part of the feature information is lost, and thus the accuracy of the estimation of the influence value would be negatively impacted, leading to the decrease of attack performance. However, how much the attack effectiveness will degrade also depends on the importance of the missing features.

\ul{When logits are not available}:
It is not straightforward to adapt our \linkteller to handle the case where logits are not available, which belongs to the ``decision based blackbox attack category'' rather than the score based. There are a few works~\cite{brendel2017decision,dong2019efficient} in the image domain that perform certain decision-based blackbox attacks. However, how to estimate the gradient/inﬂuence value in GNNs based on decision only remains an interesting future direction.

\subsubsection{Limitations to Overcome in Adapting \linkteller}
\label{sec:append-limit}
First and foremost, in order to achieve high attack effectiveness, we need to derive exact inﬂuence calculations for diﬀerent GNN structures specifically. We believe that our influence analysis based framework has the potential to perform well on diﬀerent GNN structures with the inﬂuence value calculation tailored to each of them. Another potential obstacle in the adaptation is that \linkteller cannot deal with randomized models, such as the aggregation over sampled neighbors in GraphSAGE~\cite{hamilton17graphsage}. It could be an interesting future work to take such randomness into account for inﬂuence calculation.

\subsubsection{Analysis on the Performance of \linkteller Compared with Baselines}
\label{sec:append-baseline-cmp}

First of all, we note that LSA2-X~\cite{he2020stealing} relies on measuring certain distances based on either posteriors (of the node classiﬁcation model) or node attributes to predict the connections. However, node classiﬁcation and edge inference (\ie, privacy attack goal here) are two distinct tasks, and node features (or posteriors) are useful for node classiﬁcation does not mean that they will be useful for edge inference. Thus, LSA2-X which tries to provide the attacker with diﬀerent levels of node information as prior knowledge to perform the edge re-identiﬁcation attack is not eﬀective. On the contrary, \linkteller tries to analyze the inﬂuence between nodes, which reﬂects the edge connection information based on our theoretical analysis (\Cref{thm:k-layer-gcn}) and is indeed more eﬀective for edge inference as we show empirically in~\Cref{tab:att-np} and~\Cref{tab:att-np-auc}. 

We point out that, to our best knowledge, there are no such settings where \linkteller may fail but other existing approaches (\eg, LSA2-X) may succeed. The detailed reasons are provided above. To summarize, our \linkteller leverages the edge influence information, which is more relevant for the task of edge re-identification attack than purely node level information used in LSA2-X. We then discuss two specific scenarios below.

\ul{If the model makes inferences on single nodes and not subgraphs}:  In this case, \linkteller cannot obtain influence information between nodes of interests, and thus the edge re-identification performance would be less effective. Similarly for the baselines, where they would fail to calculate the statistics of a set of nodes to compare their similarity. That is to say, if the model makes inferences on single nodes, both \linkteller and baselines may fail to effectively attack, while \linkteller may still outperform baselines given that it leverages the influence value of edges explicitly.

\ul{If the inference is transductive vs. inductive}: 
We first point out that the inductive setting is more challenging than the transductive setting. We then analyze the potential performance of \linkteller in the transductive setting. \linkteller is naturally applicable to the transductive setting--the attacker may happen to query the node in the training graph. Since these nodes are involved in model training, the influence value and the rank may be more accurate, leading to even better attack performance. 
As shown in the experimental results in~\Cref{sec:append-trans}, \linkteller indeed outperforms the baselines as well in the transductive setting.

}
\vspace{-1mm}
\subsection{Details of Evaluation}
\label{sec:append-eval-detail}
\vspace{-1mm}

\subsubsection{Dataset Statistics}
\label{sec:append-datasets}
{
\setlength{\tabcolsep}{10pt} 
\begin{table}[]
    \centering
    \caption{\small Dataset statistics (``m'' represents \textbf{m}ulti-label classification; ``s'' represents \textbf{s}ingle-label.)}\label{tab:dataset}
    \begin{subtable}[]{\linewidth}
    \centering
    \caption{\small Datasets in the \textbf{inductive} setting}\label{tab:dataset-induc}
    \resizebox{\linewidth}{!}{%
    \begin{tabular}{L{1.65cm}R{1.2cm}R{1.2cm}R{1.2cm}R{1.2cm}}
    \toprule
    \textbf{Dataset} & \textbf{Nodes} & \textbf{Edges} & \textbf{Classes} & \textbf{Features}\\\midrule
    Twitch-ES & 4,648 & 59,382 & 2 (s) & 3,170\\
    Twitch-RU & 4,385 & 37,304 & 2 (s)& 3,170\\
    Twitch-DE & 9,498 & 153,138 & 2 (s)& 3,170\\
    Twitch-FR & 6,549 & 112,666 & 2 (s) & 3,170\\
    Twitch-ENGB & 7,126 & 35,324 & 2 (s) & 3,170\\
    Twitch-PTBR & 1,912 & 31,299 & 2 (s) & 3,170\\[0.02em]\midrule[0.02em]
    PPI         & 14,755 & 225,270 & 121 (m) & 50\\
    Flickr      & 89,250 & 899,756 & 7 (s) & 500\\\bottomrule\\[0.1em]
    \end{tabular}
    }
    \end{subtable}
    \begin{subtable}[]{\linewidth}
    \centering
    \caption{\m{\small Datasets in the \textbf{transductive} setting}}\label{tab:dataset-trans}
    \resizebox{\linewidth}{!}{%
    \begin{tabular}{L{1.65cm}R{1.2cm}R{1.2cm}R{1.2cm}R{1.2cm}}
    \toprule
    \textbf{Dataset} & \textbf{Nodes} & \textbf{Edges} & \textbf{Classes} & \textbf{Features}\\\midrule
    Cora & 2,708 & 5,429 & 7 (s) & 1,433 \\
    Citeseer & 3,327 & 4,732 & 6 (s) & 3,703 \\
    Pubmed & 19,717 & 44,338 & 3 (s) & 500 \\
    \bottomrule
    \end{tabular}
    }
    \end{subtable}
\end{table}

}
We provide the dataset statistics in~\Cref{tab:dataset}.
\m{The three datasets (Cora, Citeseer, and Pubmed) in the transductive setting are all citation networks. 
Concretely, the nodes are documents/publications and the edges are the citation links between them. 
The node features are the sparse bag-of-words feature vectors for each document.}

\subsubsection{Evaluation Metrics for Model Utility}
\label{sec:append-eval-metrics}

We describe how we evaluate the utility of the trained models, including the vanilla GCN models, two DP GCN models (\edgerand and \lapgraph), and the MLP models.

We apply slightly different evaluation metrics across datasets given their varying properties.
The twitch datasets are for binary classification tasks on imbalanced datasets. Therefore, we use \textbf{F1 score of the rare class} to measure the utility of the trained GCN model. To compute the value, we first identify the minority class in the dataset and then view it as the positive class for the calculation of the F1 score.
During training, we train on twitch-ES; during inference, we evaluate the trained model on twitch-\{RU, DE, FR, ENGB, PTBR\}.
For PPI and Flickr datasets where there is no significant class imbalance, we follow previous works~\cite{hamilton17graphsage,zeng20graphsaint} and use \textbf{micro-averaged F1 score} to evaluate the classification results.

For DP GCNs particularly, in each setting, we report the averaged results over $10$ runs that use different random seeds for noise generation.

\subsubsection{Normalization Techniques}
\label{sec:append-normalize}

We followed Rong \etal~\cite{rong2020dropedge} and experimented with the techniques provided below.
$A$ is an adjacency matrix $\in\{0,1\}^{n\times n}$, $D=A+I$, and $\widehat A$ is the normalized matrix.
\begin{small}
\begin{align}
    \label{eq:norm-1}
    \widehat A&=I+D^{-1 / 2} A D^{-1 / 2}\\
    \label{eq:norm-2}
    \widehat A&=(D+I)^{-1 / 2}(A+I)(D+I)^{-1 / 2}\\
    \label{eq:norm-3}
    \widehat A&=I+(D+I)^{-1 / 2}(A+I)(D+I)^{-1 / 2}\\
    \label{eq:norm-4}
    \widehat A&=(D+I)^{-1}(A+I)
\end{align}
\end{small}
\begin{itemize}
    \item \texttt{FirstOrderGCN}: First-order GCN (Eq. \ref{eq:norm-1})
    \item \texttt{AugNormAdj}: Augmented Normalized Adjacency (Eq. \ref{eq:norm-2})
    \item \texttt{BingGeNormAdj}: Augmented Normalized Adjacency with Self-loop (Eq. \ref{eq:norm-3})
    \item \texttt{AugRWalk}: Augmented Random Walk (Eq. \ref{eq:norm-4})
\end{itemize}

\subsubsection{Search Space for the Hyper-parameters}
\label{sec:append-search-space}

In training the models, we perform an extensive grid search to find the best set of hyper-parameters. We
describe the search space of the hyper-parameters below.
\begin{itemize}
    \item learning rate (lr): \{0.005, 0.01, 0.02, 0.04, 0.05, 0.1, 0.2\}
    \item dropout rate: \{0.05, 0.1,0.2,0.3,0.5,0.8\}
    \item number of GCN layers: \{1,2,3\}
    \item number of hidden units: \{64,128,256,512\}
    \item normalization technique: \{FirstOrderGCN, AugNormAdj, BingGeNormAdj, AugRWalk\}
\end{itemize}

\subsubsection{Best Hyper-parameters for the Vanilla-GCN}
\label{sec:append-best-para}

Below, we describe the best combinations we achieve for Vanilla-GCN models. 
For twitch-ES, we use the method First-Order GCN to normalize the input graph. We train a two-layer GCN with the number of hidden units 256. The dropout rate is set to 0.5 and the learning rate is 0.01. The training epoch is 200 and the model converges within 200 epochs. For PPI, we use Augmented Normalized Adjacency with Self-loop for normalizing the adjacency matrix and train a two-layer GCN with the number of hidden layers 256. The dropout rate is 0.4 and the learning rate is 0.05. The training epoch is 200 where the model converges. For Flickr, we use Augmented Normalized Adjacency for normalization and train a two-layer GCN with the number of hidden layers 256. The dropout rate is 0.2 and the learning rate is 0.0005. The number of epochs is 200 within which the model converges.


\subsubsection{Best Hyper-parameters for the Vanilla-GAT}
\label{sec:append-best-para-gat}
We use 3-layer GATs for both PPI and Flickr datasets as described in~\Cref{sec:attack-gat}.
For PPI, the number of heads per layer are 6, 6, and 6 for the three layers. The hidden layer dimensions are 256 and 256. The skip connection is added. During training, we use dropout rate of $0$; test accuracy on the unseen node set is 0.66.
For Flickr, the number of heads per layer are 4, 4, and 4 for the three layers. The hidden layer dimensions are 256 and 256. The skip connection is added. During training, we use dropout rate of $0.5$; test accuracy on the unseen node set is 0.47.

\vspace{-1mm}
\subsection{More Evaluation Results}
\vspace{-1mm}
\subsubsection{Results for the Random Attack Baseline}
\label{sec:append-rand-baseline}
\begin{table}[t]
    \centering
    \setlength{\tabcolsep}{4pt}
    \caption{\small \textbf{Precision} (\%) of the \textit{random attack} baseline.
    }
    \label{tab:baseline}
    \begin{tabular}{crrrrrrr}
    \toprule
    Degree & \multicolumn{7}{c}{Dataset}\\
    \cmidrule(lr){2-8}
     & RU & DE & FR & ENGB & PTBR & PPI & Flickr\\\midrule
    low     &  1.7e-2 & 6.7e-3 & 7.5e-3 & 1.3e-2 & 4.5e-2 & 1.8e-2 & 4.0e-3\\
    \makecell{uncon-\\strained}  &  4.3e-1 & 3.2e-1 & 5.3e-1 & 1.5e-1 & 1.6    & 2.0e-1 & 1.0e-2\\
    high    &  1.4  & 7.5e-1 & 1.0    & 9.5e-1 & 3.4    & 1.2    & 2.6e-1\\\bottomrule
    \end{tabular}%
\end{table}

As described in~\Cref{sec:baseline-attack}, for a random classifier with Bernoulli parameter $p$,
given a set of instances containing $a$ positive examples and $b$ negative examples, its precision is $a/(a+b)$ and recall is $p$, which are \textit{density} $k$ and \textit{belief density} $\hat k$, respectively. 
We present the precision scores of the random classifier in~\Cref{tab:baseline}.
Compared with~\Cref{tab:att-np}, wee see that the \textit{precision} of \linkteller is much higher than the random attack baseline.
This reveals the significant 
advantage an attacker is able to gain 
through querying an inference API, which may 
lead to severe privacy loss. 
As for the \textit{recall} which is equal to density belief, the number $\hat{k}\in \{k/4,k/2,k,2k,4k\}$ is also extremely small compared with \linkteller.
To sum up, \linkteller significantly outperforms the random baseline.

\subsubsection{Results for a 3-layer GCN}
\label{sec:append-3-layer}

{
\setlength{\tabcolsep}{4pt} 

\begin{table}[]
    \centering
    \renewrobustcmd{\bfseries}{\fontseries{b}\selectfont}
    \sisetup{detect-weight,mode=text,group-minimum-digits = 4}

    \caption{\small \textbf{Attack Performance (Precision and Recall)} of \linkteller on twitch datasets, evaluated against a \textit{3-layer GCN}. Each table corresponds to a dataset. We sample nodes of low, unconstrained, and high degrees as our targets. Groups of rows represent different \textit{density belief} $\hat k$ of the attacker.
    }
    \label{tab:att-3-layer-1}

\begin{subtable}[]{\columnwidth}
    \centering
    \vspace{0.5em}
    \resizebox{\columnwidth}{!}{%
        \begin{tabular}{c
  S[table-format=3.1]
  @{\tiny${}\pm{}$}
  >{\tiny}S[table-format=2.1]<{\endcollectcell}
  S[table-format=3.1]
  @{\tiny${}\pm{}$}
  >{\tiny}S[table-format=2.1]<{\endcollectcell}
  S[table-format=3.1]
  @{\tiny${}\pm{}$}
  >{\tiny}S[table-format=2.1]<{\endcollectcell}
  S[table-format=3.1]
  @{\tiny${}\pm{}$}
  >{\tiny}S[table-format=2.1]<{\endcollectcell}
  S[table-format=3.1]
  @{\tiny${}\pm{}$}
  >{\tiny}S[table-format=2.1]<{\endcollectcell}
  S[table-format=3.1]
  @{\tiny${}\pm{}$}
  >{\tiny}S[table-format=2.1]<{\endcollectcell}
  S[table-format=3.1]
  @{\tiny${}\pm{}$}
  >{\tiny}S[table-format=2.1]<{\endcollectcell}
}
    \toprule
    \multicolumn{1}{c}{\textbf{RU}} &  \multicolumn{4}{c}{\small low}  & \multicolumn{4}{c}{\small unconstrained} & \multicolumn{4}{c}{\small high}\\
    \cmidrule(lr){2-5}\cmidrule(lr){6-9}\cmidrule(lr){10-13}
    \makecell{\small $\hat k$}     & \multicolumn{2}{c}{\small precision}  & \multicolumn{2}{c}{\small recall} & \multicolumn{2}{c}{\small precision} & \multicolumn{2}{c}{\small recall} & \multicolumn{2}{c}{\small precision} & \multicolumn{2}{c}{\small recall} \\\midrule
\multirow{1}{*}{$k/4$}
                & 100.0 & 0.0  & 33.0 & 2.8  & 80.8 & 4.2  & 22.1 & 1.5  & 83.9 & 2.1  & 15.4 & 1.5 \\
\multirow{1}{*}{$k/2$}
                & 84.6 & 0.0  & 51.9 & 4.3  & 65.1 & 2.1  & 35.5 & 0.9  & 72.9 & 1.1  & 26.7 & 1.9 \\
\multirow{1}{*}{$k$}
                & 69.3 & 8.2  & 81.1 & 4.2  & 45.7 & 2.2  & 50.0 & 2.8  & 55.6 & 2.8  & 40.7 & 1.6 \\
\multirow{1}{*}{$2k$}
                & 40.7 & 5.0  & 95.0 & 4.3  & 27.7 & 1.8  & 60.4 & 2.7  & 37.4 & 2.9  & 54.6 & 1.0 \\
\multirow{1}{*}{$4k$}
                & 20.3 & 2.5  & 95.0 & 4.3  & 15.8 & 1.0  & 68.8 & 3.0  & 23.0 & 2.4  & 67.0 & 2.6 \\
\bottomrule
\end{tabular}%
}
\end{subtable}

\begin{subtable}[]{\columnwidth}
    \centering
    \vspace{0.5em}
    \resizebox{\columnwidth}{!}{%
        \begin{tabular}{c
  S[table-format=3.1]
  @{\tiny${}\pm{}$}
  >{\tiny}S[table-format=2.1]<{\endcollectcell}
  S[table-format=3.1]
  @{\tiny${}\pm{}$}
  >{\tiny}S[table-format=2.1]<{\endcollectcell}
  S[table-format=3.1]
  @{\tiny${}\pm{}$}
  >{\tiny}S[table-format=2.1]<{\endcollectcell}
  S[table-format=3.1]
  @{\tiny${}\pm{}$}
  >{\tiny}S[table-format=2.1]<{\endcollectcell}
  S[table-format=3.1]
  @{\tiny${}\pm{}$}
  >{\tiny}S[table-format=2.1]<{\endcollectcell}
  S[table-format=3.1]
  @{\tiny${}\pm{}$}
  >{\tiny}S[table-format=2.1]<{\endcollectcell}
  S[table-format=3.1]
  @{\tiny${}\pm{}$}
  >{\tiny}S[table-format=2.1]<{\endcollectcell}
}
    \toprule
    \multicolumn{1}{c}{\textbf{DE}} &  \multicolumn{4}{c}{\small low}  & \multicolumn{4}{c}{\small unconstrained} & \multicolumn{4}{c}{\small high}\\
    \cmidrule(lr){2-5}\cmidrule(lr){6-9}\cmidrule(lr){10-13}
    \makecell{\small $\hat k$}     & \multicolumn{2}{c}{\small precision}  & \multicolumn{2}{c}{\small recall} & \multicolumn{2}{c}{\small precision} & \multicolumn{2}{c}{\small recall} & \multicolumn{2}{c}{\small precision} & \multicolumn{2}{c}{\small recall} \\\midrule
\multirow{1}{*}{$k/4$}
                & 91.7 & 11.8  & 29.0 & 3.4  & 75.2 & 5.8  & 18.1 & 2.6  & 71.3 & 6.7  & 18.1 & 1.8 \\
\multirow{1}{*}{$k/2$}
                & 82.1 & 12.7  & 49.6 & 6.3  & 54.6 & 2.6  & 26.3 & 3.0  & 50.3 & 4.6  & 25.5 & 2.7 \\
\multirow{1}{*}{$k$}
                & 64.6 & 7.6  & 73.0 & 9.0  & 32.7 & 1.0  & 31.3 & 3.0  & 33.0 & 2.4  & 33.4 & 2.4 \\
\multirow{1}{*}{$2k$}
                & 41.7 & 3.6  & 88.9 & 7.9  & 20.4 & 0.2  & 38.9 & 2.6  & 21.9 & 1.5  & 44.5 & 3.1 \\
\multirow{1}{*}{$4k$}
                & 22.4 & 1.7  & 94.4 & 7.9  & 13.9 & 0.4  & 53.1 & 2.1  & 14.0 & 0.6  & 56.7 & 1.4 \\
\bottomrule
\end{tabular}%
}
\end{subtable}

\begin{subtable}[]{\columnwidth}
    \centering
    \vspace{0.5em}
    \resizebox{\columnwidth}{!}{%
        \begin{tabular}{c
  S[table-format=3.1]
  @{\tiny${}\pm{}$}
  >{\tiny}S[table-format=2.1]<{\endcollectcell}
  S[table-format=3.1]
  @{\tiny${}\pm{}$}
  >{\tiny}S[table-format=2.1]<{\endcollectcell}
  S[table-format=3.1]
  @{\tiny${}\pm{}$}
  >{\tiny}S[table-format=2.1]<{\endcollectcell}
  S[table-format=3.1]
  @{\tiny${}\pm{}$}
  >{\tiny}S[table-format=2.1]<{\endcollectcell}
  S[table-format=3.1]
  @{\tiny${}\pm{}$}
  >{\tiny}S[table-format=2.1]<{\endcollectcell}
  S[table-format=3.1]
  @{\tiny${}\pm{}$}
  >{\tiny}S[table-format=2.1]<{\endcollectcell}
  S[table-format=3.1]
  @{\tiny${}\pm{}$}
  >{\tiny}S[table-format=2.1]<{\endcollectcell}
}
    \toprule
    \multicolumn{1}{c}{\textbf{FR}} &  \multicolumn{4}{c}{\small low}  & \multicolumn{4}{c}{\small unconstrained} & \multicolumn{4}{c}{\small high}\\
    \cmidrule(lr){2-5}\cmidrule(lr){6-9}\cmidrule(lr){10-13}
    \makecell{\small $\hat k$}     & \multicolumn{2}{c}{\small precision}  & \multicolumn{2}{c}{\small recall} & \multicolumn{2}{c}{\small precision} & \multicolumn{2}{c}{\small recall} & \multicolumn{2}{c}{\small precision} & \multicolumn{2}{c}{\small recall} \\\midrule
\multirow{1}{*}{$k/4$}
                & 100.0 & 0.0  & 28.3 & 2.4  & 85.4 & 5.6  & 19.9 & 1.6  & 87.9 & 3.8  & 21.3 & 1.6 \\
\multirow{1}{*}{$k/2$}
                & 100.0 & 0.0  & 50.0 & 0.0  & 71.5 & 5.5  & 33.3 & 2.9  & 70.9 & 8.5  & 34.0 & 1.5 \\
\multirow{1}{*}{$k$}
                & 78.3 & 2.4  & 78.3 & 2.4  & 50.1 & 5.1  & 46.6 & 5.0  & 48.6 & 10.1  & 46.2 & 5.3 \\
\multirow{1}{*}{$2k$}
                & 41.7 & 2.4  & 81.7 & 6.2  & 29.1 & 2.1  & 54.1 & 4.0  & 28.6 & 6.0  & 54.4 & 6.4 \\
\multirow{1}{*}{$4k$}
                & 20.8 & 1.2  & 81.7 & 6.2  & 16.3 & 1.2  & 60.7 & 5.4  & 16.6 & 2.9  & 63.4 & 5.4 \\
\bottomrule
\end{tabular}%
}
\end{subtable}

\begin{subtable}[]{\columnwidth}
    \centering
    \vspace{0.5em}
    \resizebox{\columnwidth}{!}{%
        \begin{tabular}{c
  S[table-format=3.1]
  @{\tiny${}\pm{}$}
  >{\tiny}S[table-format=2.1]<{\endcollectcell}
  S[table-format=3.1]
  @{\tiny${}\pm{}$}
  >{\tiny}S[table-format=2.1]<{\endcollectcell}
  S[table-format=3.1]
  @{\tiny${}\pm{}$}
  >{\tiny}S[table-format=2.1]<{\endcollectcell}
  S[table-format=3.1]
  @{\tiny${}\pm{}$}
  >{\tiny}S[table-format=2.1]<{\endcollectcell}
  S[table-format=3.1]
  @{\tiny${}\pm{}$}
  >{\tiny}S[table-format=2.1]<{\endcollectcell}
  S[table-format=3.1]
  @{\tiny${}\pm{}$}
  >{\tiny}S[table-format=2.1]<{\endcollectcell}
  S[table-format=3.1]
  @{\tiny${}\pm{}$}
  >{\tiny}S[table-format=2.1]<{\endcollectcell}
}
    \toprule
    \multicolumn{1}{c}{\textbf{ENGB}} &  \multicolumn{4}{c}{\small low}  & \multicolumn{4}{c}{\small unconstrained} & \multicolumn{4}{c}{\small high}\\
    \cmidrule(lr){2-5}\cmidrule(lr){6-9}\cmidrule(lr){10-13}
    \makecell{\small $\hat k$}     & \multicolumn{2}{c}{\small precision}  & \multicolumn{2}{c}{\small recall} & \multicolumn{2}{c}{\small precision} & \multicolumn{2}{c}{\small recall} & \multicolumn{2}{c}{\small precision} & \multicolumn{2}{c}{\small recall} \\\midrule
\multirow{1}{*}{$k/4$}
                & 91.7 & 11.8  & 27.7 & 5.2  & 83.1 & 3.3  & 22.9 & 4.4  & 86.7 & 1.1  & 22.1 & 0.4 \\
\multirow{1}{*}{$k/2$}
                & 85.7 & 11.7  & 47.0 & 12.7  & 69.1 & 5.0  & 37.4 & 6.1  & 71.1 & 3.0  & 36.2 & 1.7 \\
\multirow{1}{*}{$k$}
                & 66.4 & 9.3  & 68.4 & 18.9  & 48.7 & 6.9  & 51.7 & 5.9  & 50.7 & 3.0  & 51.6 & 3.3 \\
\multirow{1}{*}{$2k$}
                & 46.0 & 7.5  & 89.0 & 4.4  & 29.6 & 4.5  & 62.7 & 6.9  & 30.7 & 1.7  & 62.4 & 3.8 \\
\multirow{1}{*}{$4k$}
                & 23.7 & 3.7  & 91.6 & 3.1  & 17.0 & 3.6  & 70.9 & 3.4  & 17.5 & 0.4  & 71.2 & 1.9 \\
\bottomrule
\end{tabular}%
}
\end{subtable}

\begin{subtable}[]{\columnwidth}
    \centering
    \vspace{0.5em}
    \resizebox{\columnwidth}{!}{%
        \begin{tabular}{c
  S[table-format=3.1]
  @{\tiny${}\pm{}$}
  >{\tiny}S[table-format=2.1]<{\endcollectcell}
  S[table-format=3.1]
  @{\tiny${}\pm{}$}
  >{\tiny}S[table-format=2.1]<{\endcollectcell}
  S[table-format=3.1]
  @{\tiny${}\pm{}$}
  >{\tiny}S[table-format=2.1]<{\endcollectcell}
  S[table-format=3.1]
  @{\tiny${}\pm{}$}
  >{\tiny}S[table-format=2.1]<{\endcollectcell}
  S[table-format=3.1]
  @{\tiny${}\pm{}$}
  >{\tiny}S[table-format=2.1]<{\endcollectcell}
  S[table-format=3.1]
  @{\tiny${}\pm{}$}
  >{\tiny}S[table-format=2.1]<{\endcollectcell}
  S[table-format=3.1]
  @{\tiny${}\pm{}$}
  >{\tiny}S[table-format=2.1]<{\endcollectcell}
}
    \toprule
    \multicolumn{1}{c}{\textbf{PTBR}} &  \multicolumn{4}{c}{\small low}  & \multicolumn{4}{c}{\small unconstrained} & \multicolumn{4}{c}{\small high}\\
    \cmidrule(lr){2-5}\cmidrule(lr){6-9}\cmidrule(lr){10-13}
    \makecell{\small $\hat k$}     & \multicolumn{2}{c}{\small precision}  & \multicolumn{2}{c}{\small recall} & \multicolumn{2}{c}{\small precision} & \multicolumn{2}{c}{\small recall} & \multicolumn{2}{c}{\small precision} & \multicolumn{2}{c}{\small recall} \\\midrule
\multirow{1}{*}{$k/4$}
                & 100.0 & 0.0  & 26.7 & 1.3  & 80.8 & 4.7  & 20.9 & 4.3  & 86.7 & 1.9  & 21.3 & 2.1 \\
\multirow{1}{*}{$k/2$}
                & 91.8 & 5.8  & 48.6 & 5.7  & 65.5 & 7.7  & 33.3 & 5.3  & 73.5 & 2.0  & 36.2 & 3.4 \\
\multirow{1}{*}{$k$}
                & 71.1 & 0.9  & 74.3 & 4.0  & 46.3 & 9.5  & 46.0 & 4.1  & 53.5 & 2.3  & 52.5 & 3.9 \\
\multirow{1}{*}{$2k$}
                & 41.3 & 1.3  & 85.6 & 2.8  & 30.2 & 7.4  & 59.5 & 3.4  & 32.9 & 2.3  & 64.4 & 4.1 \\
\multirow{1}{*}{$4k$}
                & 21.4 & 1.1  & 88.7 & 0.6  & 18.2 & 4.8  & 71.4 & 3.5  & 19.1 & 1.5  & 74.7 & 3.2 \\
\bottomrule
\end{tabular}%
}
\end{subtable}

\end{table}

}
{
\setlength{\tabcolsep}{4pt} 

\begin{table}[]
    \centering
    \renewrobustcmd{\bfseries}{\fontseries{b}\selectfont}
    \sisetup{detect-weight,mode=text,group-minimum-digits = 4}    
    \caption{\small \textbf{AUC} of \linkteller on twitch datasets, evaluated against a \textit{3-layer GCN}. Each column corresponds to one dataset. Rows represent sampled nodes of varying degrees.
    }
    \label{tab:att-3-layer-auc}
\begin{subtable}[]{\columnwidth}
    \centering
    \vspace{0.5em}
    \resizebox{.8\columnwidth}{!}{%
        \begin{tabular}{c
  S[table-format=1.2]
  @{\tiny${}\pm{}$}
  >{\tiny}S[table-format=1.2]<{\endcollectcell}
  S[table-format=1.2]
  @{\tiny${}\pm{}$}
  >{\tiny}S[table-format=1.2]<{\endcollectcell}
  S[table-format=1.2]
  @{\tiny${}\pm{}$}
  >{\tiny}S[table-format=1.2]<{\endcollectcell}
  S[table-format=1.2]
  @{\tiny${}\pm{}$}
  >{\tiny}S[table-format=1.2]<{\endcollectcell}
  S[table-format=1.2]
  @{\tiny${}\pm{}$}
  >{\tiny}S[table-format=1.2]<{\endcollectcell}
}
    \toprule
    \makecell{ Degree} & \multicolumn{10}{c}{Dataset}\\
    \cmidrule(lr){2-11}
       & \multicolumn{2}{c}{RU}  & \multicolumn{2}{c}{DE} & \multicolumn{2}{c}{FR} & \multicolumn{2}{c}{ENGB} & \multicolumn{2}{c}{PTBR}  \\\midrule
{\footnotesize low }
               & 1.00 & 0.00  & 0.99 & 0.02  & 0.94 & 0.05  & 1.00 & 0.00  & 0.98 & 0.00\\
{\scriptsize \makecell{ uncon-\\strained}}
               & 0.96 & 0.01  & 0.92 & 0.01  & 0.91 & 0.02  & 0.97 & 0.01  & 0.92 & 0.01 \\
{\footnotesize high }
               & 0.93 & 0.01  & 0.88 & 0.01  & 0.90 & 0.01  & 0.95 & 0.00  & 0.90 & 0.00 \\
            \bottomrule
\end{tabular}%
}
\end{subtable}
\end{table}
}
In~\Cref{sec:attack-eval} in the main paper, we mainly evaluated 2-layer GCNs. In this section, we evaluate the performance of \linkteller on 3-layer GCNs to provide a more comprehensive view of \linkteller's capability.

\paragraph{Model}
For training the models, we follow the same principle described in~\Cref{sec:models} and use the same search space as in~\Cref{sec:append-search-space}.
The best combination of hyper-parameters/configurations are described below. 
We use the method First-Order GCN to normalize the input graph. The hidden layer dimensions are 64 and 64. 
The dropout rate is set to 0.5 and the learning rate is 0.01. 
The training epoch is 50 and the model converges.
The test F1 score on twitch-\{RU, DE, FR, ENGB, PTBR\} are 0.3419, 0.4698, 0.4926, 0.6027, and 0.5198, respectively.

\paragraph{Attack Results}
We present the attack results of \linkteller on the 3-layer GCN in~\Cref{tab:att-3-layer-1} and~\Cref{tab:att-3-layer-auc}. 
Comparing~\Cref{tab:att-3-layer-1} with~\Cref{tab:att-np}, and~\Cref{tab:att-3-layer-auc} with~\Cref{tab:att-np-auc}, we see that the performance of \linkteller on the 3-layer GCN only drops a little.
For 1-layer GCN, we know from \Cref{thm:1-layer-gcn} that \linkteller can perform a perfect attack. For GCNs with more than 3 layers, we did not bother to evaluate the attack performance since deeper GCNs suffer from over-smoothing~\cite{li2018deeper} and give poor classification results.
Thus, we can confidently conclude that \linkteller is a successful attack against most practical GCN models.

\subsubsection{Running Time of \linkteller}
\label{sec:append-runtime}

{
\setlength{\tabcolsep}{4pt} 

\begin{table}[]
    \centering
    \renewrobustcmd{\bfseries}{\fontseries{b}\selectfont}
    \sisetup{detect-weight,mode=text,group-minimum-digits = 4}    
    \caption{\small \textbf{Running time} of \linkteller on vanilla GCNs corresponding to experiments in~\Cref{sec:eval-attack}. The time unit is ``second''.
    }
    \label{tab:att-runtime}
\begin{subtable}[]{\columnwidth}
    \centering
    \vspace{0.5em}
    \resizebox{\columnwidth}{!}{%
        \begin{tabular}{c
  S[table-format=1.2]
  @{\tiny${}\pm{}$}
  >{\tiny}S[table-format=1.2]<{\endcollectcell}
  S[table-format=1.2]
  @{\tiny${}\pm{}$}
  >{\tiny}S[table-format=1.2]<{\endcollectcell}
  S[table-format=1.2]
  @{\tiny${}\pm{}$}
  >{\tiny}S[table-format=1.2]<{\endcollectcell}
  S[table-format=1.2]
  @{\tiny${}\pm{}$}
  >{\tiny}S[table-format=1.2]<{\endcollectcell}
  S[table-format=1.2]
  @{\tiny${}\pm{}$}
  >{\tiny}S[table-format=1.2]<{\endcollectcell}
  S[table-format=1.2]
  @{\tiny${}\pm{}$}
  >{\tiny}S[table-format=1.2]<{\endcollectcell}
  S[table-format=1.2]
  @{\tiny${}\pm{}$}
  >{\tiny}S[table-format=1.2]<{\endcollectcell}  
}
    \toprule
    Degree & \multicolumn{14}{c}{Dataset}\\
    \cmidrule(lr){2-15}
    & \multicolumn{2}{c}{RU}  & \multicolumn{2}{c}{DE} & \multicolumn{2}{c}{FR} & \multicolumn{2}{c}{ENGB} & \multicolumn{2}{c}{PTBR} & \multicolumn{2}{c}{PPI} & \multicolumn{2}{c}{Flickr}  \\\midrule
{\footnotesize low }
                & 12.5 & 0.0  & 16.1 & 0.1  & 13.2 & 0.0  & 12.8 & 0.0  & 11.2 & 0.1  & 14.8 & 0.1  & 30.8 & 0.1 \\
{\scriptsize \makecell{ uncon-\\strained}}
                & 12.4 & 0.1  & 16.0 & 0.1  & 13.4 & 0.1  & 12.9 & 0.1  & 11.0 & 0.1  & 14.7 & 0.1  & 30.7 & 0.2 \\
{\footnotesize high }
                & 12.4 & 0.1  & 16.1 & 0.0  & 13.0 & 0.2  & 12.6 & 0.0  & 11.5 & 0.4  & 14.8 & 0.0  & 30.5 & 0.2 \\
            \bottomrule
\end{tabular}%
}
\end{subtable}
\end{table}
}
We report the running time of \linkteller on vanilla GCNs in~\Cref{tab:att-runtime}, corresponding to the experiments in~\Cref{sec:eval-attack} in the main paper. As the table shows, \linkteller is a highly efficient attack.
On DP GCNs using \edgerand mechanism, when the graph becomes denser under smaller privacy budgets, one forward pass of the network takes longer, since the cost of matrix computation becomes larger. However, the increase of running time reflected in the attack time is only marginal, so we omit the running time for DP GCNs here.
Overall, \linkteller can efficiently and effectively attack both vanilla GCNs and DP GCNs.

\subsubsection{More Results for \linkteller on vanilla GCNs and DP GCNs}
\label{sec:append-eval-results}

{
\setlength{\tabcolsep}{4pt} 

\begin{table}[]
    \centering
    \renewrobustcmd{\bfseries}{\fontseries{b}\selectfont}
    \sisetup{detect-weight,mode=text,group-minimum-digits = 4}

    \caption{\small \textbf{Attack Performance} of \linkteller on additional datasets, compared with two baseline methods LSA2-\{post, attr\}~\cite{he2020stealing}. Each table corresponds to a dataset. We sample nodes of low, unconstrained, and high degrees as our targets. Groups of rows represent different \textit{density belief} $\hat k$ of the attacker.
    }
    \label{tab:att-np-file1}

\begin{subtable}[]{\columnwidth}
    \centering
    \vspace{0.5em}
    \resizebox{\columnwidth}{!}{%
        \begin{tabular}{cc
  S[table-format=3.1]
  @{\tiny${}\pm{}$}
  >{\tiny}S[table-format=2.1]<{\endcollectcell}
  S[table-format=3.1]
  @{\tiny${}\pm{}$}
  >{\tiny}S[table-format=2.1]<{\endcollectcell}
  S[table-format=3.1]
  @{\tiny${}\pm{}$}
  >{\tiny}S[table-format=2.1]<{\endcollectcell}
  S[table-format=3.1]
  @{\tiny${}\pm{}$}
  >{\tiny}S[table-format=2.1]<{\endcollectcell}
  S[table-format=3.1]
  @{\tiny${}\pm{}$}
  >{\tiny}S[table-format=2.1]<{\endcollectcell}
  S[table-format=3.1]
  @{\tiny${}\pm{}$}
  >{\tiny}S[table-format=2.1]<{\endcollectcell}
  S[table-format=3.1]
  @{\tiny${}\pm{}$}
  >{\tiny}S[table-format=2.1]<{\endcollectcell}
}
    \toprule
    \multicolumn{2}{c}{\textbf{twitch-DE}} &  \multicolumn{4}{c}{\small low}  & \multicolumn{4}{c}{\small unconstrained} & \multicolumn{4}{c}{\small high}\\
    \cmidrule(lr){3-6}\cmidrule(lr){7-10}\cmidrule(lr){11-14}
    \makecell{\small $\hat k$} & Method   & \multicolumn{2}{c}{\small precision}  & \multicolumn{2}{c}{\small recall} & \multicolumn{2}{c}{\small precision} & \multicolumn{2}{c}{\small recall} & \multicolumn{2}{c}{\small precision} & \multicolumn{2}{c}{\small recall} \\\midrule
\rowcolor{tabgray}
\cellcolor{white}{\multirow{3}{*}{$k/4$}}
            & Ours  & \bfseries 83.3 & 23.6  & \bfseries 26.2 & 7.0  & \bfseries 94.0 & 1.3  & \bfseries 22.5 & 1.3  & \bfseries 99.0 & 0.4  & \bfseries 25.2 & 1.2 \\
            & LSA2-post  & 0.0 & 0.0  & 0.0 & 0.0  & 0.0 & 0.0  & 0.0 & 0.0  & 0.1 & 0.2  & 0.0 & 0.1 \\
            & LSA2-attr  & 0.0 & 0.0  & 0.0 & 0.0  & 0.4 & 0.5  & 0.1 & 0.1  & 2.6 & 1.1  & 0.7 & 0.3 \\[0.05em]\midrule[0.05em]
\rowcolor{tabgray}
\cellcolor{white}{\multirow{3}{*}{$k/2$}}
            & Ours  & \bfseries 91.7 & 11.8  & \bfseries 55.2 & 3.7  & \bfseries 92.4 & 4.1  & \bfseries 44.2 & 1.6  & \bfseries 96.9 & 0.3  & \bfseries 49.2 & 2.5 \\
            & LSA2-post  & 0.0 & 0.0  & 0.0 & 0.0  & 0.0 & 0.0  & 0.0 & 0.0  & 0.1 & 0.1  & 0.0 & 0.1 \\
            & LSA2-attr  & 0.0 & 0.0  & 0.0 & 0.0  & 0.9 & 0.3  & 0.4 & 0.1  & 2.5 & 0.8  & 1.3 & 0.5 \\[0.05em]\midrule[0.05em]
\rowcolor{tabgray}
\cellcolor{white}{\multirow{3}{*}{$k$}}
            & Ours  & \bfseries 81.8 & 4.8  & \bfseries 92.5 & 5.9  & \bfseries 81.2 & 6.6  & \bfseries 77.2 & 3.4  & \bfseries 79.2 & 1.1  & \bfseries 80.4 & 3.1 \\
            & LSA2-post  & 0.0 & 0.0  & 0.0 & 0.0  & 0.0 & 0.0  & 0.0 & 0.0  & 0.0 & 0.0  & 0.0 & 0.1 \\
            & LSA2-attr  & 0.0 & 0.0  & 0.0 & 0.0  & 1.2 & 0.3  & 1.2 & 0.4  & 1.8 & 0.4  & 1.8 & 0.5 \\[0.05em]\midrule[0.05em]
\rowcolor{tabgray}
\cellcolor{white}{\multirow{3}{*}{$2k$}}
            & Ours  & \bfseries 44.7 & 3.4  & \bfseries 95.2 & 6.7  & \bfseries 49.0 & 4.7  & \bfseries 93.1 & 3.2  & \bfseries 46.9 & 2.0  & \bfseries 95.0 & 0.9 \\
            & LSA2-post  & 0.0 & 0.0  & 0.0 & 0.0  & 0.0 & 0.1  & 0.1 & 0.1  & 0.0 & 0.0  & 0.0 & 0.1 \\
            & LSA2-attr  & 0.0 & 0.0  & 0.0 & 0.0  & 0.8 & 0.1  & 1.6 & 0.2  & 1.4 & 0.3  & 2.8 & 0.8 \\[0.05em]\midrule[0.05em]
\rowcolor{tabgray}
\cellcolor{white}{\multirow{3}{*}{$4k$}}
            & Ours  & \bfseries 23.8 & 0.3  & \bfseries 100.0 & 0.0  & \bfseries 25.8 & 1.9  & \bfseries 98.1 & 0.7  & \bfseries 24.3 & 1.1  & \bfseries 98.5 & 0.5 \\
            & LSA2-post  & 0.0 & 0.0  & 0.0 & 0.0  & 0.0 & 0.1  & 0.2 & 0.2  & 0.0 & 0.0  & 0.0 & 0.1 \\
            & LSA2-attr  & 0.0 & 0.0  & 0.0 & 0.0  & 0.7 & 0.1  & 2.7 & 0.4  & 1.1 & 0.2  & 4.4 & 0.9 \\\bottomrule
\end{tabular}%
}
\end{subtable}

\begin{subtable}[]{\columnwidth}
    \centering
    \vspace{0.5em}
    \resizebox{\columnwidth}{!}{%
        \begin{tabular}{cc
  S[table-format=3.1]
  @{\tiny${}\pm{}$}
  >{\tiny}S[table-format=2.1]<{\endcollectcell}
  S[table-format=3.1]
  @{\tiny${}\pm{}$}
  >{\tiny}S[table-format=2.1]<{\endcollectcell}
  S[table-format=3.1]
  @{\tiny${}\pm{}$}
  >{\tiny}S[table-format=2.1]<{\endcollectcell}
  S[table-format=3.1]
  @{\tiny${}\pm{}$}
  >{\tiny}S[table-format=2.1]<{\endcollectcell}
  S[table-format=3.1]
  @{\tiny${}\pm{}$}
  >{\tiny}S[table-format=2.1]<{\endcollectcell}
  S[table-format=3.1]
  @{\tiny${}\pm{}$}
  >{\tiny}S[table-format=2.1]<{\endcollectcell}
  S[table-format=3.1]
  @{\tiny${}\pm{}$}
  >{\tiny}S[table-format=2.1]<{\endcollectcell}
}
    \toprule
    \multicolumn{2}{c}{\textbf{twitch-ENGB}} &  \multicolumn{4}{c}{\small low}  & \multicolumn{4}{c}{\small unconstrained} & \multicolumn{4}{c}{\small high}\\
    \cmidrule(lr){3-6}\cmidrule(lr){7-10}\cmidrule(lr){11-14}
    \makecell{\small $\hat k$} & Method   & \multicolumn{2}{c}{\small precision}  & \multicolumn{2}{c}{\small recall} & \multicolumn{2}{c}{\small precision} & \multicolumn{2}{c}{\small recall} & \multicolumn{2}{c}{\small precision} & \multicolumn{2}{c}{\small recall} \\\midrule
\rowcolor{tabgray}
\cellcolor{white}{\multirow{3}{*}{$k/4$}}
            & Ours  & \bfseries 100.0 & 0.0  & \bfseries 30.3 & 4.1  & \bfseries 92.6 & 3.0  & \bfseries 25.5 & 4.9  & \bfseries 98.8 & 0.5  & \bfseries 25.2 & 0.2 \\
            & LSA2-post  & 0.0 & 0.0  & 0.0 & 0.0  & 0.0 & 0.0  & 0.0 & 0.0  & 0.1 & 0.2  & 0.0 & 0.0 \\
            & LSA2-attr  & 0.0 & 0.0  & 0.0 & 0.0  & 0.0 & 0.0  & 0.0 & 0.0  & 1.5 & 0.4  & 0.4 & 0.1 \\[0.05em]\midrule[0.05em]
\rowcolor{tabgray}
\cellcolor{white}{\multirow{3}{*}{$k/2$}}
            & Ours  & \bfseries 100.0 & 0.0  & \bfseries 54.2 & 8.7  & \bfseries 84.3 & 5.6  & \bfseries 45.6 & 7.6  & \bfseries 96.0 & 1.2  & \bfseries 48.9 & 0.6 \\
            & LSA2-post  & 0.0 & 0.0  & 0.0 & 0.0  & 0.0 & 0.0  & 0.0 & 0.0  & 0.1 & 0.1  & 0.0 & 0.0 \\
            & LSA2-attr  & 0.0 & 0.0  & 0.0 & 0.0  & 0.3 & 0.4  & 0.2 & 0.2  & 1.8 & 0.3  & 0.9 & 0.1 \\[0.05em]\midrule[0.05em]
\rowcolor{tabgray}
\cellcolor{white}{\multirow{3}{*}{$k$}}
            & Ours  & \bfseries 83.1 & 6.6  & \bfseries 84.0 & 11.4  & \bfseries 67.9 & 6.3  & \bfseries 72.9 & 10.9  & \bfseries 81.6 & 2.7  & \bfseries 83.1 & 2.7 \\
            & LSA2-post  & 0.0 & 0.0  & 0.0 & 0.0  & 0.0 & 0.0  & 0.0 & 0.0  & 0.1 & 0.0  & 0.1 & 0.0 \\
            & LSA2-attr  & 0.0 & 0.0  & 0.0 & 0.0  & 0.7 & 0.2  & 0.7 & 0.2  & 2.0 & 0.1  & 2.0 & 0.1 \\[0.05em]\midrule[0.05em]
\rowcolor{tabgray}
\cellcolor{white}{\multirow{3}{*}{$2k$}}
            & Ours  & \bfseries 50.7 & 8.2  & \bfseries 97.9 & 2.9  & \bfseries 43.6 & 9.0  & \bfseries 91.2 & 4.9  & \bfseries 47.3 & 0.7  & \bfseries 96.3 & 1.3 \\
            & LSA2-post  & 0.0 & 0.0  & 0.0 & 0.0  & 0.0 & 0.0  & 0.0 & 0.0  & 0.4 & 0.5  & 0.8 & 1.0 \\
            & LSA2-attr  & 0.0 & 0.0  & 0.0 & 0.0  & 0.5 & 0.2  & 1.2 & 0.6  & 1.7 & 0.2  & 3.4 & 0.3 \\[0.05em]\midrule[0.05em]
\rowcolor{tabgray}
\cellcolor{white}{\multirow{3}{*}{$4k$}}
            & Ours  & \bfseries 26.0 & 4.9  & \bfseries 100.0 & 0.0  & \bfseries 23.5 & 5.9  & \bfseries 97.3 & 1.1  & \bfseries 24.2 & 0.2  & \bfseries 98.5 & 0.5 \\
            & LSA2-post  & 0.0 & 0.0  & 0.0 & 0.0  & 0.0 & 0.0  & 0.0 & 0.0  & 0.2 & 0.2  & 0.8 & 1.0 \\
            & LSA2-attr  & 0.0 & 0.0  & 0.0 & 0.0  & 0.4 & 0.2  & 1.7 & 1.0  & 1.7 & 0.1  & 6.9 & 0.3 \\\bottomrule
\end{tabular}%
}
\end{subtable}

\begin{subtable}[]{\columnwidth}
    \centering
    \vspace{0.5em}
    \resizebox{\columnwidth}{!}{%
        \begin{tabular}{cc
  S[table-format=3.1]
  @{\tiny${}\pm{}$}
  >{\tiny}S[table-format=2.1]<{\endcollectcell}
  S[table-format=3.1]
  @{\tiny${}\pm{}$}
  >{\tiny}S[table-format=2.1]<{\endcollectcell}
  S[table-format=3.1]
  @{\tiny${}\pm{}$}
  >{\tiny}S[table-format=2.1]<{\endcollectcell}
  S[table-format=3.1]
  @{\tiny${}\pm{}$}
  >{\tiny}S[table-format=2.1]<{\endcollectcell}
  S[table-format=3.1]
  @{\tiny${}\pm{}$}
  >{\tiny}S[table-format=2.1]<{\endcollectcell}
  S[table-format=3.1]
  @{\tiny${}\pm{}$}
  >{\tiny}S[table-format=2.1]<{\endcollectcell}
  S[table-format=3.1]
  @{\tiny${}\pm{}$}
  >{\tiny}S[table-format=2.1]<{\endcollectcell}
}
    \toprule
    \multicolumn{2}{c}{\textbf{twitch-PTBR}} &  \multicolumn{4}{c}{\small low}  & \multicolumn{4}{c}{\small unconstrained} & \multicolumn{4}{c}{\small high}\\
    \cmidrule(lr){3-6}\cmidrule(lr){7-10}\cmidrule(lr){11-14}
    \makecell{\small $\hat k$} & Method   & \multicolumn{2}{c}{\small precision}  & \multicolumn{2}{c}{\small recall} & \multicolumn{2}{c}{\small precision} & \multicolumn{2}{c}{\small recall} & \multicolumn{2}{c}{\small precision} & \multicolumn{2}{c}{\small recall} \\\midrule
\rowcolor{tabgray}
\cellcolor{white}{\multirow{3}{*}{$k/4$}}
            & Ours  & \bfseries 100.0 & 0.0  & \bfseries 26.7 & 1.3  & \bfseries 95.6 & 1.6  & \bfseries 25.1 & 6.5  & \bfseries 98.4 & 1.3  & \bfseries 24.2 & 2.5 \\
            & LSA2-post  & 0.0 & 0.0  & 0.0 & 0.0  & 0.0 & 0.0  & 0.0 & 0.0  & 0.0 & 0.1  & 0.0 & 0.0 \\
            & LSA2-attr  & 4.6 & 3.3  & 1.2 & 0.9  & 4.7 & 0.9  & 1.2 & 0.4  & 6.9 & 0.7  & 1.7 & 0.1 \\[0.05em]\midrule[0.05em]
\rowcolor{tabgray}
\cellcolor{white}{\multirow{3}{*}{$k/2$}}
            & Ours  & \bfseries 99.0 & 1.5  & \bfseries 52.3 & 3.3  & \bfseries 93.6 & 1.4  & \bfseries 49.0 & 12.0  & \bfseries 97.3 & 1.6  & \bfseries 47.9 & 4.7 \\
            & LSA2-post  & 0.0 & 0.0  & 0.0 & 0.0  & 0.0 & 0.0  & 0.0 & 0.0  & 0.0 & 0.0  & 0.0 & 0.0 \\
            & LSA2-attr  & 2.4 & 1.7  & 1.2 & 0.9  & 4.4 & 0.8  & 2.3 & 0.8  & 6.3 & 1.0  & 3.0 & 0.3 \\[0.05em]\midrule[0.05em]
\rowcolor{tabgray}
\cellcolor{white}{\multirow{3}{*}{$k$}}
            & Ours  & \bfseries 85.4 & 2.2  & \bfseries 89.2 & 4.2  & \bfseries 78.7 & 8.7  & \bfseries 80.3 & 13.1  & \bfseries 86.0 & 5.8  & \bfseries 84.2 & 4.0 \\
            & LSA2-post  & 0.0 & 0.0  & 0.0 & 0.0  & 0.0 & 0.0  & 0.0 & 0.0  & 0.0 & 0.0  & 0.0 & 0.0 \\
            & LSA2-attr  & 1.2 & 0.9  & 1.2 & 0.9  & 4.3 & 0.9  & 4.7 & 1.7  & 6.0 & 0.8  & 5.9 & 0.5 \\[0.05em]\midrule[0.05em]
\rowcolor{tabgray}
\cellcolor{white}{\multirow{3}{*}{$2k$}}
            & Ours  & \bfseries 48.3 & 2.7  & \bfseries 100.0 & 0.0  & \bfseries 48.7 & 12.2  & \bfseries 95.7 & 4.1  & \bfseries 50.2 & 5.4  & \bfseries 97.9 & 0.4 \\
            & LSA2-post  & 0.0 & 0.0  & 0.0 & 0.0  & 0.0 & 0.0  & 0.0 & 0.0  & 2.4 & 1.7  & 4.7 & 3.2 \\
            & LSA2-attr  & 0.6 & 0.4  & 1.2 & 0.9  & 3.8 & 0.3  & 8.0 & 2.4  & 5.0 & 0.7  & 9.8 & 0.6 \\[0.05em]\midrule[0.05em]
\rowcolor{tabgray}
\cellcolor{white}{\multirow{3}{*}{$4k$}}
            & Ours  & \bfseries 24.1 & 1.4  & \bfseries 100.0 & 0.0  & \bfseries 25.5 & 7.5  & \bfseries 99.0 & 0.6  & \bfseries 25.5 & 2.8  & \bfseries 99.5 & 0.1 \\
            & LSA2-post  & 0.0 & 0.0  & 0.0 & 0.0  & 1.3 & 0.9  & 6.5 & 4.7  & 2.5 & 0.2  & 10.1 & 1.9 \\
            & LSA2-attr  & 0.6 & 0.2  & 2.4 & 0.7  & 3.1 & 0.2  & 13.2 & 3.6  & 4.1 & 0.4  & 15.8 & 0.5 \\\bottomrule
\end{tabular}%
}
\end{subtable}

\end{table}
}

First of all, we present the additional evaluation results for \linkteller on vanilla GCNs corresponding to~\Cref{sec:eval-attack}.
The results are presented in~\Cref{tab:att-np-file1}, which are of the same format as~\Cref{tab:att-np}.

Next, we show the comprehensive evaluation results on 
a combination of $2$ DP mechanisms (\edgerand and \lapgraph), $10$ privacy budgets ($1.0,2.0,\ldots,10.0$), $3$ sampling strategies (low degree, unconstrained degree, high degree), and $5$ density beliefs ($k/4, k/2, k,2k,4k$).
We present the results in~\Crefrange{tab:att_dp_twitch-RU-er}{tab:att_dp_twitch-PTBR-lap}.
The 3 subtables in each table correspond to the 3 sampling strategies.

In~\Cref{sec:dp-gcn} of the main paper, we present the results for density belief $\hat k=k$. 
Here, we look at the results for other inexact $\hat k$ values
and find that similar observations hold.
First, the effectiveness of \linkteller will decrease as a result of increasing privacy guarantee; while when the guarantee is not sufficient (\ie, $\eps$ is large), \linkteller is not weakened by much.
Second, DP can provide better protection to nodes of low degrees.
In addition, we note that our \linkteller is not sensitive to the density belief $\hat k$ and 
achieves non-negligible success rate for all $\hat k$.


\m{

\subsubsection{\linkteller in the Transductive Setting}
\label{sec:append-trans}

{
\setlength{\tabcolsep}{4pt} 

\begin{table}[]
    \centering
    \renewrobustcmd{\bfseries}{\fontseries{b}\selectfont}
    \sisetup{detect-weight,mode=text,group-minimum-digits = 4}

    \caption{\m{\small \textbf{Attack Performance (Precision and Recall)} of \linkteller on three datasets in the \textit{transductive} setting, compared with two baseline methods LSA2-\{post, attr\}~\cite{he2020stealing}. 
    We follow He \etal~\cite{he2020stealing} to compose a balanced dataset containing an equal number of connected and unconnected node pairs.
    Groups of rows represent different \textit{density belief} $\hat k$ of the attacker.}
    }
    \label{tab:att-np-trans}

\begin{subtable}[]{\columnwidth}
    \centering
    \vspace{0.5em}
    \resizebox{\columnwidth}{!}{%
        \begin{tabular}{cc
  S[table-format=3.1]
  @{\tiny${}\pm{}$}
  >{\tiny}S[table-format=2.1]<{\endcollectcell}
  S[table-format=3.1]
  @{\tiny${}\pm{}$}
  >{\tiny}S[table-format=2.1]<{\endcollectcell}
  S[table-format=3.1]
  @{\tiny${}\pm{}$}
  >{\tiny}S[table-format=2.1]<{\endcollectcell}
  S[table-format=3.1]
  @{\tiny${}\pm{}$}
  >{\tiny}S[table-format=2.1]<{\endcollectcell}
  S[table-format=3.1]
  @{\tiny${}\pm{}$}
  >{\tiny}S[table-format=2.1]<{\endcollectcell}
  S[table-format=3.1]
  @{\tiny${}\pm{}$}
  >{\tiny}S[table-format=2.1]<{\endcollectcell}
  S[table-format=3.1]
  @{\tiny${}\pm{}$}
  >{\tiny}S[table-format=2.1]<{\endcollectcell}
}
    \toprule
    \multicolumn{2}{c}{} &  \multicolumn{4}{c}{{Cora}}  & \multicolumn{4}{c}{{Citeseer}} & \multicolumn{4}{c}{{Pubmed}}\\
    \cmidrule(lr){3-6}\cmidrule(lr){7-10}\cmidrule(lr){11-14}
    \makecell{\small $\hat k$} & Method   & \multicolumn{2}{c}{\small precision}  & \multicolumn{2}{c}{\small recall} & \multicolumn{2}{c}{\small precision} & \multicolumn{2}{c}{\small recall} & \multicolumn{2}{c}{\small precision} & \multicolumn{2}{c}{\small recall} \\\midrule
\rowcolor{tabgray}

\cellcolor{white}{\multirow{3}{*}{$k/4$}}
            & Ours  & \bfseries 99.9 & 0.1  & \bfseries 25.0 & 0.0  & \bfseries 100.0 & 0.0  & \bfseries 25.0 & 0.0  & \bfseries 100.0 & 0.0  & \bfseries 25.0 & 0.0 \\
            & LSA2-post  & 96.7 & 0.2  & 24.2 & 0.0  & 98.8 & 0.1  & 24.7 & 0.0  & 89.9 & 0.2  & 22.5 & 0.1 \\
            & LSA2-feat  & 96.9 & 0.2  & 24.2 & 0.0  & 99.8 & 0.1  & 24.9 & 0.0  & 97.8 & 0.2  & 24.4 & 0.0 \\[0.05em]\midrule[0.05em]
\rowcolor{tabgray}

\cellcolor{white}{\multirow{3}{*}{$k/2$}}
            & Ours  & \bfseries 99.9 & 0.0  & \bfseries 50.0 & 0.0  & \bfseries 100.0 & 0.0  & \bfseries 50.0 & 0.0  & \bfseries 100.0 & 0.0  & \bfseries 50.0 & 0.0 \\
            & LSA2-post  & 94.1 & 0.3  & 47.0 & 0.1  & 96.7 & 0.0  & 48.4 & 0.0  & 86.8 & 0.1  & 43.4 & 0.0 \\
            & LSA2-feat  & 90.4 & 0.5  & 45.2 & 0.2  & 97.4 & 0.1  & 48.7 & 0.1  & 95.2 & 0.1  & 47.6 & 0.0 \\[0.05em]\midrule[0.05em]
\rowcolor{tabgray}

\cellcolor{white}{\multirow{3}{*}{$k$}}
            & Ours  & \bfseries 99.5 & 0.1  & \bfseries 99.5 & 0.1  & \bfseries 99.7 & 0.0  & \bfseries 99.7 & 0.0  & \bfseries 99.7 & 0.0  & \bfseries 99.7 & 0.0 \\
            & LSA2-post  & 86.7 & 0.2  & 86.7 & 0.2  & 90.1 & 0.2  & 90.1 & 0.2  & 78.8 & 0.1  & 78.8 & 0.1 \\
            & LSA2-feat  & 73.6 & 0.1  & 73.6 & 0.1  & 80.9 & 0.1  & 80.9 & 0.1  & 82.4 & 0.1  & 82.4 & 0.1 \\[0.05em]\midrule[0.05em]
\rowcolor{tabgray}

\cellcolor{white}{\multirow{3}{*}{$1.5k$}}
            & Ours  & \bfseries 66.7 & 0.0  & \bfseries 100.0 & 0.0  & \bfseries 66.7 & 0.0  & \bfseries 100.0 & 0.0  & \bfseries 66.6 & 0.0  & \bfseries 99.9 & 0.0 \\
            & LSA2-post  & 66.0 & 0.0  & 99.1 & 0.0  & 66.4 & 0.0  & 99.6 & 0.0  & 65.3 & 0.0  & 98.0 & 0.0 \\
            & LSA2-feat  & 59.9 & 0.2  & 89.8 & 0.2  & 63.2 & 0.1  & 94.7 & 0.2  & 64.0 & 0.0  & 96.0 & 0.0 \\\bottomrule
\end{tabular}%
}
\end{subtable}

\end{table}
}
{
\setlength{\tabcolsep}{4pt} 

\begin{table}[]
    \centering
    \renewrobustcmd{\bfseries}{\fontseries{b}\selectfont}
    \sisetup{detect-weight,mode=text,group-minimum-digits = 4}    
    \caption{\m{\small \textbf{AUC} of \linkteller comparing with two baselines LSA2-\{post, attr\}~\cite{he2020stealing} in the \textit{transductive} setting. Each column corresponds to one dataset. 
    Different rows represent different methods.}
    }
    \label{tab:att-np-trans-auc}
\begin{subtable}[]{.5\columnwidth}
    \centering
    \vspace{0.5em}
    \resizebox{\columnwidth}{!}{%
        \begin{tabular}{c
  S[table-format=1.2]
  @{\tiny${}\pm{}$}
  >{\tiny}S[table-format=1.2]<{\endcollectcell}
  S[table-format=1.2]
  @{\tiny${}\pm{}$}
  >{\tiny}S[table-format=1.2]<{\endcollectcell}
  S[table-format=1.2]
  @{\tiny${}\pm{}$}
  >{\tiny}S[table-format=1.2]<{\endcollectcell}
}
    \toprule
    Method & \multicolumn{6}{c}{Dataset}\\
    \cmidrule(lr){2-7}
         & \multicolumn{2}{c}{ {Cora}}  & \multicolumn{2}{c}{ {Citeseer}} & \multicolumn{2}{c}{ {Pubmed}} \\\midrule
\rowcolor{tabgray}
            Ours  & \bfseries 1.00 & 0.00  & \bfseries 1.00 & 0.00  & \bfseries 1.00 & 0.00 \\
            LSA2-post  & 0.93 & 0.00  & 0.96 & 0.00  & 0.87 & 0.00 \\
            LSA2-attr  & 0.81 & 0.00  & 0.89 & 0.00  & 0.90 & 0.00 \\\bottomrule
\end{tabular}%
}
\end{subtable}
\end{table}
}

In the main paper, we mainly evaluate the performance of \linkteller in the \textit{inductive} setting. 
As analyzed in~\Cref{sec:append-baseline-cmp}, in the \textit{transductive} setting, \linkteller is expected to achieve better performance and retain its advantage over LSA2-X.
Here, we aim to provide evaluations on the performance of \linkteller in the \textit{transductive} setting to support the analysis.

We compare \linkteller to LSA2-\{post, attr\}~\cite{he2020stealing} in the \textit{transductive} setting using three datasets (Cora, Citeseer, and Pubmed) from their paper.
We also follow the same setup (as in their Paragraph ``Datasets Configuration'' in Section 5.1) to compose the \textit{balanced} set of node pairs to be attacked which contains an equal number of connected and unconnected pairs.
We follow the hyper-parameters in Kipf \etal~\cite{Kipf2016gcn} to train the GCN models on these datasets and then perform \linkteller attack and LSA2-\{post, attr\} attacks on the trained models.

We report the attack performance (Precision and Recall) in~\Cref{tab:att-np-trans} and the AUC scores in~\Cref{tab:att-np-trans-auc}.
\textbf{First}, as a sanity check, our results in~\Cref{tab:att-np-trans-auc} matches the Figure 4 in He \etal~\cite{he2020stealing} on these three datasets.
\textbf{Second}, we evaluate the density belief $\hat k$ only up to $1.5k$ in~\Cref{tab:att-np-trans}, since $2k$ corresponds to the case where all node pairs are predicted positive by the attacker, leading to $50\%$ precision and $100\%$ recall for \textit{all} methods.
\textbf{Overall}, as shown in the two tables, \linkteller invariably outperforms LSA2-\{post, attr\} in the transductive setting.

}

\m{

\subsubsection{Choosing $\eps$ on a Validation Dataset}
\label{sec:append-choice}

{
\setlength{\tabcolsep}{4pt} 

\begin{table}[]
    \centering
    \renewrobustcmd{\bfseries}{\fontseries{b}\selectfont}
    \sisetup{detect-weight,mode=text,group-minimum-digits = 4}    
    \caption{\m{\small (a) \textbf{Model utility} and (b) \textbf{attack effectiveness} on different models. 
    Each column corresponds to a dataset. We consider four types of models: vanilla GCN, MLP, \edgerand, and \lapgraph.}}
    \label{tab:chc-dp}
\begin{subtable}[]{\columnwidth}
    \centering
    \caption{\small \m{\textbf{Model utility} (F1 score)}}\label{tab:chc-dp-acc}
    \resizebox{\columnwidth}{!}{%
        \begin{tabular}{c
  S[table-format=1.3]
  @{\tiny${}\pm{}$}
  >{\tiny}S[table-format=1.3]<{\endcollectcell}
  S[table-format=1.3]
  @{\tiny${}\pm{}$}
  >{\tiny}S[table-format=1.3]<{\endcollectcell}
  S[table-format=1.3]
  @{\tiny${}\pm{}$}
  >{\tiny}S[table-format=1.3]<{\endcollectcell}
  S[table-format=1.3]
  @{\tiny${}\pm{}$}
  >{\tiny}S[table-format=1.3]<{\endcollectcell}
  S[table-format=1.3]
  @{\tiny${}\pm{}$}
  >{\tiny}S[table-format=1.3]<{\endcollectcell}
  S[table-format=1.3]
  @{\tiny${}\pm{}$}
  >{\tiny}S[table-format=1.3]<{\endcollectcell}
}
    \toprule
    Model & \multicolumn{12}{c}{Dataset}\\
    \cmidrule(lr){2-13}
    & \multicolumn{2}{c}{RU}  & \multicolumn{2}{c}{DE} & \multicolumn{2}{c}{FR} & \multicolumn{2}{c}{ENGB} & \multicolumn{2}{c}{PTBR} & \multicolumn{2}{c}{Flickr}  \\\midrule
{\footnotesize vanilla GCN}
                & \multicolumn{2}{c}{0.319}  & \multicolumn{2}{c}{0.551}  & \multicolumn{2}{c}{0.404}  & \multicolumn{2}{c}{0.601} & \multicolumn{2}{c}{0.411}  & \multicolumn{2}{c}{0.515}  \\
                \cmidrule(lr){2-13}
{\footnotesize MLP }
                & \multicolumn{2}{c}{0.290}  & \multicolumn{2}{c}{0.545}  & \multicolumn{2}{c}{0.373}  & \multicolumn{2}{c}{0.598} & \multicolumn{2}{c}{0.358}  & \multicolumn{2}{c}{0.463}  \\
                \cmidrule(lr){2-11} \cmidrule(lr){12-13}
\multirow{2}{*}{\footnotesize \edgerand}
                & \multicolumn{10}{c}{$\eps=7$} & \multicolumn{2}{c}{$\eps=6$} \\
                 & 0.299  & 0.006  & 0.545  & 0.003  & 0.321  & 0.027  & 0.607  & 0.000  & 0.423  & 0.018  & 0.459  & 0.002   \\
                \cmidrule(lr){2-11} \cmidrule(lr){12-13}
\multirow{2}{*}{\footnotesize \lapgraph }
                & \multicolumn{10}{c}{$\eps=8$} & \multicolumn{2}{c}{$\eps=9$} \\
                 & 0.292  & 0.011  & 0.546  & 0.001  & 0.299  & 0.017  & 0.601  & 0.001  & 0.401  & 0.010  & 0.467  & 0.002   \\
            \bottomrule\\[0.5em]
\end{tabular}%
}
\end{subtable}
\begin{subtable}[]{\columnwidth}
    \centering
    \caption{\small \m{\textbf{Attack effectiveness} (F1 score) on different node degree distributions}}\label{tab:chc-dp-att}
    \resizebox{\columnwidth}{!}{%
        \begin{tabular}{cc
  S[table-format=2.1]
  @{\tiny${}\pm{}$}
  >{\tiny}S[table-format=2.1]<{\endcollectcell}
  S[table-format=2.1]
  @{\tiny${}\pm{}$}
  >{\tiny}S[table-format=2.1]<{\endcollectcell}
  S[table-format=2.1]
  @{\tiny${}\pm{}$}
  >{\tiny}S[table-format=2.1]<{\endcollectcell}
  S[table-format=2.1]
  @{\tiny${}\pm{}$}
  >{\tiny}S[table-format=2.1]<{\endcollectcell}
  S[table-format=2.1]
  @{\tiny${}\pm{}$}
  >{\tiny}S[table-format=2.1]<{\endcollectcell}
  S[table-format=2.1]
  @{\tiny${}\pm{}$}
  >{\tiny}S[table-format=2.1]<{\endcollectcell}
}
    \toprule
 Degree &   Model & \multicolumn{12}{c}{Dataset}\\
    \cmidrule(lr){3-14}
&    & \multicolumn{2}{c}{RU}  & \multicolumn{2}{c}{DE} & \multicolumn{2}{c}{FR} & \multicolumn{2}{c}{ENGB} & \multicolumn{2}{c}{PTBR} & \multicolumn{2}{c}{Flickr}  \\\midrule
{\footnotesize \multirow{3}{*}{low}}
& vanilla GCN &  84.9 &  1.2 &  86.8 &  5.1 &  92.5 &  5.4 &  82.9 &  4.9 &  86.6 &  1.3 &  52.1 &  5.8 \\
& \edgerand &  18.9 &  10.8 &  4.4 &  6.3 &  0.0 &  0.0 &  19.0 &  12.0 &  32.9 &  2.5 &  0.0 &  0.0 \\
& \lapgraph &  22.9 &  3.3 &  5.3 &  7.5 &  13.3 &  12.5 &  22.0 &  1.0 &  23.8 &  2.2 &  26.8 &  10.1 \\
[0.05em]\midrule[0.05em]
{\footnotesize \multirow{3}{*}{\makecell{uncon-\\strained}}}
& vanilla GCN &  74.7 &  1.5 &  78.5 &  4.5 &  80.9 &  2.0 &  69.5 &  2.5 &  77.9 &  3.5 &  32.9 &  13.3 \\
& \edgerand &  58.1 &  2.2 &  60.1 &  5.0 &  67.1 &  4.6 &  41.6 &  8.1 &  73.2 &  5.1 &  0.0 &  0.0 \\
& \lapgraph &  59.6 &  1.2 &  59.6 &  2.0 &  67.1 &  2.9 &  46.9 &  3.7 &  68.4 &  5.8 &  2.4 &  3.4 \\
[0.05em]\midrule[0.05em]
{\footnotesize \multirow{3}{*}{high} }
& vanilla GCN &  75.8 &  2.3 &  78.9 &  0.8 &  83.0 &  3.7 &  82.2 &  3.4 &  84.8 &  1.6 &  18.3 &  5.2 \\
& \edgerand &  72.6 &  1.5 &  76.5 &  1.8 &  78.1 &  2.7 &  82.3 &  1.5 &  83.7 &  1.0 &  16.9 &  2.9 \\
& \lapgraph &  69.6 &  1.0 &  68.5 &  1.1 &  73.4 &  2.8 &  68.0 &  1.7 &  78.4 &  1.4 &  15.7 &  2.6 \\
            \bottomrule
\end{tabular}%
}
\end{subtable}
\end{table}
}

In~\Cref{sec:dp-gcn}, we present the model utility and attack effectiveness under a range of $\eps$ (1-10) in~\Cref{fig:dp-attack}, and provide corresponding discussions regarding the tradeoff between model utility and privacy as well as its dependencies
in~\Cref{sec:eval-dp-util} and~\Cref{sec:tradeoff}.

In this part, we take on a new perspective and focus on a practical approach for selecting the appropriate parameters (mainly the privacy budget $\eps$) to train DP GCN models with reasonable model utility--we select the parameter $\eps$ on the validation dataset, and then report the final performance on the testing set. 
Concretely, when selecting $\eps$ on the \textit{validation set}, we set a lower threshold (\eg, F1 score of the MLP model) for the model utility, and select the smallest $\eps$ satisfying the condition that the utility of the trained DP GCN on the validation set is higher than the given threshold.
Then, we evaluate both the model utility and attack effectiveness under the selected $\varepsilon$ on the \textit{testing set}.

We report the evaluation results in~\Cref{tab:chc-dp}.
We do not include results for the PPI dataset, since none of the DP GCN models could attain utility higher than the MLP model. (In fact, even the vanilla GCN model cannot match the performance of the MLP model, which have been concretely discussed in~\Cref{sec:tradeoff}.)
For the twitch datasets and Flickr dataset, we select $\eps \in \{1.0,2.0,\ldots, 10.0\}$ on the validation dataset and include the numbers in~\Cref{tab:chc-dp}(a).
Specifically, for the twitch datasets, the model is trained on twitch-ES (with the parameters selected on the validation set) and then directly transferred to twitch-\{RU, DE, FR, ENGB, PTBR\}, so we report only one number for all five countries.

We next discuss the conclusions from~\Cref{tab:chc-dp}.
\textbf{First}, regarding the model utility in~\Cref{tab:chc-dp}(a), with the same constraint on model utility on the same dataset, we end up with different $\eps$ for different DP mechanisms, \eg, $\eps=6$ for \edgerand and $\eps=9$ for \lapgraph on the Flickr dataset. This implies that the level of noise that can be tolerated by different DP mechanisms is different.
\m{\textbf{Second}, in terms of the attack effectiveness in~\Cref{tab:chc-dp}(b), the main conclusion is that the theoretical guarantee provided by DP cannot translate into sufficient protection against \linkteller while maintaining a reasonable level of model utility. Specifically, with $\eps=6$ for \edgerand and $\eps=9$ for \lapgraph, the theoretical upper bound of precision is greater than 1 for all the datasets based on~\Cref{thm:dp-bound}, which means that the DP GCN methods do \emph{not} provide any reasonable \emph{theoretical} privacy protection against \linkteller. \textbf{However}, the attack effectiveness presented in~\Cref{tab:chc-dp}(b) suggests that, in low degree settings, differential privacy can \emph{empirically} protect against \linkteller although} the level of protection is heavily data-dependent, varying a lot across different \textit{datasets} and different \textit{node degree distributions}.
Relevant discussions on the impact of node degree distributions specifically are provided in~\Cref{sec:tradeoff}.

}

{

\setlength{\tabcolsep}{4pt} 

\begin{table}
    \centering
    \caption{twitch-RU (\edgerand)}
    \label{tab:att_dp_twitch-RU-er}    
    \begin{subtable}[]{\linewidth}
    \centering
    \vspace{0.5em}
    \resizebox{\linewidth}{!}{%
\begin{tabular}{cc
  >{\collectcell\num}R{2.18em}<{\endcollectcell}
  @{\tiny${}\pm{}$}
  >{\tiny\collectcell\num}R{1.308em}<{\endcollectcell}
  >{\collectcell\num}R{2.18em}<{\endcollectcell}
  @{\tiny${}\pm{}$}
  >{\tiny\collectcell\num}R{1.308em}<{\endcollectcell}
  >{\collectcell\num}R{2.18em}<{\endcollectcell}
  @{\tiny${}\pm{}$}
  >{\tiny\collectcell\num}R{1.308em}<{\endcollectcell}
  >{\collectcell\num}R{2.18em}<{\endcollectcell}
  @{\tiny${}\pm{}$}
  >{\tiny\collectcell\num}R{1.308em}<{\endcollectcell}
  >{\collectcell\num}R{2.18em}<{\endcollectcell}
  @{\tiny${}\pm{}$}
  >{\tiny\collectcell\num}R{1.308em}<{\endcollectcell}
}
    \toprule
\multicolumn{2}{c}{\footnotesize \textbf{low}}&  \multicolumn{2}{c}{\scriptsize $\hat k=k/4$}  
    &  \multicolumn{2}{c}{\scriptsize $\hat k=k/2$}  
    &  \multicolumn{2}{c}{\scriptsize $\hat k=k$}  
    &  \multicolumn{2}{c}{\scriptsize $\hat k=2k$}  
    &  \multicolumn{2}{c}{\scriptsize $\hat k=4k$}  \\\midrule
\multirow{2}{*}{$\varepsilon=1$} & \multicolumn{1}{c}{\scriptsize precision} & 0.0 & 0.0 & 0.0 & 0.0 & 0.0 & 0.0 & 0.0 & 0.0 & 0.0 & 0.0\\
& \multicolumn{1}{c}{\scriptsize recall} & 0.0 & 0.0 & 0.0 & 0.0 & 0.0 & 0.0 & 0.0 & 0.0 & 0.0 & 0.0\\\midrule
\multirow{2}{*}{$\varepsilon=2$} & \multicolumn{1}{c}{\scriptsize precision} & 0.0 & 0.0 & 0.0 & 0.0 & 0.0 & 0.0 & 0.0 & 0.0 & 0.0 & 0.0\\
& \multicolumn{1}{c}{\scriptsize recall} & 0.0 & 0.0 & 0.0 & 0.0 & 0.0 & 0.0 & 0.0 & 0.0 & 0.0 & 0.0\\\midrule
\multirow{2}{*}{$\varepsilon=3$} & \multicolumn{1}{c}{\scriptsize precision} & 0.0 & 0.0 & 0.0 & 0.0 & 0.0 & 0.0 & 0.0 & 0.0 & 0.0 & 0.0\\
& \multicolumn{1}{c}{\scriptsize recall} & 0.0 & 0.0 & 0.0 & 0.0 & 0.0 & 0.0 & 0.0 & 0.0 & 0.0 & 0.0\\\midrule
\multirow{2}{*}{$\varepsilon=4$} & \multicolumn{1}{c}{\scriptsize precision} & 0.0 & 0.0 & 0.0 & 0.0 & 1.3 & 1.9 & 1.3 & 1.9 & 1.3 & 0.5\\
\bottomrule
    \end{tabular}%
}

\end{subtable}

    \begin{subtable}[]{\linewidth}
    \centering
    \vspace{0.5em}
    \resizebox{\linewidth}{!}{%
        \begin{tabular}{cc
  >{\collectcell\num}R{2.18em}<{\endcollectcell}
  @{\tiny${}\pm{}$}
  >{\tiny\collectcell\num}R{1.308em}<{\endcollectcell}
  >{\collectcell\num}R{2.18em}<{\endcollectcell}
  @{\tiny${}\pm{}$}
  >{\tiny\collectcell\num}R{1.308em}<{\endcollectcell}
  >{\collectcell\num}R{2.18em}<{\endcollectcell}
  @{\tiny${}\pm{}$}
  >{\tiny\collectcell\num}R{1.308em}<{\endcollectcell}
  >{\collectcell\num}R{2.18em}<{\endcollectcell}
  @{\tiny${}\pm{}$}
  >{\tiny\collectcell\num}R{1.308em}<{\endcollectcell}
  >{\collectcell\num}R{2.18em}<{\endcollectcell}
  @{\tiny${}\pm{}$}
  >{\tiny\collectcell\num}R{1.308em}<{\endcollectcell}
}
    \toprule
\multicolumn{2}{c}{\footnotesize \textbf{unconstrained}}&  \multicolumn{2}{c}{\scriptsize $\hat k=k/4$}  
    &  \multicolumn{2}{c}{\scriptsize $\hat k=k/2$}  
    &  \multicolumn{2}{c}{\scriptsize $\hat k=k$}  
    &  \multicolumn{2}{c}{\scriptsize $\hat k=2k$}  
    &  \multicolumn{2}{c}{\scriptsize $\hat k=4k$}  \\\midrule
\multirow{2}{*}{$\varepsilon=1$} & \multicolumn{1}{c}{\scriptsize precision} & 0.7 & 0.1 & 0.6 & 0.1 & 1.2 & 0.4 & 1.6 & 0.6 & 1.5 & 0.5\\
& \multicolumn{1}{c}{\scriptsize recall} & 0.2 & 0.0 & 0.3 & 0.1 & 1.3 & 0.4 & 3.4 & 1.3 & 6.4 & 1.9\\\midrule
\multirow{2}{*}{$\varepsilon=2$} & \multicolumn{1}{c}{\scriptsize precision} & 1.1 & 1.5 & 2.1 & 1.3 & 2.3 & 1.2 & 2.1 & 0.7 & 2.1 & 0.6\\
& \multicolumn{1}{c}{\scriptsize recall} & 0.3 & 0.4 & 1.2 & 0.7 & 2.5 & 1.3 & 4.5 & 1.6 & 9.2 & 2.2\\\midrule
\multirow{2}{*}{$\varepsilon=3$} & \multicolumn{1}{c}{\scriptsize precision} & 5.2 & 4.3 & 5.8 & 2.6 & 7.2 & 1.8 & 6.4 & 1.7 & 6.4 & 1.5\\
& \multicolumn{1}{c}{\scriptsize recall} & 1.4 & 1.1 & 3.1 & 1.3 & 7.8 & 1.7 & 13.8 & 3.1 & 27.8 & 5.3\\\midrule
\multirow{2}{*}{$\varepsilon=4$} & \multicolumn{1}{c}{\scriptsize precision} & 5.5 & 2.7 & 7.5 & 2.2 & 9.0 & 2.0 & 11.6 & 2.0 & 15.4 & 2.3\\
\bottomrule
    \end{tabular}%
}

\end{subtable}

    \begin{subtable}[]{\linewidth}
    \centering     \vspace{0.5em}
    \resizebox{\linewidth}{!}{%
        \begin{tabular}{cc
  >{\collectcell\num}R{2.18em}<{\endcollectcell}
  @{\tiny${}\pm{}$}
  >{\tiny\collectcell\num}R{1.308em}<{\endcollectcell}
  >{\collectcell\num}R{2.18em}<{\endcollectcell}
  @{\tiny${}\pm{}$}
  >{\tiny\collectcell\num}R{1.308em}<{\endcollectcell}
  >{\collectcell\num}R{2.18em}<{\endcollectcell}
  @{\tiny${}\pm{}$}
  >{\tiny\collectcell\num}R{1.308em}<{\endcollectcell}
  >{\collectcell\num}R{2.18em}<{\endcollectcell}
  @{\tiny${}\pm{}$}
  >{\tiny\collectcell\num}R{1.308em}<{\endcollectcell}
  >{\collectcell\num}R{2.18em}<{\endcollectcell}
  @{\tiny${}\pm{}$}
  >{\tiny\collectcell\num}R{1.308em}<{\endcollectcell}
}
    \toprule
\multicolumn{2}{c}{\footnotesize \textbf{high}}&  \multicolumn{2}{c}{\scriptsize $\hat k=k/4$}  
    &  \multicolumn{2}{c}{\scriptsize $\hat k=k/2$}  
    &  \multicolumn{2}{c}{\scriptsize $\hat k=k$}  
    &  \multicolumn{2}{c}{\scriptsize $\hat k=2k$}  
    &  \multicolumn{2}{c}{\scriptsize $\hat k=4k$}  \\\midrule
\multirow{2}{*}{$\varepsilon=1$} & \multicolumn{1}{c}{\scriptsize precision} & 15.6 & 9.1 & 11.2 & 4.6 & 9.2 & 3.7 & 7.1 & 2.5 & 5.5 & 1.4\\
& \multicolumn{1}{c}{\scriptsize recall} & 2.8 & 1.4 & 4.0 & 1.4 & 6.6 & 2.2 & 10.2 & 3.0 & 15.7 & 3.0\\\midrule
\multirow{2}{*}{$\varepsilon=2$} & \multicolumn{1}{c}{\scriptsize precision} & 22.3 & 11.9 & 15.5 & 5.8 & 12.4 & 3.0 & 11.3 & 2.2 & 10.0 & 1.5\\
& \multicolumn{1}{c}{\scriptsize recall} & 4.0 & 1.9 & 5.6 & 1.7 & 9.0 & 1.6 & 16.4 & 2.1 & 28.9 & 2.5\\\midrule
\multirow{2}{*}{$\varepsilon=3$} & \multicolumn{1}{c}{\scriptsize precision} & 27.5 & 7.7 & 25.3 & 4.9 & 24.6 & 5.4 & 22.6 & 3.2 & 20.3 & 2.3\\
& \multicolumn{1}{c}{\scriptsize recall} & 5.0 & 1.1 & 9.2 & 1.3 & 17.8 & 2.7 & 32.9 & 2.6 & 59.2 & 3.7\\\midrule
\multirow{2}{*}{$\varepsilon=4$} & \multicolumn{1}{c}{\scriptsize precision} & 38.7 & 8.9 & 36.9 & 5.9 & 37.6 & 4.5 & 39.4 & 3.4 & 29.8 & 1.9\\
\bottomrule
    \end{tabular}%
}

\end{subtable}
\end{table}
\newpage

\begin{table}
    \centering
    \caption{twitch-RU (\lapgraph)}
    \label{tab:att_dp_twitch-RU-lap}
    \begin{subtable}[]{\linewidth}
    \centering
    \vspace{0.5em}
    \resizebox{\linewidth}{!}{%
        \begin{tabular}{cc
  >{\collectcell\num}R{2.18em}<{\endcollectcell}
  @{\tiny${}\pm{}$}
  >{\tiny\collectcell\num}R{1.308em}<{\endcollectcell}
  >{\collectcell\num}R{2.18em}<{\endcollectcell}
  @{\tiny${}\pm{}$}
  >{\tiny\collectcell\num}R{1.308em}<{\endcollectcell}
  >{\collectcell\num}R{2.18em}<{\endcollectcell}
  @{\tiny${}\pm{}$}
  >{\tiny\collectcell\num}R{1.308em}<{\endcollectcell}
  >{\collectcell\num}R{2.18em}<{\endcollectcell}
  @{\tiny${}\pm{}$}
  >{\tiny\collectcell\num}R{1.308em}<{\endcollectcell}
  >{\collectcell\num}R{2.18em}<{\endcollectcell}
  @{\tiny${}\pm{}$}
  >{\tiny\collectcell\num}R{1.308em}<{\endcollectcell}
}
    \toprule
\multicolumn{2}{c}{\footnotesize \textbf{low}}&  \multicolumn{2}{c}{\scriptsize $\hat k=k/4$}  
    &  \multicolumn{2}{c}{\scriptsize $\hat k=k/2$}  
    &  \multicolumn{2}{c}{\scriptsize $\hat k=k$}  
    &  \multicolumn{2}{c}{\scriptsize $\hat k=2k$}  
    &  \multicolumn{2}{c}{\scriptsize $\hat k=4k$}  \\\midrule
\multirow{2}{*}{$\varepsilon=1$} & \multicolumn{1}{c}{\scriptsize precision} & 0.0 & 0.0 & 0.0 & 0.0 & 0.0 & 0.0 & 0.0 & 0.0 & 0.0 & 0.0\\
& \multicolumn{1}{c}{\scriptsize recall} & 0.0 & 0.0 & 0.0 & 0.0 & 0.0 & 0.0 & 0.0 & 0.0 & 0.0 & 0.0\\\midrule
\multirow{2}{*}{$\varepsilon=2$} & \multicolumn{1}{c}{\scriptsize precision} & 0.0 & 0.0 & 0.0 & 0.0 & 0.0 & 0.0 & 0.0 & 0.0 & 0.0 & 0.0\\
& \multicolumn{1}{c}{\scriptsize recall} & 0.0 & 0.0 & 0.0 & 0.0 & 0.0 & 0.0 & 0.0 & 0.0 & 0.0 & 0.0\\\midrule
\multirow{2}{*}{$\varepsilon=3$} & \multicolumn{1}{c}{\scriptsize precision} & 0.0 & 0.0 & 0.0 & 0.0 & 0.0 & 0.0 & 0.0 & 0.0 & 0.0 & 0.0\\
& \multicolumn{1}{c}{\scriptsize recall} & 0.0 & 0.0 & 0.0 & 0.0 & 0.0 & 0.0 & 0.0 & 0.0 & 0.0 & 0.0\\\midrule
\multirow{2}{*}{$\varepsilon=4$} & \multicolumn{1}{c}{\scriptsize precision} & 0.0 & 0.0 & 0.0 & 0.0 & 0.0 & 0.0 & 0.0 & 0.0 & 0.3 & 0.5\\
\bottomrule
    \end{tabular}%
}
\end{subtable}

    \begin{subtable}[]{\linewidth}
    \centering\vspace{0.5em}
    \resizebox{\linewidth}{!}{%
        \begin{tabular}{cc
  >{\collectcell\num}R{2.18em}<{\endcollectcell}
  @{\tiny${}\pm{}$}
  >{\tiny\collectcell\num}R{1.308em}<{\endcollectcell}
  >{\collectcell\num}R{2.18em}<{\endcollectcell}
  @{\tiny${}\pm{}$}
  >{\tiny\collectcell\num}R{1.308em}<{\endcollectcell}
  >{\collectcell\num}R{2.18em}<{\endcollectcell}
  @{\tiny${}\pm{}$}
  >{\tiny\collectcell\num}R{1.308em}<{\endcollectcell}
  >{\collectcell\num}R{2.18em}<{\endcollectcell}
  @{\tiny${}\pm{}$}
  >{\tiny\collectcell\num}R{1.308em}<{\endcollectcell}
  >{\collectcell\num}R{2.18em}<{\endcollectcell}
  @{\tiny${}\pm{}$}
  >{\tiny\collectcell\num}R{1.308em}<{\endcollectcell}
}
    \toprule
\multicolumn{2}{c}{\footnotesize \textbf{unconstrained}}&  \multicolumn{2}{c}{\scriptsize $\hat k=k/4$}  
    &  \multicolumn{2}{c}{\scriptsize $\hat k=k/2$}  
    &  \multicolumn{2}{c}{\scriptsize $\hat k=k$}  
    &  \multicolumn{2}{c}{\scriptsize $\hat k=2k$}  
    &  \multicolumn{2}{c}{\scriptsize $\hat k=4k$}  \\\midrule
\multirow{2}{*}{$\varepsilon=1$} & \multicolumn{1}{c}{\scriptsize precision} & 1.8 & 0.7 & 1.8 & 0.4 & 1.2 & 0.2 & 0.9 & 0.1 & 0.7 & 0.1\\
& \multicolumn{1}{c}{\scriptsize recall} & 0.5 & 0.2 & 1.0 & 0.2 & 1.3 & 0.2 & 2.0 & 0.3 & 2.9 & 0.3\\\midrule
\multirow{2}{*}{$\varepsilon=2$} & \multicolumn{1}{c}{\scriptsize precision} & 3.9 & 0.1 & 3.1 & 0.4 & 2.4 & 0.7 & 1.4 & 0.4 & 1.0 & 0.1\\
& \multicolumn{1}{c}{\scriptsize recall} & 1.1 & 0.1 & 1.7 & 0.3 & 2.6 & 0.9 & 3.1 & 1.1 & 4.5 & 0.4\\\midrule
\multirow{2}{*}{$\varepsilon=3$} & \multicolumn{1}{c}{\scriptsize precision} & 6.4 & 1.6 & 7.6 & 1.3 & 5.9 & 0.7 & 3.4 & 0.5 & 2.1 & 0.3\\
& \multicolumn{1}{c}{\scriptsize recall} & 1.7 & 0.4 & 4.2 & 0.7 & 6.5 & 0.9 & 7.5 & 1.4 & 9.1 & 1.5\\\midrule
\multirow{2}{*}{$\varepsilon=4$} & \multicolumn{1}{c}{\scriptsize precision} & 12.4 & 2.6 & 13.2 & 1.6 & 13.1 & 1.5 & 7.6 & 0.8 & 4.3 & 0.3\\
\bottomrule
    \end{tabular}%
}
\end{subtable}

    \begin{subtable}[]{\linewidth}
    \centering\vspace{0.5em}
    \resizebox{\linewidth}{!}{%
        \begin{tabular}{cc
  >{\collectcell\num}R{2.18em}<{\endcollectcell}
  @{\tiny${}\pm{}$}
  >{\tiny\collectcell\num}R{1.308em}<{\endcollectcell}
  >{\collectcell\num}R{2.18em}<{\endcollectcell}
  @{\tiny${}\pm{}$}
  >{\tiny\collectcell\num}R{1.308em}<{\endcollectcell}
  >{\collectcell\num}R{2.18em}<{\endcollectcell}
  @{\tiny${}\pm{}$}
  >{\tiny\collectcell\num}R{1.308em}<{\endcollectcell}
  >{\collectcell\num}R{2.18em}<{\endcollectcell}
  @{\tiny${}\pm{}$}
  >{\tiny\collectcell\num}R{1.308em}<{\endcollectcell}
  >{\collectcell\num}R{2.18em}<{\endcollectcell}
  @{\tiny${}\pm{}$}
  >{\tiny\collectcell\num}R{1.308em}<{\endcollectcell}
}
    \toprule
\multicolumn{2}{c}{\footnotesize \textbf{high}}&  \multicolumn{2}{c}{\scriptsize $\hat k=k/4$}  
    &  \multicolumn{2}{c}{\scriptsize $\hat k=k/2$}  
    &  \multicolumn{2}{c}{\scriptsize $\hat k=k$}  
    &  \multicolumn{2}{c}{\scriptsize $\hat k=2k$}  
    &  \multicolumn{2}{c}{\scriptsize $\hat k=4k$}  \\\midrule
\multirow{2}{*}{$\varepsilon=1$} & \multicolumn{1}{c}{\scriptsize precision} & 3.6 & 0.5 & 3.3 & 1.1 & 2.5 & 0.7 & 2.0 & 0.3 & 1.8 & 0.2\\
& \multicolumn{1}{c}{\scriptsize recall} & 0.7 & 0.1 & 1.2 & 0.3 & 1.8 & 0.4 & 2.9 & 0.2 & 5.3 & 0.3\\\midrule
\multirow{2}{*}{$\varepsilon=2$} & \multicolumn{1}{c}{\scriptsize precision} & 10.5 & 1.2 & 8.5 & 1.9 & 5.3 & 1.1 & 3.6 & 0.6 & 2.5 & 0.3\\
& \multicolumn{1}{c}{\scriptsize recall} & 1.9 & 0.2 & 3.1 & 0.5 & 3.9 & 0.6 & 5.2 & 0.5 & 7.4 & 0.6\\\midrule
\multirow{2}{*}{$\varepsilon=3$} & \multicolumn{1}{c}{\scriptsize precision} & 22.2 & 3.5 & 19.6 & 2.0 & 11.5 & 1.6 & 6.9 & 1.1 & 4.3 & 0.6\\
& \multicolumn{1}{c}{\scriptsize recall} & 4.1 & 0.6 & 7.1 & 0.7 & 8.3 & 0.9 & 10.1 & 1.1 & 12.7 & 1.2\\\midrule
\multirow{2}{*}{$\varepsilon=4$} & \multicolumn{1}{c}{\scriptsize precision} & 39.3 & 3.9 & 38.4 & 3.4 & 23.6 & 3.2 & 13.3 & 1.9 & 8.3 & 1.1\\
\bottomrule
    \end{tabular}%
}
\end{subtable}
\end{table}
\newpage

\begin{table}
    \centering
    \caption{twitch-DE (\edgerand)}
    \label{tab:att_dp_twitch-DE-er}    
    \begin{subtable}[]{\linewidth}
    \centering
    \vspace{0.5em}
    \resizebox{\linewidth}{!}{%
        \begin{tabular}{cc
  >{\collectcell\num}R{2.18em}<{\endcollectcell}
  @{\tiny${}\pm{}$}
  >{\tiny\collectcell\num}R{1.308em}<{\endcollectcell}
  >{\collectcell\num}R{2.18em}<{\endcollectcell}
  @{\tiny${}\pm{}$}
  >{\tiny\collectcell\num}R{1.308em}<{\endcollectcell}
  >{\collectcell\num}R{2.18em}<{\endcollectcell}
  @{\tiny${}\pm{}$}
  >{\tiny\collectcell\num}R{1.308em}<{\endcollectcell}
  >{\collectcell\num}R{2.18em}<{\endcollectcell}
  @{\tiny${}\pm{}$}
  >{\tiny\collectcell\num}R{1.308em}<{\endcollectcell}
  >{\collectcell\num}R{2.18em}<{\endcollectcell}
  @{\tiny${}\pm{}$}
  >{\tiny\collectcell\num}R{1.308em}<{\endcollectcell}
}
    \toprule
\multicolumn{2}{c}{\footnotesize \textbf{low}}&  \multicolumn{2}{c}{\scriptsize $\hat k=k/4$}  
    &  \multicolumn{2}{c}{\scriptsize $\hat k=k/2$}  
    &  \multicolumn{2}{c}{\scriptsize $\hat k=k$}  
    &  \multicolumn{2}{c}{\scriptsize $\hat k=2k$}  
    &  \multicolumn{2}{c}{\scriptsize $\hat k=4k$}  \\\midrule
\multirow{2}{*}{$\varepsilon=1$} & \multicolumn{1}{c}{\scriptsize precision} & 0.0 & 0.0 & 0.0 & 0.0 & 0.0 & 0.0 & 0.0 & 0.0 & 0.0 & 0.0\\
& \multicolumn{1}{c}{\scriptsize recall} & 0.0 & 0.0 & 0.0 & 0.0 & 0.0 & 0.0 & 0.0 & 0.0 & 0.0 & 0.0\\\midrule
\multirow{2}{*}{$\varepsilon=2$} & \multicolumn{1}{c}{\scriptsize precision} & 0.0 & 0.0 & 0.0 & 0.0 & 0.0 & 0.0 & 0.0 & 0.0 & 0.0 & 0.0\\
& \multicolumn{1}{c}{\scriptsize recall} & 0.0 & 0.0 & 0.0 & 0.0 & 0.0 & 0.0 & 0.0 & 0.0 & 0.0 & 0.0\\\midrule
\multirow{2}{*}{$\varepsilon=3$} & \multicolumn{1}{c}{\scriptsize precision} & 0.0 & 0.0 & 0.0 & 0.0 & 0.0 & 0.0 & 0.0 & 0.0 & 0.0 & 0.0\\
& \multicolumn{1}{c}{\scriptsize recall} & 0.0 & 0.0 & 0.0 & 0.0 & 0.0 & 0.0 & 0.0 & 0.0 & 0.0 & 0.0\\\midrule
\multirow{2}{*}{$\varepsilon=4$} & \multicolumn{1}{c}{\scriptsize precision} & 0.0 & 0.0 & 0.0 & 0.0 & 0.0 & 0.0 & 0.0 & 0.0 & 0.0 & 0.0\\
\bottomrule
    \end{tabular}%
}

\end{subtable}

    \begin{subtable}[]{\linewidth}
    \centering
    \vspace{0.5em}
    \resizebox{\linewidth}{!}{%
        \begin{tabular}{cc
  >{\collectcell\num}R{2.18em}<{\endcollectcell}
  @{\tiny${}\pm{}$}
  >{\tiny\collectcell\num}R{1.308em}<{\endcollectcell}
  >{\collectcell\num}R{2.18em}<{\endcollectcell}
  @{\tiny${}\pm{}$}
  >{\tiny\collectcell\num}R{1.308em}<{\endcollectcell}
  >{\collectcell\num}R{2.18em}<{\endcollectcell}
  @{\tiny${}\pm{}$}
  >{\tiny\collectcell\num}R{1.308em}<{\endcollectcell}
  >{\collectcell\num}R{2.18em}<{\endcollectcell}
  @{\tiny${}\pm{}$}
  >{\tiny\collectcell\num}R{1.308em}<{\endcollectcell}
  >{\collectcell\num}R{2.18em}<{\endcollectcell}
  @{\tiny${}\pm{}$}
  >{\tiny\collectcell\num}R{1.308em}<{\endcollectcell}
}
    \toprule
\multicolumn{2}{c}{\footnotesize \textbf{unconstrained}}&  \multicolumn{2}{c}{\scriptsize $\hat k=k/4$}  
    &  \multicolumn{2}{c}{\scriptsize $\hat k=k/2$}  
    &  \multicolumn{2}{c}{\scriptsize $\hat k=k$}  
    &  \multicolumn{2}{c}{\scriptsize $\hat k=2k$}  
    &  \multicolumn{2}{c}{\scriptsize $\hat k=4k$}  \\\midrule
\multirow{2}{*}{$\varepsilon=1$} & \multicolumn{1}{c}{\scriptsize precision} & 3.9 & 4.8 & 2.3 & 2.2 & 1.4 & 1.0 & 1.2 & 0.4 & 1.1 & 0.1\\
& \multicolumn{1}{c}{\scriptsize recall} & 1.0 & 1.2 & 1.2 & 1.2 & 1.4 & 1.0 & 2.3 & 0.8 & 4.3 & 0.6\\\midrule
\multirow{2}{*}{$\varepsilon=2$} & \multicolumn{1}{c}{\scriptsize precision} & 5.7 & 6.6 & 3.7 & 3.6 & 3.2 & 0.9 & 2.8 & 0.5 & 2.7 & 0.6\\
& \multicolumn{1}{c}{\scriptsize recall} & 1.4 & 1.7 & 1.8 & 1.9 & 3.1 & 1.1 & 5.4 & 0.7 & 10.2 & 1.7\\\midrule
\multirow{2}{*}{$\varepsilon=3$} & \multicolumn{1}{c}{\scriptsize precision} & 6.4 & 6.3 & 5.1 & 3.5 & 5.3 & 2.0 & 5.6 & 1.6 & 6.0 & 1.6\\
& \multicolumn{1}{c}{\scriptsize recall} & 1.6 & 1.6 & 2.5 & 1.8 & 5.1 & 2.1 & 10.5 & 2.6 & 22.5 & 4.7\\\midrule
\multirow{2}{*}{$\varepsilon=4$} & \multicolumn{1}{c}{\scriptsize precision} & 8.9 & 5.3 & 9.0 & 4.6 & 10.1 & 2.0 & 9.4 & 0.9 & 11.6 & 1.5\\
\bottomrule
    \end{tabular}%
}

\end{subtable}

    \begin{subtable}[]{\linewidth}
    \centering
    \vspace{0.5em}
    \resizebox{\linewidth}{!}{%
        \begin{tabular}{cc
  >{\collectcell\num}R{2.18em}<{\endcollectcell}
  @{\tiny${}\pm{}$}
  >{\tiny\collectcell\num}R{1.308em}<{\endcollectcell}
  >{\collectcell\num}R{2.18em}<{\endcollectcell}
  @{\tiny${}\pm{}$}
  >{\tiny\collectcell\num}R{1.308em}<{\endcollectcell}
  >{\collectcell\num}R{2.18em}<{\endcollectcell}
  @{\tiny${}\pm{}$}
  >{\tiny\collectcell\num}R{1.308em}<{\endcollectcell}
  >{\collectcell\num}R{2.18em}<{\endcollectcell}
  @{\tiny${}\pm{}$}
  >{\tiny\collectcell\num}R{1.308em}<{\endcollectcell}
  >{\collectcell\num}R{2.18em}<{\endcollectcell}
  @{\tiny${}\pm{}$}
  >{\tiny\collectcell\num}R{1.308em}<{\endcollectcell}
}
    \toprule
\multicolumn{2}{c}{\footnotesize \textbf{high}}&  \multicolumn{2}{c}{\scriptsize $\hat k=k/4$}  
    &  \multicolumn{2}{c}{\scriptsize $\hat k=k/2$}  
    &  \multicolumn{2}{c}{\scriptsize $\hat k=k$}  
    &  \multicolumn{2}{c}{\scriptsize $\hat k=2k$}  
    &  \multicolumn{2}{c}{\scriptsize $\hat k=4k$}  \\\midrule
\multirow{2}{*}{$\varepsilon=1$} & \multicolumn{1}{c}{\scriptsize precision} & 3.6 & 2.4 & 3.1 & 0.9 & 2.9 & 0.7 & 2.4 & 0.2 & 2.5 & 0.4\\
& \multicolumn{1}{c}{\scriptsize recall} & 0.9 & 0.6 & 1.6 & 0.5 & 3.0 & 0.8 & 4.9 & 0.5 & 10.3 & 1.8\\\midrule
\multirow{2}{*}{$\varepsilon=2$} & \multicolumn{1}{c}{\scriptsize precision} & 7.2 & 2.0 & 6.0 & 1.1 & 6.0 & 0.5 & 5.9 & 0.7 & 5.6 & 0.7\\
& \multicolumn{1}{c}{\scriptsize recall} & 1.8 & 0.5 & 3.1 & 0.7 & 6.1 & 0.6 & 12.1 & 1.9 & 22.8 & 3.8\\\midrule
\multirow{2}{*}{$\varepsilon=3$} & \multicolumn{1}{c}{\scriptsize precision} & 12.9 & 0.8 & 11.6 & 1.4 & 12.4 & 0.4 & 13.0 & 0.1 & 12.4 & 0.3\\
& \multicolumn{1}{c}{\scriptsize recall} & 3.3 & 0.3 & 5.9 & 1.0 & 12.6 & 0.8 & 26.4 & 1.1 & 50.3 & 3.8\\\midrule
\multirow{2}{*}{$\varepsilon=4$} & \multicolumn{1}{c}{\scriptsize precision} & 20.2 & 2.2 & 21.0 & 0.9 & 23.3 & 0.9 & 23.9 & 0.8 & 20.3 & 0.5\\
\bottomrule
    \end{tabular}%
}

\end{subtable}
\end{table}
\newpage

\begin{table}
    \centering
    \caption{twitch-DE (\lapgraph)}
    \label{tab:att_dp_twitch-DE-lap} 
    \begin{subtable}[]{\linewidth}
    \centering\vspace{0.5em}
    \resizebox{\linewidth}{!}{%
        \begin{tabular}{cc
  >{\collectcell\num}R{2.18em}<{\endcollectcell}
  @{\tiny${}\pm{}$}
  >{\tiny\collectcell\num}R{1.308em}<{\endcollectcell}
  >{\collectcell\num}R{2.18em}<{\endcollectcell}
  @{\tiny${}\pm{}$}
  >{\tiny\collectcell\num}R{1.308em}<{\endcollectcell}
  >{\collectcell\num}R{2.18em}<{\endcollectcell}
  @{\tiny${}\pm{}$}
  >{\tiny\collectcell\num}R{1.308em}<{\endcollectcell}
  >{\collectcell\num}R{2.18em}<{\endcollectcell}
  @{\tiny${}\pm{}$}
  >{\tiny\collectcell\num}R{1.308em}<{\endcollectcell}
  >{\collectcell\num}R{2.18em}<{\endcollectcell}
  @{\tiny${}\pm{}$}
  >{\tiny\collectcell\num}R{1.308em}<{\endcollectcell}
}
    \toprule
    \multicolumn{2}{c}{\footnotesize \textbf{low}}&  \multicolumn{2}{c}{\scriptsize $\hat k=k/4$}  
    &  \multicolumn{2}{c}{\scriptsize $\hat k=k/2$}  
    &  \multicolumn{2}{c}{\scriptsize $\hat k=k$}  
    &  \multicolumn{2}{c}{\scriptsize $\hat k=2k$}  
    &  \multicolumn{2}{c}{\scriptsize $\hat k=4k$}  \\\midrule
\multirow{2}{*}{$\varepsilon=1$} & \multicolumn{1}{c}{\scriptsize precision} & 0.0 & 0.0 & 0.0 & 0.0 & 0.0 & 0.0 & 0.0 & 0.0 & 0.0 & 0.0\\
& \multicolumn{1}{c}{\scriptsize recall} & 0.0 & 0.0 & 0.0 & 0.0 & 0.0 & 0.0 & 0.0 & 0.0 & 0.0 & 0.0\\\midrule
\multirow{2}{*}{$\varepsilon=2$} & \multicolumn{1}{c}{\scriptsize precision} & 0.0 & 0.0 & 0.0 & 0.0 & 0.0 & 0.0 & 0.0 & 0.0 & 0.0 & 0.0\\
& \multicolumn{1}{c}{\scriptsize recall} & 0.0 & 0.0 & 0.0 & 0.0 & 0.0 & 0.0 & 0.0 & 0.0 & 0.0 & 0.0\\\midrule
\multirow{2}{*}{$\varepsilon=3$} & \multicolumn{1}{c}{\scriptsize precision} & 0.0 & 0.0 & 0.0 & 0.0 & 0.0 & 0.0 & 0.0 & 0.0 & 0.0 & 0.0\\
& \multicolumn{1}{c}{\scriptsize recall} & 0.0 & 0.0 & 0.0 & 0.0 & 0.0 & 0.0 & 0.0 & 0.0 & 0.0 & 0.0\\\midrule
\multirow{2}{*}{$\varepsilon=4$} & \multicolumn{1}{c}{\scriptsize precision} & 0.0 & 0.0 & 0.0 & 0.0 & 0.0 & 0.0 & 0.0 & 0.0 & 0.0 & 0.0\\
\bottomrule
    \end{tabular}%
}
\end{subtable}

    \begin{subtable}[]{\linewidth}
    \centering\vspace{0.5em}
    \resizebox{\linewidth}{!}{%
        \begin{tabular}{cc
  >{\collectcell\num}R{2.18em}<{\endcollectcell}
  @{\tiny${}\pm{}$}
  >{\tiny\collectcell\num}R{1.308em}<{\endcollectcell}
  >{\collectcell\num}R{2.18em}<{\endcollectcell}
  @{\tiny${}\pm{}$}
  >{\tiny\collectcell\num}R{1.308em}<{\endcollectcell}
  >{\collectcell\num}R{2.18em}<{\endcollectcell}
  @{\tiny${}\pm{}$}
  >{\tiny\collectcell\num}R{1.308em}<{\endcollectcell}
  >{\collectcell\num}R{2.18em}<{\endcollectcell}
  @{\tiny${}\pm{}$}
  >{\tiny\collectcell\num}R{1.308em}<{\endcollectcell}
  >{\collectcell\num}R{2.18em}<{\endcollectcell}
  @{\tiny${}\pm{}$}
  >{\tiny\collectcell\num}R{1.308em}<{\endcollectcell}
}
    \toprule
\multicolumn{2}{c}{\footnotesize \textbf{unconstrained}}&  \multicolumn{2}{c}{\scriptsize $\hat k=k/4$}  
    &  \multicolumn{2}{c}{\scriptsize $\hat k=k/2$}  
    &  \multicolumn{2}{c}{\scriptsize $\hat k=k$}  
    &  \multicolumn{2}{c}{\scriptsize $\hat k=2k$}  
    &  \multicolumn{2}{c}{\scriptsize $\hat k=4k$}  \\\midrule
\multirow{2}{*}{$\varepsilon=1$} & \multicolumn{1}{c}{\scriptsize precision} & 1.1 & 0.9 & 0.7 & 0.3 & 0.9 & 0.3 & 0.8 & 0.5 & 0.6 & 0.3\\
& \multicolumn{1}{c}{\scriptsize recall} & 0.3 & 0.2 & 0.3 & 0.1 & 0.9 & 0.4 & 1.5 & 0.8 & 2.3 & 0.9\\\midrule
\multirow{2}{*}{$\varepsilon=2$} & \multicolumn{1}{c}{\scriptsize precision} & 2.5 & 1.3 & 2.0 & 0.9 & 1.9 & 0.8 & 1.4 & 0.7 & 1.0 & 0.5\\
& \multicolumn{1}{c}{\scriptsize recall} & 0.6 & 0.3 & 0.9 & 0.4 & 1.7 & 0.6 & 2.6 & 1.1 & 3.7 & 1.6\\\midrule
\multirow{2}{*}{$\varepsilon=3$} & \multicolumn{1}{c}{\scriptsize precision} & 6.4 & 1.7 & 5.7 & 1.3 & 4.8 & 0.6 & 3.0 & 0.5 & 1.8 & 0.4\\
& \multicolumn{1}{c}{\scriptsize recall} & 1.5 & 0.3 & 2.7 & 0.4 & 4.6 & 0.2 & 5.7 & 0.6 & 7.0 & 1.3\\\midrule
\multirow{2}{*}{$\varepsilon=4$} & \multicolumn{1}{c}{\scriptsize precision} & 11.0 & 1.8 & 12.2 & 2.6 & 12.3 & 1.5 & 7.3 & 0.8 & 4.0 & 0.6\\
\bottomrule
    \end{tabular}%
}

\end{subtable}

    \begin{subtable}[]{\linewidth}
    \centering\vspace{0.5em}
    \resizebox{\linewidth}{!}{%
        \begin{tabular}{cc
  >{\collectcell\num}R{2.18em}<{\endcollectcell}
  @{\tiny${}\pm{}$}
  >{\tiny\collectcell\num}R{1.308em}<{\endcollectcell}
  >{\collectcell\num}R{2.18em}<{\endcollectcell}
  @{\tiny${}\pm{}$}
  >{\tiny\collectcell\num}R{1.308em}<{\endcollectcell}
  >{\collectcell\num}R{2.18em}<{\endcollectcell}
  @{\tiny${}\pm{}$}
  >{\tiny\collectcell\num}R{1.308em}<{\endcollectcell}
  >{\collectcell\num}R{2.18em}<{\endcollectcell}
  @{\tiny${}\pm{}$}
  >{\tiny\collectcell\num}R{1.308em}<{\endcollectcell}
  >{\collectcell\num}R{2.18em}<{\endcollectcell}
  @{\tiny${}\pm{}$}
  >{\tiny\collectcell\num}R{1.308em}<{\endcollectcell}
}
    \toprule
\multicolumn{2}{c}{\footnotesize \textbf{high}}&  \multicolumn{2}{c}{\scriptsize $\hat k=k/4$}  
    &  \multicolumn{2}{c}{\scriptsize $\hat k=k/2$}  
    &  \multicolumn{2}{c}{\scriptsize $\hat k=k$}  
    &  \multicolumn{2}{c}{\scriptsize $\hat k=2k$}  
    &  \multicolumn{2}{c}{\scriptsize $\hat k=4k$}  \\\midrule
\multirow{2}{*}{$\varepsilon=1$} & \multicolumn{1}{c}{\scriptsize precision} & 2.1 & 1.0 & 1.9 & 0.6 & 1.6 & 0.5 & 1.3 & 0.3 & 1.1 & 0.1\\
& \multicolumn{1}{c}{\scriptsize recall} & 0.5 & 0.2 & 1.0 & 0.3 & 1.6 & 0.4 & 2.6 & 0.6 & 4.5 & 0.4\\\midrule
\multirow{2}{*}{$\varepsilon=2$} & \multicolumn{1}{c}{\scriptsize precision} & 5.5 & 1.3 & 5.2 & 0.4 & 3.2 & 0.2 & 2.2 & 0.0 & 1.6 & 0.0\\
& \multicolumn{1}{c}{\scriptsize recall} & 1.4 & 0.4 & 2.7 & 0.2 & 3.2 & 0.1 & 4.4 & 0.2 & 6.3 & 0.4\\\midrule
\multirow{2}{*}{$\varepsilon=3$} & \multicolumn{1}{c}{\scriptsize precision} & 13.7 & 1.4 & 12.7 & 0.3 & 7.6 & 0.3 & 4.3 & 0.3 & 2.6 & 0.2\\
& \multicolumn{1}{c}{\scriptsize recall} & 3.5 & 0.5 & 6.5 & 0.5 & 7.8 & 0.6 & 8.7 & 0.9 & 10.6 & 1.2\\\midrule
\multirow{2}{*}{$\varepsilon=4$} & \multicolumn{1}{c}{\scriptsize precision} & 26.1 & 1.2 & 26.2 & 0.7 & 15.4 & 1.1 & 8.5 & 0.5 & 4.8 & 0.2\\
\bottomrule
    \end{tabular}%
}

\end{subtable}
\end{table}
\newpage

\begin{table}
    \centering
    \caption{twitch-FR (\edgerand)}
    \label{tab:att_dp_twitch-FR-er}    
    \begin{subtable}[]{\linewidth}
    \centering
    \vspace{0.5em}
    \resizebox{\linewidth}{!}{%
        \begin{tabular}{cc
  >{\collectcell\num}R{2.18em}<{\endcollectcell}
  @{\tiny${}\pm{}$}
  >{\tiny\collectcell\num}R{1.308em}<{\endcollectcell}
  >{\collectcell\num}R{2.18em}<{\endcollectcell}
  @{\tiny${}\pm{}$}
  >{\tiny\collectcell\num}R{1.308em}<{\endcollectcell}
  >{\collectcell\num}R{2.18em}<{\endcollectcell}
  @{\tiny${}\pm{}$}
  >{\tiny\collectcell\num}R{1.308em}<{\endcollectcell}
  >{\collectcell\num}R{2.18em}<{\endcollectcell}
  @{\tiny${}\pm{}$}
  >{\tiny\collectcell\num}R{1.308em}<{\endcollectcell}
  >{\collectcell\num}R{2.18em}<{\endcollectcell}
  @{\tiny${}\pm{}$}
  >{\tiny\collectcell\num}R{1.308em}<{\endcollectcell}
}
    \toprule
    
\multicolumn{2}{c}{\footnotesize \textbf{low}}&  \multicolumn{2}{c}{\scriptsize $\hat k=k/4$}  
    &  \multicolumn{2}{c}{\scriptsize $\hat k=k/2$}  
    &  \multicolumn{2}{c}{\scriptsize $\hat k=k$}  
    &  \multicolumn{2}{c}{\scriptsize $\hat k=2k$}  
    &  \multicolumn{2}{c}{\scriptsize $\hat k=4k$}  \\\midrule
\multirow{2}{*}{$\varepsilon=1$} & \multicolumn{1}{c}{\scriptsize precision} & 0.0 & 0.0 & 0.0 & 0.0 & 0.0 & 0.0 & 0.0 & 0.0 & 0.0 & 0.0\\
& \multicolumn{1}{c}{\scriptsize recall} & 0.0 & 0.0 & 0.0 & 0.0 & 0.0 & 0.0 & 0.0 & 0.0 & 0.0 & 0.0\\\midrule
\multirow{2}{*}{$\varepsilon=2$} & \multicolumn{1}{c}{\scriptsize precision} & 0.0 & 0.0 & 0.0 & 0.0 & 0.0 & 0.0 & 0.0 & 0.0 & 0.0 & 0.0\\
& \multicolumn{1}{c}{\scriptsize recall} & 0.0 & 0.0 & 0.0 & 0.0 & 0.0 & 0.0 & 0.0 & 0.0 & 0.0 & 0.0\\\midrule
\multirow{2}{*}{$\varepsilon=3$} & \multicolumn{1}{c}{\scriptsize precision} & 0.0 & 0.0 & 0.0 & 0.0 & 0.0 & 0.0 & 0.0 & 0.0 & 0.0 & 0.0\\
& \multicolumn{1}{c}{\scriptsize recall} & 0.0 & 0.0 & 0.0 & 0.0 & 0.0 & 0.0 & 0.0 & 0.0 & 0.0 & 0.0\\\midrule
\multirow{2}{*}{$\varepsilon=4$} & \multicolumn{1}{c}{\scriptsize precision} & 0.0 & 0.0 & 0.0 & 0.0 & 0.0 & 0.0 & 0.0 & 0.0 & 0.0 & 0.0\\
\bottomrule
    \end{tabular}%
}

\end{subtable}

    \begin{subtable}[]{\linewidth}
    \centering
    \vspace{0.5em}
    \resizebox{\linewidth}{!}{%
        \begin{tabular}{cc
  >{\collectcell\num}R{2.18em}<{\endcollectcell}
  @{\tiny${}\pm{}$}
  >{\tiny\collectcell\num}R{1.308em}<{\endcollectcell}
  >{\collectcell\num}R{2.18em}<{\endcollectcell}
  @{\tiny${}\pm{}$}
  >{\tiny\collectcell\num}R{1.308em}<{\endcollectcell}
  >{\collectcell\num}R{2.18em}<{\endcollectcell}
  @{\tiny${}\pm{}$}
  >{\tiny\collectcell\num}R{1.308em}<{\endcollectcell}
  >{\collectcell\num}R{2.18em}<{\endcollectcell}
  @{\tiny${}\pm{}$}
  >{\tiny\collectcell\num}R{1.308em}<{\endcollectcell}
  >{\collectcell\num}R{2.18em}<{\endcollectcell}
  @{\tiny${}\pm{}$}
  >{\tiny\collectcell\num}R{1.308em}<{\endcollectcell}
}
    \toprule
\multicolumn{2}{c}{\footnotesize \textbf{unconstrained}}&  \multicolumn{2}{c}{\scriptsize $\hat k=k/4$}  
    &  \multicolumn{2}{c}{\scriptsize $\hat k=k/2$}  
    &  \multicolumn{2}{c}{\scriptsize $\hat k=k$}  
    &  \multicolumn{2}{c}{\scriptsize $\hat k=2k$}  
    &  \multicolumn{2}{c}{\scriptsize $\hat k=4k$}  \\\midrule
\multirow{2}{*}{$\varepsilon=1$} & \multicolumn{1}{c}{\scriptsize precision} & 7.1 & 7.2 & 3.8 & 3.4 & 2.5 & 2.0 & 1.8 & 1.0 & 1.5 & 0.6\\
& \multicolumn{1}{c}{\scriptsize recall} & 1.7 & 1.8 & 1.8 & 1.7 & 2.4 & 1.9 & 3.4 & 2.1 & 5.5 & 2.6\\\midrule
\multirow{2}{*}{$\varepsilon=2$} & \multicolumn{1}{c}{\scriptsize precision} & 9.1 & 9.4 & 5.4 & 4.8 & 4.5 & 3.1 & 3.7 & 1.9 & 3.6 & 1.3\\
& \multicolumn{1}{c}{\scriptsize recall} & 2.2 & 2.3 & 2.6 & 2.4 & 4.3 & 3.1 & 6.9 & 3.9 & 13.5 & 5.3\\\midrule
\multirow{2}{*}{$\varepsilon=3$} & \multicolumn{1}{c}{\scriptsize precision} & 11.2 & 10.3 & 9.1 & 6.1 & 8.1 & 4.2 & 9.1 & 2.8 & 9.6 & 2.7\\
& \multicolumn{1}{c}{\scriptsize recall} & 2.7 & 2.5 & 4.3 & 3.0 & 7.7 & 4.3 & 17.0 & 5.7 & 35.9 & 11.2\\\midrule
\multirow{2}{*}{$\varepsilon=4$} & \multicolumn{1}{c}{\scriptsize precision} & 15.0 & 12.7 & 16.7 & 5.4 & 17.0 & 4.9 & 18.7 & 4.0 & 18.3 & 2.8\\
\bottomrule
    \end{tabular}%
}

\end{subtable}

    \begin{subtable}[]{\linewidth}
    \centering
    \vspace{0.5em}
    \resizebox{\linewidth}{!}{%
        \begin{tabular}{cc
  >{\collectcell\num}R{2.18em}<{\endcollectcell}
  @{\tiny${}\pm{}$}
  >{\tiny\collectcell\num}R{1.308em}<{\endcollectcell}
  >{\collectcell\num}R{2.18em}<{\endcollectcell}
  @{\tiny${}\pm{}$}
  >{\tiny\collectcell\num}R{1.308em}<{\endcollectcell}
  >{\collectcell\num}R{2.18em}<{\endcollectcell}
  @{\tiny${}\pm{}$}
  >{\tiny\collectcell\num}R{1.308em}<{\endcollectcell}
  >{\collectcell\num}R{2.18em}<{\endcollectcell}
  @{\tiny${}\pm{}$}
  >{\tiny\collectcell\num}R{1.308em}<{\endcollectcell}
  >{\collectcell\num}R{2.18em}<{\endcollectcell}
  @{\tiny${}\pm{}$}
  >{\tiny\collectcell\num}R{1.308em}<{\endcollectcell}
}
    \toprule
\multicolumn{2}{c}{\footnotesize \textbf{high}}&  \multicolumn{2}{c}{\scriptsize $\hat k=k/4$}  
    &  \multicolumn{2}{c}{\scriptsize $\hat k=k/2$}  
    &  \multicolumn{2}{c}{\scriptsize $\hat k=k$}  
    &  \multicolumn{2}{c}{\scriptsize $\hat k=2k$}  
    &  \multicolumn{2}{c}{\scriptsize $\hat k=4k$}  \\\midrule
\multirow{2}{*}{$\varepsilon=1$} & \multicolumn{1}{c}{\scriptsize precision} & 4.6 & 4.2 & 3.7 & 2.4 & 3.4 & 0.9 & 3.2 & 1.0 & 3.2 & 0.5\\
& \multicolumn{1}{c}{\scriptsize recall} & 1.1 & 1.1 & 1.8 & 1.2 & 3.3 & 1.0 & 6.4 & 2.4 & 12.5 & 2.9\\\midrule
\multirow{2}{*}{$\varepsilon=2$} & \multicolumn{1}{c}{\scriptsize precision} & 8.7 & 2.4 & 8.5 & 2.5 & 8.5 & 3.2 & 8.1 & 2.4 & 7.7 & 1.2\\
& \multicolumn{1}{c}{\scriptsize recall} & 2.1 & 0.7 & 4.2 & 1.5 & 8.5 & 3.7 & 15.9 & 5.9 & 30.1 & 6.6\\\midrule
\multirow{2}{*}{$\varepsilon=3$} & \multicolumn{1}{c}{\scriptsize precision} & 14.9 & 1.6 & 18.3 & 1.0 & 19.2 & 0.2 & 18.2 & 0.4 & 16.8 & 0.5\\
& \multicolumn{1}{c}{\scriptsize recall} & 3.6 & 0.5 & 8.8 & 0.5 & 18.6 & 1.5 & 35.3 & 2.8 & 65.0 & 5.1\\\midrule
\multirow{2}{*}{$\varepsilon=4$} & \multicolumn{1}{c}{\scriptsize precision} & 31.8 & 3.0 & 32.6 & 2.2 & 32.2 & 0.4 & 32.9 & 0.3 & 21.9 & 0.5\\
\bottomrule
    \end{tabular}%
}

\end{subtable}
\end{table}
\newpage

\begin{table}
    \centering
    \caption{twitch-FR (\lapgraph)}
    \label{tab:att_dp_twitch-FR-lap}    
    \begin{subtable}[]{\linewidth}
    \centering\vspace{0.5em}
    \resizebox{\linewidth}{!}{%
        \begin{tabular}{cc
  >{\collectcell\num}R{2.18em}<{\endcollectcell}
  @{\tiny${}\pm{}$}
  >{\tiny\collectcell\num}R{1.308em}<{\endcollectcell}
  >{\collectcell\num}R{2.18em}<{\endcollectcell}
  @{\tiny${}\pm{}$}
  >{\tiny\collectcell\num}R{1.308em}<{\endcollectcell}
  >{\collectcell\num}R{2.18em}<{\endcollectcell}
  @{\tiny${}\pm{}$}
  >{\tiny\collectcell\num}R{1.308em}<{\endcollectcell}
  >{\collectcell\num}R{2.18em}<{\endcollectcell}
  @{\tiny${}\pm{}$}
  >{\tiny\collectcell\num}R{1.308em}<{\endcollectcell}
  >{\collectcell\num}R{2.18em}<{\endcollectcell}
  @{\tiny${}\pm{}$}
  >{\tiny\collectcell\num}R{1.308em}<{\endcollectcell}
}
    \toprule
\multicolumn{2}{c}{\footnotesize \textbf{low}}&  \multicolumn{2}{c}{\scriptsize $\hat k=k/4$}  
    &  \multicolumn{2}{c}{\scriptsize $\hat k=k/2$}  
    &  \multicolumn{2}{c}{\scriptsize $\hat k=k$}  
    &  \multicolumn{2}{c}{\scriptsize $\hat k=2k$}  
    &  \multicolumn{2}{c}{\scriptsize $\hat k=4k$}  \\\midrule
\multirow{2}{*}{$\varepsilon=1$} & \multicolumn{1}{c}{\scriptsize precision} & 0.0 & 0.0 & 0.0 & 0.0 & 0.0 & 0.0 & 0.0 & 0.0 & 0.0 & 0.0\\
& \multicolumn{1}{c}{\scriptsize recall} & 0.0 & 0.0 & 0.0 & 0.0 & 0.0 & 0.0 & 0.0 & 0.0 & 0.0 & 0.0\\\midrule
\multirow{2}{*}{$\varepsilon=2$} & \multicolumn{1}{c}{\scriptsize precision} & 0.0 & 0.0 & 0.0 & 0.0 & 0.0 & 0.0 & 0.0 & 0.0 & 1.1 & 1.6\\
& \multicolumn{1}{c}{\scriptsize recall} & 0.0 & 0.0 & 0.0 & 0.0 & 0.0 & 0.0 & 0.0 & 0.0 & 4.2 & 5.9\\\midrule
\multirow{2}{*}{$\varepsilon=3$} & \multicolumn{1}{c}{\scriptsize precision} & 0.0 & 0.0 & 0.0 & 0.0 & 0.0 & 0.0 & 2.2 & 3.1 & 1.9 & 1.4\\
& \multicolumn{1}{c}{\scriptsize recall} & 0.0 & 0.0 & 0.0 & 0.0 & 0.0 & 0.0 & 4.2 & 5.9 & 7.5 & 5.4\\\midrule
\multirow{2}{*}{$\varepsilon=4$} & \multicolumn{1}{c}{\scriptsize precision} & 0.0 & 0.0 & 0.0 & 0.0 & 0.0 & 0.0 & 0.0 & 0.0 & 1.1 & 1.6\\
\bottomrule
    \end{tabular}%
}

\end{subtable}

    \begin{subtable}[]{\linewidth}
    \centering\vspace{0.5em}
    \resizebox{\linewidth}{!}{%
        \begin{tabular}{cc
  >{\collectcell\num}R{2.18em}<{\endcollectcell}
  @{\tiny${}\pm{}$}
  >{\tiny\collectcell\num}R{1.308em}<{\endcollectcell}
  >{\collectcell\num}R{2.18em}<{\endcollectcell}
  @{\tiny${}\pm{}$}
  >{\tiny\collectcell\num}R{1.308em}<{\endcollectcell}
  >{\collectcell\num}R{2.18em}<{\endcollectcell}
  @{\tiny${}\pm{}$}
  >{\tiny\collectcell\num}R{1.308em}<{\endcollectcell}
  >{\collectcell\num}R{2.18em}<{\endcollectcell}
  @{\tiny${}\pm{}$}
  >{\tiny\collectcell\num}R{1.308em}<{\endcollectcell}
  >{\collectcell\num}R{2.18em}<{\endcollectcell}
  @{\tiny${}\pm{}$}
  >{\tiny\collectcell\num}R{1.308em}<{\endcollectcell}
}
    \toprule
\multicolumn{2}{c}{\footnotesize \textbf{unconstrained}}&  \multicolumn{2}{c}{\scriptsize $\hat k=k/4$}  
    &  \multicolumn{2}{c}{\scriptsize $\hat k=k/2$}  
    &  \multicolumn{2}{c}{\scriptsize $\hat k=k$}  
    &  \multicolumn{2}{c}{\scriptsize $\hat k=2k$}  
    &  \multicolumn{2}{c}{\scriptsize $\hat k=4k$}  \\\midrule
\multirow{2}{*}{$\varepsilon=1$} & \multicolumn{1}{c}{\scriptsize precision} & 1.8 & 0.3 & 2.1 & 0.4 & 2.0 & 0.4 & 1.4 & 0.2 & 1.1 & 0.2\\
& \multicolumn{1}{c}{\scriptsize recall} & 0.4 & 0.1 & 1.0 & 0.2 & 1.9 & 0.5 & 2.7 & 0.6 & 4.0 & 0.8\\\midrule
\multirow{2}{*}{$\varepsilon=2$} & \multicolumn{1}{c}{\scriptsize precision} & 3.8 & 0.8 & 4.3 & 1.4 & 4.1 & 0.9 & 2.7 & 0.3 & 1.6 & 0.3\\
& \multicolumn{1}{c}{\scriptsize recall} & 0.9 & 0.2 & 2.0 & 0.7 & 3.8 & 1.0 & 5.0 & 0.6 & 6.0 & 1.2\\\midrule
\multirow{2}{*}{$\varepsilon=3$} & \multicolumn{1}{c}{\scriptsize precision} & 7.5 & 0.8 & 7.7 & 2.2 & 10.0 & 0.8 & 6.1 & 0.1 & 3.4 & 0.2\\
& \multicolumn{1}{c}{\scriptsize recall} & 1.8 & 0.2 & 3.6 & 1.2 & 9.3 & 1.1 & 11.4 & 0.4 & 12.7 & 0.9\\\midrule
\multirow{2}{*}{$\varepsilon=4$} & \multicolumn{1}{c}{\scriptsize precision} & 15.1 & 5.6 & 17.9 & 4.8 & 18.6 & 2.8 & 12.2 & 0.4 & 6.6 & 0.2\\
\bottomrule
    \end{tabular}%
}

\end{subtable}

    \begin{subtable}[]{\linewidth}
    \centering\vspace{0.5em}
    \resizebox{\linewidth}{!}{%
        \begin{tabular}{cc
  >{\collectcell\num}R{2.18em}<{\endcollectcell}
  @{\tiny${}\pm{}$}
  >{\tiny\collectcell\num}R{1.308em}<{\endcollectcell}
  >{\collectcell\num}R{2.18em}<{\endcollectcell}
  @{\tiny${}\pm{}$}
  >{\tiny\collectcell\num}R{1.308em}<{\endcollectcell}
  >{\collectcell\num}R{2.18em}<{\endcollectcell}
  @{\tiny${}\pm{}$}
  >{\tiny\collectcell\num}R{1.308em}<{\endcollectcell}
  >{\collectcell\num}R{2.18em}<{\endcollectcell}
  @{\tiny${}\pm{}$}
  >{\tiny\collectcell\num}R{1.308em}<{\endcollectcell}
  >{\collectcell\num}R{2.18em}<{\endcollectcell}
  @{\tiny${}\pm{}$}
  >{\tiny\collectcell\num}R{1.308em}<{\endcollectcell}
}
    \toprule
\multicolumn{2}{c}{\footnotesize \textbf{high}}&  \multicolumn{2}{c}{\scriptsize $\hat k=k/4$}  
    &  \multicolumn{2}{c}{\scriptsize $\hat k=k/2$}  
    &  \multicolumn{2}{c}{\scriptsize $\hat k=k$}  
    &  \multicolumn{2}{c}{\scriptsize $\hat k=2k$}  
    &  \multicolumn{2}{c}{\scriptsize $\hat k=4k$}  \\\midrule
\multirow{2}{*}{$\varepsilon=1$} & \multicolumn{1}{c}{\scriptsize precision} & 3.0 & 1.0 & 3.2 & 1.2 & 2.7 & 0.7 & 2.4 & 0.3 & 1.7 & 0.2\\
& \multicolumn{1}{c}{\scriptsize recall} & 0.7 & 0.2 & 1.5 & 0.4 & 2.6 & 0.4 & 4.5 & 0.4 & 6.6 & 0.2\\\midrule
\multirow{2}{*}{$\varepsilon=2$} & \multicolumn{1}{c}{\scriptsize precision} & 7.8 & 0.6 & 8.4 & 0.9 & 5.3 & 0.7 & 3.5 & 0.6 & 2.6 & 0.3\\
& \multicolumn{1}{c}{\scriptsize recall} & 1.9 & 0.0 & 4.1 & 0.2 & 5.1 & 0.2 & 6.7 & 0.5 & 9.8 & 0.5\\\midrule
\multirow{2}{*}{$\varepsilon=3$} & \multicolumn{1}{c}{\scriptsize precision} & 16.3 & 1.2 & 17.3 & 0.8 & 10.8 & 1.4 & 6.2 & 0.8 & 3.9 & 0.4\\
& \multicolumn{1}{c}{\scriptsize recall} & 3.9 & 0.3 & 8.3 & 0.4 & 10.3 & 0.4 & 11.8 & 0.5 & 14.8 & 0.2\\\midrule
\multirow{2}{*}{$\varepsilon=4$} & \multicolumn{1}{c}{\scriptsize precision} & 30.2 & 0.6 & 33.9 & 1.4 & 22.9 & 1.7 & 12.9 & 1.3 & 7.4 & 0.6\\
\bottomrule
    \end{tabular}%
}

\end{subtable}
\end{table}
\newpage

\begin{table}
    \centering
    \caption{twitch-ENGB (\edgerand)}
    \label{tab:att_dp_twitch-ENGB-er}    
    \begin{subtable}[]{\linewidth}
    \centering
    \vspace{0.5em}
    \resizebox{\linewidth}{!}{%
        \begin{tabular}{cc
  >{\collectcell\num}R{2.18em}<{\endcollectcell}
  @{\tiny${}\pm{}$}
  >{\tiny\collectcell\num}R{1.308em}<{\endcollectcell}
  >{\collectcell\num}R{2.18em}<{\endcollectcell}
  @{\tiny${}\pm{}$}
  >{\tiny\collectcell\num}R{1.308em}<{\endcollectcell}
  >{\collectcell\num}R{2.18em}<{\endcollectcell}
  @{\tiny${}\pm{}$}
  >{\tiny\collectcell\num}R{1.308em}<{\endcollectcell}
  >{\collectcell\num}R{2.18em}<{\endcollectcell}
  @{\tiny${}\pm{}$}
  >{\tiny\collectcell\num}R{1.308em}<{\endcollectcell}
  >{\collectcell\num}R{2.18em}<{\endcollectcell}
  @{\tiny${}\pm{}$}
  >{\tiny\collectcell\num}R{1.308em}<{\endcollectcell}
}
    \toprule
\multicolumn{2}{c}{\footnotesize \textbf{low}}&  \multicolumn{2}{c}{\scriptsize $\hat k=k/4$}  
    &  \multicolumn{2}{c}{\scriptsize $\hat k=k/2$}  
    &  \multicolumn{2}{c}{\scriptsize $\hat k=k$}  
    &  \multicolumn{2}{c}{\scriptsize $\hat k=2k$}  
    &  \multicolumn{2}{c}{\scriptsize $\hat k=4k$}  \\\midrule
\multirow{2}{*}{$\varepsilon=1$} & \multicolumn{1}{c}{\scriptsize precision} & 0.0 & 0.0 & 0.0 & 0.0 & 0.0 & 0.0 & 0.0 & 0.0 & 0.0 & 0.0\\
& \multicolumn{1}{c}{\scriptsize recall} & 0.0 & 0.0 & 0.0 & 0.0 & 0.0 & 0.0 & 0.0 & 0.0 & 0.0 & 0.0\\\midrule
\multirow{2}{*}{$\varepsilon=2$} & \multicolumn{1}{c}{\scriptsize precision} & 0.0 & 0.0 & 0.0 & 0.0 & 0.0 & 0.0 & 0.0 & 0.0 & 0.0 & 0.0\\
& \multicolumn{1}{c}{\scriptsize recall} & 0.0 & 0.0 & 0.0 & 0.0 & 0.0 & 0.0 & 0.0 & 0.0 & 0.0 & 0.0\\\midrule
\multirow{2}{*}{$\varepsilon=3$} & \multicolumn{1}{c}{\scriptsize precision} & 0.0 & 0.0 & 0.0 & 0.0 & 0.0 & 0.0 & 0.0 & 0.0 & 0.0 & 0.0\\
& \multicolumn{1}{c}{\scriptsize recall} & 0.0 & 0.0 & 0.0 & 0.0 & 0.0 & 0.0 & 0.0 & 0.0 & 0.0 & 0.0\\\midrule
\multirow{2}{*}{$\varepsilon=4$} & \multicolumn{1}{c}{\scriptsize precision} & 0.0 & 0.0 & 0.0 & 0.0 & 0.0 & 0.0 & 0.7 & 0.9 & 2.3 & 1.2\\
\bottomrule
    \end{tabular}%
}

\end{subtable}

    \begin{subtable}[]{\linewidth}
    \centering
    \resizebox{\linewidth}{!}{%
        \begin{tabular}{cc
  >{\collectcell\num}R{2.18em}<{\endcollectcell}
  @{\tiny${}\pm{}$}
  >{\tiny\collectcell\num}R{1.308em}<{\endcollectcell}
  >{\collectcell\num}R{2.18em}<{\endcollectcell}
  @{\tiny${}\pm{}$}
  >{\tiny\collectcell\num}R{1.308em}<{\endcollectcell}
  >{\collectcell\num}R{2.18em}<{\endcollectcell}
  @{\tiny${}\pm{}$}
  >{\tiny\collectcell\num}R{1.308em}<{\endcollectcell}
  >{\collectcell\num}R{2.18em}<{\endcollectcell}
  @{\tiny${}\pm{}$}
  >{\tiny\collectcell\num}R{1.308em}<{\endcollectcell}
  >{\collectcell\num}R{2.18em}<{\endcollectcell}
  @{\tiny${}\pm{}$}
  >{\tiny\collectcell\num}R{1.308em}<{\endcollectcell}
}
    \toprule
\multicolumn{2}{c}{\footnotesize \textbf{unconstrained}}&  \multicolumn{2}{c}{\scriptsize $\hat k=k/4$}  
    &  \multicolumn{2}{c}{\scriptsize $\hat k=k/2$}  
    &  \multicolumn{2}{c}{\scriptsize $\hat k=k$}  
    &  \multicolumn{2}{c}{\scriptsize $\hat k=2k$}  
    &  \multicolumn{2}{c}{\scriptsize $\hat k=4k$}  \\\midrule
\multirow{2}{*}{$\varepsilon=1$} & \multicolumn{1}{c}{\scriptsize precision} & 1.1 & 1.5 & 0.8 & 0.7 & 0.5 & 0.4 & 0.6 & 0.5 & 0.4 & 0.4\\
& \multicolumn{1}{c}{\scriptsize recall} & 0.4 & 0.5 & 0.5 & 0.4 & 0.7 & 0.5 & 1.5 & 1.2 & 2.2 & 1.8\\\midrule
\multirow{2}{*}{$\varepsilon=2$} & \multicolumn{1}{c}{\scriptsize precision} & 1.1 & 1.5 & 1.3 & 1.0 & 1.3 & 0.8 & 0.9 & 0.2 & 0.7 & 0.2\\
& \multicolumn{1}{c}{\scriptsize recall} & 0.3 & 0.5 & 0.7 & 0.6 & 1.3 & 0.5 & 1.9 & 0.4 & 3.3 & 1.3\\\midrule
\multirow{2}{*}{$\varepsilon=3$} & \multicolumn{1}{c}{\scriptsize precision} & 0.0 & 0.0 & 0.8 & 1.1 & 1.2 & 0.9 & 1.5 & 0.6 & 1.9 & 0.8\\
& \multicolumn{1}{c}{\scriptsize recall} & 0.0 & 0.0 & 0.5 & 0.7 & 1.5 & 1.1 & 3.6 & 1.8 & 8.7 & 4.5\\\midrule
\multirow{2}{*}{$\varepsilon=4$} & \multicolumn{1}{c}{\scriptsize precision} & 2.1 & 2.0 & 1.6 & 1.7 & 3.2 & 1.3 & 3.3 & 1.0 & 4.5 & 1.1\\
\bottomrule
    \end{tabular}%
}

\end{subtable}

    \begin{subtable}[]{\linewidth}
    \centering
    \resizebox{\linewidth}{!}{%
        \begin{tabular}{cc
  >{\collectcell\num}R{2.18em}<{\endcollectcell}
  @{\tiny${}\pm{}$}
  >{\tiny\collectcell\num}R{1.308em}<{\endcollectcell}
  >{\collectcell\num}R{2.18em}<{\endcollectcell}
  @{\tiny${}\pm{}$}
  >{\tiny\collectcell\num}R{1.308em}<{\endcollectcell}
  >{\collectcell\num}R{2.18em}<{\endcollectcell}
  @{\tiny${}\pm{}$}
  >{\tiny\collectcell\num}R{1.308em}<{\endcollectcell}
  >{\collectcell\num}R{2.18em}<{\endcollectcell}
  @{\tiny${}\pm{}$}
  >{\tiny\collectcell\num}R{1.308em}<{\endcollectcell}
  >{\collectcell\num}R{2.18em}<{\endcollectcell}
  @{\tiny${}\pm{}$}
  >{\tiny\collectcell\num}R{1.308em}<{\endcollectcell}
}
    \toprule
\multicolumn{2}{c}{\footnotesize \textbf{high}}&  \multicolumn{2}{c}{\scriptsize $\hat k=k/4$}  
    &  \multicolumn{2}{c}{\scriptsize $\hat k=k/2$}  
    &  \multicolumn{2}{c}{\scriptsize $\hat k=k$}  
    &  \multicolumn{2}{c}{\scriptsize $\hat k=2k$}  
    &  \multicolumn{2}{c}{\scriptsize $\hat k=4k$}  \\\midrule
\multirow{2}{*}{$\varepsilon=1$} & \multicolumn{1}{c}{\scriptsize precision} & 4.2 & 0.9 & 4.1 & 1.2 & 4.0 & 1.5 & 3.0 & 0.8 & 2.9 & 0.5\\
& \multicolumn{1}{c}{\scriptsize recall} & 1.1 & 0.2 & 2.1 & 0.6 & 4.1 & 1.5 & 6.0 & 1.7 & 11.9 & 2.1\\\midrule
\multirow{2}{*}{$\varepsilon=2$} & \multicolumn{1}{c}{\scriptsize precision} & 6.1 & 2.1 & 7.2 & 2.9 & 7.0 & 1.6 & 6.7 & 0.7 & 6.0 & 0.6\\
& \multicolumn{1}{c}{\scriptsize recall} & 1.5 & 0.5 & 3.7 & 1.5 & 7.1 & 1.6 & 13.7 & 1.4 & 24.4 & 2.2\\\midrule
\multirow{2}{*}{$\varepsilon=3$} & \multicolumn{1}{c}{\scriptsize precision} & 19.9 & 5.3 & 18.6 & 1.5 & 17.0 & 0.9 & 16.8 & 0.6 & 15.2 & 0.6\\
& \multicolumn{1}{c}{\scriptsize recall} & 5.1 & 1.3 & 9.5 & 0.7 & 17.3 & 0.9 & 34.3 & 1.0 & 61.9 & 2.2\\\midrule
\multirow{2}{*}{$\varepsilon=4$} & \multicolumn{1}{c}{\scriptsize precision} & 29.7 & 3.7 & 29.5 & 2.2 & 31.2 & 1.7 & 31.8 & 1.6 & 21.4 & 0.3\\
\bottomrule
    \end{tabular}%
}

\end{subtable}
\end{table}
\newpage

\begin{table}
    \centering
    \caption{twitch-ENGB (\lapgraph)}
    \label{tab:att_dp_twitch-ENGB-lap}    
    \begin{subtable}[]{\linewidth}
    \centering\vspace{0.5em}
    \resizebox{\linewidth}{!}{%
        \begin{tabular}{cc
  >{\collectcell\num}R{2.18em}<{\endcollectcell}
  @{\tiny${}\pm{}$}
  >{\tiny\collectcell\num}R{1.308em}<{\endcollectcell}
  >{\collectcell\num}R{2.18em}<{\endcollectcell}
  @{\tiny${}\pm{}$}
  >{\tiny\collectcell\num}R{1.308em}<{\endcollectcell}
  >{\collectcell\num}R{2.18em}<{\endcollectcell}
  @{\tiny${}\pm{}$}
  >{\tiny\collectcell\num}R{1.308em}<{\endcollectcell}
  >{\collectcell\num}R{2.18em}<{\endcollectcell}
  @{\tiny${}\pm{}$}
  >{\tiny\collectcell\num}R{1.308em}<{\endcollectcell}
  >{\collectcell\num}R{2.18em}<{\endcollectcell}
  @{\tiny${}\pm{}$}
  >{\tiny\collectcell\num}R{1.308em}<{\endcollectcell}
}
    \toprule
\multicolumn{2}{c}{\footnotesize \textbf{low}}&  \multicolumn{2}{c}{\scriptsize $\hat k=k/4$}  
    &  \multicolumn{2}{c}{\scriptsize $\hat k=k/2$}  
    &  \multicolumn{2}{c}{\scriptsize $\hat k=k$}  
    &  \multicolumn{2}{c}{\scriptsize $\hat k=2k$}  
    &  \multicolumn{2}{c}{\scriptsize $\hat k=4k$}  \\\midrule
\multirow{2}{*}{$\varepsilon=1$} & \multicolumn{1}{c}{\scriptsize precision} & 0.0 & 0.0 & 0.0 & 0.0 & 0.0 & 0.0 & 0.0 & 0.0 & 0.0 & 0.0\\
& \multicolumn{1}{c}{\scriptsize recall} & 0.0 & 0.0 & 0.0 & 0.0 & 0.0 & 0.0 & 0.0 & 0.0 & 0.0 & 0.0\\\midrule
\multirow{2}{*}{$\varepsilon=2$} & \multicolumn{1}{c}{\scriptsize precision} & 0.0 & 0.0 & 0.0 & 0.0 & 0.0 & 0.0 & 0.0 & 0.0 & 0.0 & 0.0\\
& \multicolumn{1}{c}{\scriptsize recall} & 0.0 & 0.0 & 0.0 & 0.0 & 0.0 & 0.0 & 0.0 & 0.0 & 0.0 & 0.0\\\midrule
\multirow{2}{*}{$\varepsilon=3$} & \multicolumn{1}{c}{\scriptsize precision} & 0.0 & 0.0 & 0.0 & 0.0 & 0.0 & 0.0 & 0.0 & 0.0 & 0.0 & 0.0\\
& \multicolumn{1}{c}{\scriptsize recall} & 0.0 & 0.0 & 0.0 & 0.0 & 0.0 & 0.0 & 0.0 & 0.0 & 0.0 & 0.0\\\midrule
\multirow{2}{*}{$\varepsilon=4$} & \multicolumn{1}{c}{\scriptsize precision} & 0.0 & 0.0 & 0.0 & 0.0 & 0.0 & 0.0 & 0.0 & 0.0 & 0.0 & 0.0\\
\bottomrule
    \end{tabular}%
}

\end{subtable}

    \begin{subtable}[]{\linewidth}
    \centering\vspace{0.5em}
    \resizebox{\linewidth}{!}{%
        \begin{tabular}{cc
  >{\collectcell\num}R{2.18em}<{\endcollectcell}
  @{\tiny${}\pm{}$}
  >{\tiny\collectcell\num}R{1.308em}<{\endcollectcell}
  >{\collectcell\num}R{2.18em}<{\endcollectcell}
  @{\tiny${}\pm{}$}
  >{\tiny\collectcell\num}R{1.308em}<{\endcollectcell}
  >{\collectcell\num}R{2.18em}<{\endcollectcell}
  @{\tiny${}\pm{}$}
  >{\tiny\collectcell\num}R{1.308em}<{\endcollectcell}
  >{\collectcell\num}R{2.18em}<{\endcollectcell}
  @{\tiny${}\pm{}$}
  >{\tiny\collectcell\num}R{1.308em}<{\endcollectcell}
  >{\collectcell\num}R{2.18em}<{\endcollectcell}
  @{\tiny${}\pm{}$}
  >{\tiny\collectcell\num}R{1.308em}<{\endcollectcell}
}
    \toprule
\multicolumn{2}{c}{\footnotesize \textbf{unconstrained}}&  \multicolumn{2}{c}{\scriptsize $\hat k=k/4$}  
    &  \multicolumn{2}{c}{\scriptsize $\hat k=k/2$}  
    &  \multicolumn{2}{c}{\scriptsize $\hat k=k$}  
    &  \multicolumn{2}{c}{\scriptsize $\hat k=2k$}  
    &  \multicolumn{2}{c}{\scriptsize $\hat k=4k$}  \\\midrule
\multirow{2}{*}{$\varepsilon=1$} & \multicolumn{1}{c}{\scriptsize precision} & 0.5 & 0.7 & 0.3 & 0.4 & 0.1 & 0.2 & 0.2 & 0.2 & 0.2 & 0.2\\
& \multicolumn{1}{c}{\scriptsize recall} & 0.2 & 0.2 & 0.2 & 0.2 & 0.2 & 0.2 & 0.5 & 0.4 & 1.0 & 1.0\\\midrule
\multirow{2}{*}{$\varepsilon=2$} & \multicolumn{1}{c}{\scriptsize precision} & 1.1 & 1.5 & 0.5 & 0.8 & 0.4 & 0.6 & 0.3 & 0.4 & 0.3 & 0.1\\
& \multicolumn{1}{c}{\scriptsize recall} & 0.3 & 0.5 & 0.3 & 0.5 & 0.5 & 0.7 & 0.7 & 0.9 & 1.2 & 0.6\\\midrule
\multirow{2}{*}{$\varepsilon=3$} & \multicolumn{1}{c}{\scriptsize precision} & 3.7 & 2.7 & 2.7 & 0.7 & 2.1 & 0.8 & 1.3 & 0.6 & 0.8 & 0.3\\
& \multicolumn{1}{c}{\scriptsize recall} & 0.9 & 0.6 & 1.4 & 0.4 & 2.1 & 0.4 & 2.5 & 0.7 & 3.4 & 0.9\\\midrule
\multirow{2}{*}{$\varepsilon=4$} & \multicolumn{1}{c}{\scriptsize precision} & 5.3 & 1.5 & 6.1 & 2.5 & 4.9 & 1.5 & 3.0 & 1.2 & 1.8 & 0.6\\
\bottomrule
    \end{tabular}%
}

\end{subtable}

    \begin{subtable}[]{\linewidth}
    \centering\vspace{0.5em}
    \resizebox{\linewidth}{!}{%
        \begin{tabular}{cc
  >{\collectcell\num}R{2.18em}<{\endcollectcell}
  @{\tiny${}\pm{}$}
  >{\tiny\collectcell\num}R{1.308em}<{\endcollectcell}
  >{\collectcell\num}R{2.18em}<{\endcollectcell}
  @{\tiny${}\pm{}$}
  >{\tiny\collectcell\num}R{1.308em}<{\endcollectcell}
  >{\collectcell\num}R{2.18em}<{\endcollectcell}
  @{\tiny${}\pm{}$}
  >{\tiny\collectcell\num}R{1.308em}<{\endcollectcell}
  >{\collectcell\num}R{2.18em}<{\endcollectcell}
  @{\tiny${}\pm{}$}
  >{\tiny\collectcell\num}R{1.308em}<{\endcollectcell}
  >{\collectcell\num}R{2.18em}<{\endcollectcell}
  @{\tiny${}\pm{}$}
  >{\tiny\collectcell\num}R{1.308em}<{\endcollectcell}
}
    \toprule
\multicolumn{2}{c}{\footnotesize \textbf{high}}&  \multicolumn{2}{c}{\scriptsize $\hat k=k/4$}  
    &  \multicolumn{2}{c}{\scriptsize $\hat k=k/2$}  
    &  \multicolumn{2}{c}{\scriptsize $\hat k=k$}  
    &  \multicolumn{2}{c}{\scriptsize $\hat k=2k$}  
    &  \multicolumn{2}{c}{\scriptsize $\hat k=4k$}  \\\midrule
\multirow{2}{*}{$\varepsilon=1$} & \multicolumn{1}{c}{\scriptsize precision} & 1.7 & 0.4 & 1.2 & 0.3 & 1.1 & 0.3 & 0.9 & 0.2 & 0.4 & 0.1\\
& \multicolumn{1}{c}{\scriptsize recall} & 0.4 & 0.1 & 0.6 & 0.1 & 1.1 & 0.4 & 1.7 & 0.3 & 1.7 & 0.3\\\midrule
\multirow{2}{*}{$\varepsilon=2$} & \multicolumn{1}{c}{\scriptsize precision} & 3.2 & 0.5 & 2.0 & 0.4 & 1.5 & 0.2 & 1.1 & 0.2 & 0.6 & 0.1\\
& \multicolumn{1}{c}{\scriptsize recall} & 0.8 & 0.1 & 1.0 & 0.2 & 1.6 & 0.2 & 2.3 & 0.3 & 2.3 & 0.3\\\midrule
\multirow{2}{*}{$\varepsilon=3$} & \multicolumn{1}{c}{\scriptsize precision} & 9.3 & 2.1 & 5.4 & 1.2 & 3.4 & 0.8 & 2.2 & 0.4 & 1.1 & 0.2\\
& \multicolumn{1}{c}{\scriptsize recall} & 2.4 & 0.5 & 2.8 & 0.6 & 3.5 & 0.8 & 4.5 & 0.7 & 4.5 & 0.7\\\midrule
\multirow{2}{*}{$\varepsilon=4$} & \multicolumn{1}{c}{\scriptsize precision} & 24.7 & 2.5 & 13.2 & 1.4 & 7.9 & 0.8 & 5.1 & 0.5 & 3.1 & 0.9\\
\bottomrule
    \end{tabular}%
}

\end{subtable}
\end{table}
\newpage

\begin{table}
    \centering
    \caption{twitch-PTBR (\edgerand)}
    \label{tab:att_dp_twitch-PTBR-er}    

    \begin{subtable}[]{\linewidth}
    \centering
    \vspace{0.5em}
    \resizebox{\linewidth}{!}{%
        %
%
}

\end{subtable}
\end{table}
\newpage

\begin{table}
    \centering
    \caption{twitch-PTBR (\lapgraph)}
    \label{tab:att_dp_twitch-PTBR-lap}    
    \begin{subtable}[]{\linewidth}
    \centering\vspace{0.5em}
    \resizebox{\linewidth}{!}{%
        %
%
}

\end{subtable}
\end{table}
\newpage

\begin{table}
    \centering
    \caption{PPI (\edgerand)}
    \label{tab:att_dp_twitch-PTBR-er}    
    \begin{subtable}[]{\linewidth}
    \centering
    \resizebox{\linewidth}{!}{%
        %
%
}

\end{subtable}
\end{table}
\newpage

\begin{table}
    \centering
    \caption{PPI (\lapgraph)}
    \label{tab:att_dp_twitch-PTBR-lap}    
    \begin{subtable}[]{\linewidth}
    \centering\vspace{0.5em}
    \resizebox{\linewidth}{!}{%
        %
%
}

\end{subtable}
\end{table}
\newpage

\begin{table}
    \centering
    \caption{Flickr (\edgerand)}
    \label{tab:att_dp_twitch-PTBR-er}    
    \begin{subtable}[]{\linewidth}
    \centering
    \vspace{0.5em}
    \resizebox{\linewidth}{!}{%
        %
%
}

\end{subtable}
\end{table}
\newpage

\begin{table}
    \centering
    \caption{Flickr (\lapgraph)}
    \label{tab:att_dp_twitch-PTBR-lap}    
    \begin{subtable}[]{\linewidth}
    \centering\vspace{0.5em}
    \resizebox{\linewidth}{!}{%
        %
%
}

\end{subtable}
\end{table}
\newpage

}




\end{document}